\documentclass[sn-mathphys-num]{sn-jnl}

\usepackage{lmodern}  
\usepackage{graphicx}%
\usepackage{multirow}%
\usepackage{amsmath,amssymb,amsfonts}%
\usepackage{amsthm}%
\usepackage{mathrsfs}%
\usepackage[title]{appendix}%
\usepackage{textcomp}%
\usepackage{manyfoot}%
\usepackage{booktabs}%
\usepackage{algorithm}%
\usepackage{algorithmicx}%
\usepackage{algpseudocode}%
\usepackage{listings}%
\usepackage[dvipsnames]{xcolor}
\usepackage[utf8]{inputenc} 
\usepackage[T1]{fontenc}    
\usepackage{url}            
\usepackage{nicefrac}       
\usepackage{microtype}      
\usepackage{multirow}
\usepackage{bbm}
\usepackage{rotating}
\usepackage{xspace}

\microtypecontext{expansion=none}  

\theoremstyle{thmstyleone}%
\theoremstyle{thmstyletwo}%

\theoremstyle{thmstylethree}%
\makeatletter
\DeclareRobustCommand\onedot{\futurelet\@let@token\@onedot}
\def\@onedot{\ifx\@let@token.\else.\null\fi\xspace}

\def\eg{\emph{e.g}\onedot} 
\def\ie{\emph{i.e}\onedot} 
 
 \def\vs{\emph{vs}\onedot}
\def\wrt{w.r.t\onedot} 
\def\etal{\emph{et al}\onedot}
\makeatother
\DeclareMathOperator*{\argmin}{arg\,min}

\definecolor{green}{RGB}{0,150,10}

\newcommand*{\pd}[3][]{\ensuremath{\frac{\partial^{#1} #2}{\partial #3}}}

\newcommand{\figLabel}{Fig.~}
\newcommand{\eqLabel}[1]{{Eq (#1)}}
\newcommand{\secLabel}{Section~}

\newcommand{\mysection}[1]{\noindent\textbf{#1.}}
\newcommand{\supp}{\textbf{supplementary material\xspace} }
\newcommand{\specialcell}[2][c]{%
  \begin{tabular}[#1]{@{}c@{}}#2\end{tabular}}

\DisableLigatures[f]{encoding = *}
\microtypecontext{spacing=nonfrench}

\begin{document}

\title[MVTN: Learning Multi-View Transformations for 3D Understanding]{MVTN: Learning Multi-View Transformations for 3D Understanding}

\author{Abdullah Hamdi \quad\quad Faisal AlZahrani \quad\quad Silvio Giancola \quad \quad Bernard Ghanem\\  
King Abdullah University of Science and Technology (KAUST), Thuwal, Saudi Arabia\\
\small{\{abdullah.hamdi, faisal.zahrani.1, silvio.giancola, bernard.ghanem\}@kaust.edu.sa}
}




\abstract{Multi-view projection techniques have shown themselves to be highly effective in achieving top-performing results in the recognition of 3D shapes. These methods involve learning how to combine information from multiple view-points. However, the camera view-points from which these views are obtained are often fixed for all shapes. To overcome the static nature of current multi-view techniques, we propose learning these view-points. Specifically, we introduce the Multi-View Transformation Network (MVTN), which uses differentiable rendering to determine optimal view-points for 3D shape recognition. As a result, MVTN can be trained end-to-end with any multi-view network for 3D shape classification. We integrate MVTN into a novel adaptive multi-view pipeline that is capable of rendering both 3D meshes and point clouds. Our approach demonstrates state-of-the-art performance in 3D classification and shape retrieval on several benchmarks (ModelNet40, ScanObjectNN, ShapeNet Core55). Further analysis indicates that our approach exhibits improved robustness to occlusion compared to other methods. We also investigate additional aspects of MVTN, such as 2D pretraining and its use for segmentation. To support further research in this area, we have released MVTorch, a PyTorch library for 3D understanding and generation using multi-view projections.}




\maketitle
Given its success in the 2D realm, deep learning naturally expanded to the 3D vision domain. Deep learning networks have achieved impressive results in 3D tasks including classification, segmentation, and detection. These 3D deep learning pipelines generally operate directly on 3D data, which can be represented as point clouds \cite{pointnet,pointnet++,dgcn}, meshes \cite{meshnet,meshcnn}, or voxels \cite{voxnet,minkosky,sparseconv}. However, another approach is to represent 3D information through the rendering of multiple 2D views of objects or scenes, as seen in multi-view methods such as MVCNN \cite{mvcnn}. This approach more closely resembles how the human visual system processes information, as it receives streams of rendered images rather than more elaborate 3D representations.

\begin{figure}[t]
    \centering
    \includegraphics[trim= 5cm 3cm 8cm 3cm , clip,width=0.7\linewidth ]{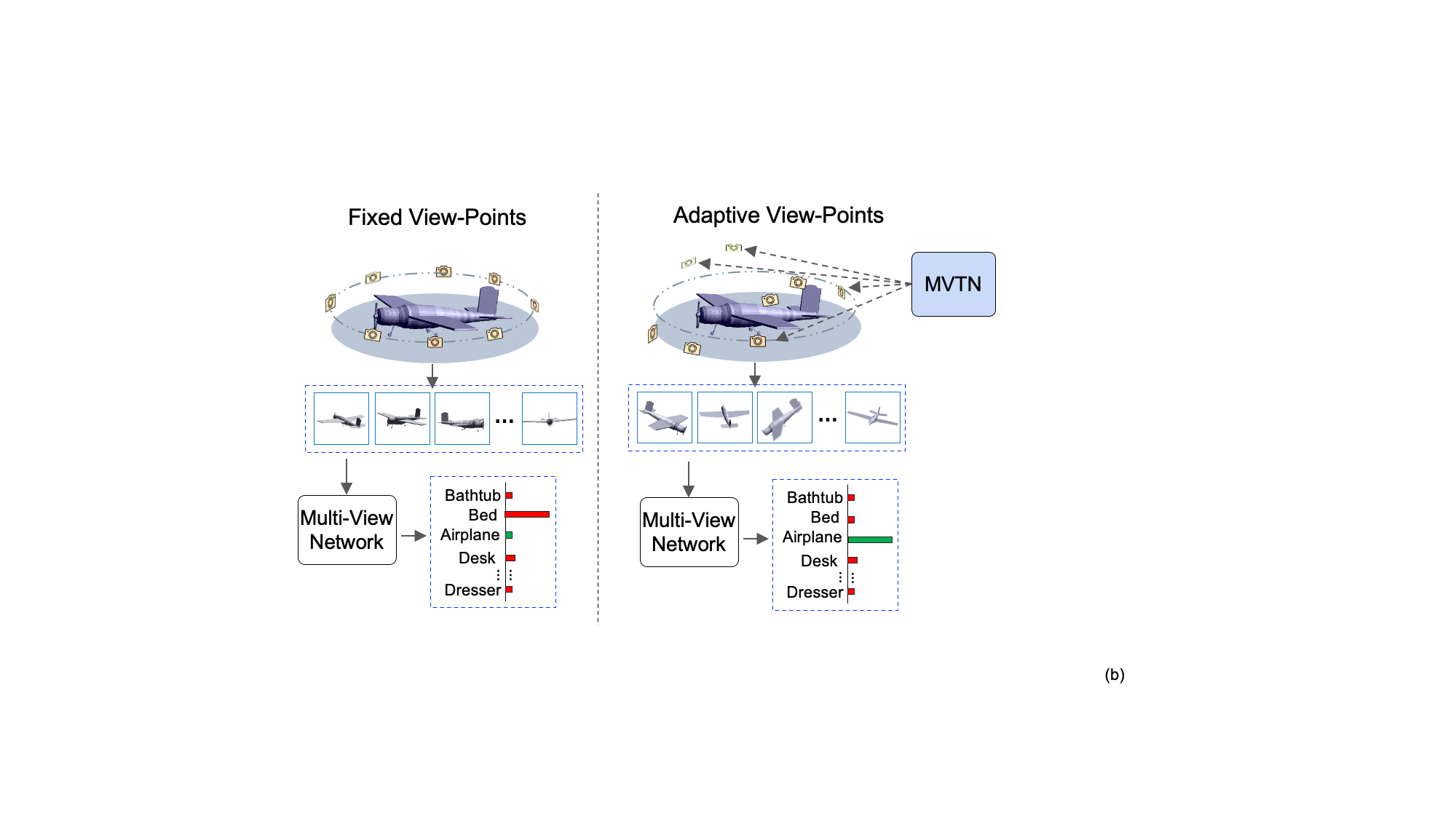}
    \caption{   \textbf{Multi-View Transformation Network (MVTN).} We propose a differentiable module that predicts the best view-points for a task-specific multi-view network. MVTN is trained jointly with this network without any extra training supervision, while improving the performance on 3D classification and shape retrieval.}
    \label{fig:pullingFigure}
\end{figure}

Recent multi-view methods have shown great performance in 3D shape classification and segmentation, often achieving state-of-the-art results \cite{mvrotationnet,mvviewgcn,mvvirtualsceneseg,mvshapeseg,mvsceneseg}. These approaches use 2D convolutional architectures to solve 3D tasks by rendering multiple views of a given 3D shape and leveraging the rendered images. This allows them to build upon advances in 2D deep learning and make use of larger image datasets, such as ImageNet \cite{IMAGENET}, to compensate for the lack of labeled 3D data. However, the choice of rendering view-points for these methods is often based on heuristics, such as random sampling \cite{mvvirtualsceneseg} or predefined view-points, rather than being optimized for the task at hand. To address this issue, we propose the Multi-View Transformation Network (MVTN), which learns to regress suitable view-points for a given task and trains a downstream task-specific network in an end-to-end fashion using these views. As shown in \figLabel{\ref{fig:pullingFigure}}, MVTN learns to regress view-points, renders those views with a differentiable renderer, and trains the downstream task-specific network in an end-to-end fashion, thus leading to the most suitable views for the task. This approach is inspired by the Spatial Transformer Network (STN) \cite{stn}, which performs a similar function in the 2D image domain. Both MVTN and STN learn spatial transformations for the input without requiring additional supervision or adjustments to the learning process.

The concept of perception through the prediction of the best environment parameters that generated an image is known as Vision as Inverse Graphics (VIG) \cite{old-vision1,vig-cinvg,vig-nsd,vig-reinforce,vig-inverse-render-net}. One approach to VIG is to make the rendering process invertible or differentiable \cite{vig-open-dr,vig-nmr,soft-rasterizer,vig-bid-r,vig-monte-carlo-raytrace}. In this paper, we use the Multi-View Transformation Network (MVTN) to take advantage of differentiable rendering \cite{vig-nmr,soft-rasterizer,pytorch3d} in order to train models end-to-end for a specific 3D vision task, with the view-points (\ie camera poses) being inferred by MVTN in the same forward pass. To the best of our knowledge, we are the first to integrate a learnable approach to view-point prediction in multi-view methods by using a differentiable renderer and establishing an end-to-end pipeline that works for both mesh and 3D point cloud classification and retrieval.

\vspace{2pt}\noindent\textbf{Contributions:} \textbf{(1)} We propose a Multi-View Transformation Network (MVTN) that regresses better view-points for multi-view methods. Our MVTN leverages a differentiable renderer that enables end-to-end training for 3D shape recognition tasks. 
\textbf{(2)} Combining MVTN with multi-view approaches leads to state-of-the-art results in 3D classification and shape retrieval on standard benchmarks ModelNet40 \cite{modelnet}, ShapeNet Core55 \cite{shapenet,shrek17}, and ScanObjectNN \cite{scanobjectnn}. 
\textbf{(3)} Additional analysis shows that MVTN improves the robustness of multi-view approaches to rotation and occlusion.
\textbf{(4)} We investigate an optimization alternative to MVTN, study different 2D pretraining strategies on MVTN, and study extending MVTN for 3D segmentation. To wrap up the work, we release MVTorch, a modular Pytroch library for multi-view research

A preliminary version of this work was published at ICCV 2021 \cite{mvtn}.
This journal manuscript extends the initial version in several
aspects. First, we investigate and experiment with a logical alternative to MVTN by treating the problem as an optimization of the scene parameters instead of learning a transformation network. 
Second, we study the effect of different pretraining strategies of the 2D backbone, which was shown in \cite{mvtn} to play an important role in MVTN's success.  
Third, we extend MVTN to the 3D part segmentation task and show promise in learning views beyond classification pipelines.
Finally, to ensure the reproducibility of our experiments and to contribute to the 3D understanding/generation research community, we have published \textit{MVTorch}, a modular Pytorch library for training, testing, and visualization of multi-view deep learning pipelines.

\section{Related Work} \label{sec:related}
\mysection{Deep Learning on 3D Data}
PointNet \cite{pointnet} was the first deep learning algorithm to operate directly on 3D point clouds. It computed point features independently and aggregated them using an order invariant function such as max-pooling. Subsequent works focused on finding neighborhoods of points in order to define point convolutional operations \cite{pointnet++,dgcn,pc_li2018pointcnn,pc_landrieu2018large,pc_landrieu2019point,pc_wang2018sgpn}. Voxel-based deep networks enable 3D CNNs, but they suffer from cubic memory complexity \cite{voxnet,minkosky,sparseconv}. Some recent works have combined point cloud representations with other 3D modalities, such as voxels \cite{pvoxelcnn} or multi-view images \cite{pvnet,mvpnet}. In this paper, we use a point encoder to predict optimal view-points, from which images are rendered and fed to a multi-view network.

\mysection{Multi-View 3D Shape Classification}
The use of 2D images to recognize 3D objects was first proposed by Bradski~\etal~\cite{bradski1994recognition}. Two decades later, MVCNN \cite{mvcnn} emerged as the first application of deep 2D CNNs for 3D object recognition. MVCNN used max pooling to aggregate features from different views. Subsequent works proposed different strategies for assigning weights to views in order to perform weighted average pooling of view-specific features \cite{mvnhbn,mvrelations,mvgvcnn,mvvram}. RotationNet \cite{mvrotationnet} classified views and objects jointly, while Equivariant MV-Network \cite{mvequivariant} used a rotation equivariant convolution operation on multiple views with rotation group convolutions \cite{groupconv}. ViewGCN \cite{mvviewgcn} used dynamic graph convolution operations to adaptively pool features from fixed views for 3D shape classification. Previous methods relied on fixed rendered datasets of 3D objects. The work of \cite{mvvram} attempted to adaptively select views through reinforcement learning and RNNs, but it had limited success and required a complex training process. In this paper, we propose the Multi-View Transformation Network (MVTN) for predicting optimal view-points in a multi-view setup, by jointly training MVTN with a multi-view task-specific network without requiring any additional supervision or adjustments to the learning process.

\begin{figure*}
    \centering
    \includegraphics[trim= 0cm 2.4cm 0cm 5.5cm , clip,width=0.99\linewidth]{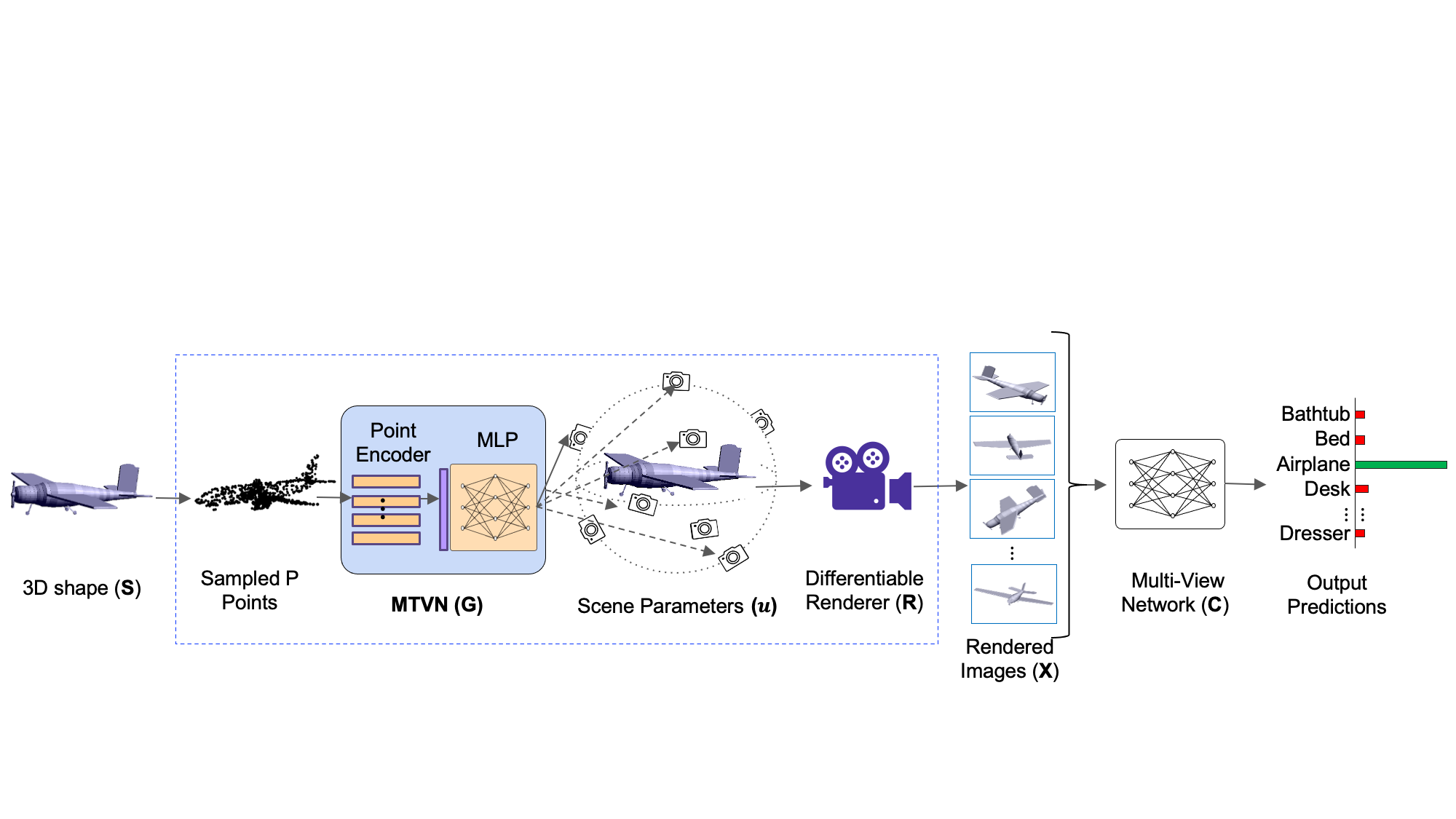}
    \caption{\textbf{End-to-End Learning Pipeline for Multi-View Recognition.} To learn adaptive scene parameters $\mathbf{u}$ that maximize the performance of a multi-view network $\mathbf{C}$ for every 3D object shape $\mathbf{S}$, we use a differentiable renderer $\mathbf{R}$. MVTN extracts coarse features from $\mathbf{S}$ by a point encoder and regresses the adaptive scene parameters for that object. In this example, the parameters $\mathbf{u}$ are the azimuth and elevation angles of cameras pointing towards the center of the object. The MVTN pipeline is optimized end-to-end for the task loss. 
    }
    \label{fig:pipeline}
\end{figure*}

\mysection{3D Shape Retrieval}
Early methods in the literature compared the distribution of hand-crafted descriptors to retrieve similar 3D shapes. These shape signatures could represent either geometric~\cite{osada2002shape} or visual~\cite{chen2003visual} cues. Traditional geometric methods estimated the distributions of certain characteristics (such as distances, angles, areas, or volumes) to measure the similarity between shapes \cite{akgul20093d,chaudhuri2010data,bronstein2011shape}. Gao~\etal~\cite{gao2011camera} used multiple camera projections, and Wu~\etal~\cite{wu20153d} used a voxel grid to extract model-based signatures. Su~\etal~\cite{mvcnn} introduced a deep learning pipeline for multi-view classification, with aggregated features achieving high retrieval performance. They used a low-rank Mahalanobis metric on top of extracted multi-view features to improve retrieval performance. This work on multi-view learning was extended for retrieval with volumetric-based descriptors~\cite{qi2016volumetric}, hierarchical view-group architectures~\cite{mvgvcnn}, and triplet-center loss~\cite{he2018triplet}. Jiang~\etal~\cite{mlvcnn} investigated better views for retrieval using many loops of circular cameras around the three principal axes. However, these approaches considered fixed camera view-points, as opposed to the learnable view-points of MVTN.

\mysection{Vision as Inverse Graphics (VIG)}
A key challenge in Vision as Inverse Graphics (VIG) is the non-differentiability of the classical graphics pipeline. Recent VIG approaches have focused on making graphics operations differentiable, allowing gradients to flow directly from the image to the rendering parameters \cite{vig-open-dr,vig-nmr,soft-rasterizer,vig-monte-carlo-raytrace,vig-bid-r,}. NMR \cite{vig-nmr} approximates non-differentiable rasterization by smoothing edge rendering, while SoftRas \cite{soft-rasterizer} assigns a probability for all mesh triangles to every pixel in the image. Synsin \cite{synsin} proposes an alpha-blending mechanism for differentiable point cloud rendering. Pytorch3D \cite{pytorch3d} improves the speed and modularity of SoftRas and Synsin, and allows for customized shaders and point cloud rendering. MVTN takes advantage of these advances in differentiable rendering to jointly train with the multi-view network in an end-to-end fashion. By using both mesh and point cloud differentiable rendering, MVTN can work with 3D CAD models and more readily available 3D point cloud data

\section{Methodology} \label{sec:methodology}
We illustrate our proposed multi-view pipeline using MVTN in \figLabel{\ref{fig:pipeline}}. MVTN is a generic module that learns camera view-point transformations for specific 3D multi-view tasks, \eg 3D shape classification. In this section, we review a generic framework for common multi-view pipelines, introduce MVTN details, and present an integration of MVTN for 3D shape classification and retrieval.

\subsection{Overview of Multi-View 3D Recognition}
3D multi-view recognition defines $M$ different images $\{\mathbf{x}_i\}_{i=1}^M$ rendered from multiple view-points of the same shape $\mathbf{S}$. The views are fed into the same backbone network $\mathbf{f}$ that extracts discriminative features per view. These features are then aggregated among views to describe the entire shape and used for downstream tasks such as classification or retrieval. 
Specifically, a multi-view network $\mathbf{C}$ with parameters $\boldsymbol{\theta}_{\mathbf{C}}$
operates on an input set of images $\mathbf{X} \in \mathbb{R}^{M  \times h\times w \times c }$ to obtain a softmax probability vector for the shape $\mathbf{S}$.

\mysection{Training Multi-View Networks}
The simplest deep multi-view classifier is MVCNN, where $\mathbf{C} = \text{MLP}\left( \max_{i} \mathbf{f}(\mathbf{x}_i)\right) $ with $\mathbf{f} : \mathbb{R}^{h \times w \times c} \rightarrow{\mathbb{R}^{d}}$ being a 2D CNN backbone (\eg ResNet \cite{resnet}) applied individually on each rendered image. A more recent method like ViewGCN would be described as $\mathbf{C} = \text{MLP}\left( \text{cat}_{\text{GCN}}\left( \mathbf{f}(\mathbf{x}_i)\right)\right) $, where $\text{cat}_{\text{GCN}}$ is an aggregation of views' features learned from a graph convolutional network. In general, learning a task-specific multi-view network on a labeled 3D dataset is formulated as:
\begin{equation}
\begin{aligned} 
 &\argmin_{\boldsymbol{\theta}_{\mathbf{C}}}~~ \sum_{n}^{N} L~\big( \mathbf{C} (\mathbf{X}_n)~,~y_n \big) \\ =~
 &\argmin_{\boldsymbol{\theta}_{\mathbf{C}}} \sum_{n}^{N} L~\Big( \mathbf{C} \big(\mathbf{R}(\mathbf{S}_n,\mathbf{u}_0)\big)~,~y_n \Big),
\label{eq:mv-objectiive}
\end{aligned} 
\end{equation}
\noindent where $L$ is a task-specific loss defined over $N$ 3D shapes in the dataset,
$y_n$ is the label for the $n^{\text{th}}$ 3D shape $\mathbf{S}_n$, and $\mathbf{u}_0 \in \mathbb{R}^{\tau}$ is a set of $\tau$ fixed scene parameters for the entire dataset. These parameters represent properties that affect the rendered image, including camera view-point, light, object color, and background. $\mathbf{R}$ is the renderer that takes as input a shape $\mathbf{S}_n$ and the parameters $\mathbf{u}_0$ to produce $M$ multi-view images $\mathbf{X}_n$ per shape. 
In our experiments, we choose the scene parameters $\mathbf{u}$ to be the azimuth and elevation angles of the camera view-points pointing towards the object center, thus setting $\tau = 2M$.

\mysection{Canonical Views}
Previous multi-view methods rely on scene parameters $\mathbf{u}_0$ that are pre-defined for the entire 3D dataset. In particular, the fixed camera view-points are usually selected based on the alignment of the 3D models in the dataset. The most common view configurations are \textit{circular} that aligns view-points on a circle around the object~\cite{mvcnn,mvnhbn} and \textit{spherical} that aligns equally spaced view-points on a sphere surrounding the object~\cite{mvviewgcn,mvrotationnet}. 
Fixing those canonical views for all 3D objects can be misleading for some classes. For example, looking at a bed from the bottom could confuse a 3D classifier.
In contrast, MVTN learns to regress per-shape view-points, as illustrated in \figLabel{\ref{fig:views-types}}.

\subsection{Multi-View Transformation Network (MVTN)}
Previous multi-view methods take the multi-view image $\mathbf{X}$ as the only representation for the 3D shape, where $\mathbf{X}$ is rendered using fixed scene parameters $\mathbf{u}_0$. In contrast, we consider a more general case, where $\mathbf{u}$ is \textit{variable} yet within bounds $\pm \mathbf{u}_{\text{bound}}$.
Here, $\mathbf{u}_{\text{bound}}$ is positive and it defines the permissible range for the scene parameters.
We set $\mathbf{u}_{\text{bound}} $ to  $180^\circ$ and $90^\circ$ for each azimuth and elevation angle.

\begin{figure} [t] 
\centering
\tabcolsep=0.03cm
          \resizebox{0.85\linewidth}{!}{%
\begin{tabular}{c|c|c}  
\textbf{Circular} & \textbf{Spherical}  &    \textbf{MVTN} \\  

 \includegraphics[trim= 4cm 2.7cm 4cm 2.2cm , clip, width = 0.333\linewidth]{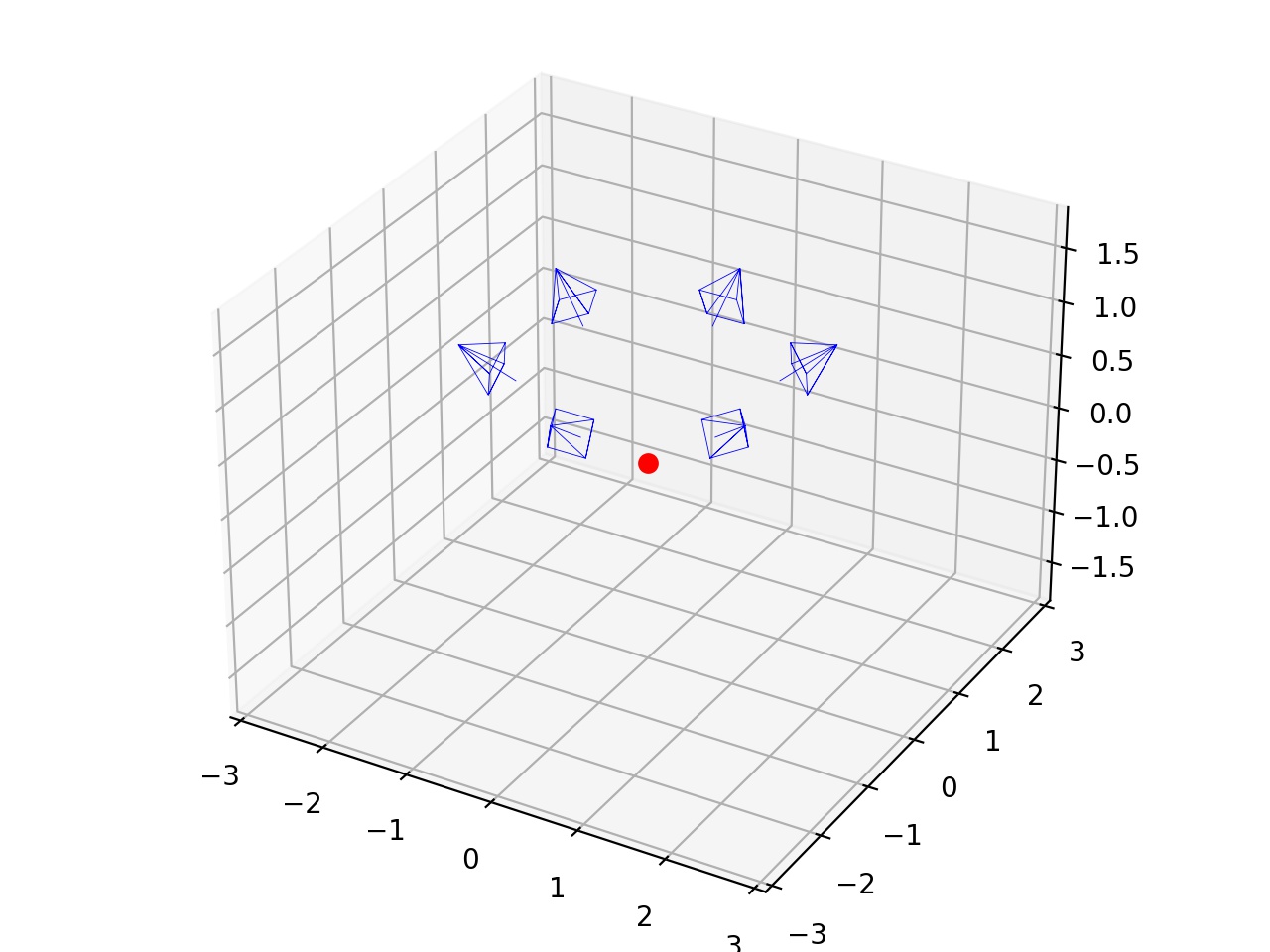} & \includegraphics[trim= 4cm 2.7cm 4cm 2.2cm , clip, width = 0.333\linewidth]{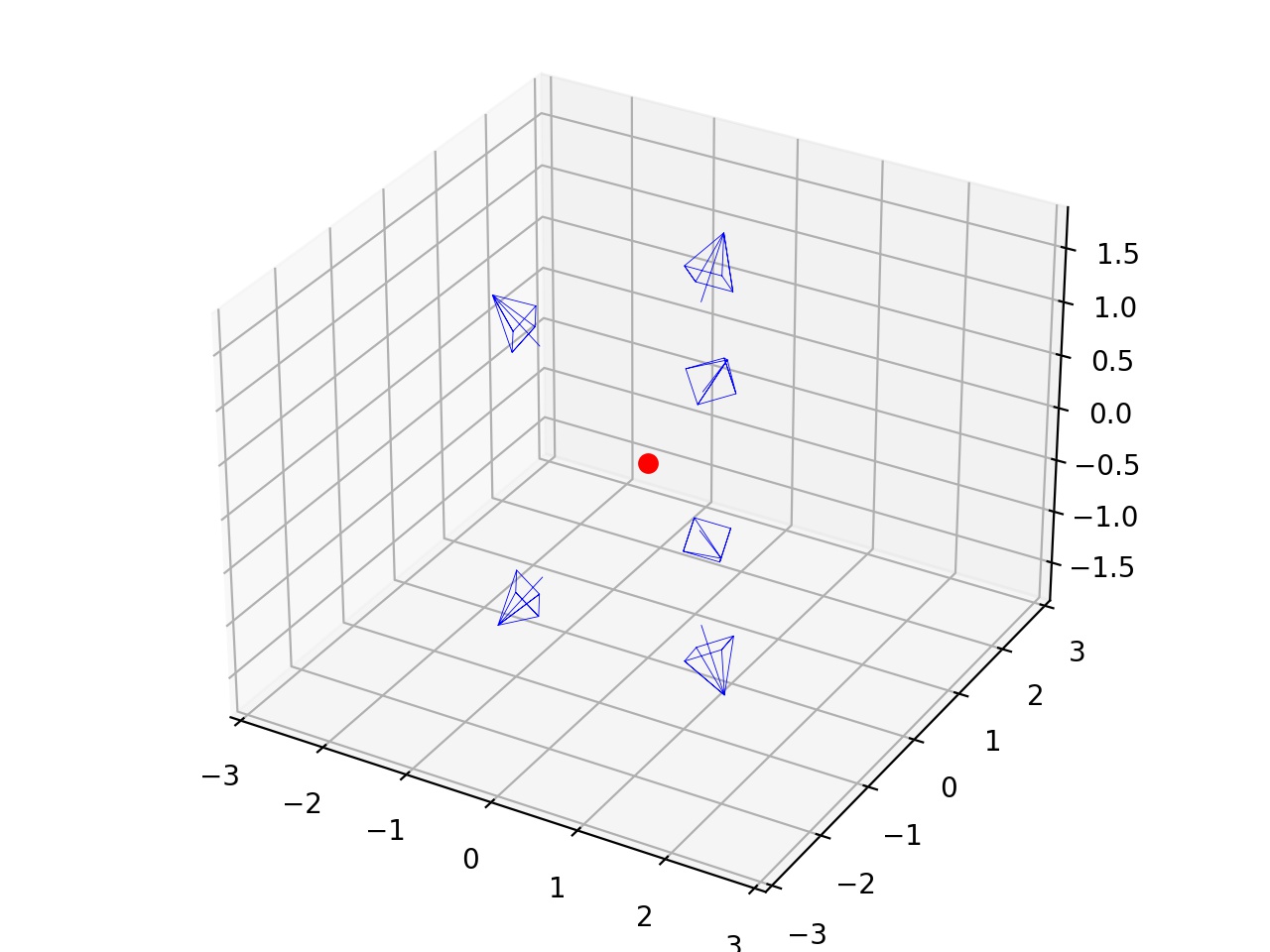} 
&\includegraphics[trim= 4cm 2.7cm 4cm 2.2cm , clip, width = 0.333\linewidth]{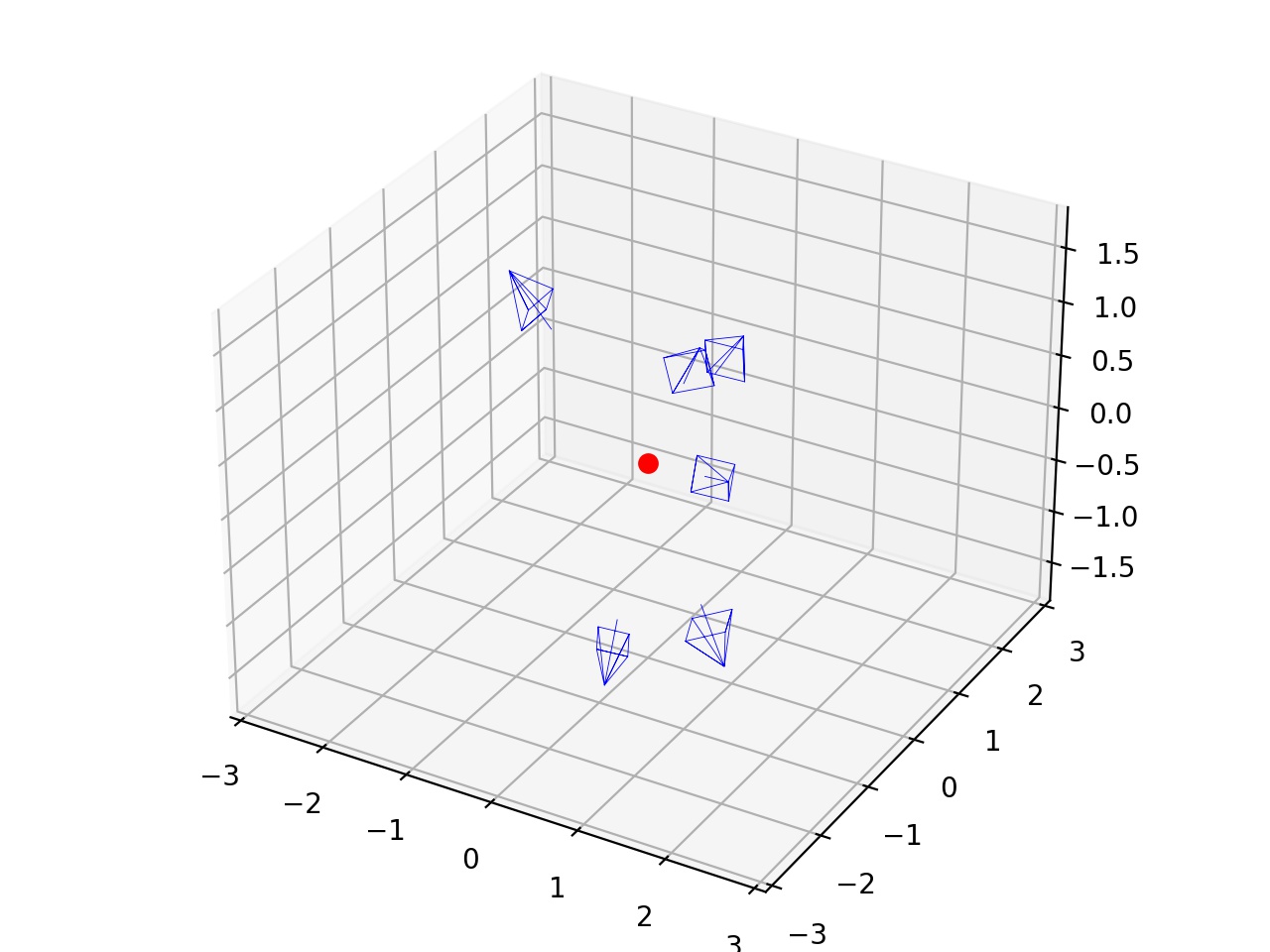} \\
  \includegraphics[trim= 0cm 0cm 24cm 0cm , clip, width = 0.333\linewidth]{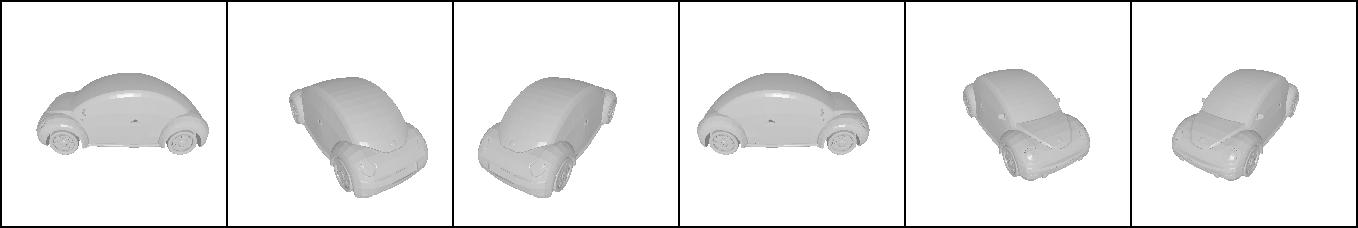} & \includegraphics[trim= 0cm 0cm 24cm 0cm , clip, width = 0.333\linewidth]{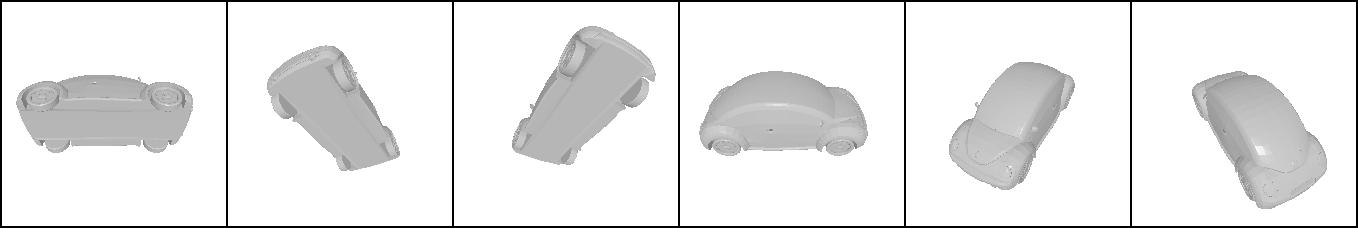} & 
\includegraphics[trim= 0cm 0cm 24cm 0cm , clip, width = 0.333\linewidth]{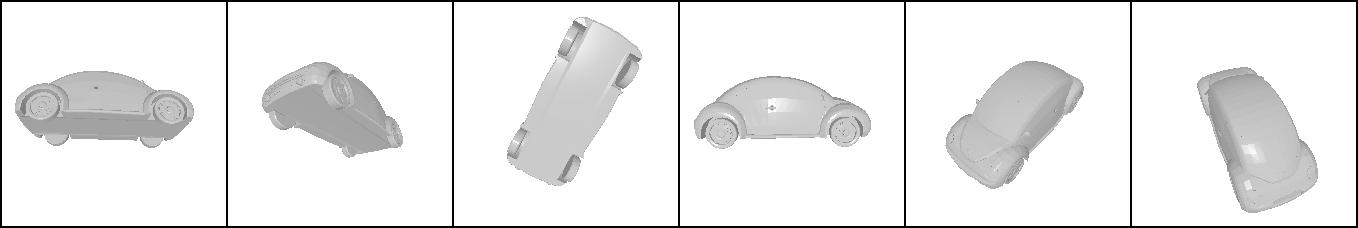}  \\
  \includegraphics[trim= 24cm 0cm 0cm 0cm , clip, width = 0.333\linewidth]{images_qualitative_views_circular_rend.jpg} & \includegraphics[trim= 24cm 0cm 0cm 0cm , clip, width = 0.333\linewidth]{images_qualitative_views_spherical_rend.jpg} & 
\includegraphics[trim= 24cm 0cm 0cm 0cm , clip, width = 0.333\linewidth]{images_qualitative_views_mvt_spherical_rend.jpg}  \\ 

\end{tabular}
}
\caption{  \textbf{Multi-View Camera Configurations}: The view setups commonly used in the multi-view literature are circular \cite{mvcnn} or spherical \cite{mvviewgcn,mvrotationnet}. Our MVTN learns to predict specific view-points for each object shape at inference time. The shape's center is shown as a red dot, and the view-points as blue cameras with their mesh renderings shown at the bottom.
}
    \label{fig:views-types}
\end{figure}

\begin{figure}[t]
    \centering
    \includegraphics[width=0.8\linewidth]{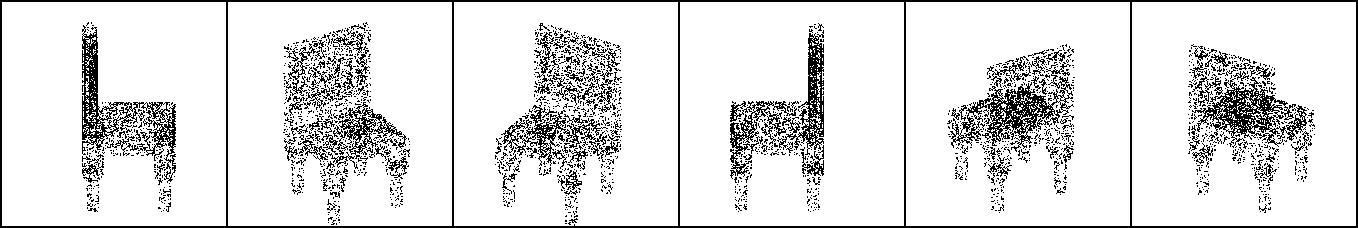}\\
    \includegraphics[width=0.8\linewidth]{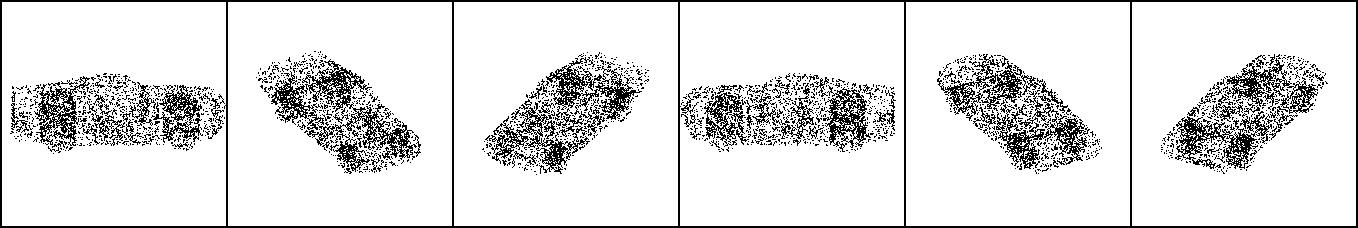} 
    \caption{\textbf{Multi-View Point Cloud Renderings.} We show some examples of point cloud renderings used in our pipeline. Note how point cloud renderings offer more information about content hidden from the camera view-point (\eg car wheels from the occluded side), which can be useful for recognition.}
    \label{fig:point-rendring}
\end{figure}

\vspace{1pt}\mysection{Differentiable Renderer}
A renderer $\mathbf{R}$ takes a 3D shape $\mathbf{S}$ (mesh or point cloud) and scene parameters $\mathbf{u}$ as inputs, and outputs the corresponding $M$ rendered images $\{\mathbf{x}_i\}_{i=1}^M$. 
Since $\mathbf{R}$ is differentiable, gradients $\pd{\mathbf{x}_i}{\mathbf{u}}{}$ can propagate backward from each rendered image to the scene parameters, thus establishing a framework that suits end-to-end deep learning pipelines.
When $\mathbf{S}$ is represented as a 3D mesh, $\mathbf{R}$ has two components: a \textit{rasterizer} and a \textit{shader}. First, the rasterizer transforms meshes from the world to view coordinates given the camera view-point and assigns faces to pixels. Using these face assignments, the shader creates multiple values for each pixel then blends them. On the other hand, if $\mathbf{S}$ is represented by a 3D point cloud, $\mathbf{R}$ would use an alpha-blending mechanism instead \cite{synsin}.
\figLabel{\ref{fig:views-types}} and \figLabel{\ref{fig:point-rendring}} illustrate examples of mesh and point cloud renderings used in MVTN. 

\vspace{1pt}\mysection{View-Points Conditioned on 3D Shape}
We design $\mathbf{u}$ to be a function of the 3D shape by learning a Multi-View Transformation Network (MVTN), denoted as $\mathbf{G} \in \mathbb{R}^{P\times 3} \rightarrow{\mathbb{R}^{\tau}} $ and parameterized by $\boldsymbol{\theta}_{\mathbf{G}}$, where $P$ is the number of points sampled from shape $\mathbf{S}$. 
Unlike \eqLabel{\ref{eq:mv-objectiive}} that relies on constant rendering parameters, MVTN predicts $\mathbf{u}$ adaptively for each object shape $\mathbf{S}$ and is optimized along with the classifier $ \mathbf{C}$. The pipeline is trained end-to-end to minimize the following loss on a dataset of N objects:
\begin{equation}
\begin{aligned} 
 \argmin_{\boldsymbol{\theta}_{\mathbf{C}}, \boldsymbol{\theta}_{\mathbf{G}}} &\sum_{n}^{N} L~\Big( \mathbf{C} \big(\mathbf{R}(\mathbf{S}_n,\mathbf{u}_n)\big)~,~y_n \Big), \\& \text{s. t.} \quad \mathbf{u}_n = ~ \mathbf{u}_{\text{bound}}.\text{tanh}\big( \mathbf{G}(\mathbf{S}_n)\big)
\label{eq:mvt-objective}
\end{aligned} 
\end{equation}
Here, $\mathbf{G}$ encodes a 3D shape to predict its optimal view-points for the task-specific multi-view network $\mathbf{C}$. Since the goal of $\mathbf{G}$ is only to predict view-points and not classify objects (as opposed to $\mathbf{C}$), its architecture is designed to be simple and light-weight. 
As such, we use a simple point encoder (\eg shared MLP as in PointNet \cite{pointnet}) that processes $P$ points from $\mathbf{S}$ and produces coarse shape features of dimension $b$. 
Then, a shallow MLP regresses the scene parameters $\mathbf{u}_n$ from the global shape features.
To force the predicted parameters $\mathbf{u}$ to be within a permissible range $\pm\mathbf{u}_{\text{bound}}$, we use a hyperbolic tangent function scaled by $\mathbf{u}_{\text{bound}}$.

\mysection{MVTN for 3D Shape Classification} \label{sec:mmtd-cls}
To train MVTN for 3D shape classification, 
we define a cross-entropy loss in \eqLabel{\ref{eq:mvt-objective}},
yet other losses and regularizers can be used here as well. 
The multi-view network ($\mathbf{C}$) and the MVTN ($\mathbf{G}$) are trained jointly on the same loss. 
One merit of our multi-view pipeline is its ability to seamlessly handle 3D point clouds, which is absent in previous multi-view methods. When $\mathbf{S}$ is a 3D point cloud, we simply define $\mathbf{R}$ as a differentiable point cloud renderer.

\vspace{1pt}\mysection{MVTN for 3D Shape Retrieval} \label{sec:mmtd-retr}
The shape retrieval task is defined as follows: given a query shape $\mathbf{S}_q$, find the most similar shapes in a broader set of size $N$.
For this task, we follow the retrieval setup of MVCNN~\cite{mvcnn}. In particular, we consider the deep feature representation of the last layer before the classifier in $\mathbf{C}$. We project those features into a more expressive space using LFDA reduction \cite{sugiyama2007dimensionality} and consider the reduced feature as the signature to describe a shape. At test time, shape signatures are used to retrieve (in order) the most similar shapes in the training set.

\section{Experiments} \label{sec:experiments}
We evaluate MVTN for the tasks of 3D shape classification and retrieval on  ModelNet40 \cite{modelnet}, ShapeNet Core55 \cite{shapenet}, and the more realistic ScanObjectNN \cite{scanobjectnn}.
\subsection{Datasets}
\mysection{ModelNet40}
The ModelNet40 dataset \cite{modelnet} consists of 12,311 3D objects with 40 object classes, with a split of 9,843 objects in the training set and 2,468 in the testing set. Due to hardware limitations, the meshes in this dataset have been simplified to 20,000 vertices using the official Blender API \cite{blender,mesh-simplify}.

\mysection{ShapeNet Core55}
The ShapeNet Core55 dataset \cite{shapenet} is a subset of ShapeNet comprising 51,162 3D mesh objects with 55 object classes. It was created for the shape retrieval challenge SHREK \cite{shrek17} and includes 35764 objects in the training set, 5133 in the validation set, and 10265 in the test set.

\mysection{ScanObjectNN}
The ScanObjectNN dataset \cite{scanobjectnn} is a more realistic and challenging point cloud dataset for 3D classification, including background and occlusions. It consists of 2,902 point clouds divided into 15 object categories and has three main variants: object only, object with background, and the hardest perturbed variant (PB\_T50\_RS variant). These variants are used in the 3D Scene Understanding Benchmark associated with the ScanObjectNN dataset and offer a more challenging evaluation of the generalization capabilities of 3D deep learning models in realistic scenarios compared to ModelNet40.

\subsection{Metrics}
\mysection{Classification Accuracy} 
The standard evaluation metric in 3D classification is accuracy. We report overall accuracy (percentage of correctly classified test samples) and average per-class accuracy (mean of all true class accuracies).

\mysection{Retrieval mAP} 
Shape retrieval is evaluated by mean Average Precision (mAP) over test queries. For every query shape $\mathbf{S}_q$ from the test set, AP is defined as $AP= \frac{1}{\text{GTP}} \sum_{n}^{N}\frac{\mathbbm{1}(\mathbf{S}_n)}{n} $, where $GTP$ is the number of ground truth positives, $N$ is the size of the ordered training set, and $\mathbbm{1}(\mathbf{S}_n) = 1$ if the shape $\mathbf{S}_n$ is from the same class label of query $\mathbf{S}_q$. We average the retrieval AP over the test set to measure retrieval mAP.

\subsection{Baselines}
\mysection{Voxel Networks}
We choose VoxNet \cite{voxnet}, DLAN \cite{dlanretr}, and 3DShapeNets \cite{modelnet} as baselines that use voxels.

\mysection{Point Cloud Networks}
We select PointNet \cite{pointnet}, PointNet++ \cite{pointnet++}, DGCNN \cite{dgcn}, PVNet \cite{pvnet}, and KPConv \cite{kpconv} as baselines that use point clouds.
These methods leverage different convolution operators on point clouds by aggregating local and global point information.

\mysection{Multi-view Networks}
We compare against MVCNN \cite{mvcnn}, RotationNet \cite{mvrotationnet}, GVCNN \cite{mvgvcnn} and ViewGCN \cite{mvviewgcn} as representative multi-view methods.
These methods are limited to meshes, pre-rendered from canonical view-points.
 \subsection{MVTN Details}
\mysection{Rendering}
In our pipeline, we utilize the differentiable mesh and point cloud renderers from Pytorch3D \cite{pytorch3d} for their compatibility with Pytorch libraries \cite{paszke2017pytorch} and fast processing speed. Examples of rendered images for meshes and point clouds can be seen in Figures \ref{fig:views-types} and \ref{fig:point-rendring}, respectively. Each rendered image has a size of 224$\times$224. 
For ModelNet40, we utilize the differentiable mesh renderer and apply augmentation during training by randomly directing the light and assigning a random color to the object. In testing, we fix the light direction towards the center of the object and color the object white for stable performance. For ShapeNet Core55 and ScanObjectNN, we use the differentiable point cloud renderer with 2048 and 5000 points, respectively. The use of a point cloud renderer offers a lighter alternative to mesh rendering when the mesh contains a large number of faces, which can hinder the training of the MVTN pipeline.

\begin{table}[t]
\tabcolsep=0.07cm
    \centering
\resizebox{0.6\linewidth}{!}{\begin{tabular}{rccc}
\toprule
 &   & \multicolumn{2}{c}{Classification Accuracy} \\
\multicolumn{1}{c}{Method}       & Data Type & (\textbf{Per-Class})   & (\textbf{Overall}) \\ \midrule
VoxNet \cite{voxnet}     & Voxels                 & 83.0 & 85.9      \\
PointNet \cite{pointnet} &  Points                  &       86.2 & 89.2      \\
PointNet++ \cite{pointnet++}   & Points                & - & 91.9      \\
PointCNN \cite{pc_li2018pointcnn}  & Points            &   88.1   & 91.8      \\
DGCNN \cite{dgcn}             & Points                &  90.2     & 92.2      \\
KPConv\cite{kpconv}  &  Points & -  & 92.9 \\ 
MVCNN  \cite{mvcnn}         & 12 Views                   & 90.1  & 90.1 \\
GVCNN \cite{mvgvcnn}         & 12 Views                   & 90.7 & 93.1 \\
ViewGCN \cite{mvviewgcn}  & 20 Views   & \textbf{96.5} & \textbf{97.6} \\ 

\midrule
ViewGCN \cite{mvviewgcn}$^*$& 12 views &    90.7   &93.0 \\
ViewGCN \cite{mvviewgcn}$^*$& 20 views &    91.3   &93.3 \\
MVTN (ours)$^*$  & 12 Views       & 92.0 & \textbf{93.8} \\
MVTN (ours)$^*$  & 20 Views       & \textbf{92.2} & 93.5 \\
\bottomrule
\end{tabular}
}
\vspace{2pt}
    \caption{\textbf{3D Shape Classification on ModelNet40}. We compare MVTN against other methods in 3D classification on ModelNet40 \cite{modelnet}. $^*$ indicates results from our rendering setup (differentiable pipeline), while other multi-view results are reported from pre-rendered views. \textbf{Bold} denotes the best result in its setup.}
    \label{tab:ModelNet40-cls}
\end{table}
\begin{table}[t]
\tabcolsep=0.08cm
    \centering
\resizebox{0.65\linewidth}{!}{\begin{tabular}{rccc}
\toprule
 &  \multicolumn{3}{c}{Classification Overall Accuracy } \\
\multicolumn{1}{c}{Method}& \textbf{OBJ\_BG}  & \textbf{OBJ\_ONLY} & \textbf{Hardest}  \\ \midrule
3DMFV \cite{3Dmfv} &  68.2                  &  73.8  &  63.0  \\
PointNet \cite{pointnet}   & 73.3                &   79.2  & 68.0 \\
SpiderCNN \cite{pc_xu2018spidercnn}&    77.1                &    79.5   &  73.7    \\
PointNet ++ \cite{pointnet++}            & 82.3 & 84.3  &   77.9   \\
PointCNN \cite{pc_li2018pointcnn}  & 86.1 & 85.5  & 78.5 \\
DGCNN \cite{dgcn}  & 82.8 & 86.2  & 78.1 \\ 
SimpleView \cite{simpleview}& - & - & 79.5 \\
BGA-DGCNN \cite{scanobjectnn}   & - & - & 79.7 \\
BGA-PN++ \cite{scanobjectnn}   & - & - & 80.2 \\
\midrule
MVTN (ours)  & \textbf{92.6}    & \textbf{92.3} & \textbf{82.8} \\
\bottomrule
\end{tabular}
}
\vspace{2pt}
    \caption{\textbf{3D Point Cloud Classification on ScanObjectNN}. We compare the performance of MVTN in 3D point cloud classification on three different variants of ScanObjectNN \cite{scanobjectnn}. The variants include object with background, object only, and the hardest variant.}
    \label{tab:Scanobjectnn}
\end{table}

\mysection{View-Point Prediction}
As shown in \eqLabel{\ref{eq:mvt-objective}}, the MVTN $\mathbf{G}$ network learns to predict the view-points directly (\textit{MVTN-direct}). Alternatively, MVTN can learn relative offsets \wrt initial parameters $\mathbf{u}_0$. In this case, we concatenate the point features extracted in $\mathbf{G}$ with $\mathbf{u}_0$ to predict the offsets to apply on $\mathbf{u}_0$.
The learned view-points $\mathbf{u}_n$ in \eqLabel{\ref{eq:mvt-objective}} are defined as: $~ \mathbf{u}_n = \mathbf{u}_0 + \mathbf{u}_{\text{bound}}.\text{tanh}\big( \mathbf{G}(\mathbf{u}_0~,~\mathbf{S}_n)\big)$. We take $\mathbf{u}_0$ to be the circular or spherical configurations commonly used in multi-view classification pipelines~\cite{mvcnn,mvrotationnet,mvviewgcn}.
We refer to these learnable variants as \textit{MVTN-circular} and \textit{MVTN-spherical}, accordingly. For MVTN-circular, the initial elevations for the views are  30$^\circ$, and the azimuth angles are equally distributed over 360$^\circ$ following \cite{mvcnn}. For MVTN-spherical, we follow the method from \cite{spherical-config} that places equally-spaced view-points on a sphere for an arbitrary number of views, which is similar to the ``dodecahedral'' configuration in ViewGCN. 

\begin{table}[t]
\tabcolsep=0.07cm
    \centering
\resizebox{0.72\linewidth}{!}{\begin{tabular}{rccc}
\toprule
 &   & \multicolumn{2}{c}{ Shape Retrieval (mAP)} \\
\multicolumn{1}{c}{Method} & Data Type  & \textbf{ModelNet40}  & \textbf{ShapeNet Core} \\ \midrule
DLAN \cite{dlanretr}   & Voxels                &   - & 66.3      \\
LFD     \cite{lfd}                &  Voxels                 & 40.9 & - \\ 
3D ShapeNets \cite{modelnet} &  Voxels                 & 49.2 & - \\
PVNet\cite{pvnet}  &  Points & 89.5 & - \\ 
MVCNN  \cite{mvcnn}         & 12 Views                   & 80.2 & 73.5 \\
GIFT \cite{giftretr}&    20 Views                &     - & 64.0      \\
MVFusionNet \cite{mvfusionnet}            & 12 Views                &      - & 62.2      \\
ReVGG \cite{shrek17}  & 20 Views            &     - & 74.9      \\
RotNet \cite{mvrotationnet}  & 20 Views   & - & 77.2 \\ 
ViewGCN \cite{mvviewgcn}  & 20 Views   & - &  78.4 \\ 
MLVCNN \cite{mlvcnn} &   24 Views               & 92.2      & -\\
\midrule
MVTN (ours)  & 12 Views         & \textbf{92.9} &  \textbf{82.9} \\
\bottomrule
\end{tabular}
}
\vspace{2pt}
    \caption{\textbf{3D Shape Retrieval}. We benchmark the shape retrieval mAP of  MVTN on ModelNet40 \cite{modelnet} and ShapeNet Core55 \cite{shapenet,shrek17}. MVTN achieves the best retrieval performance among recent state-of-the-art methods on both datasets with only 12 views.}
     \label{tab:retrieval}
\end{table}

\mysection{Architecture}
In our MVTN pipeline, we select MVCNN \cite{mvcnn} and ViewGCN \cite{mvviewgcn} as our multi-view networks of choice. In our experiments, we use PointNet \cite{pointnet} as the 3D point encoder network $\mathbf{G}$ and experiment with DGCNN in Section \ref{sec:ablation}.
We sample 2048 points from each mesh as input to the point encoder and use a 5-layer MLP for the regression network, which takes as input the point features extracted by the point encoder of size $b=40$.
All MVTN variants and the baseline multi-view networks utilize a ResNet-18 \cite{resnet} network, pretrained on ImageNet \cite{IMAGENET}, as the multi-view backbone in $\mathbf{C}$, with output features of size $d=1024$. The main classification and retrieval results are based on MVTN-spherical with ViewGCN \cite{mvviewgcn} as the multi-view network $\mathbf{C}$, unless otherwise specified in \secLabel{\ref{sec:exp-robust}} and \ref{sec:ablation}.

\mysection{Training Setup}
To avoid gradient instability introduced by the renderer, we use gradient clipping in the MVTN network $\mathbf{G}$. We clip the gradient updates such that the $\ell_2$ norm of the gradients does not exceed 30. We use a learning rate of $0.001$ but refrain from fine-tuning the hyper-parameters introduced in MVCNN~ \cite{mvcnn} and View-GCN~\cite{mvviewgcn}. More details about the training procedure are in the \supp\hspace{-2pt}. 

\begin{figure} [t] 
  \centering
  \tabcolsep=0.03cm
  \resizebox{0.7\linewidth}{!}{
  \begin{tabular}{c|ccccc}

  \includegraphics[width = 0.16666666666666666\linewidth]{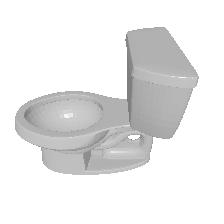} &
  \includegraphics[width = 0.16666666666666666\linewidth]{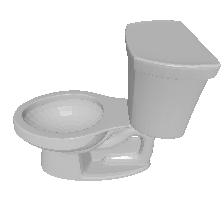} &
  \includegraphics[width = 0.16666666666666666\linewidth]{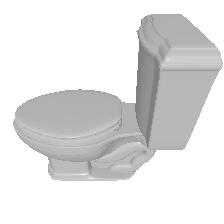} &
  \includegraphics[width = 0.16666666666666666\linewidth]{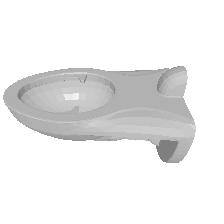} &
  \includegraphics[width = 0.16666666666666666\linewidth]{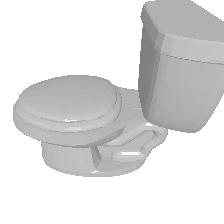} &
  \includegraphics[width = 0.16666666666666666\linewidth]{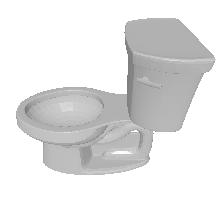} \\ \midrule

  \includegraphics[width = 0.16666666666666666\linewidth]{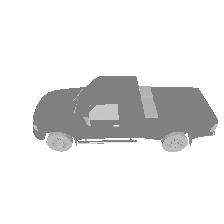} &
  \includegraphics[width = 0.16666666666666666\linewidth]{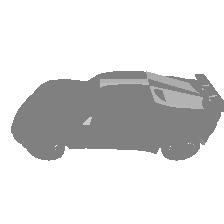} &
  \includegraphics[width = 0.16666666666666666\linewidth]{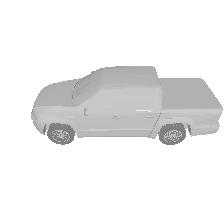} &
  \includegraphics[width = 0.16666666666666666\linewidth]{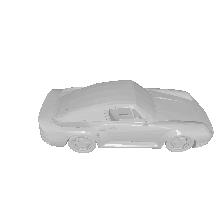} &
  \includegraphics[width = 0.16666666666666666\linewidth]{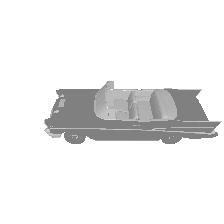} &
  \includegraphics[width = 0.16666666666666666\linewidth]{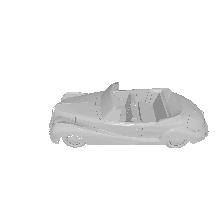} \\ \midrule

  \includegraphics[width = 0.16666666666666666\linewidth]{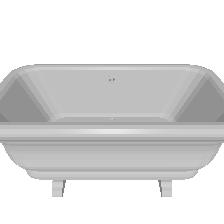} &
  \includegraphics[width = 0.16666666666666666\linewidth]{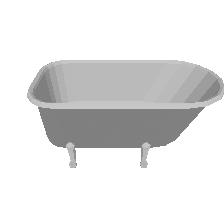} &
  \includegraphics[width = 0.16666666666666666\linewidth]{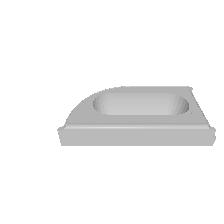} &
  \includegraphics[width = 0.16666666666666666\linewidth]{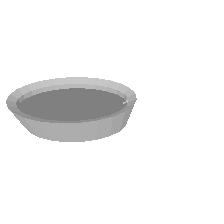} &
  \includegraphics[width = 0.16666666666666666\linewidth]{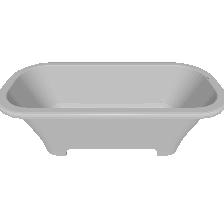} &
  \includegraphics[width = 0.16666666666666666\linewidth]{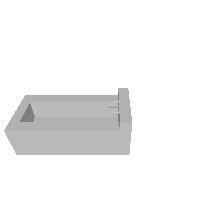} \\ \midrule 
  
  \includegraphics[width = 0.16666666666666666\linewidth]{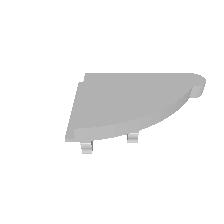} &
  \includegraphics[width = 0.16666666666666666\linewidth]{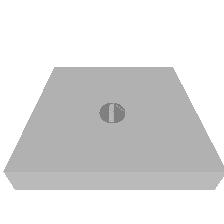} &
  \includegraphics[width = 0.16666666666666666\linewidth]{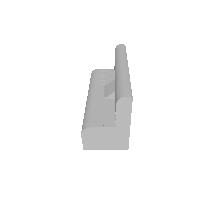} &
  \fbox{ \includegraphics[width = 0.16666666666666666\linewidth]{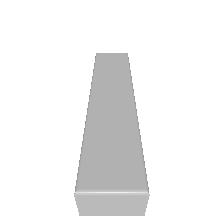}} &
  \includegraphics[width = 0.16666666666666666\linewidth]{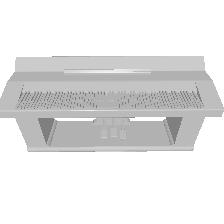} &
  \includegraphics[width = 0.16666666666666666\linewidth]{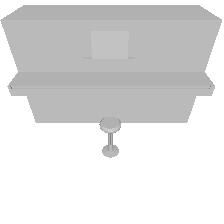} \\ \bottomrule
  \end{tabular}
  }
  \vspace{2pt}
  \caption{  \textbf{Qualitative Examples for Object Retrieval}: \textit{(left):} we show some query objects from the test set. \textit{(right)}: we show top five retrieved objects by our MVTN from the training set. Images of negative retrieved objects are framed.}
      \label{fig:imgs-retr}
  \end{figure}
  \section{Results} \label{sec:results}
  The main results of MVTN are summarized in Tables \ref{tab:ModelNet40-cls}, \ref{tab:Scanobjectnn}, \ref{tab:retrieval} and \ref{tbl:y-robustness}. We achieve state-of-the-art performance in 3D classification on ScanObjectNN by a large margin (up to 6\%) and achieve a competitive test accuracy of \textbf{93.8}\% on ModelNet40. On shape retrieval, we achieve state-of-the-art performance on both ShapeNet Core55 (\textbf{82.9} mAP) and ModelNet40 (\textbf{92.9} mAP). 
  Following the common practice, we report the best results out of four runs in benchmark tables, but detailed results are in \supp\hspace{-2pt}.
  
  \subsection{3D Shape Classification} \label{sec:exp-classification}
  Table \ref{tab:ModelNet40-cls} compares the performance of MVTN against other methods on ModelNet40 \cite{modelnet}. Our MVTN achieves a competitive test accuracy of 93.8\% compared to all previous methods. ViewGCN~\cite{mvviewgcn} achieves higher classification performance by relying on higher quality images from a more advanced yet non-differentiable OpenGL \cite{opengl} renderer. For a fair comparison, we report with an $^*$ the performance of ViewGCN using images generated by the renderer used in MVTN. Using the same rendering process, regressing views with MVTN improves the classification performance of the baseline ViewGCN at 12 and 20 views. We believe future advances in differentiable rendering would bridge the gap between our rendered images and the original high-quality pre-rendered ones.

  
  Table \ref{tab:Scanobjectnn} reports the classification accuracy of a 12 view MVTN on the realistic ScanObjectNN benchmark \cite{scanobjectnn}. MVTN improves performance on different variants of the dataset. The most difficult variant of ScanObjectNN (PB\_T50\_RS) includes challenging scenarios of objects undergoing translation and rotation. Our MVTN achieves state-of-the-art results (+2.6\%) on this variant, highlighting the merits of MVTN for realistic 3D point cloud scans. Also, note how adding background points (in OBJ\_BG) does not hurt MVTN, contrary to most other classifiers. . 
  
  
  \subsection{3D Shape Retrieval} 
  \label{sec:exp-retr}
  Table \ref{tab:retrieval} presents the retrieval mean average precision (mAP) of MVTN compared to recent methods on ModelNet40 \cite{modelnet} and ShapeNet Core55 \cite{shapenet}. The results for the latter methods are taken from \cite{mlvcnn,mvviewgcn,pvnet}.
  MVTN demonstrates state-of-the-art retrieval performance, achieving a mAP of 92.9\% on ModelNet40. It also significantly improves upon the state-of-the-art on ShapeNet, using only 12 views. It is worth noting that the baselines in Table \ref{tab:retrieval} include strong, recently developed methods specifically trained for retrieval, such as MLVCNN \cite{mlvcnn}.
   \figLabel{\ref{fig:imgs-retr}} shows qualitative examples of objects retrieved using MVTN.

  \begin{table}[t]
  
  \tabcolsep=0.3cm
  \centering
  \resizebox{0.62\linewidth}{!}{
  \begin{tabular}{rcccc} 
  \toprule
  & \multicolumn{3}{c}{Rotation Perturbations Range} \\
  \multicolumn{1}{c}{Method}  &$0^\circ$ & $\pm90^\circ$ & $\pm180^\circ$ \\ 
  \midrule
  PointNet \cite{pointnet}&  88.7  &  42.5   &  38.6 \\
  PointNet ++ \cite{pointnet++} &  88.2   & 47.9   & 39.7   \\
  RSCNN \cite{rspointcloud}  &  90.3 & 90.3    &  90.3  \\ \midrule
  MVTN (ours) &  \textbf{91.7} & \textbf{90.8}  & \textbf{91.2}  \\
  
  \bottomrule
  \end{tabular}
  }
  \vspace{2pt}
  \caption{  \textbf{Rotation Robustness on ModelNet40.} 
  At test time, we randomly rotate objects in ModelNet40 around the Y-axis (gravity) with different ranges and report the overall accuracy. MVTN displays strong robustness to such Y-rotations.}
  %
  \label{tbl:y-robustness}
  \end{table}
  \subsection{Rotation Robustness} \label{sec:exp-robust}
  Evaluating the robustness of trained models to perturbations at test time is a common practice in the 3D shape classification literature. Following the same procedure as \cite{rspointcloud,sada}, we perturb the shapes with random rotations around the Y-axis (gravity-axis) within the range of $\pm90^\circ$ and $\pm180^\circ$.
  We repeat the inference process ten times for each setup and report the average performance in Table \ref{tbl:y-robustness}. The MVTN-circular variant (using MVCNN) achieves state-of-the-art performance in rotation robustness (91.2\% test accuracy) compared to more advanced methods trained in the same setup. The baseline method, RSCNN \cite{rspointcloud}, is a strong model designed to be invariant to translation and rotation, while MVTN is learned in a simpler setup using MVCNN without targeting rotation invariance.

  \subsection{Occlusion Robustness} \label{sec:occlusion}
\begin{figure} [] 
  \centering
  \tabcolsep=0.03cm
  \resizebox{0.85\linewidth}{!}{
  \begin{tabular}{c|ccccc}
  
   & \multicolumn{5}{c}{Occlusion Ratio} \\ 
   dir.& 0.1 & 0.2 & 0.3 & 0.5 & 0.75 \\ \midrule
  
  +X &
  \includegraphics[trim= 0cm 2cm 0cm 2cm , clip, width = 0.19\linewidth]{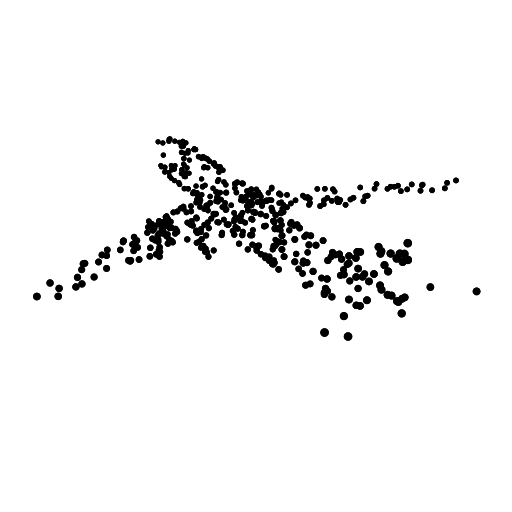} &
  \includegraphics[trim= 0cm 2cm 0cm 2cm , clip, width = 0.19\linewidth]{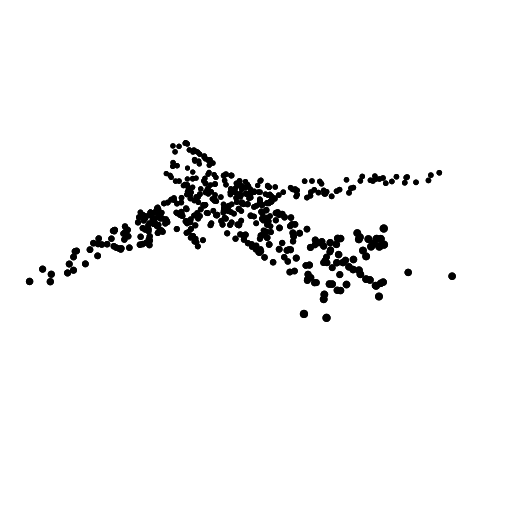} &
  \includegraphics[trim= 0cm 2cm 0cm 2cm , clip, width = 0.19\linewidth]{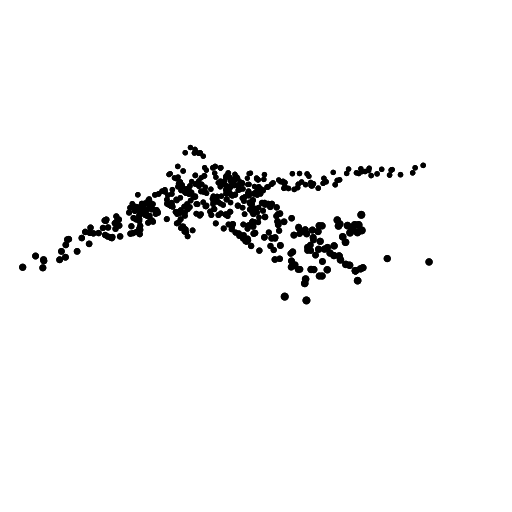} &
  \includegraphics[trim= 0cm 2cm 0cm 2cm , clip, width = 0.19\linewidth]{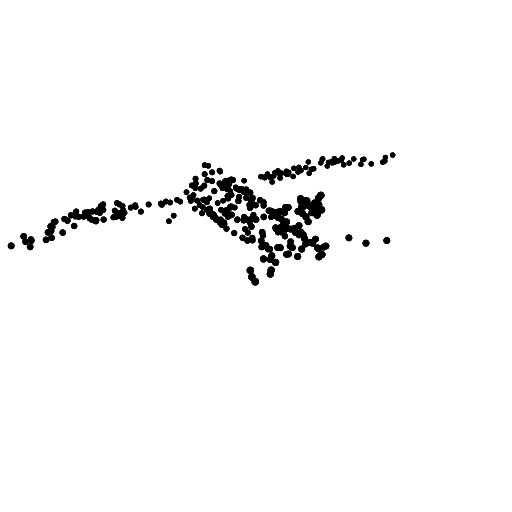} &
  \includegraphics[trim= 0cm 2cm 0cm 2cm , clip, width = 0.19\linewidth]{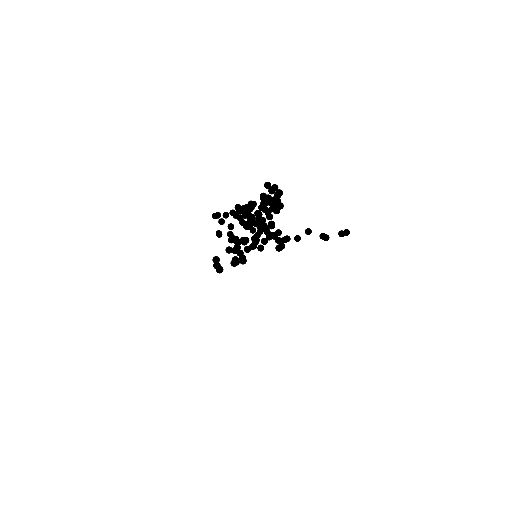} \\ \hline
  
  -X &
  \includegraphics[trim= 0cm 2cm 0cm 2cm , clip, width = 0.19\linewidth]{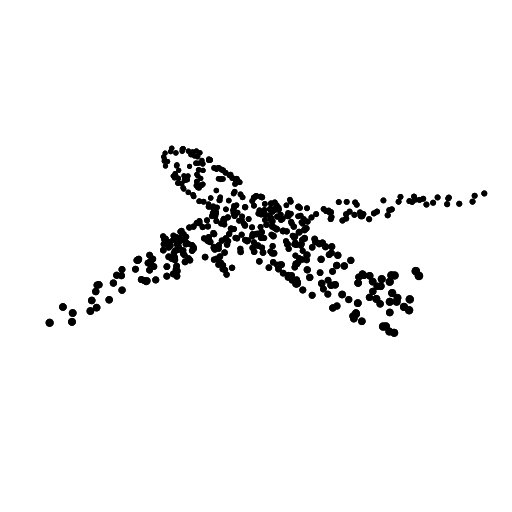} &
  \includegraphics[trim= 0cm 2cm 0cm 2cm , clip, width = 0.19\linewidth]{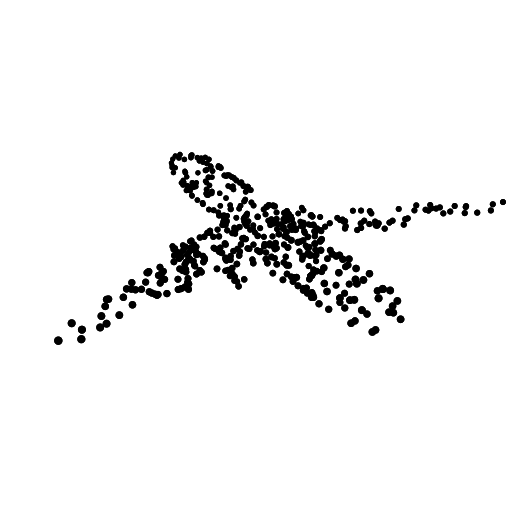} &
  \includegraphics[trim= 0cm 2cm 0cm 2cm , clip, width = 0.19\linewidth]{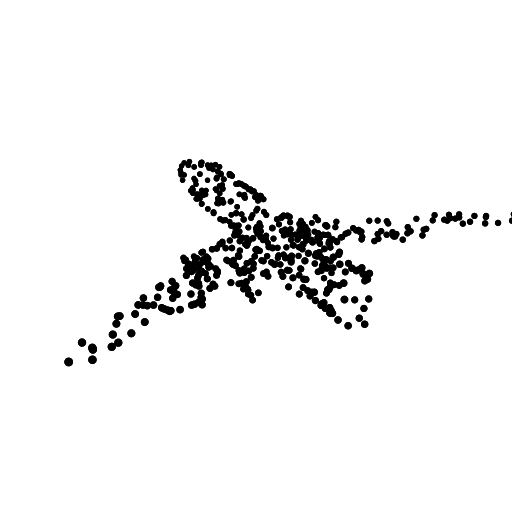} &
  \includegraphics[trim= 0cm 2cm 0cm 2cm , clip, width = 0.19\linewidth]{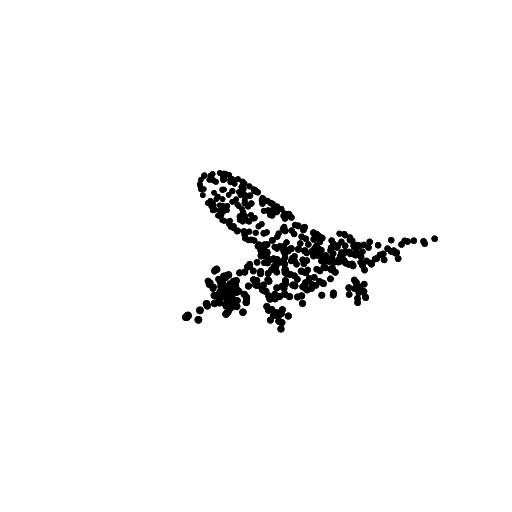} &
  \includegraphics[trim= 0cm 2cm 0cm 2cm , clip, width = 0.19\linewidth]{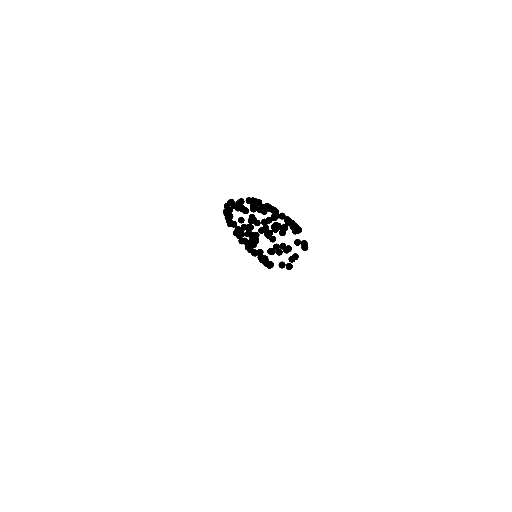} \\ \hline
  
  +Y &
  \includegraphics[trim= 0cm 2cm 0cm 2cm , clip, width = 0.19\linewidth]{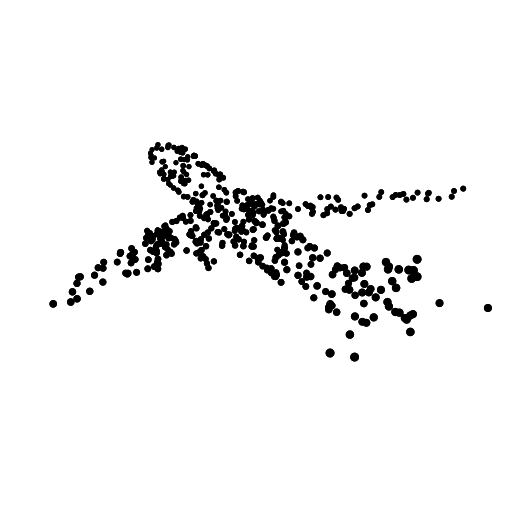} &
  \includegraphics[trim= 0cm 2cm 0cm 2cm , clip, width = 0.19\linewidth]{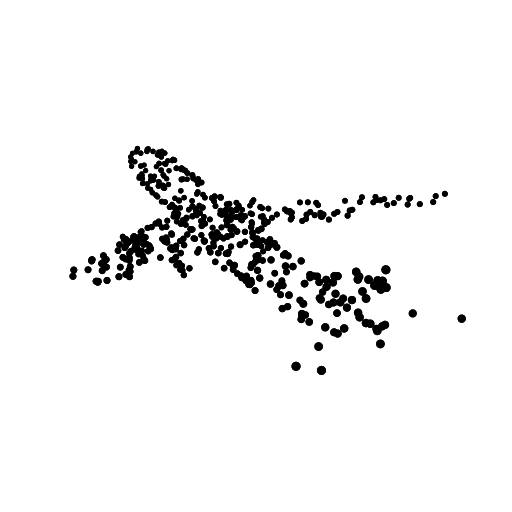} &
  \includegraphics[trim= 0cm 2cm 0cm 2cm , clip, width = 0.19\linewidth]{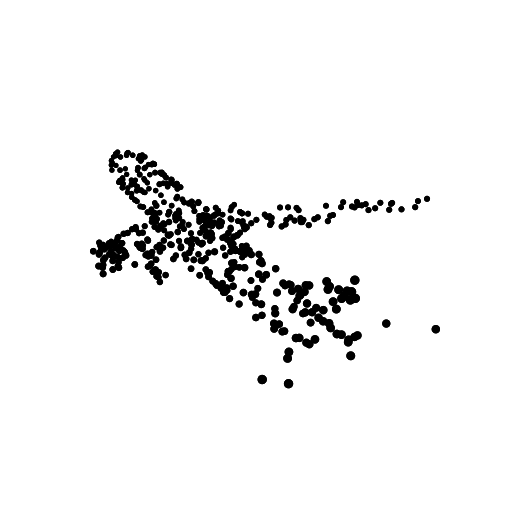} &
  \includegraphics[trim= 0cm 2cm 0cm 2cm , clip, width = 0.19\linewidth]{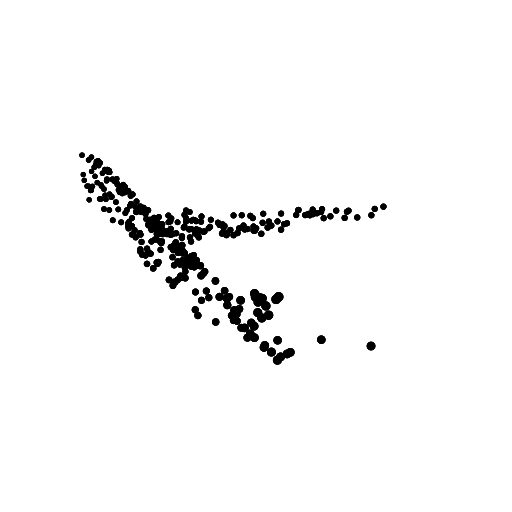} &
  \includegraphics[trim= 0cm 2cm 0cm 2cm , clip, width = 0.19\linewidth]{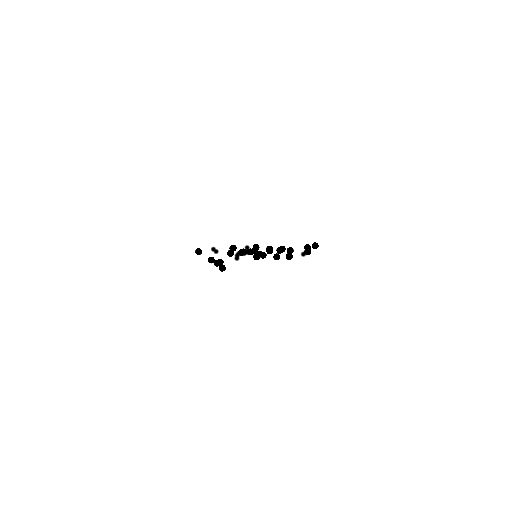} \\ \hline
  
  -Y &
  \includegraphics[trim= 0cm 2cm 0cm 2cm , clip, width = 0.19\linewidth]{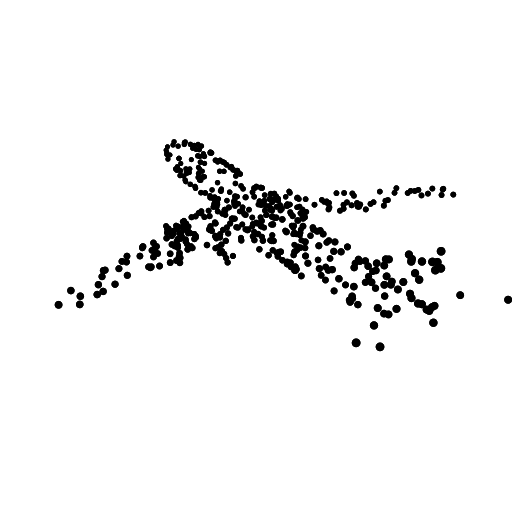} &
  \includegraphics[trim= 0cm 2cm 0cm 2cm , clip, width = 0.19\linewidth]{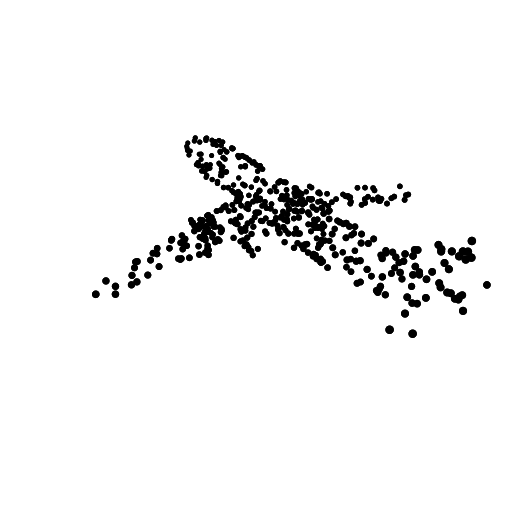} &
  \includegraphics[trim= 0cm 2cm 0cm 2cm , clip, width = 0.19\linewidth]{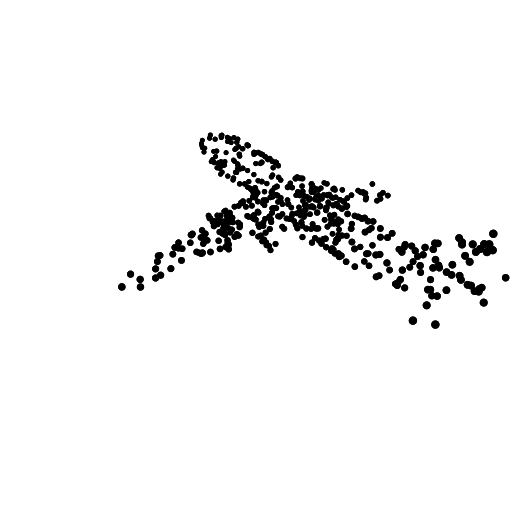} &
  \includegraphics[trim= 0cm 2cm 0cm 2cm , clip, width = 0.19\linewidth]{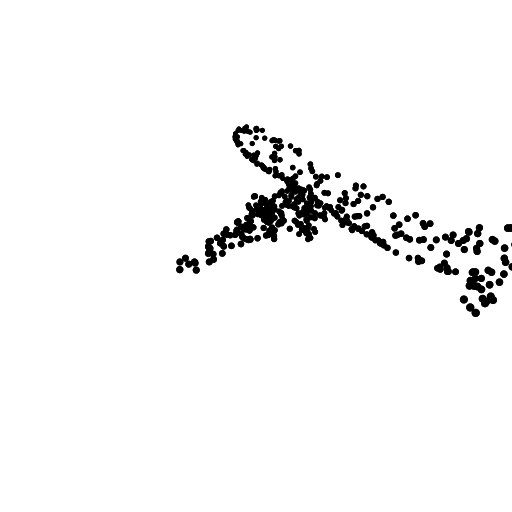} &
  \includegraphics[trim= 0cm 2cm 0cm 2cm , clip, width = 0.19\linewidth]{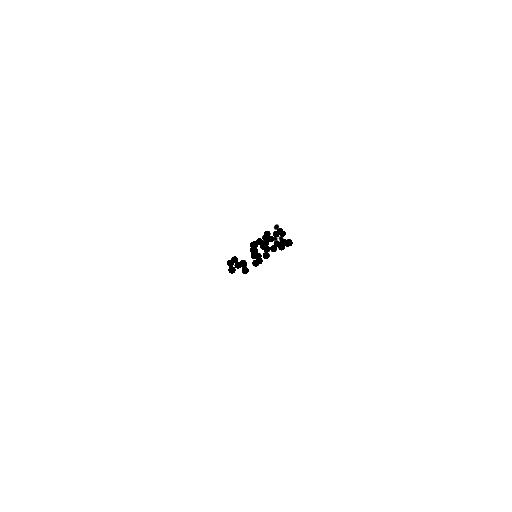} \\ \bottomrule
  \end{tabular}
  }
  \vspace{2pt}
  \caption{  \textbf{Occlusion of 3D Objects}: We simulate realistic occlusion scenarios in 3D point clouds by cropping a percentage of the object along canonical directions. Here, we show an object occluded with different ratios and from different directions. 
  }
      \label{fig:occlusion-qual}
  \end{figure}

  To assess the practical usefulness of MVTN in realistic scenarios, we investigate the problem of occlusion in 3D computer vision, particularly in 3D point cloud scans. Occlusion can be caused by various factors, such as the view angle to the object, the sampling density of the sensor (e.g., LiDAR), or the presence of noise in the sensor. In these realistic scenarios, deep learning models often struggle. To evaluate the effect of occlusion due to the viewing angle of the 3D sensor in our 3D classification setup, we simulate realistic occlusion by cropping the object from canonical directions. We train PointNet \cite{pointnet}, DGCNN \cite{dgcn}, and MVTN on the ModelNet40 point cloud dataset, and then crop a portion of the object (from 0\% occlusion to 100\%) along the $\pm$X, $\pm$Y, and $\pm$Z directions at test time. \figLabel{\ref{fig:occlusion-qual}} illustrates examples of this occlusion effect at different occlusion ratios. Table \ref{tbl:occlusion} reports the average test accuracy of the six cropping directions for the baselines and MVTN. MVTN exhibits high test accuracy even when large portions of the object are cropped. In fact, MVTN outperforms PointNet \cite{pointnet} by 13\% in test accuracy when half of the object is occluded, despite PointNet's reputation for robustness \cite{pointnet,advpc}. This result highlights the effectiveness of MVTN in handling occlusion.
  
  \begin{table}[t]
  
  \tabcolsep=0.15cm
  \centering
  \resizebox{0.65\linewidth}{!}{
  \begin{tabular}{rcccccc} 
  \toprule
  & \multicolumn{6}{c}{Occlusion Ratio} \\
  \multicolumn{1}{c}{Method} & 0 & 0.1 &0.2 & 0.3 & 0.5 & 0.75 \\ 
  \midrule
  PointNet \cite{pointnet}& 89.1 &  88.2 &  86.1 &  81.6 &  53.5 &  4.7\\
  DGCNN \cite{dgcn} &  92.1 &  77.1 &  74.5 &  71.2 &  30.1 &  4.3\\ \midrule
  MVTN (ours)& \textbf{92.3} &  \textbf{90.3} &  \textbf{89.9} &  \textbf{88.3} &  \textbf{67.1} &  \textbf{9.5}   \\
  \bottomrule
  \end{tabular}
  }
  \vspace{2pt}
  \caption{ \textbf{Occlusion Robustness of 3D Methods.} We report the test accuracy on point cloud ModelNet40 for different occlusion ratios of the data to measure occlusion robustness of different 3D methods. MVTN achieves 13\% better accuracy than PointNet (a robust network) when half of the object is occluded.}
  \label{tbl:occlusion}
  \end{table}

  \subsection{Optimizing Scene Parameters}
  As an alternative to MVTN that leverages the end-to-end differentiable pipeline, we can treat the view selection problem as an optimization objective for scene parameters instead of learning a network. To do this, we experiment with optimizing the azimuth and elevation angles used for rendering during training by either maximizing or minimizing the Cross-Entropy (CE) loss while the main multi-view network is training. To accomplish this, we run a varying number of iterations for each batch in training, during which we render the input scene parameters, calculate the loss, and update the parameters using SGD. We test whether this optimization will outperform the baseline model without learning views and the baseline model with random noise added to the input view parameters. We also try minimizing and maximizing the norm of the rendered images, which decreases or increases coverage, respectively (Coverage Loss). Additionally, we attempt to maximize the distance between the top two logits from the output of the images (Adversarial Loss \cite{carlini}). The CE loss optimization is only applied during training, while the coverage and adversarial losses are optimized during both training and test time.
  
  \mysection{Initial Setup}
  We use the regular pipeline without learning views, a batch size of 20, 8 spherical views, black backgrounds, and white point cloud renderings to train on the three variants of the ScanObjectNN \cite{scanobjectnn} dataset. We optimize MVCNN \cite{mvcnn} using AdamW \cite{adamw} and a learning rate of 0.001, and use ResNet-18 \cite{resnet} as the backbone CNN. The baseline with noise adds a sample drawn from a normal distribution with a mean of 0 and standard deviation of 18 to each azimuth angle, and a mean of 0 and standard deviation of 9 to the elevation angles.
  
  \mysection{Optimization Results}
  To determine the optimal parameters for our optimization approach, we test different numbers of optimization iterations for each batch and various values for the learning rate. We also try using a ResNet-18 \cite{resnet} backbone that is either pretrained or not pretrained. We find that the best results are obtained with a pretrained backbone, a learning rate of 50, and 10 optimization iterations using the cross entropy loss. For the coverage and adversarial losses, we use a pretrained backbone, a learning rate of 25, and 10 optimization iterations. The results are presented in Table \ref{tab:input_opt}.
  
  \begin{table}[t]
  \tabcolsep=0.08cm
      \centering
  \resizebox{0.78\linewidth}{!}{\begin{tabular}{rccc}
  \toprule
   &  \multicolumn{3}{c}{Classification Overall Accuracy } \\
  \multicolumn{1}{c}{Method}& \textbf{OBJ\_BG}  & \textbf{OBJ\_ONLY} & \textbf{Hardest}  \\ \midrule
  Baseline Without Noise & 85.42 &  84.56 & 72.83  \\
  Baseline With Noise  & 85.59 & 84.56 & 78.52 \\
  Maximizing CE Loss & 83.53 & 86.28 & 74.95\\
  Minimizing CE Loss & 84.91 & 85.08 & 75.19\\
  Maximizing Coverage Loss & 86.28 & 85.25 & 75.75\\
  Minimizing Coverage Loss & 84.73 & 85.08 & 74.39\\
  Maximizing Adversarial Loss & 86.11 & 85.42 & 76.89 \\
  \bottomrule
  \end{tabular}
  }
  \vspace{2pt}
      \caption{\textbf{Input Scene Parameters Optimization}. We show the classification accuracy utilizing input scene optimization strategy using the different methods on the three ScanObjectNN \cite{scanobjectnn} variants. The optimization methods either maximize or minimize different losses, and we compare them to the baseline fixed spherical models with and without added parameter noise.}
      \label{tab:input_opt}
  \end{table}
  

  \section{Analysis and Insights} \label{sec:analysis}
  \subsection{Ablation Study} \label{sec:ablation}
  This section performs a comprehensive ablation study on the different components of MVTN and their effect on the overall test accuracy on ModelNet40 \cite{modelnet}.
  
\begin{figure}[t]
      \centering
      \includegraphics[,trim=0 0 1.5cm 1.8cm, clip,width=0.65\linewidth]{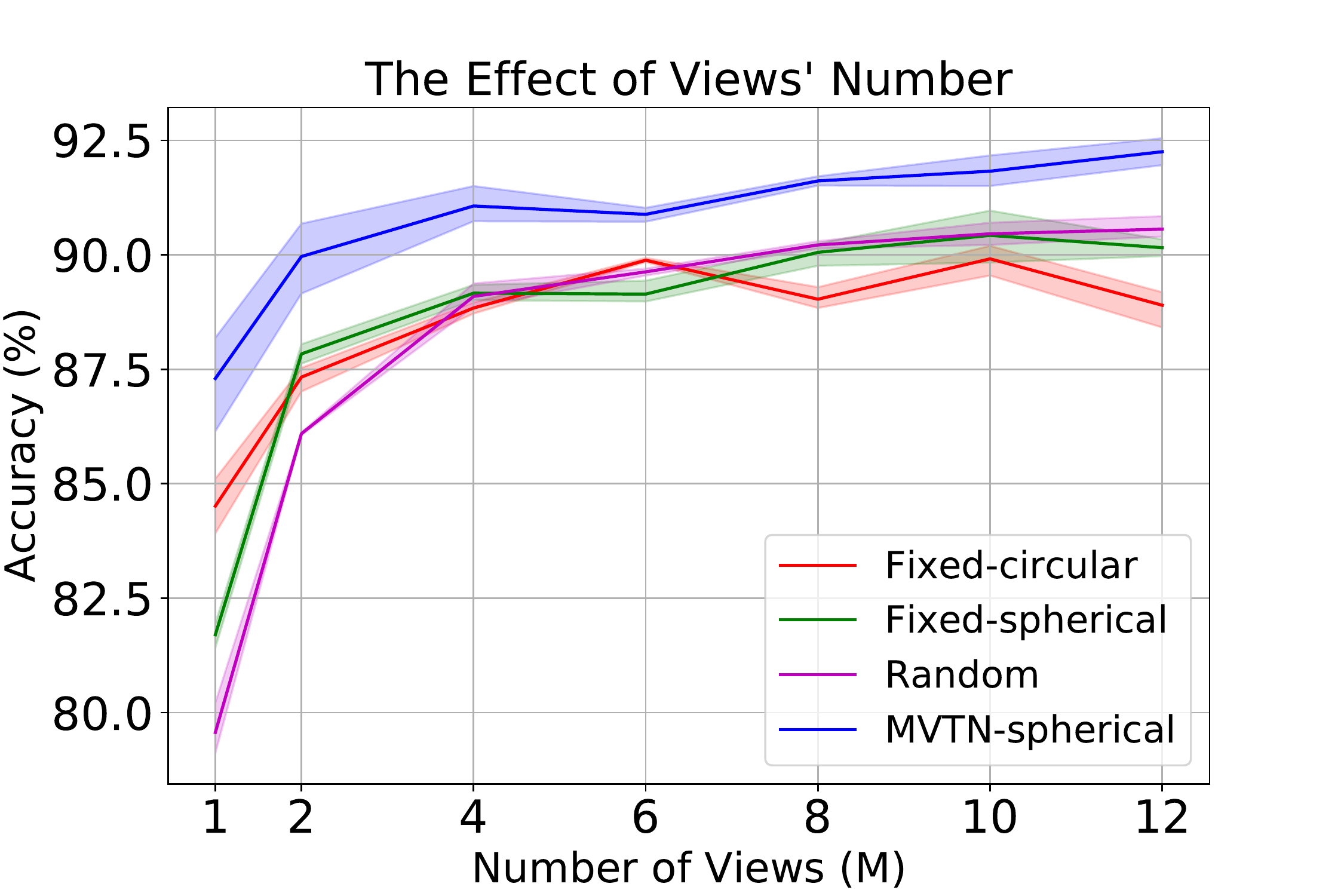}
      \caption{\textbf{Effect of the Number of Views.} We plot the test accuracy \vs the number of views (M) used to train MVCNN on fixed, random, and learned MVTN view configurations. 
      We observe a consistent 2\% improvement with MVTN over a variety of views.}
      \label{fig:classification}
  \end{figure}

  \mysection{Number of Views} \label{sec:views}
  We study the effect of the number of views $M$ on the performance of MVCNN when using fixed views (circular/spherical), learned views (MVTN), and random views.
  The experiments are repeated four times, and the average test accuracies with confidence intervals are shown in  \figLabel{\ref{fig:classification}}. 
   The plots show how learned MVTN-spherical achieves consistently superior performance across a different number of views. 
  
  \mysection{Choice of Backbone and Point Encoders}
  Throughout our main MVTN experiments, we use ResNet-18 as the backbone and PointNet as the point feature extractor. However, different choices could be made for both.
  We investigate the use of DGCNN \cite{dgcn} as an alternative point encoder and ResNet-34 as an alternative 2D backbone in ViewGCN. The results of these ablation studies are presented in Table \ref{tbl:ablation}. We observe that using more complex CNN backbones and shape feature extractors in the MVTN setup does not significantly improve the results, which justifies our use of the simpler combination in our main experiments.
  
  
  \mysection{Late Fusion}
  In the MVTN pipeline, we use a point encoder and a multi-view network. One can argue that an easy way to combine them would be to fuse them later in the architecture. For example, PointNet \cite{pointnet} and MVCNN \cite{mvcnn} can be max pooled together at the last layers and trained jointly. We train such a setup and compare it to MVTN. We observe that  MVTN achieves $91.8\%$ compared to $88.4\%$ by late fusion.
  
  \mysection{Effect of Object Color}
  Our main experiments used random colors for the objects during training and fixed them to white in testing. We tried different coloring approaches, like using a fixed color during training and test. The results are illustrated in Table \ref{tbl:color}. 
  
  \mysection{Other Factors Affecting MVTN} 
  We study the effect of the light direction in the renderer and the camera's distance to the object.
  We also study the transferability of the learned views from one multi-view network to another, and the performance of MVTN variants. More details are provided in the \supp\hspace{-2pt}.
  
  \subsection{Visualizing MVTN Learned Views}
\begin{figure}[t]
      \centering
         \begin{turn}{90} 
  \   \quad \quad Probability Density
  \end{turn} \includegraphics[width=0.65\linewidth ,trim=0 0 0 10.2cm, clip]{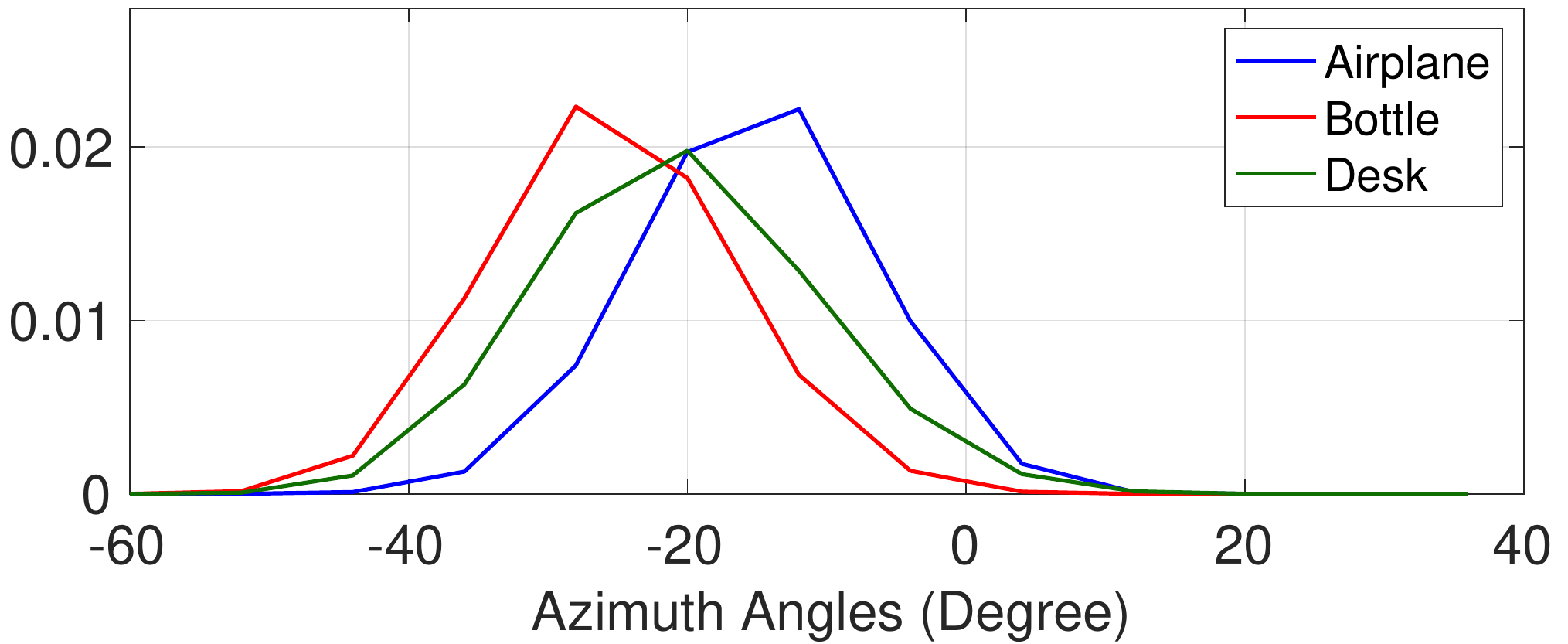}
      \caption{\textbf{Visualizing MVTN Learned Views.} We visualize the distribution of azimuth angles predicted by MVTN for three different classes. Note that MVTN learns inter-class variations (between different classes) and intra-class variations (in the same class).}
      \label{fig:distribution}
  \end{figure}

  In \figLabel{\ref{fig:distribution}}, we visualize the distribution of views predicted by MVTN for shapes from three different object classes. We use the MVTN-direct variant with $M=1$ to study the behavior of a single view. We histogram the learned views for the entire test set of ModelNet40. 
  In this setups, we observe that MVTN is learning a per-instance view and \textit{not} regressing the same view for shapes belonging to the same class. We see from \figLabel{\ref{fig:distribution}} that the distribution of the MVTN views varies from one class to another, and the views from the same class exhibit some variance between instances. 
  
  \begin{table}[t]
  \footnotesize
  \setlength{\tabcolsep}{6pt} 
  \renewcommand{\arraystretch}{1.1} 
  \centering
  \resizebox{0.63\hsize}{!}{
  \begin{tabular}{c|c|c|c} 
  \toprule
   \multicolumn{1}{c|}{\textbf{Backbone}}& \multicolumn{1}{c|}{\textbf{Point}}& \multicolumn{1}{c|}{\textbf{MVTN}}&   \multicolumn{1}{c}{\textbf{Results}}\\
   \textbf{Network} & \textbf{Encoder}  & \textbf{Setup} &  \textbf{Accuracy}     \\
  \midrule
   \multirow{4}{*}{ResNet-18} & \multirow{2}{*}{PointNet} &  circular  &  92.83 $\pm$ 0.06 \\ 
    & &  spherical  &  93.41 $\pm$ 0.13  \\ 
   & \multirow{2}{*}{DGCNN} & circular  &  93.03 $\pm$ 0.15  \\ 
   & &  spherical   &  93.26 $\pm$ 0.04 \\  \hline
  
     \multirow{4}{*}{ResNet-34} & \multirow{2}{*}{PointNet} &  circular &  92.72 $\pm$ 0.16 \\ 
     &  &  spherical  &  92.83 $\pm$ 0.12 \\ 
     & \multirow{2}{*}{DGCNN} & circular  & 92.72 $\pm$ 0.03 \\ 
     &  &  spherical   &  92.63 $\pm$ 0.15 \\ 
   \bottomrule
  \end{tabular}
  }
  \vspace{2pt}
  \caption{  \textbf{Ablation Study}. We analyze the effect of ablating different MVTN components on test accuracy in ModelNet40. Namely, we observe that using deeper backbone CNNs or a more complex point encoder do not increase the test accuracy.}
  \label{tbl:ablation}
  \end{table}
  \begin{table}[t]
  \tabcolsep=0.25cm
  \centering
  \resizebox{0.6\linewidth}{!}{
  \begin{tabular}{c|cc} 
  \toprule
   & \multicolumn{2}{c}{Object Color}  \\ 
  Method & White & Random \\
  \midrule
  Fixed views & 92.8 $\pm$  0.1  & 92.8 $\pm$  0.1     \\
  MVTN (learned)   &93.3 $\pm$  0.1 & \textbf{93.4} $\pm$  0.1    \\
  \bottomrule
  \end{tabular}
  }
  \vspace{2pt}
  \caption{  \textbf{Effect of Color Selection}. We ablate selecting the color of the object in training our MVTN and when views are fixed in the spherical configuration. Fixed white color is compared to random colors in training. Note how randomizing the color helps in improving the test accuracy on ModelNet40 a little bit. }
  \label{tbl:color}
  \end{table}
  
  
  \subsection{2D Pretraining Strategy}
  We aim to evaluate the impact of 2D pretraining on the performance of our pipeline. We focus on two pretraining paradigms: Self-Supervised pretraining (SSL) and Full-Supervised pretraining (FSL). We conduct experiments using MVTN with ResNet-50 \cite{resnet} and ViT \cite{vit} as backbone networks, starting from scratch or using weights from ImageNet \cite{IMAGENET}, Dino \cite{caron2021emerging}, and Masked Autoencoders (MAE) \cite{mae} as initial weights. We also test FSL as pretraining for our model setup.
  
  \mysection{Experiment Setup}
  Using 8 learned spherical views, batch size of 20, black backgrounds, and white point cloud renderings. We train for 100 epochs using ResNet-50 \cite{resnet} and ViT \cite{vit} on the three variants of the ScanObjectNN \cite{scanobjectnn} dataset. For each backbone network, we run training from scratch without pretraining, and we run experiments with different pretraining methods. The pretraining methods include using ImageNet \cite{IMAGENET} weights and using SSL weights from Dino \cite{caron2021emerging}, and MAE \cite{mae}. For FSL pretraining, we first train for 100 epochs from scratch using 1 random view of each shape in the dataset and then use the resulting weights as our initial weights in the experiment setup mentioned before.
  
  \mysection{Results}
  The results of these experiments can be found in Tables \ref{tab:ssl_and_fsl_vit} and \ref{tab:ssl_and_fsl_resnet} for ViT \cite{vit} and ResNet-50 \cite{resnet}, respectively. 
  
  \begin{table}[t]
  \tabcolsep=0.08cm
      \centering
  \resizebox{0.6\linewidth}{!}{\begin{tabular}{rccc}
  \toprule
   &  \multicolumn{3}{c}{Classification Overall Accuracy } \\
  \multicolumn{1}{c}{Method}& \textbf{OBJ\_BG}  & \textbf{OBJ\_ONLY} & \textbf{Hardest}  \\ \midrule
  Scratch ViT & 68.78 & 75.47 & 61.97  \\
  ImageNet ViT & 91.25 & 90.74 & 84.21  \\
  Dino ViT & 90.91 & 91.08 & 81.37  \\
  MAE ViT & 86.11 & 88.34 & 80.85  \\
  FSL ViT & 62.61 & 71.18 & 57.53  \\
  \bottomrule
  \end{tabular}
  }
  \vspace{2pt}
      \caption{\textbf{Effect of ViT Pretraining Strategy on MVTN}. We show the impact of using different pretraining methods when training MVTN with ViT \cite{vit} as the backbone network.}
      \label{tab:ssl_and_fsl_vit}
  \end{table}
  
  \begin{table}[t]
  \tabcolsep=0.08cm
      \centering
  \resizebox{0.63\linewidth}{!}{\begin{tabular}{rccc}
  \toprule
   &  \multicolumn{3}{c}{Classification Overall Accuracy } \\
  \multicolumn{1}{c}{Method}& \textbf{OBJ\_BG}  & \textbf{OBJ\_ONLY} & \textbf{Hardest}  \\ \midrule
  Scratch ResNet-50  & 63.64 & 71.01 & 61.21  \\
  ImageNet ResNet-50  & 88.16 & 89.02 & 81.26  \\
  Dino ResNet-50  & 81.13 & 84.39 & 77.69  \\
  FSL ResNet-50  & 58.66 & 65.18 & 52.22  \\
  \bottomrule
  \end{tabular}
  }
  \vspace{2pt}
      \caption{\textbf{Effect of ResNet50 Pretraining Strategy on MVTN}. We show the impact of using different pretraining methods when training MVTN with ResNet-50 \cite{resnet} as the backbone network.}
      \label{tab:ssl_and_fsl_resnet}
  \end{table}
  
  \subsection{MVTN for Part Segmentation}
  \mysection{Experiment Setup}
  We investigate applying MVTN for part segmentation on ShapeNet Parts \cite{shapenetparts}. Similar to Voint Cloud \cite{voint}, we use DeepLabV3 \cite{deeplabv3} as the 2D backbone and follow similar procedures in uplifting the predictions from 2D to 3D (mode un-projection). We add MVTN before the rendering process and optimize the MVTN network controlling the azimuth and elevation angles as before but using the 2D segmentation loss instead of the classification loss. 
  
\begin{figure}[t]
      \centering
      \includegraphics[,clip,width=0.65\linewidth]{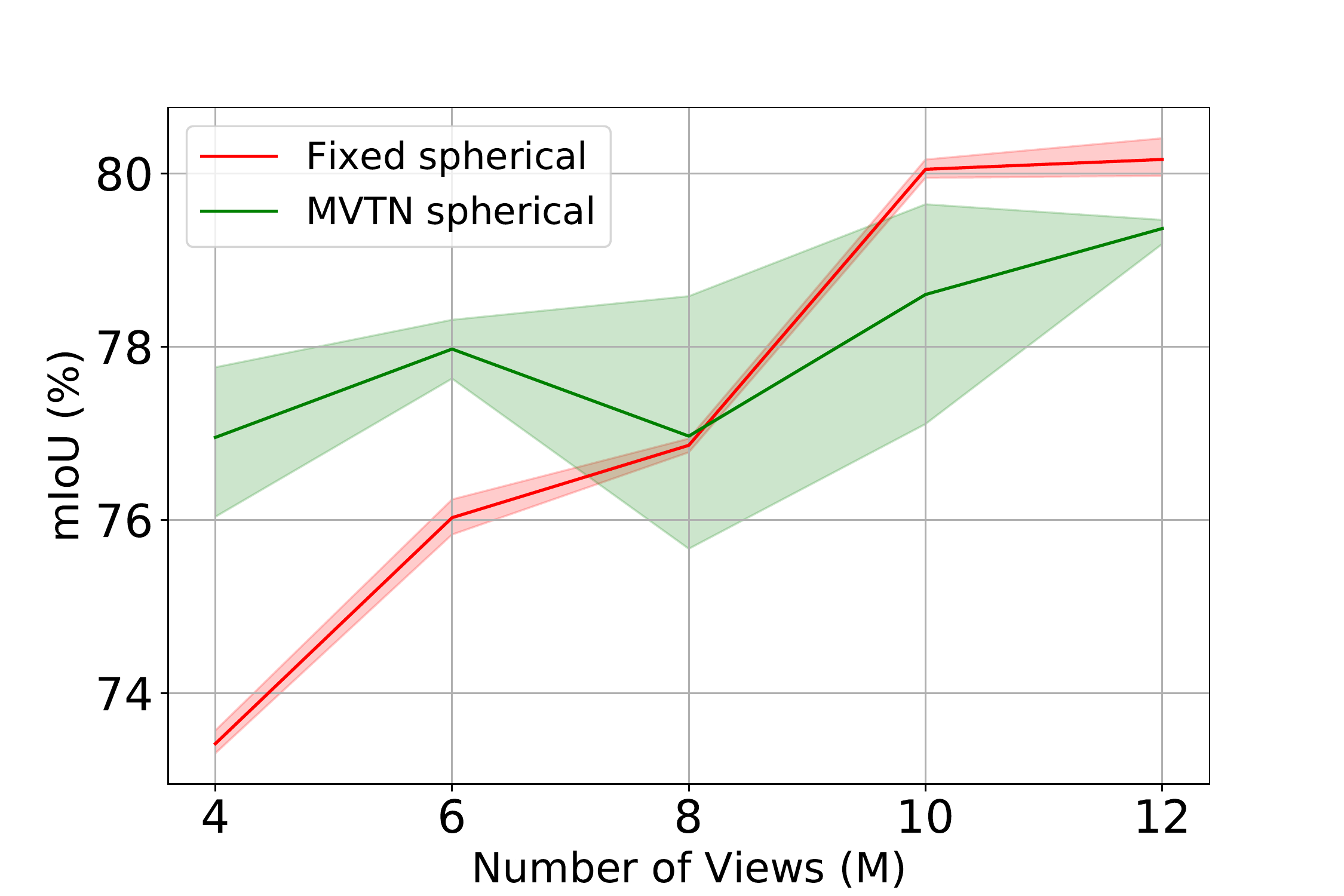}
      \caption{\textbf{MVTN for Part Segmentation.} We plot the 3D segmentation Instance mIOU vs. the number of views used in training the pipeline. }
      \label{fig:segment}
  \end{figure}

  \mysection{Results} 
  The results for segmentation instance mean IOU of the pipeline with MVTN and without MVTN are reported in \figLabel{\ref{fig:segment}}. It shows that as the number of views increases,  the performance increases for both the baseline and MVTN, with marginal improvement using MVTN. Similar to the classification case, as the number of views becomes large, the margin of MVTN improvement becomes smaller. We also show in Table \ref{tab:shapenetparts} the performance of different methods tested on a similar setup. Note how point methods despite performing well, don't have the utility of increasing the performance by increasing the views, while MVTN tiny network is giving competitive performance to that of the recent and heavy VointNet \cite{voint}, especially for few views (4 views for example).

  \begin{table}[t]
          \centering
  \resizebox{0.67\linewidth}{!}{\begin{tabular}{lccc}
  \toprule
   &   \multicolumn{3}{c}{\textbf{3D Part Segmentation}} \\
  \multicolumn{1}{c}{Method}       & Data Type & \# of Views   & Inst. mIoU \\ \midrule
  PointNet \citep{pointnet}&  Points & - &   80.1     \\
  DGCNN \citep{dgcn}&  Points & -  & 80.1      \\ 
  Label Fuse \citep{mvlabeldifusion}& M-View  & 4  & 75.2  \\
  Label Fuse \citep{mvlabeldifusion}& M-View  & 12  & 80.0  \\
  VointNet \cite{voint} & Voints & 4  &  77.8 \\ 
  VointNet \cite{voint} & Voints & 12  &  \textbf{81.2} \\ \midrule
  MVTN (ours) & M-View & 4  &  77.0 \\
  MVTN (ours) & M-View & 12  &  79.4 \\
  \bottomrule
  \end{tabular}
  }
  \vspace{2pt}
      \caption{  \textbf{3D Part Segmentation on ShapeNet Parts}. We compare the Inst. mIoU of MVTN simple pipeline against other methods in 3D segmentation on ShapeNet Parts baselines, including the recent VointNet \cite{voint}. All the results are reproduced in our setup.}
      \label{tab:shapenetparts}
  \end{table}
  
  \subsection{large-Scale Scene Processing}
\begin{figure*} [h] 
  \centering
  \resizebox{0.85\linewidth}{!}{\begin{tabular}{c}
  \includegraphics[,trim=0.8cm 0 0.8cm 0, clip,width = 0.99\linewidth]{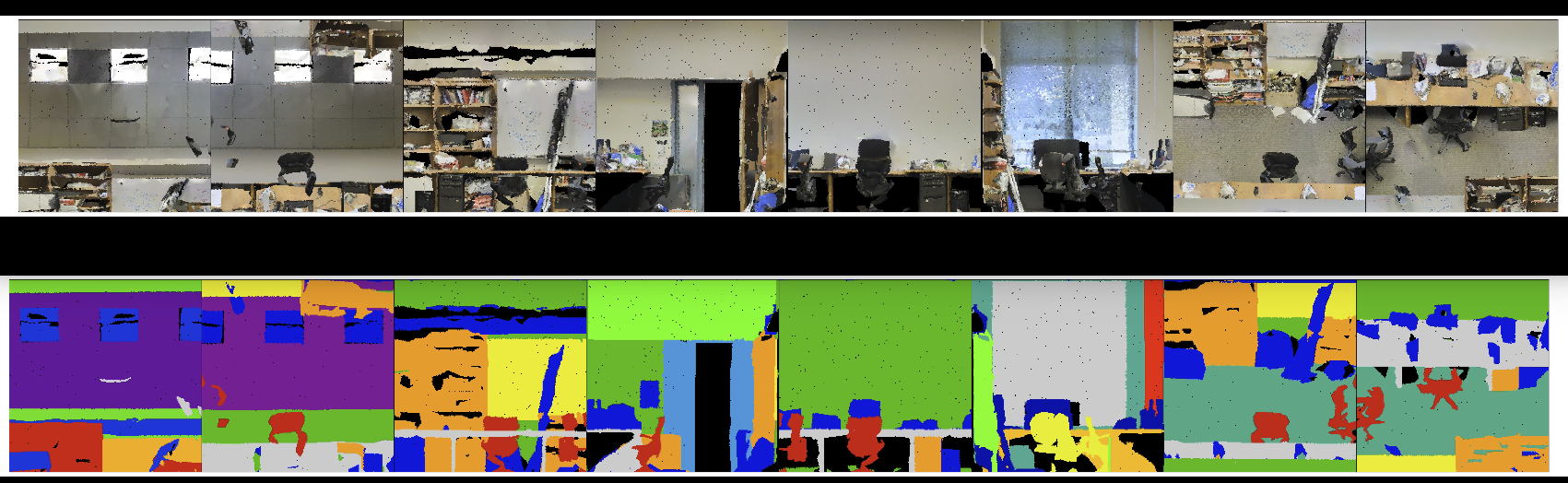} \\ \midrule
  \includegraphics[,trim=0.8cm 0 0.8cm 0, clip,width = 0.99\linewidth]{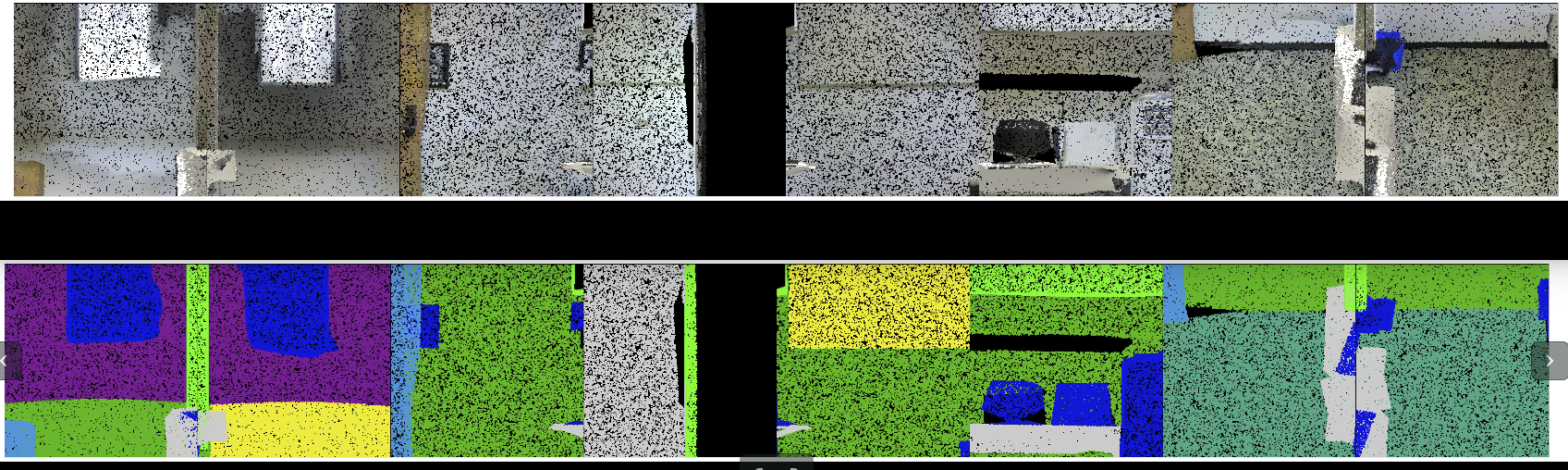} \\ \midrule
  \includegraphics[,trim=0.8cm 0 0.8cm 0, clip,width = 0.99\linewidth]{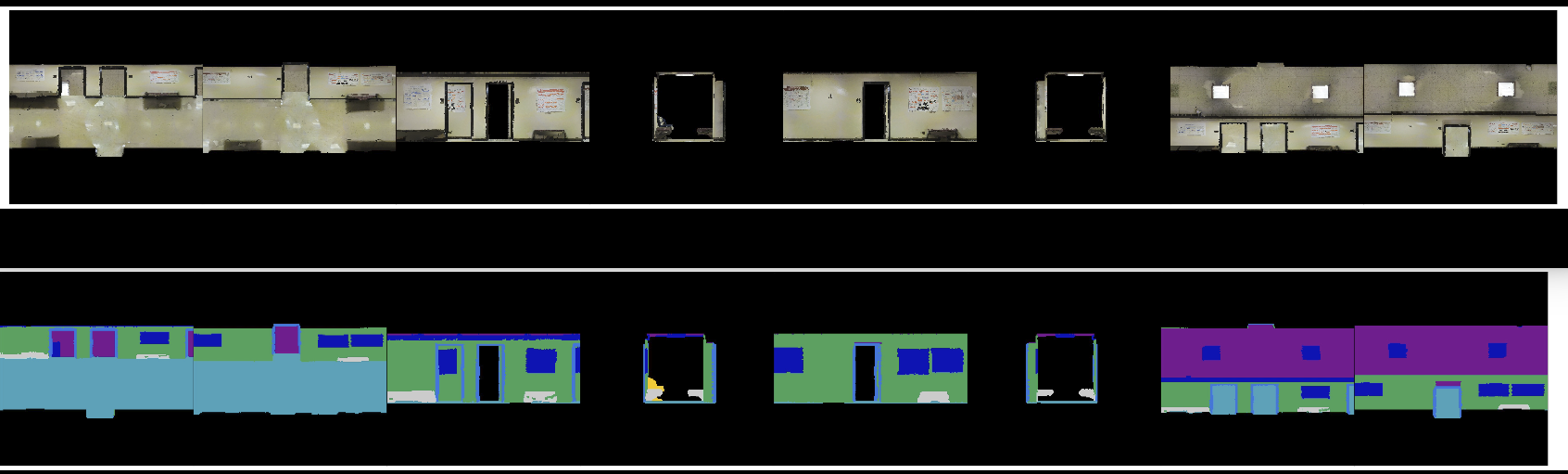} 
  \end{tabular}
  }
  \vspace{2pt}
  \caption{\textbf{Multi-View Renderings on Large 3D Scenes.} We show examples of multi-view renderings of 3D point clouds from the S3DIS \cite{s3dis} dataset of three different scenes on each of the three rows. Each row has the RGB rendering (\textit{upper part}) and the 3D labels rendering (\textit{lower part}). Note that because of the scale and structure differences between the different scenes, it is tricky to place the camera viewpoints such as to cover the entire scene and extract relevant semantic information from the multi-view network.      
  }
      \label{fig:s3dis}
  \end{figure*}

  Large-scale outdoor 3D segmentation (\eg Semantic-Kitty dataset \cite{semantickitty}) is a challenging task for view-based methods due to the difficulty in structuring LiDAR point clouds into reasonable images for processing. The points are sparse and far apart, limiting the benefits of view features in 3D. View selection is also more complex as there is no clear ``object of interest''. This is also a problem in 3D indoor scene segmentation where the point density is higher than outdoor but the ambiguity of picking the views is similar.  
  We show examples of multi-view renderings of 3D point clouds from the S3DIS \cite{s3dis} dataset with different scenes in \figLabel{\ref{fig:s3dis}}. Note that because of the scale and structure differences between the different scenes, it is tricky to place the camera viewpoints such as to cover the entire scene and extract relevant semantic information from the multi-view network.  NVTN can in theory mitigate this sever effect by adjusting the views to increase the segmentation performance. Additional constraints can be added to regularize the unconstrained views like the total points covered and total normal norm maximization \cite{simpleview}.

  \subsection{Time and Memory of MVTN}
  In this study, we evaluated the impact of the MVTN module on the time and memory requirements of our pipeline. To do this, we measured the number of floating point operations (FLOPs) and multiply-accumulate (MAC) operations for each module, as well as the time required for a single input sample to pass through each module and the number of parameters for each module. Our results, presented in Table \ref{tbl:speed-supp}, show that the MVTN module has a negligible impact on the overall time and memory requirements of the multi-view networks and 3D point encoders.
  \begin{table}[t]
  \tabcolsep=0.2cm
  \centering
  \resizebox{0.65\linewidth}{!}{
  \begin{tabular}{l|cccc} 
  \toprule
  Network & FLOPs & MACs & Params. \# & Time\\
  \midrule
  PointNet &  1.78 G   & 0.89 G    & 3.49 M   &3.34 ms    \\
  DGCNN &  10.42 G  &  5.21 G & 0.95 M  & 16.35 ms\\
  MVCNN &  43.72 G  & 21.86 G  &  11.20 M &  39.89 ms\\
  ViewGCN &  44.19 G &  22.09 G & 23.56 M  & 26.06 ms \\ \midrule
  MVTN$^*$ &  \textbf{18.52 K}  & \textbf{9.26 K}  & \textbf{9.09 K}  & \textbf{0.9 ms}\\
  MVTN$^\circ$ & 1.78 G & 0.89 G & 4.24 M   & 3.50 ms \\
  \bottomrule
  \end{tabular}
  }
  \vspace{2pt}
  \caption{  \textbf{Time and Memory Requirements}. We assess the contribution of the MVTN module to time and memory requirements in the multi-view pipeline. MVTN$^*$ refers to MVTN's regressor excluding the point encoder, while MVTN$^\circ$ refers to the full MVTN module including PointNet as a point encoder.}
  \label{tbl:speed-supp}
  \end{table}
  
  \subsection{MVTorch: a Library for Multi-view Deep Learning Research} \label{sec:mvtorch}
  We propose \href{https://github.com/ajhamdi/mvtorch}{\textit{MVTorch}}, a PyTorch library that enables efficient, modular development of multi-view 3D computer vision and graphics research. This library is built using PyTorch \cite{paszke2017pytorch} and Pytorch3D \cite{pytorch3d} and offers a variety of features for working with 3D data and multi-view images. These features include the ability to render differentiable multi-view images from 3D meshes and point clouds, data loaders for 3D data and multi-view images, visualizations of 3D meshes, point clouds, and multi-view images, modular training of multi-view networks for different 3D tasks, and input/output capabilities for 3D data and multi-view images. These features are implemented using PyTorch tensors and are designed to be used with mini-batches of heterogeneous data, allowing them to be easily differentiated and run on GPUs for acceleration.
  
  \begin{table}[h]
  \centering
  \tabcolsep=0.12cm
      \centering
  \resizebox{0.8\linewidth}{!}{
  \begin{tabular}{l|c|c|c}
  \toprule
  \textbf{Module} & \textbf{Time (ms)} & \textbf{Params (M)} & \textbf{GFLOPs} \\
  \midrule
  MVRenderer Points 4-V  &   10.74  &   0.00  &   - \\
  MVRenderer Points 6-V  &   12.61  &   0.00  &   - \\
  MVRenderer Points 8-V  &   15.42  &   0.00  &   - \\
  MVRenderer Points 10-V  &   17.45  &   0.00  &   - \\
  MVRenderer Points 12-V  &   20.16  &   0.00  &   - \\
  MVRenderer Meshes 4-V  &   19.15  &   0.00  &   - \\
  MVRenderer Meshes 6-V  &   25.11  &   0.00  &   - \\
  MVRenderer Meshes 8-V  &   28.83  &   0.00  &   - \\
  MVRenderer Meshes 10-V  &   32.94  &   0.00  &   - \\
  MVRenderer Meshes 12-V  &   37.55  &   0.00  &   - \\ \hline 
  MVCNN ResNet-18 4-V  &   2.59  &   11.20  &   14.57 \\
  MVCNN ResNet-18 6-V  &   2.75  &   11.20  &   21.86 \\
  MVCNN ResNet-18 8-V  &   3.13  &   11.20  &   29.15 \\
  MVCNN ResNet-18 10-V  &   3.50  &   11.20  &   36.43 \\
  MVCNN ResNet-18 12-V  &   4.01  &   11.20  &   43.72 \\ \hline 
  MVTN PointNet 4-V  &   3.46  &   3.47  &   1.77 \\
  MVTN PointNet 6-V  &   3.44  &   3.47  &   1.77 \\
  MVTN PointNet 8-V  &   3.46  &   3.47  &   1.77 \\
  MVTN PointNet 10-V  &   3.44  &   3.48  &   1.77 \\
  MVTN PointNet 12-V  &   3.42  &   3.48  &   1.77 \\
  MVTN DGCNN 4-V  &   4.09  &   0.95  &   10.42 \\
  MVTN DGCNN 6-V  &   4.10  &   0.96  &   10.42 \\
  MVTN DGCNN 8-V  &   4.10  &   0.96  &   10.42 \\
  MVTN DGCNN 10-V  &   4.10  &   0.96  &   10.42 \\
  MVTN DGCNN 12-V  &   4.10  &   0.96  &   10.42 \\
  \hline 
  MVPartSeg deeplab 4-V  &   54.00  &   61.00  &   387.74 \\
  MVPartSeg deeplab 6-V  &   67.15  &   61.00  &   581.61 \\
  MVPartSeg deeplab 8-V  &   91.04  &   61.00  &   775.48 \\
  MVPartSeg deeplab 10-V  &   112.52  &   61.00  &   969.35 \\
  MVPartSeg deeplab 12-V  &   121.49  &   61.00  &   1163.21 \\
  \bottomrule 
  \end{tabular}
  }
  \vspace{2pt}
  \caption{  \textbf{Performance Benchmarking of MVTorch.} We show different pipelines and configurations of the modules in MVtorch with the speed and compute cost associated with them on a varying number of views. The first block includes only the rendering pipeline. The second and third blocks are for 3D classification using MVCNN \cite{mvcnn} with and without MVTN respectively. The last block is for the segmentation pipeline with DeepLabV3 \cite{deeplabv3}.}
  \label{tab:perf}
  \end{table}
  
  Tutorials and documentation for MVTorch can be found on the project's GitHub page (\href{https://github.com/ajhamdi/mvtorch}{github.com/ajhamdi/mvtorch}). his documentation includes examples of how to use MVTorch for tasks such as 3D classification, retrieval, segmentation, and generating 3D meshes from text \cite{text2mesh}, and Neural Radiance Fields \cite{nerf} examples.  Key classes in MVTorch include \textit{MVRenderer} (for rendering both point clouds and meshes), \textit{MVNetwork} (which allows any 2D network to be input and outputs its multi-view version), visualizer (for handling multi-view and 3D visualization), \textit{MVDataloader} (for loading any dataset including ModelNet, ShapeNet, ScanObjectNN, ShapeNet Parts, and S3DIS), \textit{view-selector} (\eg MVTN, random, circular, etc.), and \textit{MVAggregate}, which aggregates multi-view data into 3D representations (such as through maxpooling, meanpooling, point aggregation, and lifting). Table \ref{tab:perf} shows the performance of different modules and configurations in MVTorch for classification and segmentation tasks.

  \subsection{MVTN and Modern Multi-View Approaches}
  Multi-view methods are currently constituting the mainstream research direction for zero-shot 3D understanding \cite{astar}, 3D generation \cite{text2mesh,DreamFusion,Zero-1-to-3,magic123}, and even novel view synthesis \cite{nerf,gaussiansplatter,ges}. Neural Radiance Fields (NeRFs) \cite{nerf} learn a view-dependant implicit representation through MLPs sampled along rays to allow for realistic novel views synthesis, while 3D Gaussian Splatting \cite{gaussiansplatter} utilize explicit 3D splats to speed up the rendering and optimization. With the success of 2D generative models \cite{DALL-E,LDM}, several approaches used 2D CLIP \cite{CLIP} or 2D-Diffusion priors \cite{LDM} in order to improve the 3D generation performance tremendously \cite{text2mesh,DreamFusion}.
  We have included examples of these modern approaches (\eg NeRF \cite{nerf} and Text2Mesh \cite{text2mesh} ) as part of the MVTorch library we propose in this work in \secLabel{\ref{sec:mvtorch}}. 
  
  While the majority of the MVTN work conducted in this study preceded this late wave of multi-view deep learning, there exist a lot of opportunities for MVTN to play in the problem of view selection as follows. For example, the distribution and density of views in NeRFs and Gaussian Splatting is still an active area of research \cite{barf,sparf2023}. Even for Diffusion-based methods that are fine-tuned on large datasets like Zero-1-2-3 \cite{Zero-1-to-3}, the choice of views is very critical and will affect the performance greatly. MVTN can play a role here by learning an affordable refinement network that changes the viewpoints to improve performance.   
  
  \section{Conclusions} \label{sec:conclusion}
   Current multi-view methods rely on fixed views aligned with the dataset. We propose MVTN that learns to regress view-points for any multi-view network in a fully differentiable pipeline. MVTN harnesses recent developments in differentiable rendering and does not require any extra training supervision. Empirical results highlight the benefits of MVTN in 3D classification and 3D shape retrieval. 
   
   \mysection{Acknowledgements}
 This work was supported by the King Abdullah University of Science and Technology (KAUST) Office of Sponsored Research through the Visual Computing Center (VCC) funding.
 
\clearpage
\begin{appendices}

\section{Details of the IJCV Journal Extension of this Manuscript}
A preliminary version of this work was published at ICCV 2021 \cite{mvtn}.
This journal manuscript extends the initial version in several
aspects as follows.
\begin{itemize}
    \item We investigate and experiment with a logical alternative to MVTN by treating the problem as an optimization of the scene parameters instead of learning a transformation network. 
    \item  We study the effect of different pretraining strategies of the 2D backbone, which was shown in \cite{mvtn} to play an important role in MVTN's success. 
    \item We extend MVTN to the 3D part segmentation task and show promise in learning views beyond classification pipelines.
    \item To ensure the reproducibility of our experiments and to contribute to the 3D understanding/generation research community, we have published \textit{MVTorch}, a modular Pytorch library for training, testing, and visualization of multi-view deep learning pipelines.

\end{itemize}
 \clearpage \clearpage 
 
\section{Detailed Experimental Setup}
\subsection{Datasets}
\begin{figure*} [h] 
\centering
\includegraphics[width = 0.9\linewidth]{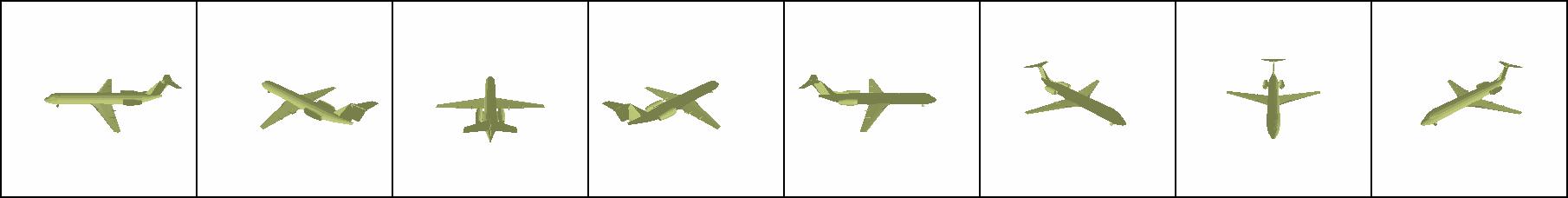} \\
\includegraphics[width = 0.9\linewidth]{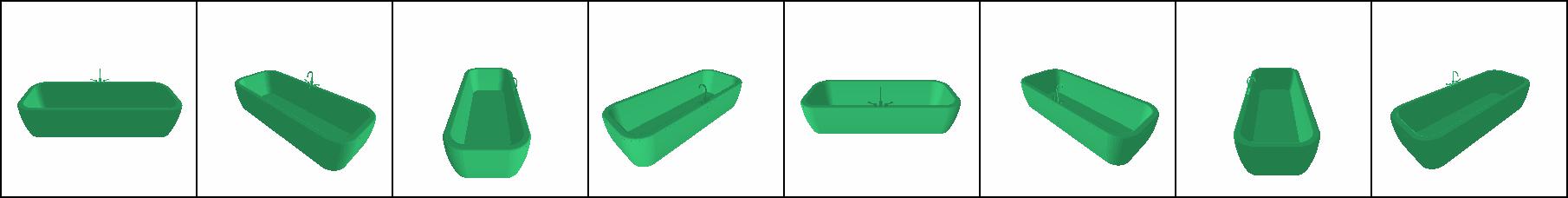} \\
\includegraphics[width = 0.9\linewidth]{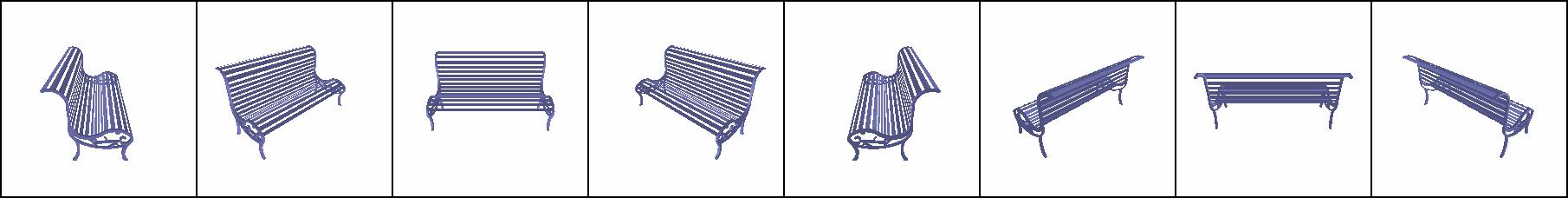} \\
\includegraphics[width = 0.9\linewidth]{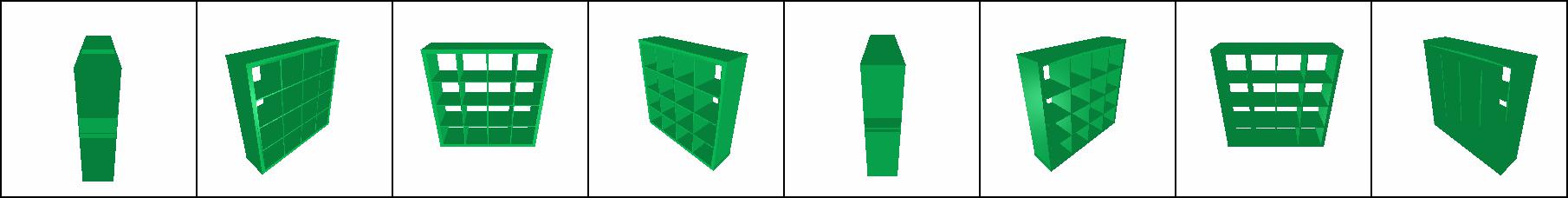} \\
\includegraphics[width = 0.9\linewidth]{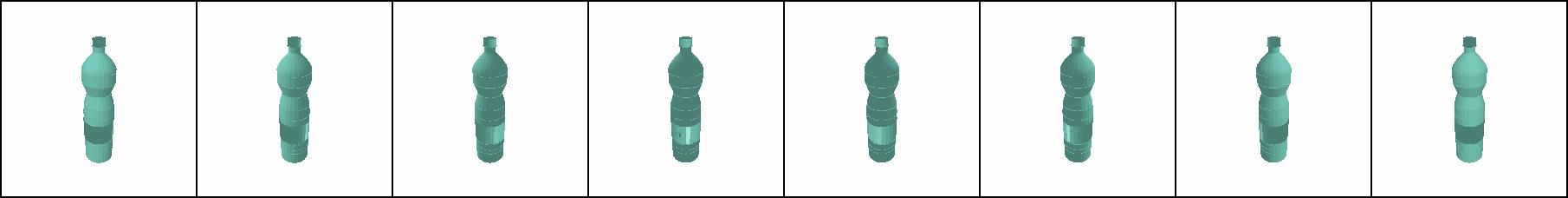} \\
\includegraphics[width = 0.9\linewidth]{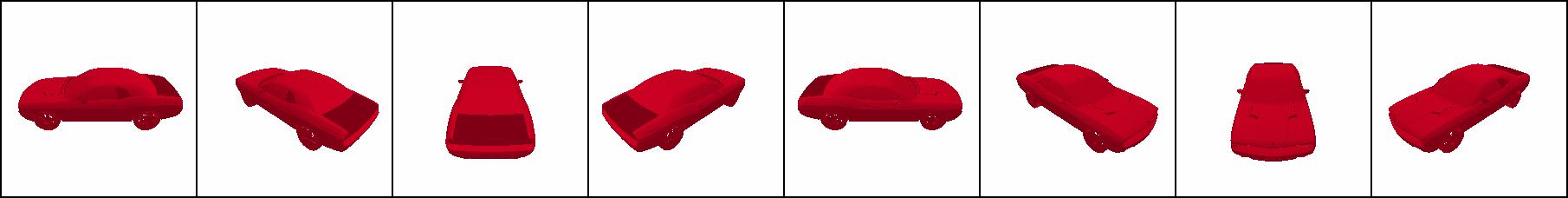} \\
\includegraphics[width = 0.9\linewidth]{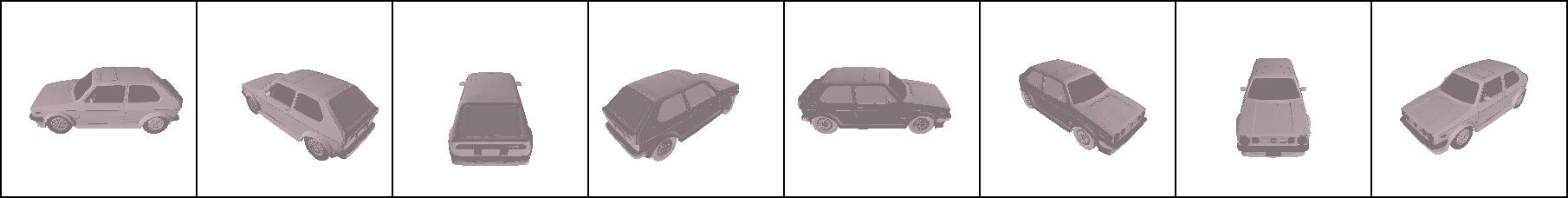} \\
\includegraphics[width = 0.9\linewidth]{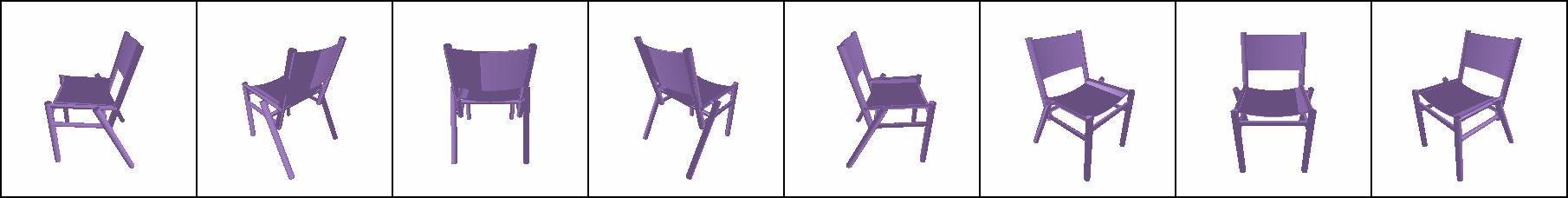} \\
\includegraphics[width = 0.9\linewidth]{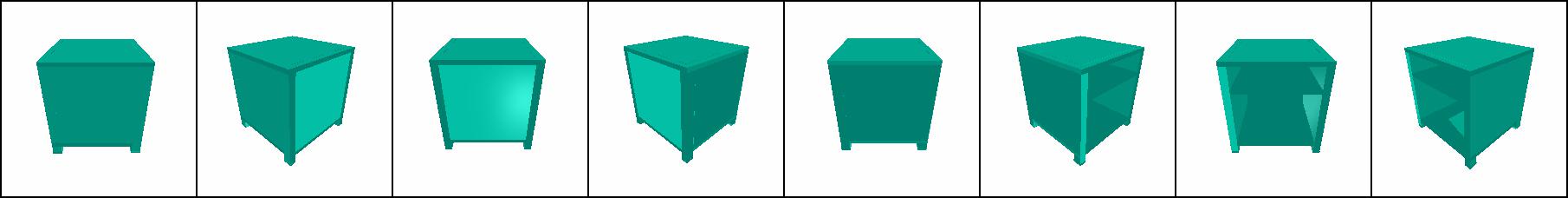} \\
\includegraphics[width = 0.9\linewidth]{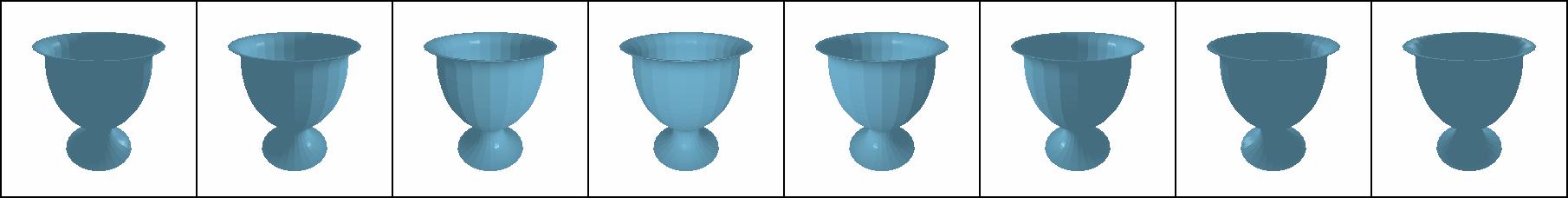} \\

\vspace{2pt}
\caption{  \textbf{Training Data with Randomized Color and Lighting.} We show examples of mesh renderings of ModelNet40 used in training our MVTN. The color of the object and the light's direction are randomized during training for augmentation purposes and fixed in testing for stable performance. For this figure, eight circular views are shown for each 3D shape.  
}
    \label{fig:colorful-sup}
\end{figure*}

\mysection{ModelNet40}
We show in \figLabel{\ref{fig:colorful-sup}} examples of the mesh renderings of ModelNet40 used in training our MVTN. Note that the color of the object and the light direction are randomized in training for augmentation but are fixed in testing for stable performance.   

\begin{figure*}[t]
    \centering
    \includegraphics[width=0.9\linewidth]{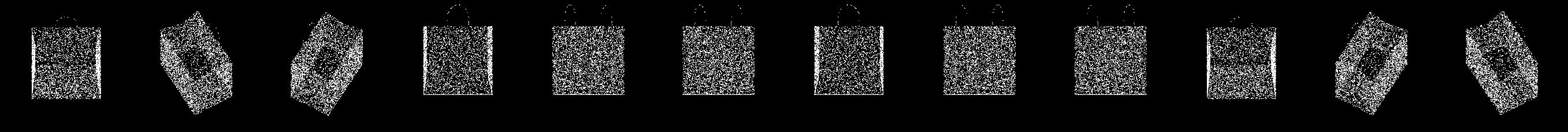}\\
    \includegraphics[width=0.9\linewidth]{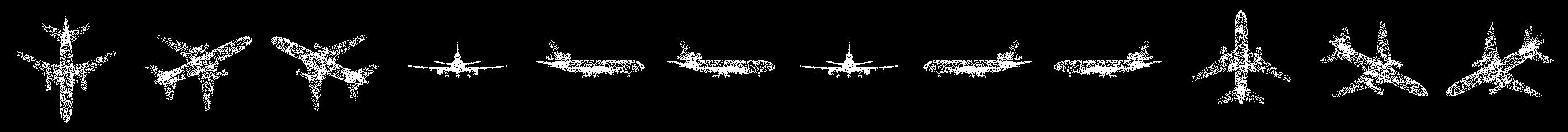}\\
    \includegraphics[width=0.9\linewidth]{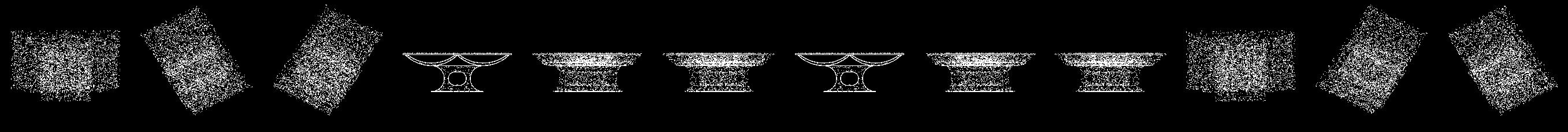}\\
    \includegraphics[width=0.9\linewidth]{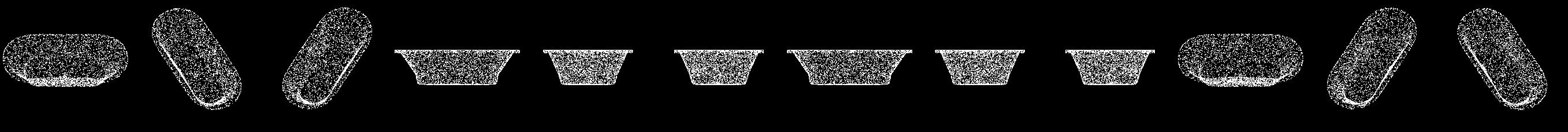}\\
    \includegraphics[width=0.9\linewidth]{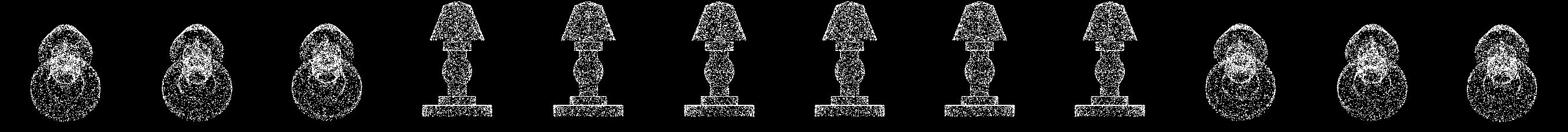}\\
    \includegraphics[width=0.9\linewidth]{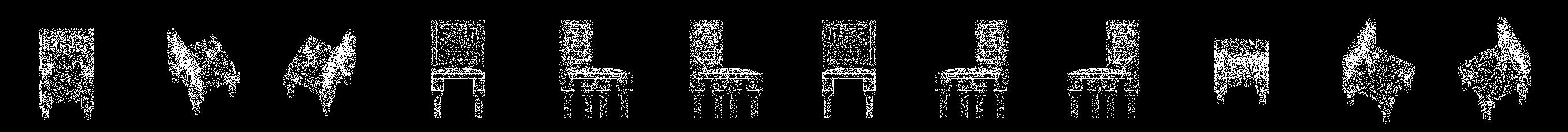}\\
    \includegraphics[width=0.9\linewidth]{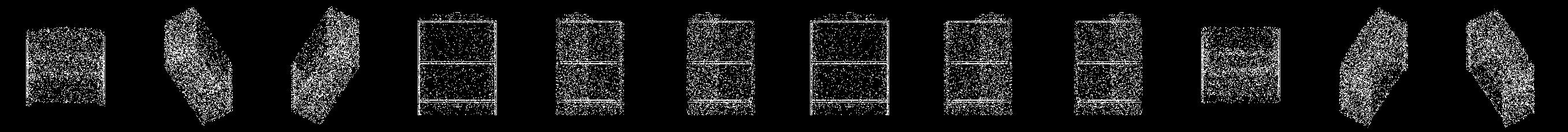}\\
    \includegraphics[width=0.9\linewidth]{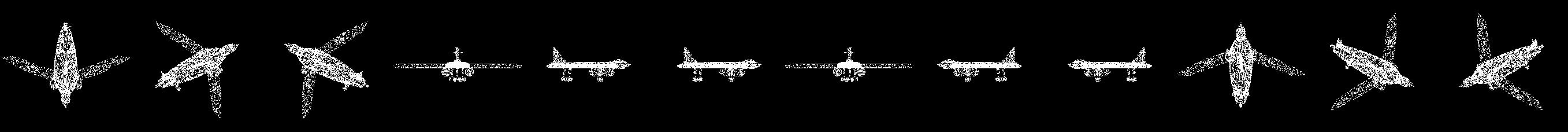}\\
    \includegraphics[width=0.9\linewidth]{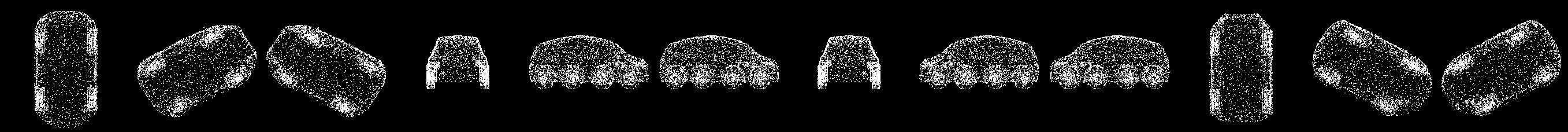}\\
    \includegraphics[width=0.9\linewidth]{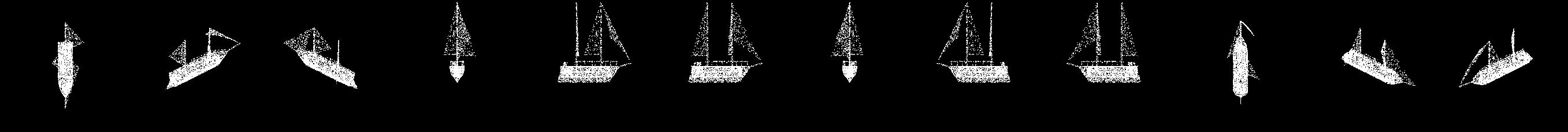}
    \caption{\textbf{ShapeNet Core55.} We show some examples of point cloud renderings of ShapeNet Core55 \cite{shapenet} used in training MVTN. Note how point cloud renderings offer more information about content hidden from the camera view-point (\eg car wheels from the occluded side), which can be useful for recognition. For this figure, 12 spherical views are shown for each 3D shape.}
    \label{fig:point-rendring-supp}
\end{figure*}

\mysection{ShapeNet Core55}
In \figLabel{\ref{fig:point-rendring-supp}}, we show  examples of the point cloud renderings of ShapeNet Core55 \cite{shapenet,shrek17} used in training MVTN. Note how point cloud renderings offer more information about content hidden from the camera view-point, which can be useful for recognition. White color is used in training and testing for all point cloud renderings. For visualization purposes, colors are inverted in the main paper examples (Fig. 4 in the main paper).

\begin{figure*} [h] 
\tabcolsep=0.03cm
\resizebox{0.9\linewidth}{!}{
\begin{tabular}{c|c}

\multirow{3}{*}{\textbf{Object Only}} &
\includegraphics[width = 0.7\linewidth]{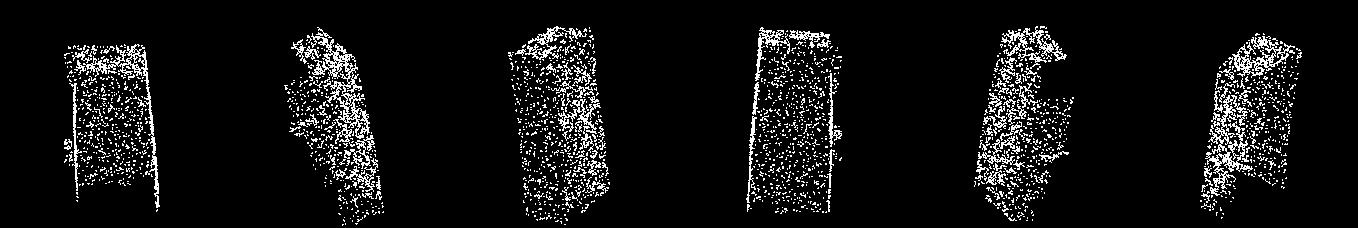} \\
&\includegraphics[width = 0.7\linewidth]{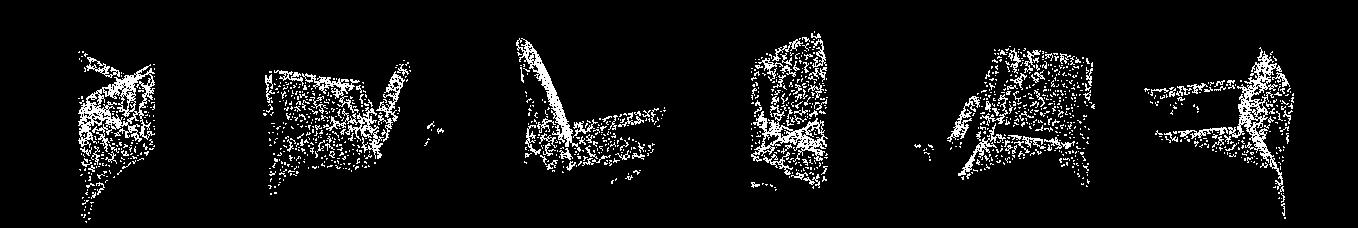} \\
&\includegraphics[width = 0.7\linewidth]{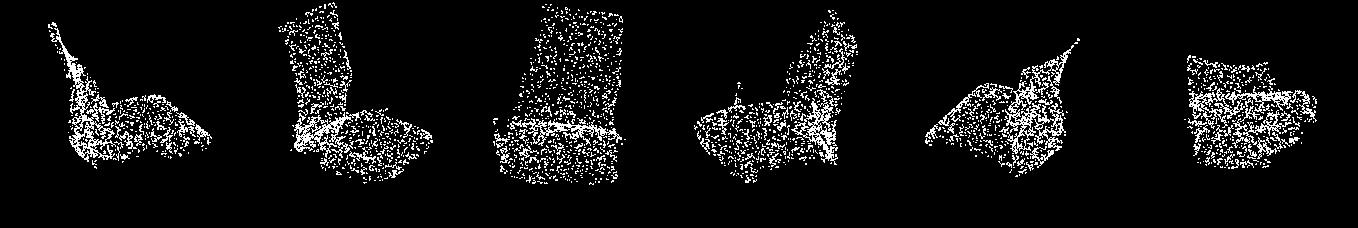} \\
\midrule
\multirow{3}{*}{\textbf{With Background}} &
\includegraphics[width = 0.7\linewidth]{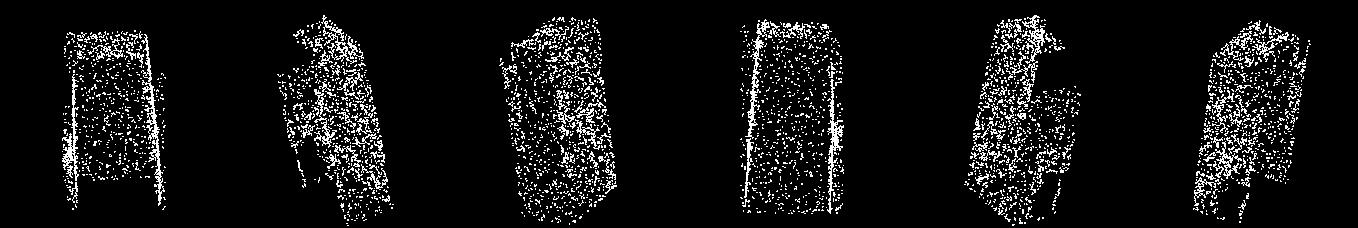} \\
&\includegraphics[width = 0.7\linewidth]{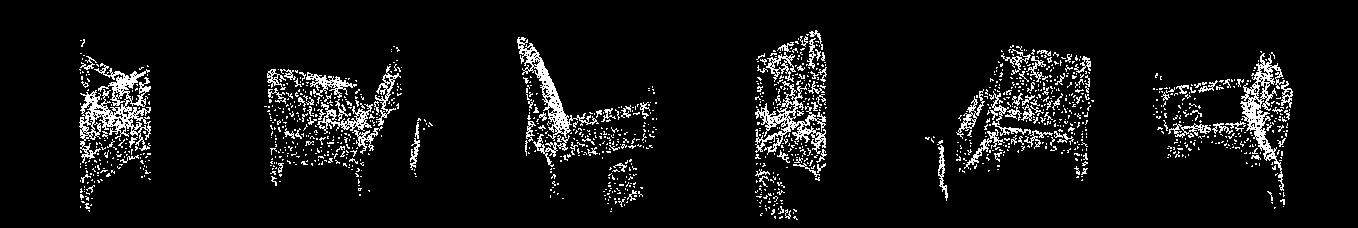} \\
&\includegraphics[width = 0.7\linewidth]{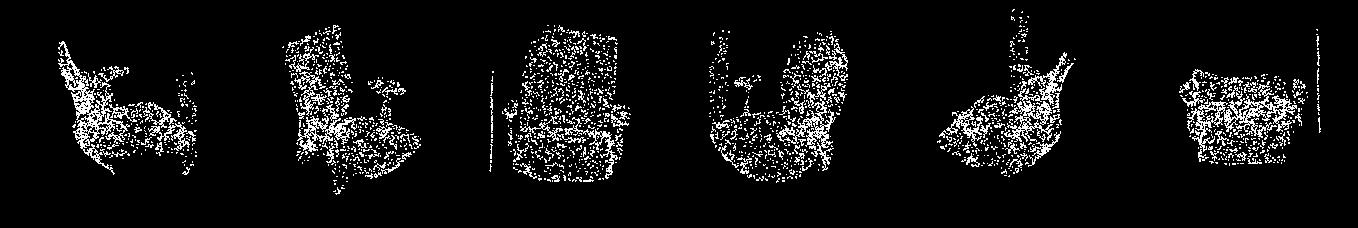} \\
\midrule
\multirow{3}{*}{\textbf{PB\_T50\_RS (Hardest)}} &
\includegraphics[width = 0.7\linewidth]{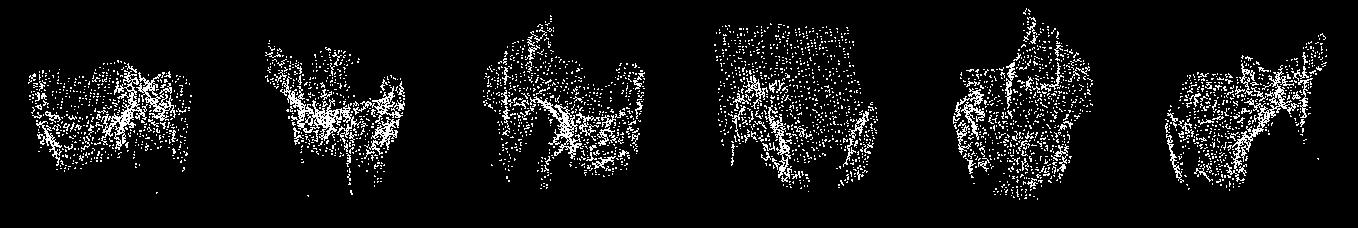} \\
&\includegraphics[width = 0.7\linewidth]{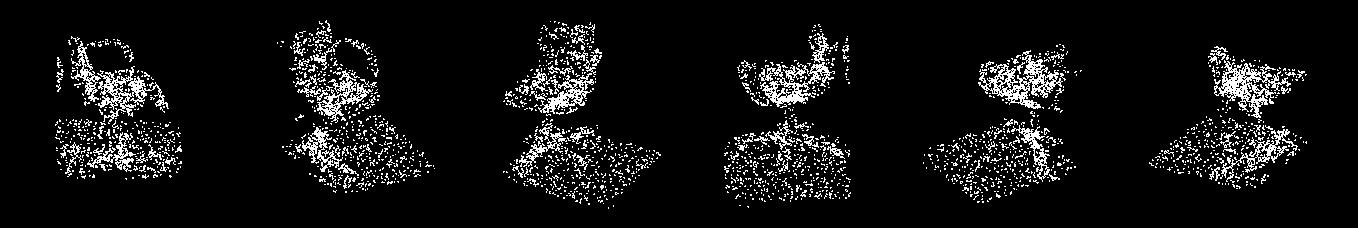} \\
&\includegraphics[width = 0.7\linewidth]{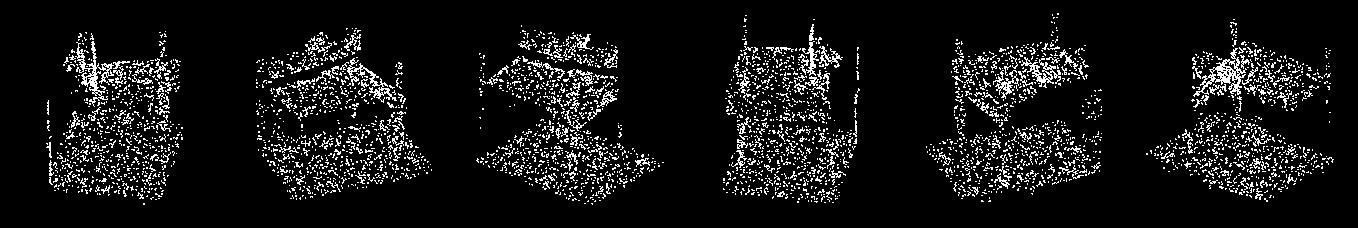} \\
\midrule

\end{tabular}
}
\vspace{2pt}
\caption{  \textbf{ScanObjectNN Variants.} We show examples of point cloud renderings of different variants of the ScanObjectNN \cite{scanobjectnn} point cloud dataset used to train MVTN. The variants are: object only, object with background, and the hardest perturbed variant (with rotation and translation). For this figure, six circular views are shown for each 3D shape.     
}
    \label{fig:scanobjectnn-sup}
\end{figure*}

\mysection{ScanObjectNN}
ScanObjectNN \cite{scanobjectnn} has three main variants: object only, object with background, and the PB\_T50\_RS variant (hardest perturbed variant). \figLabel{\ref{fig:scanobjectnn-sup}} show examples of multi-view renderings of different samples of the dataset from its three variants. Note that adding the background points to the rendering gives some clues to our MVTN about the object, which explains why adding background improves the performance of MVTN in Table \ref{tab:Scanobjectnn-supp}. 

\subsection{MVTN Details}

\mysection{MVTN Rendering}
 Point cloud rendering offers a light alternative to mesh rendering in ShapeNet because its meshes contain large numbers of faces that hinders training the MVTN pipeline. Simplifying theses "high-poly" meshes (similar to ModelNet40) results in corrupted shapes that lose their main visual clues. Therefore, we use point cloud rendering for ShapeNet, allowing to process all shapes with equal memory requirements. Another benefit of point cloud rendering is making it possible to train MVTN with a large batch size on the same GPU (bath size of 30 on V100 GPU).    
 
\begin{figure*}
\centering
        \includegraphics[width=0.75\linewidth]{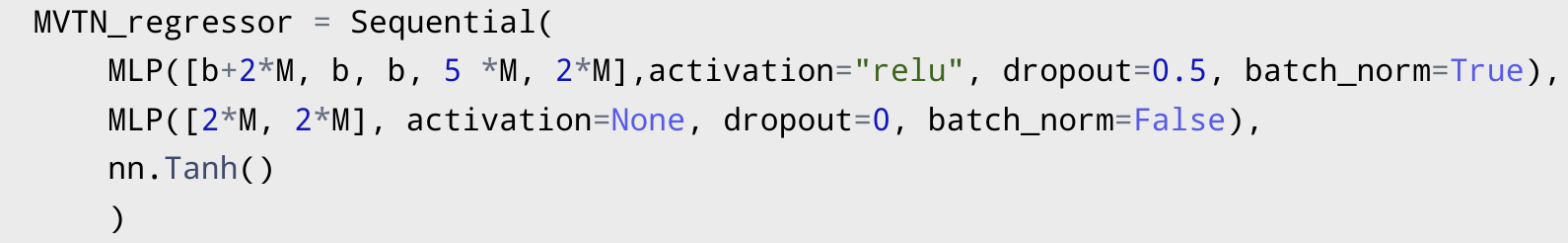} \vspace{8pt} \\
\quad $b+2\times M$ \quad \quad  $b$\quad \quad\quad \quad  $b$\quad \quad\quad \quad $5\times M$ \quad\quad  $2\times M$ \quad \quad  $2\times M$ \\
    \includegraphics[trim= 0 0 0 2cm , clip,width=0.75\linewidth]{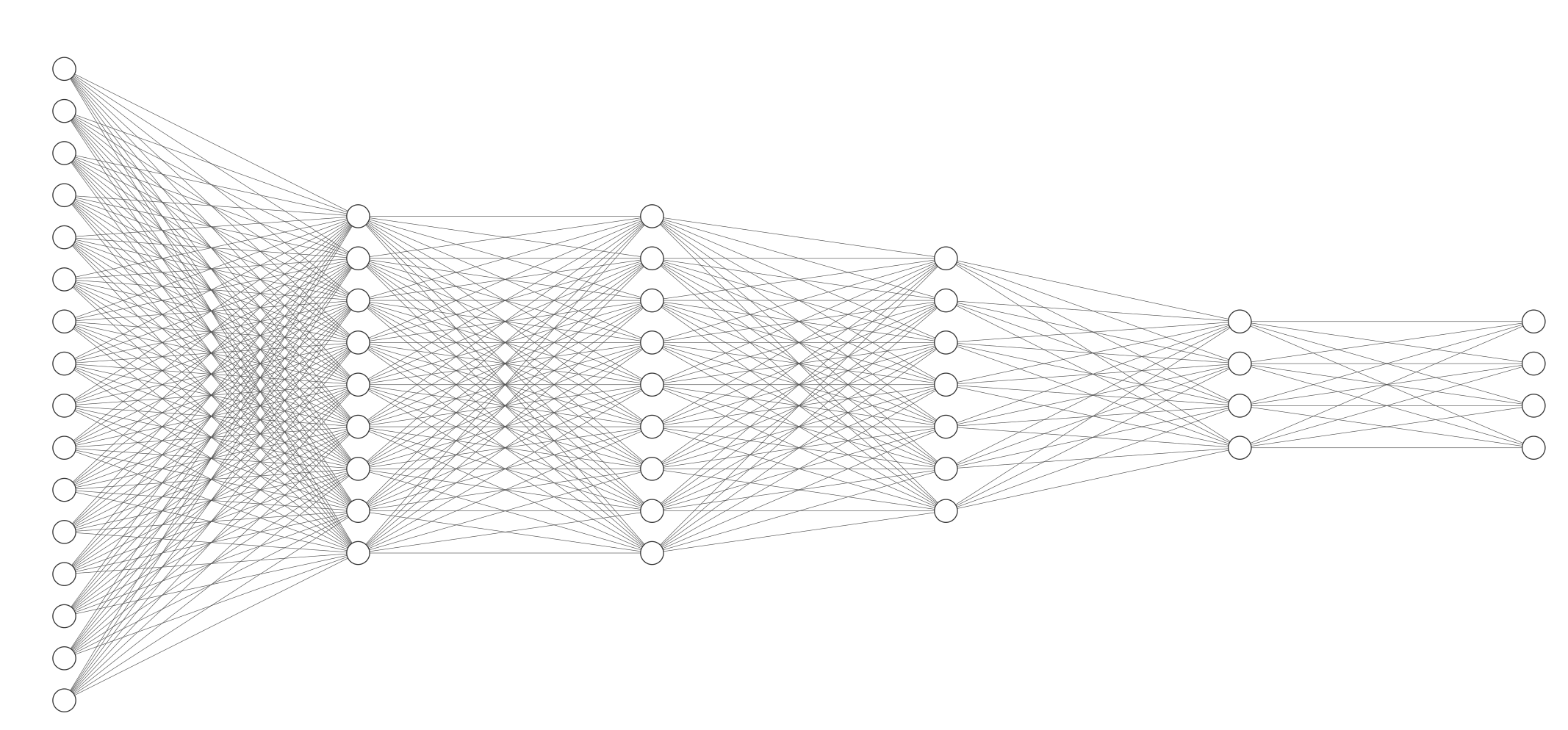} \\
    \caption{\textbf{MVTN Network Architecture.} We show a schematic and a code snippet for MVTN-spherical/MVTN-circular regression architectures used, where $b$ is the size of the point features extracted by the point encoder of MVTN and $M$ is the number of views learned. In most of our experiments, $b=40$, while the output is the azimuth and elevation angles for all the $M$ views used. The network is drawn using \cite{nndraw}}
    \label{fig:architecture-sup}
\end{figure*}

\mysection{MVTN Architecture}
We incorporate our MVTN into MVCNN \cite{mvcnn} and ViewGCN \cite{mvviewgcn}. 
In our experiments, we select PointNet \cite{pointnet} as the default point encoder of MVTN.
All MVTNs and their baseline multi-view networks use ResNet18 \cite{resnet} as backbone in our main experiments with output feature size $d=1024$. The azimuth angle maximum range ($\mathbf{u}_{\text{bound}}$) is $\frac{180^\circ}{M}$  for MVTN-circular and MVTN-spherical, while it is $180^\circ$ for MVTN-direct. On the other hand, the elevation angle maximum range ($\mathbf{u}_{\text{bound}}$) is $90^\circ$.
We use a 4-layer MLP for MVTN's regression network $\mathbf{G}$. For MVTN-spherical/MVTN-spherical, the regression network takes as input $M$ azimuth angles, $M$ elevation angles, and the point features of shape $\mathbf{S}$ of size $b=40$. The widths of the MVTN networks are illustrated in \figLabel{\ref{fig:architecture-sup}}.MVTN concatenates all of its inputs, and the MLP outputs the offsets to the initial $2\times M$ azimuth and elevation angles. The size of the MVTN network (with $b=40$) is $14 M^2 + 211 M + 3320$ parameters, where $M$ is the number of views. It is a shallow network of only around 9K parameters when $M=12$.  

\begin{figure*} [h] 
\tabcolsep=0.03cm
\resizebox{0.95\linewidth}{!}{
\begin{tabular}{c|ccccccc}
 & \textbf{1 view} & \textbf{2 views}& \textbf{4 views}& \textbf{6 views}& \textbf{8 views}& \textbf{10 views}& \textbf{12 views} \\
\textbf{(a)} & 
\includegraphics[trim= 4cm 2.7cm 4cm 2.2cm , clip, width = 0.135\linewidth]{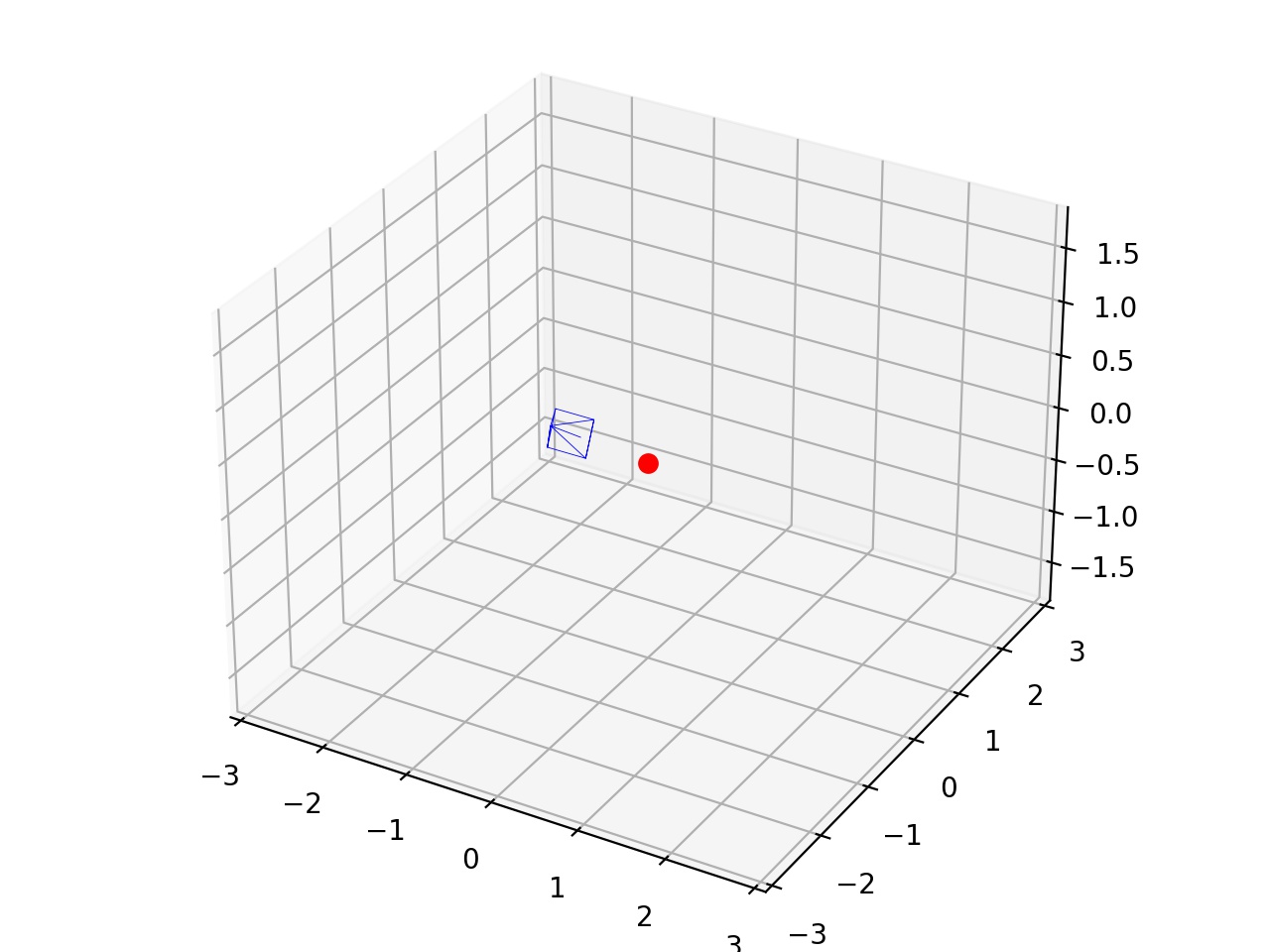} &
\includegraphics[trim= 4cm 2.7cm 4cm 2.2cm , clip, width = 0.135\linewidth]{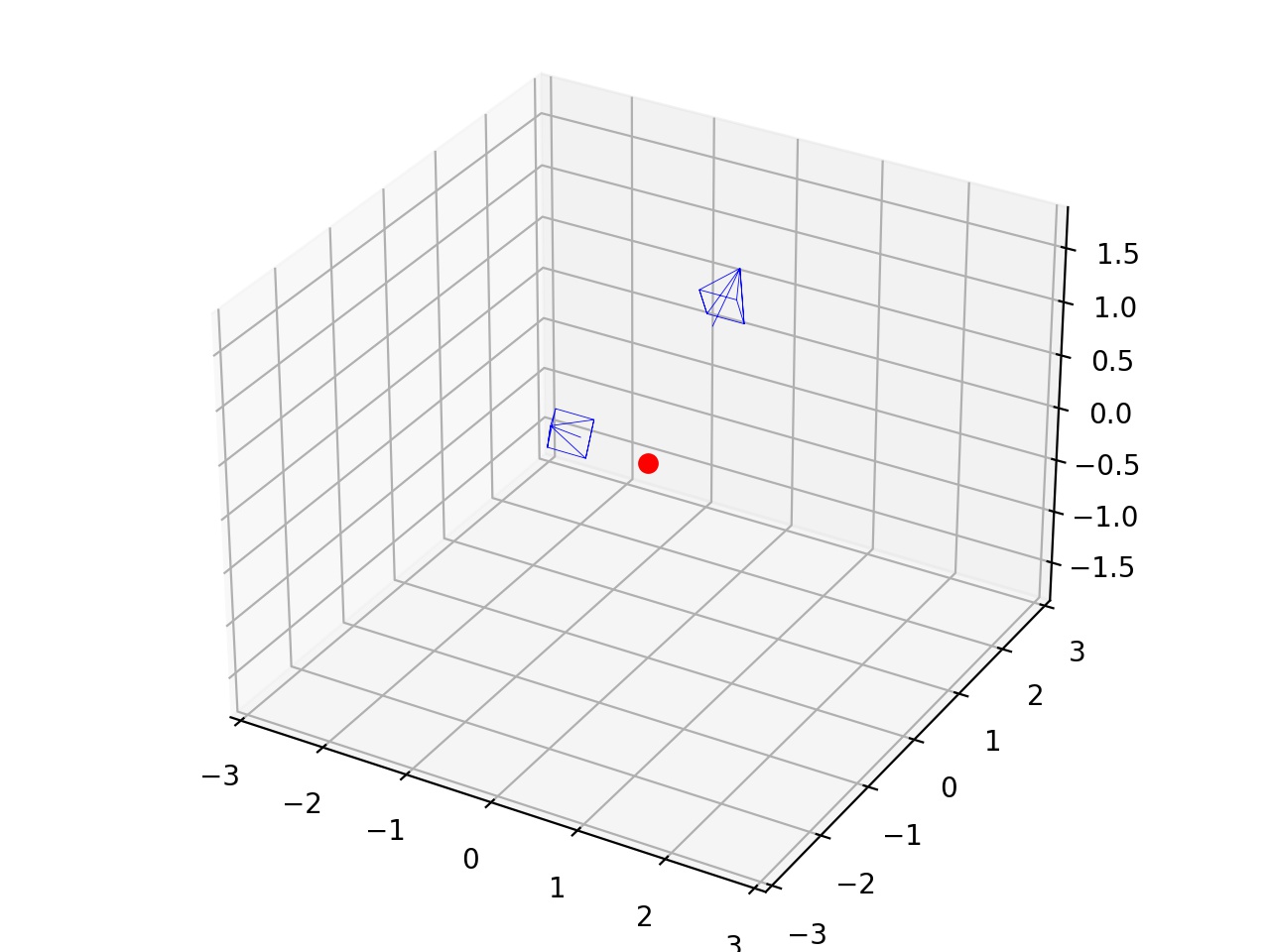} &
\includegraphics[trim= 4cm 2.7cm 4cm 2.2cm , clip, width = 0.135\linewidth]{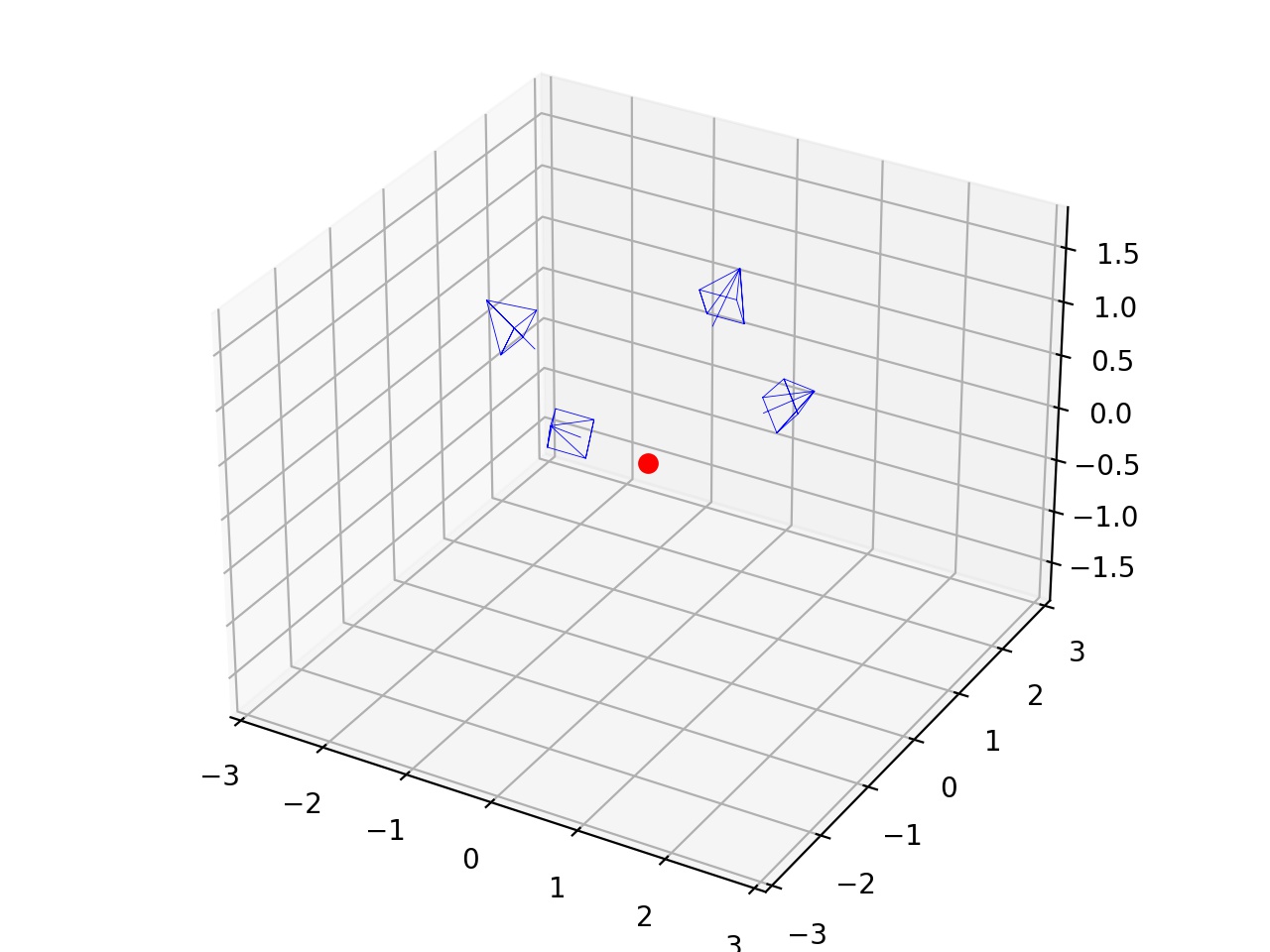} &
\includegraphics[trim= 4cm 2.7cm 4cm 2.2cm , clip, width = 0.135\linewidth]{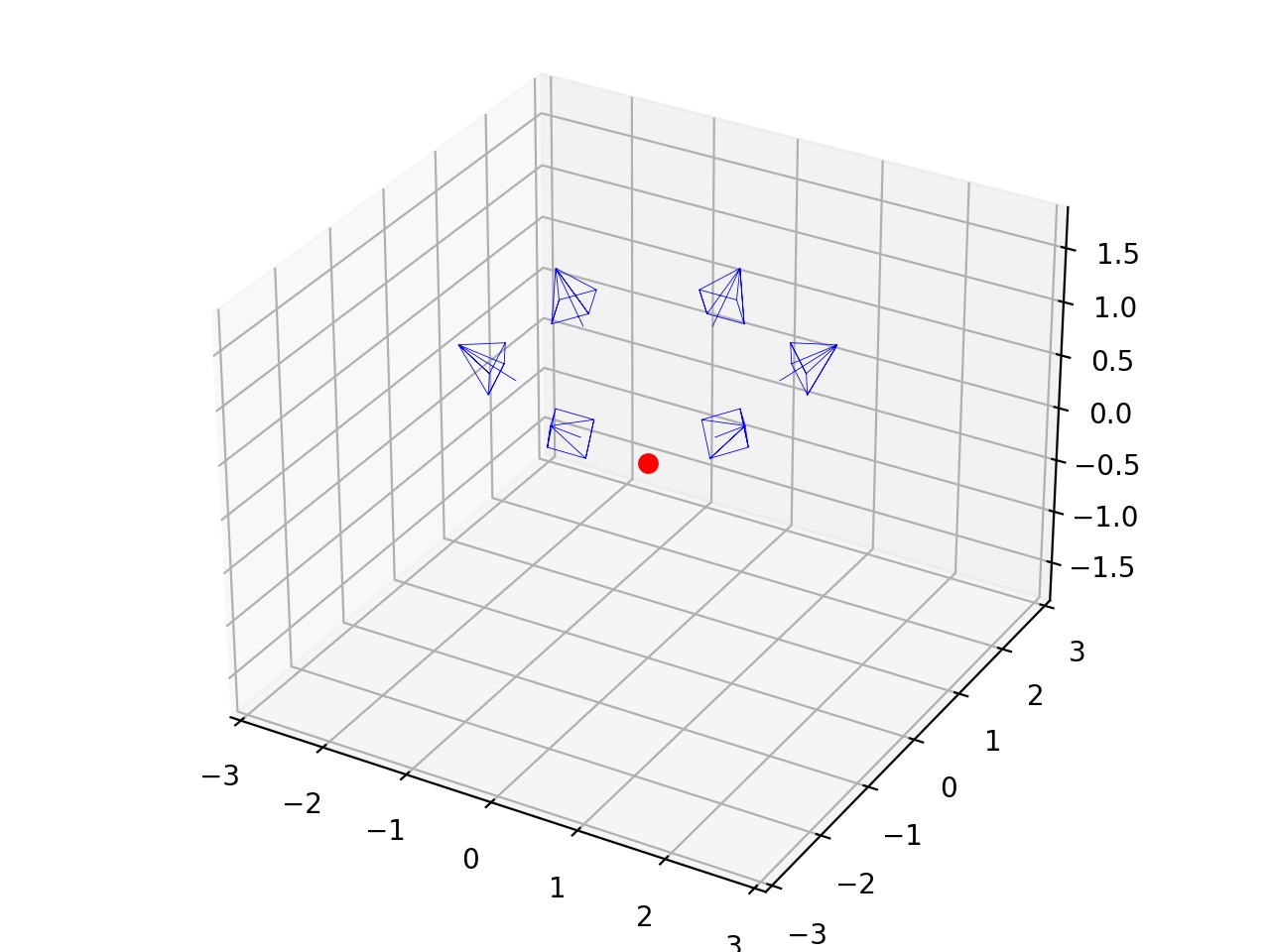} &
\includegraphics[trim= 4cm 2.7cm 4cm 2.2cm , clip, width = 0.135\linewidth]{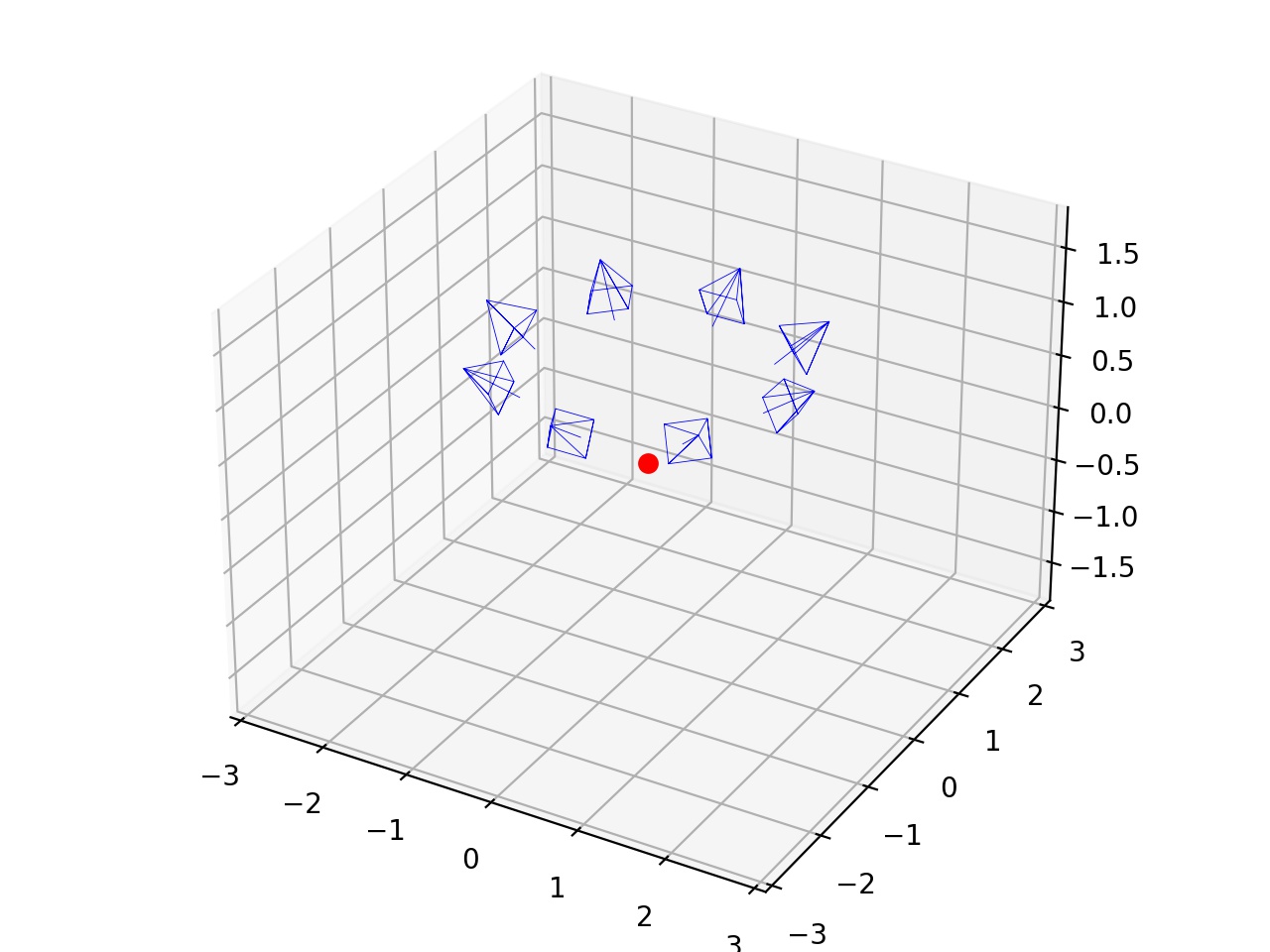} &
\includegraphics[trim= 4cm 2.7cm 4cm 2.2cm , clip, width = 0.135\linewidth]{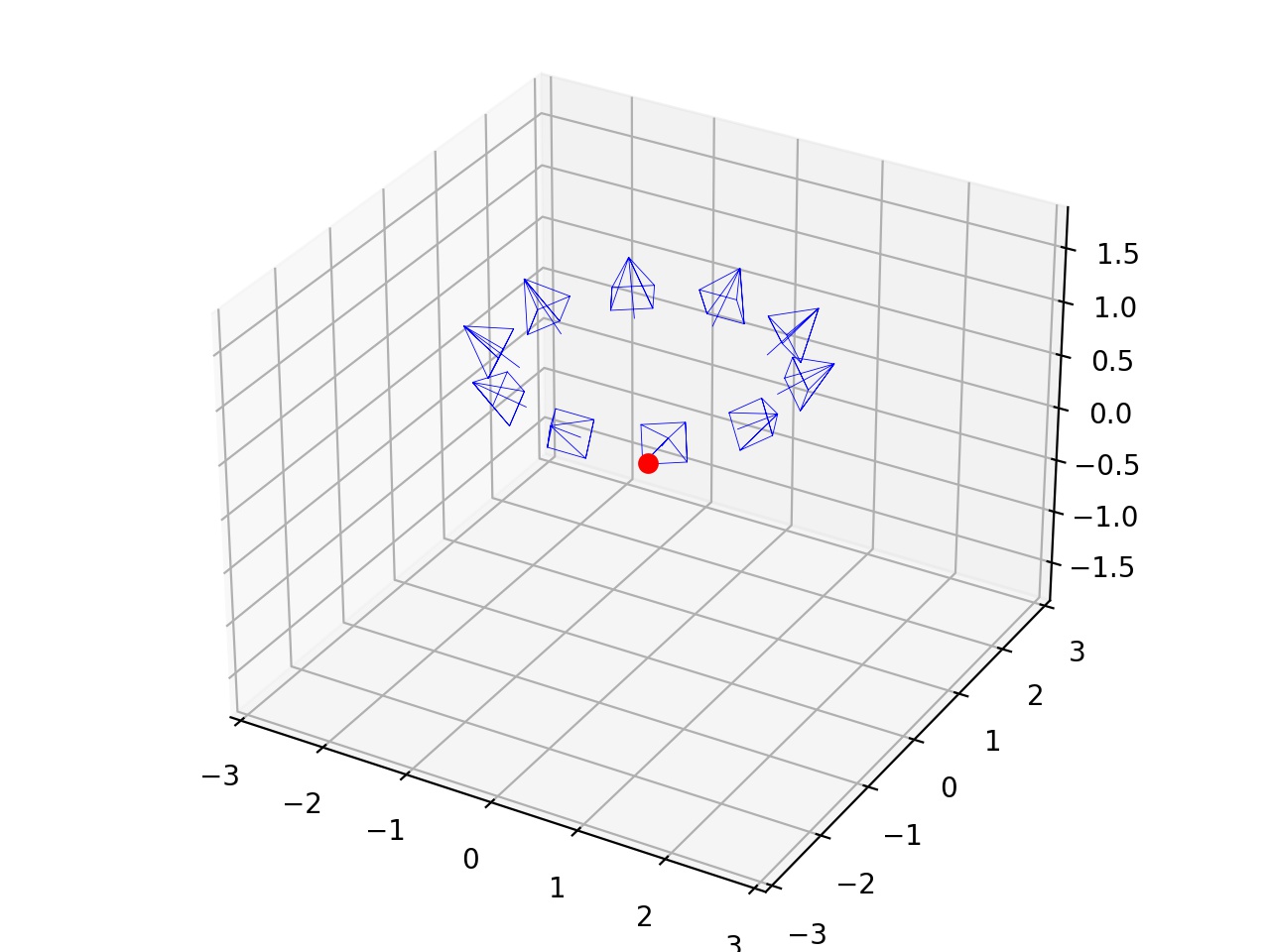} &
\includegraphics[trim= 4cm 2.7cm 4cm 2.2cm , clip, width = 0.135\linewidth]{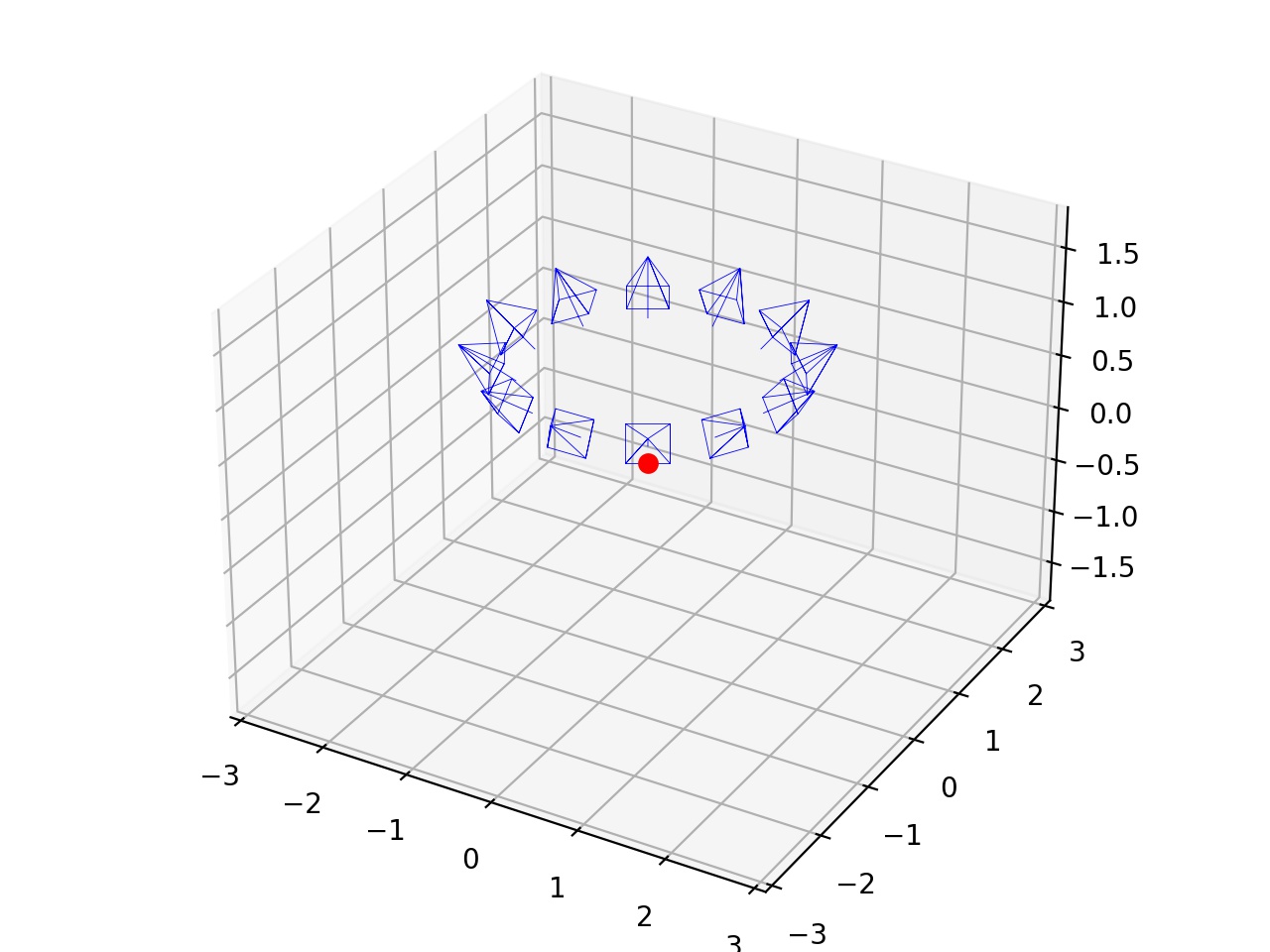} \\
\textbf{(b)} & 
\includegraphics[trim= 4cm 2.7cm 4cm 2.2cm , clip, width = 0.135\linewidth]{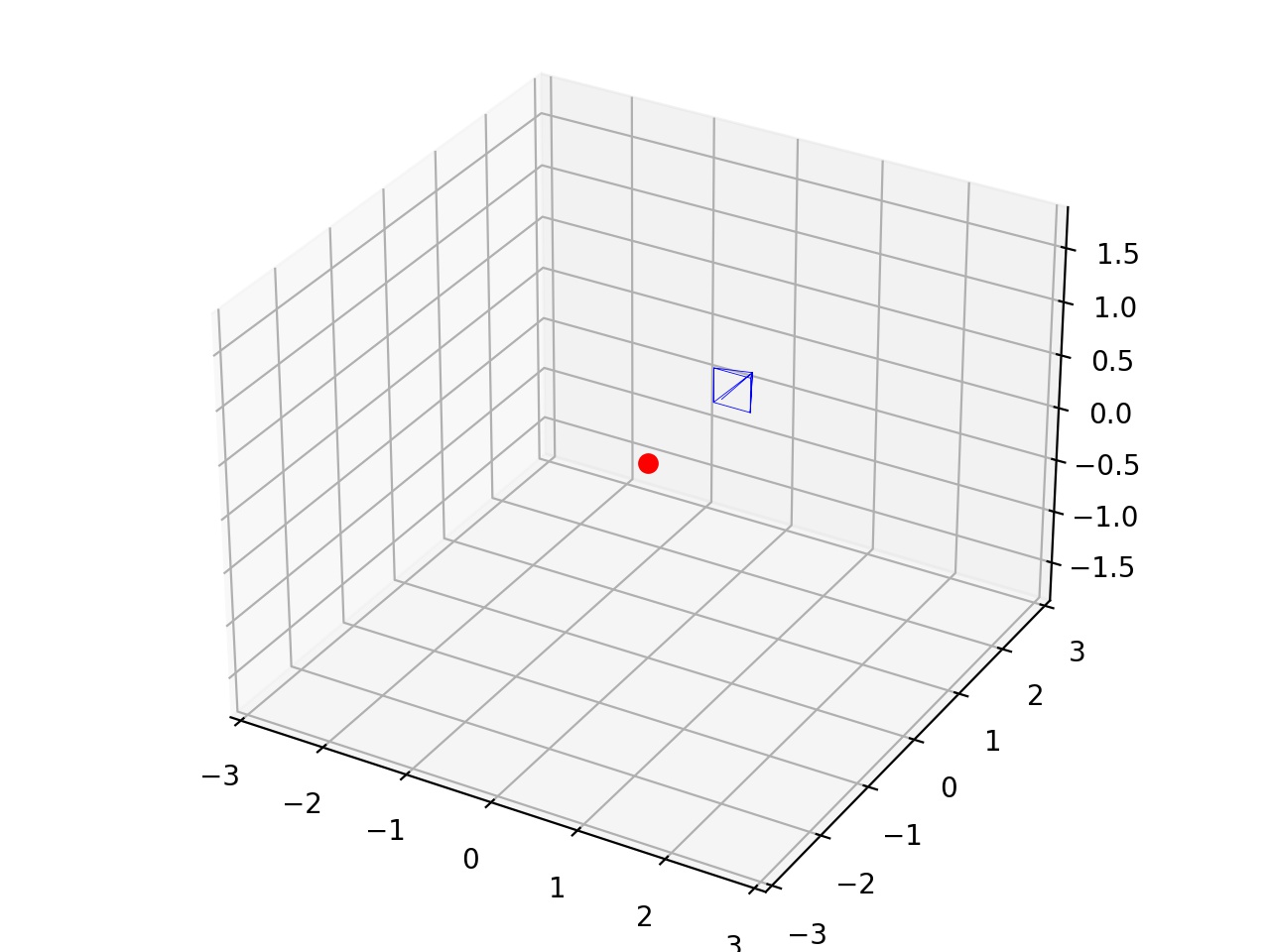} &
\includegraphics[trim= 4cm 2.7cm 4cm 2.2cm , clip, width = 0.135\linewidth]{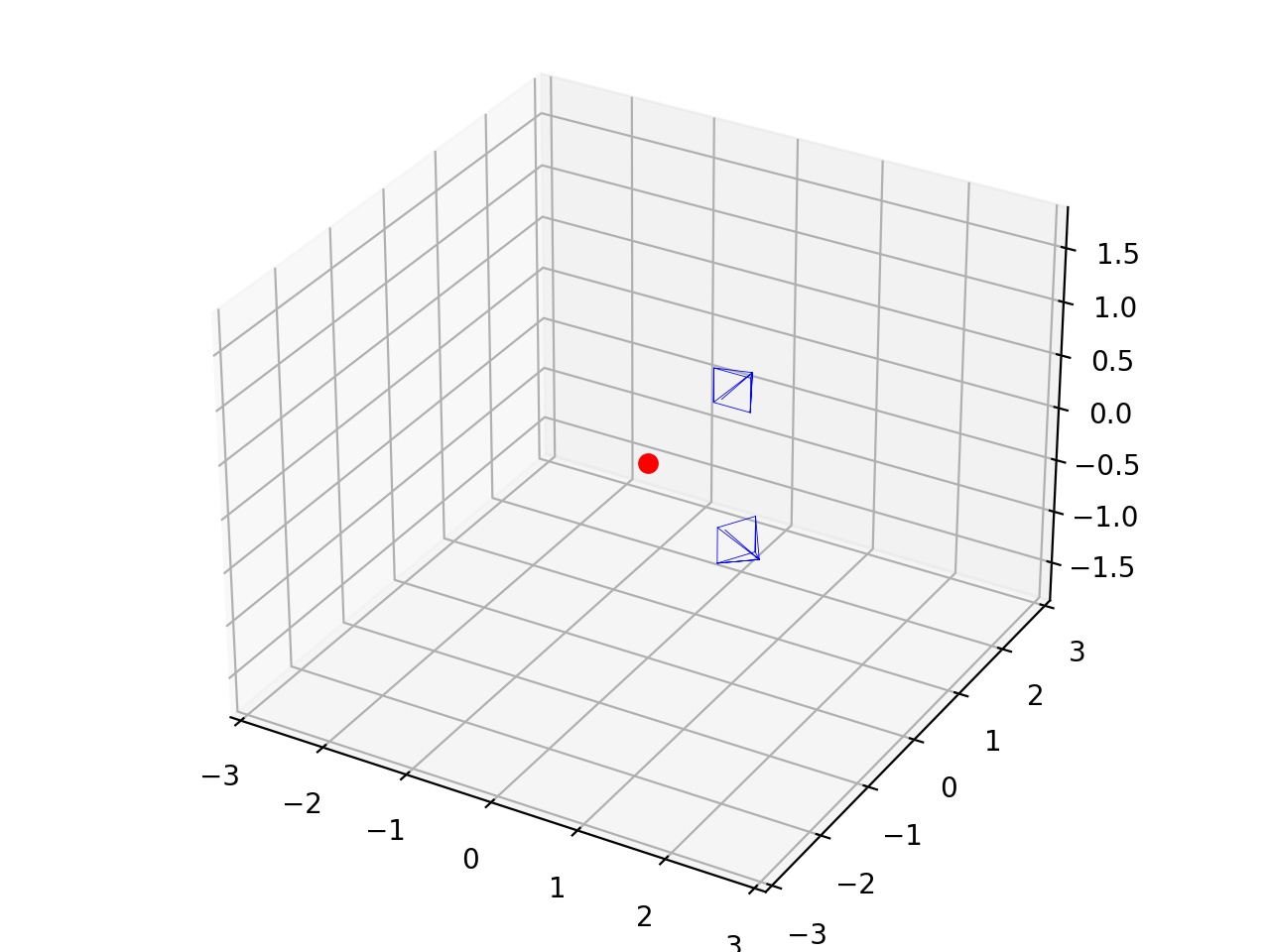} &
\includegraphics[trim= 4cm 2.7cm 4cm 2.2cm , clip, width = 0.135\linewidth]{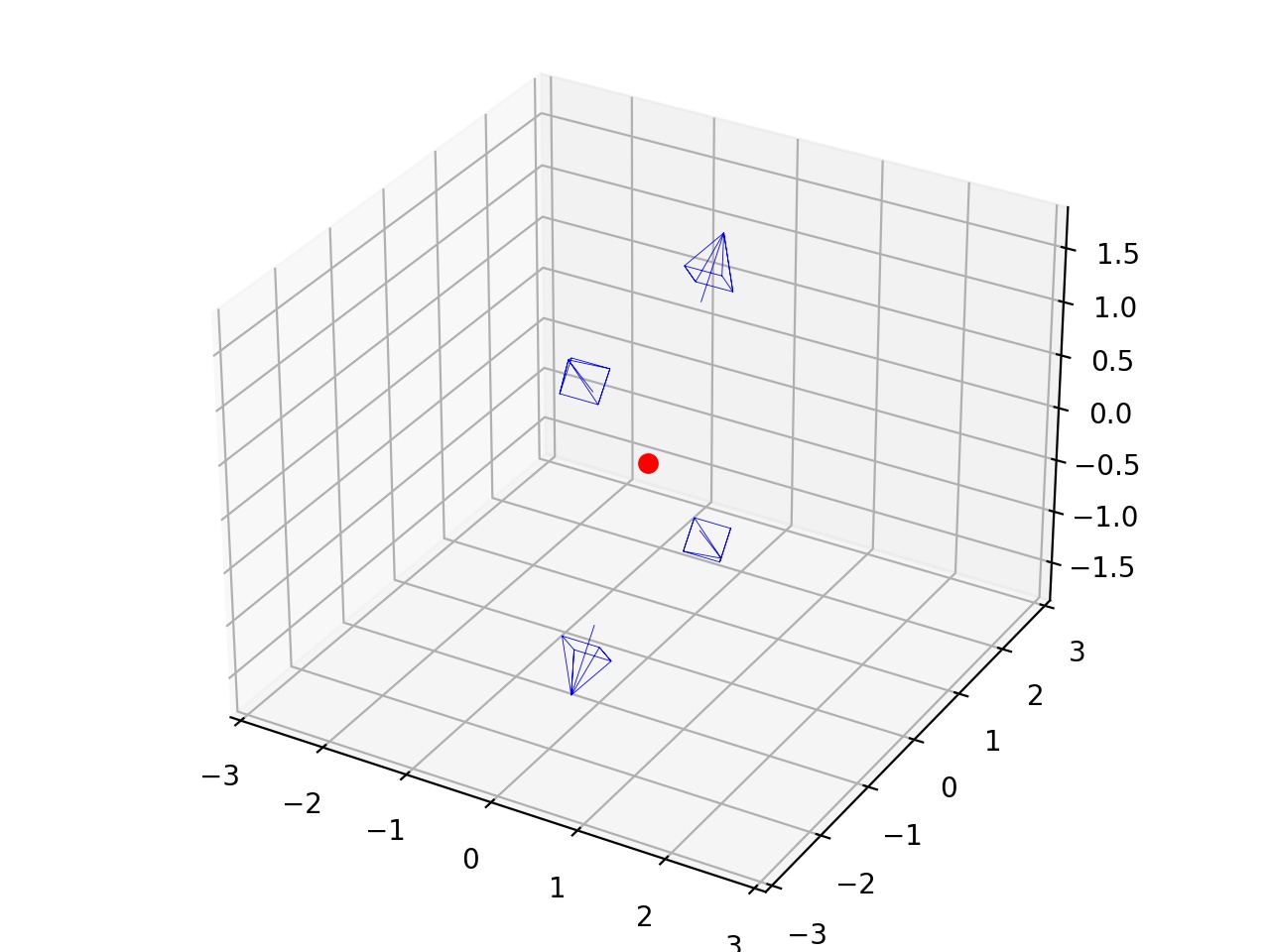} &
\includegraphics[trim= 4cm 2.7cm 4cm 2.2cm , clip, width = 0.135\linewidth]{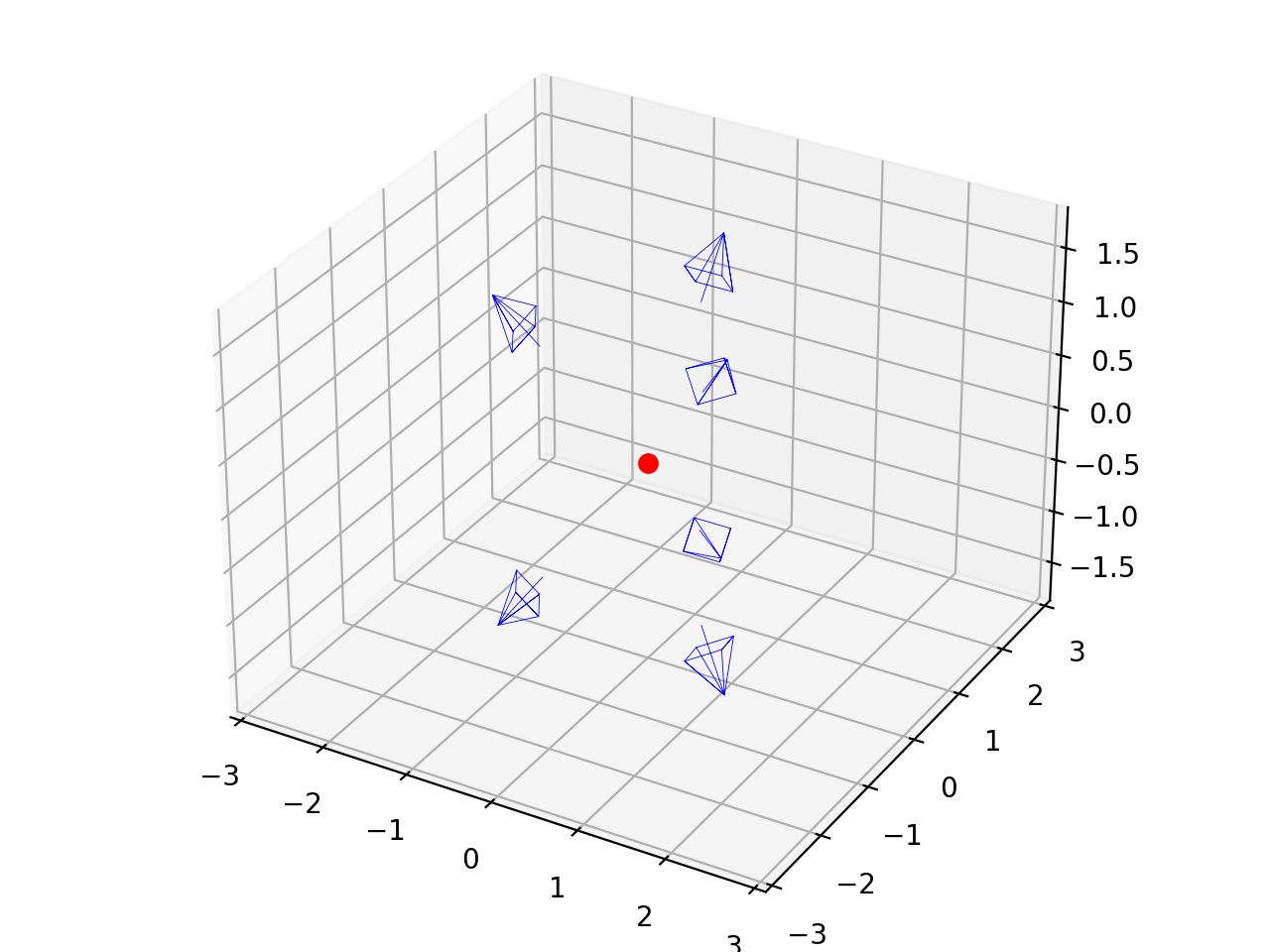} &
\includegraphics[trim= 4cm 2.7cm 4cm 2.2cm , clip, width = 0.135\linewidth]{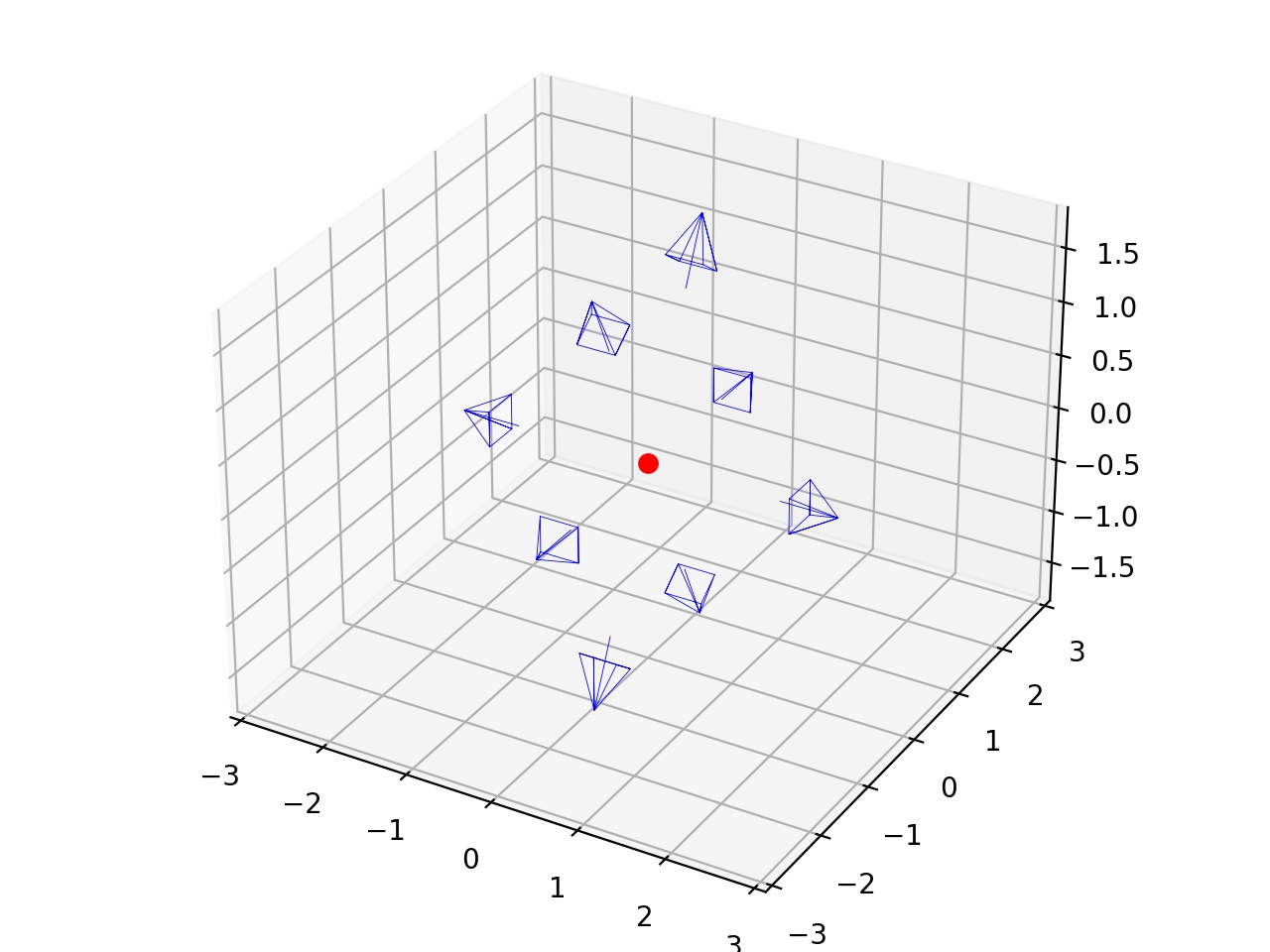} &
\includegraphics[trim= 4cm 2.7cm 4cm 2.2cm , clip, width = 0.135\linewidth]{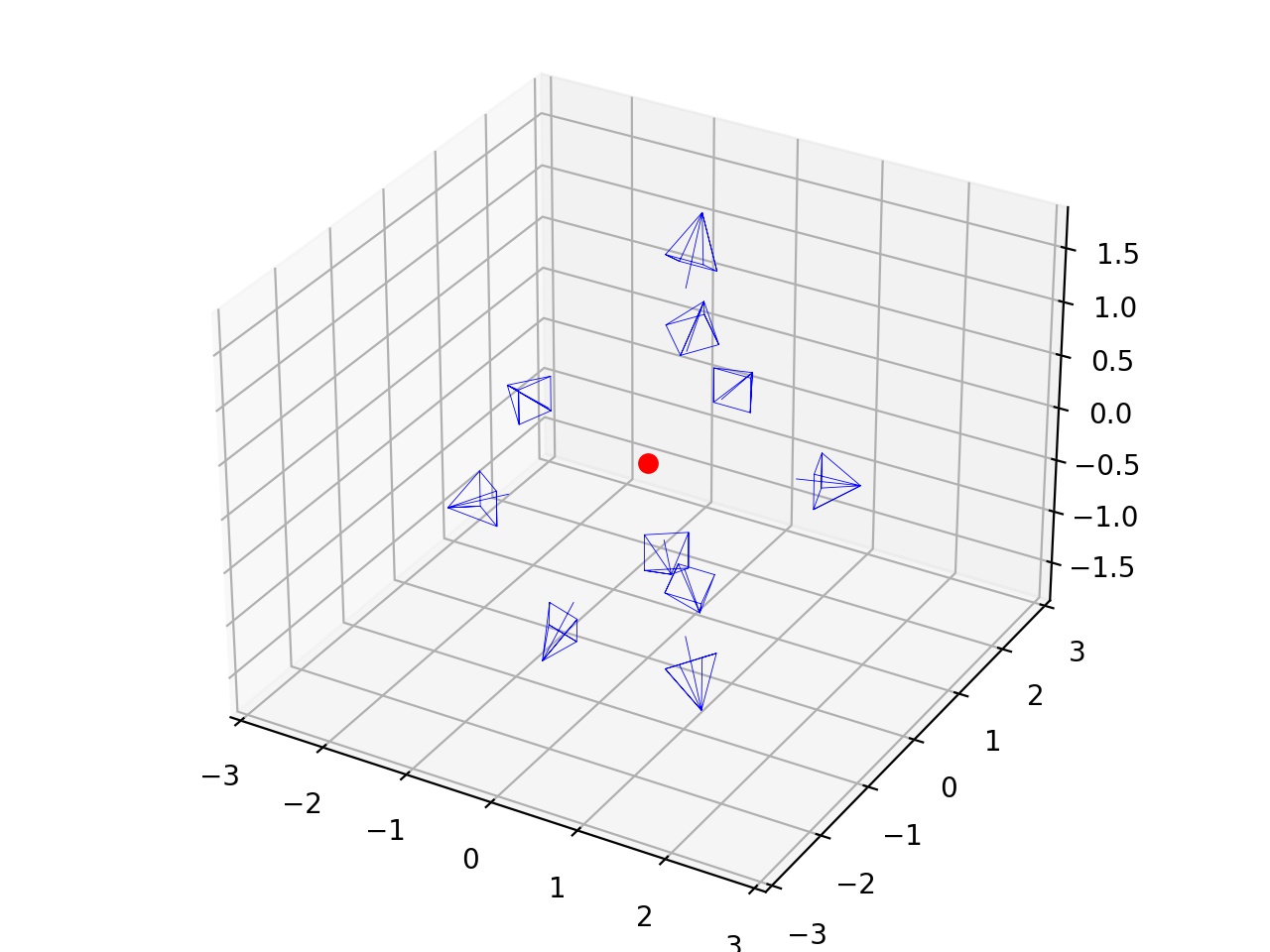} &
\includegraphics[trim= 4cm 2.7cm 4cm 2.2cm , clip, width = 0.135\linewidth]{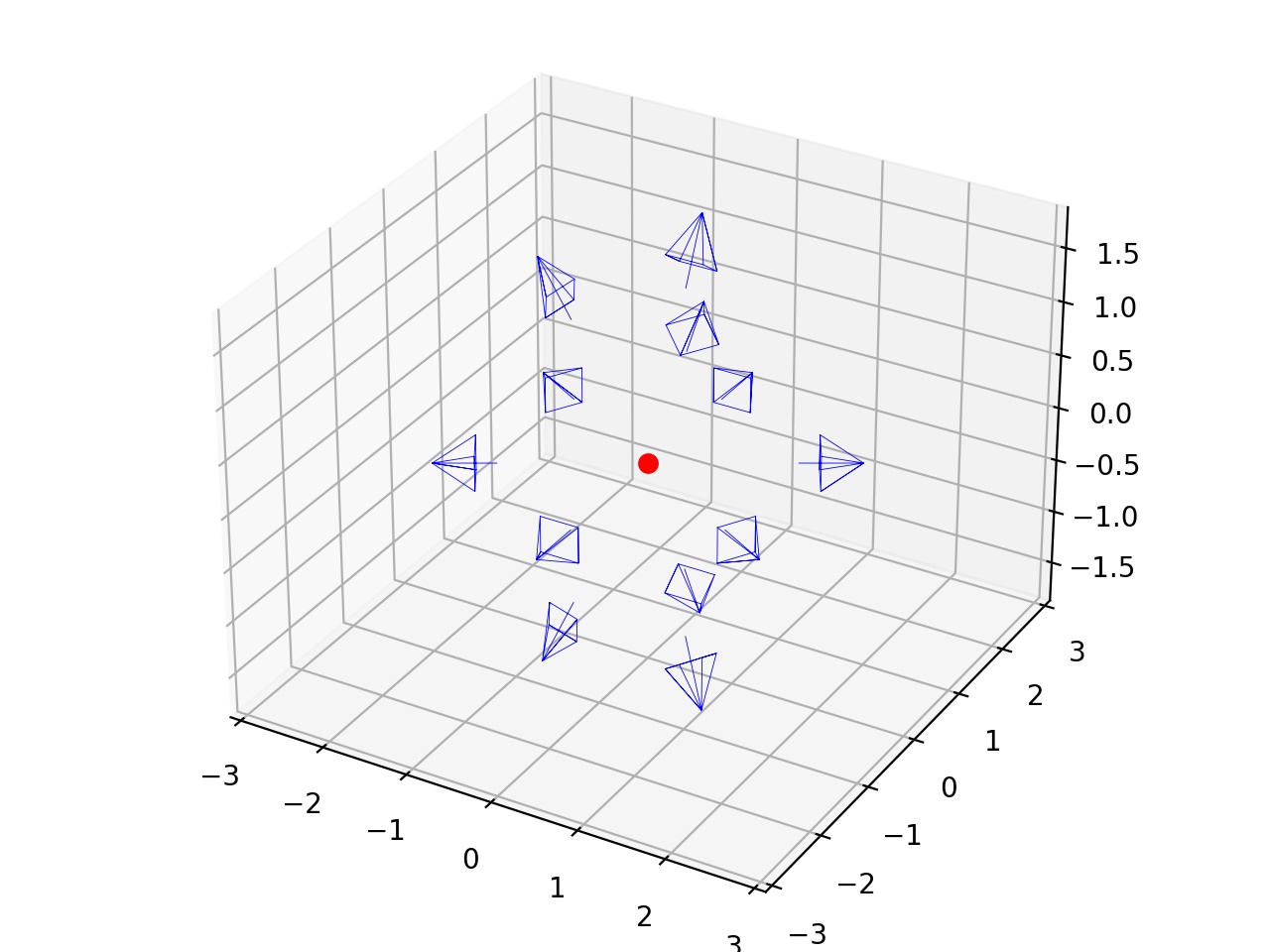} \\
\textbf{(c)} & 
\includegraphics[trim= 4cm 2.7cm 4cm 2.2cm , clip, width = 0.135\linewidth]{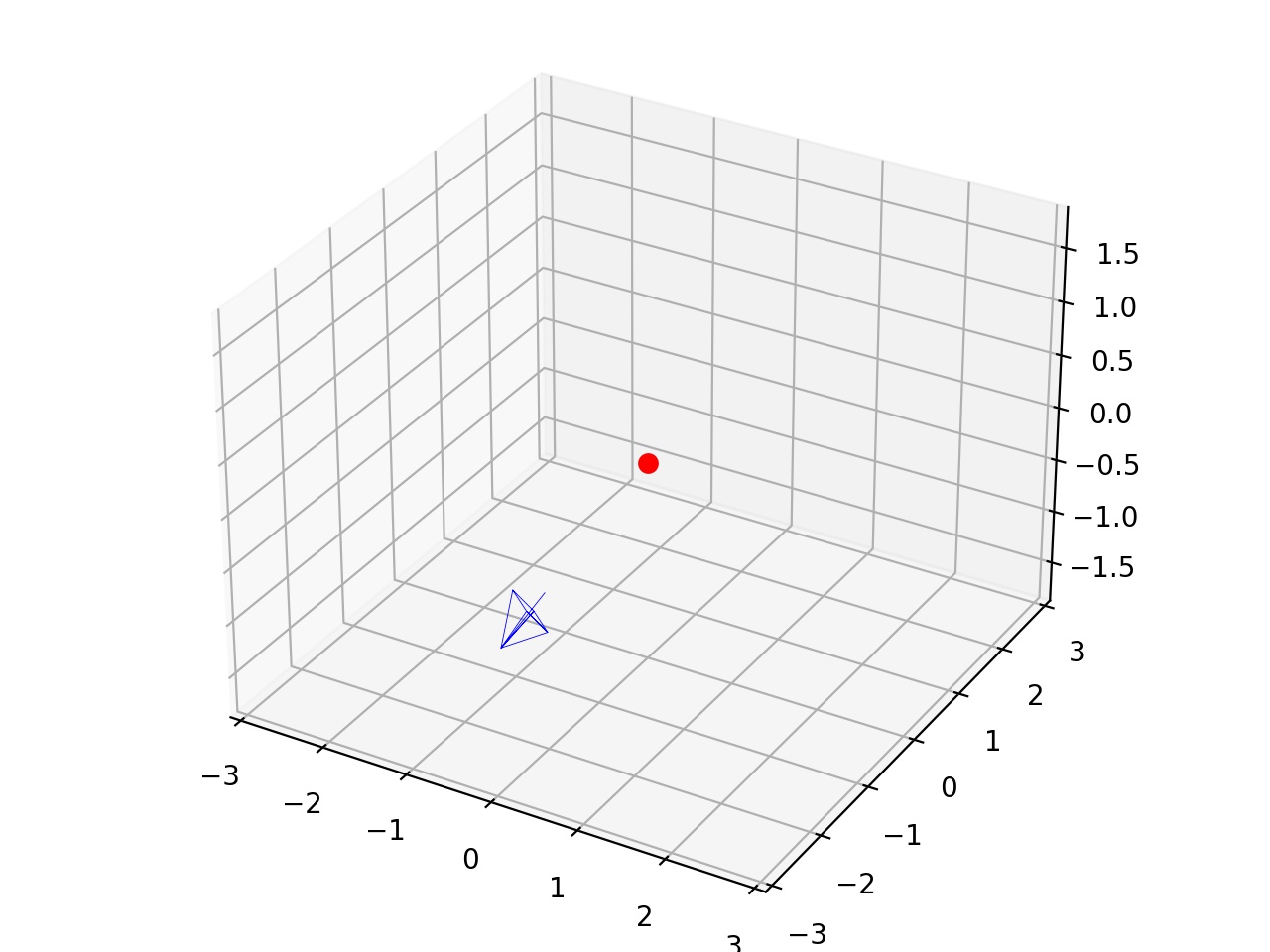} &
\includegraphics[trim= 4cm 2.7cm 4cm 2.2cm , clip, width = 0.135\linewidth]{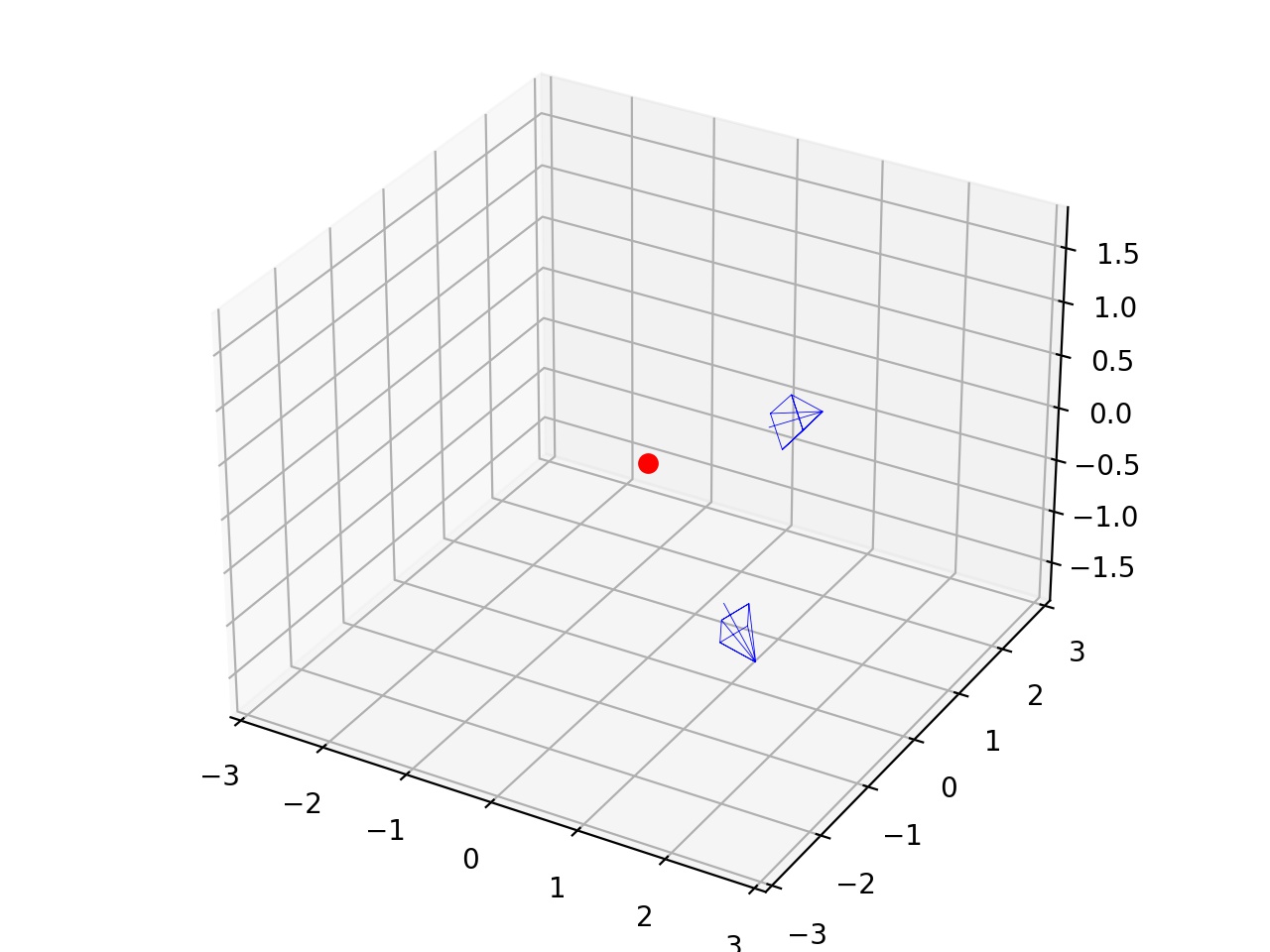} &
\includegraphics[trim= 4cm 2.7cm 4cm 2.2cm , clip, width = 0.135\linewidth]{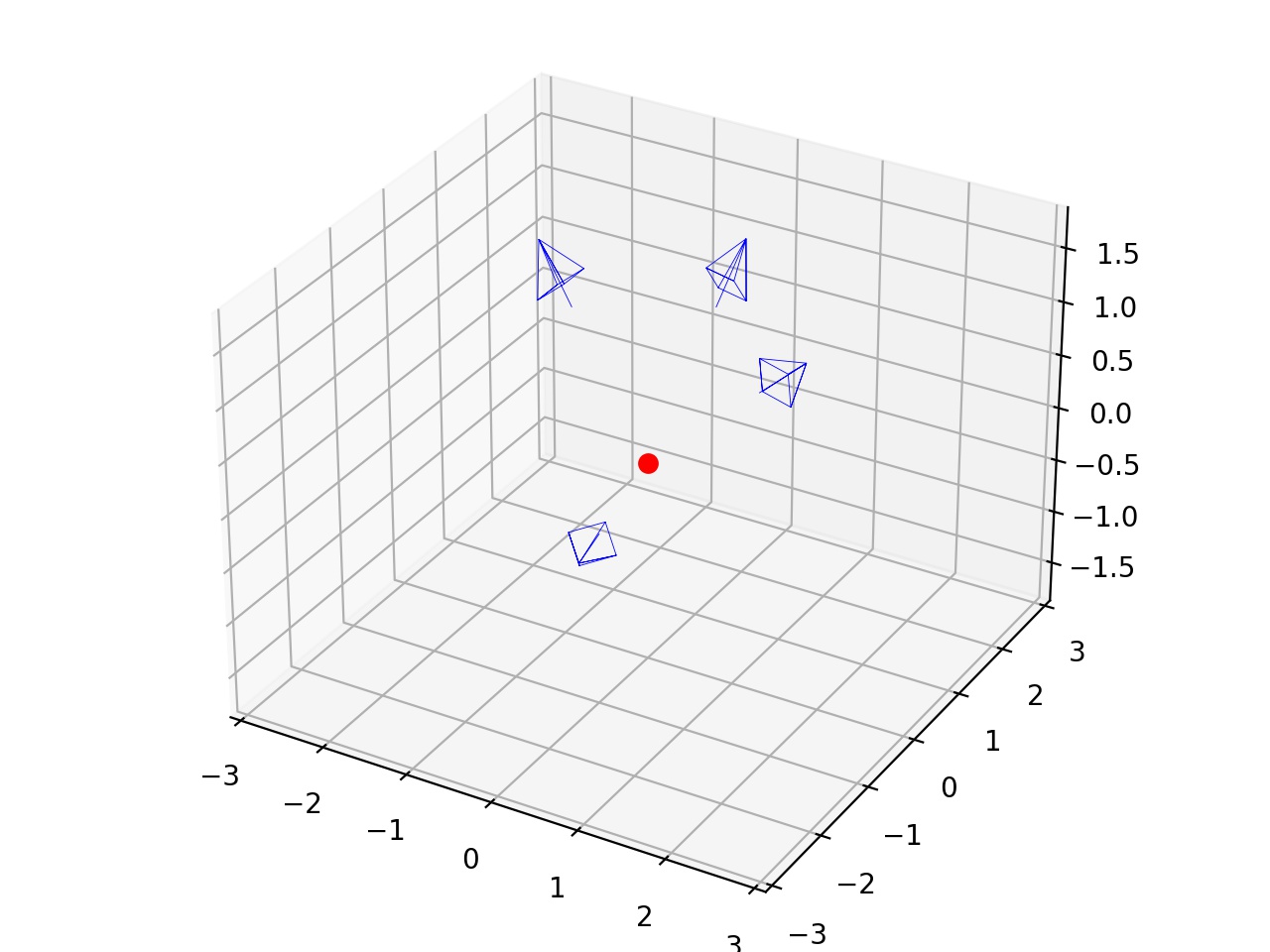} &
\includegraphics[trim= 4cm 2.7cm 4cm 2.2cm , clip, width = 0.135\linewidth]{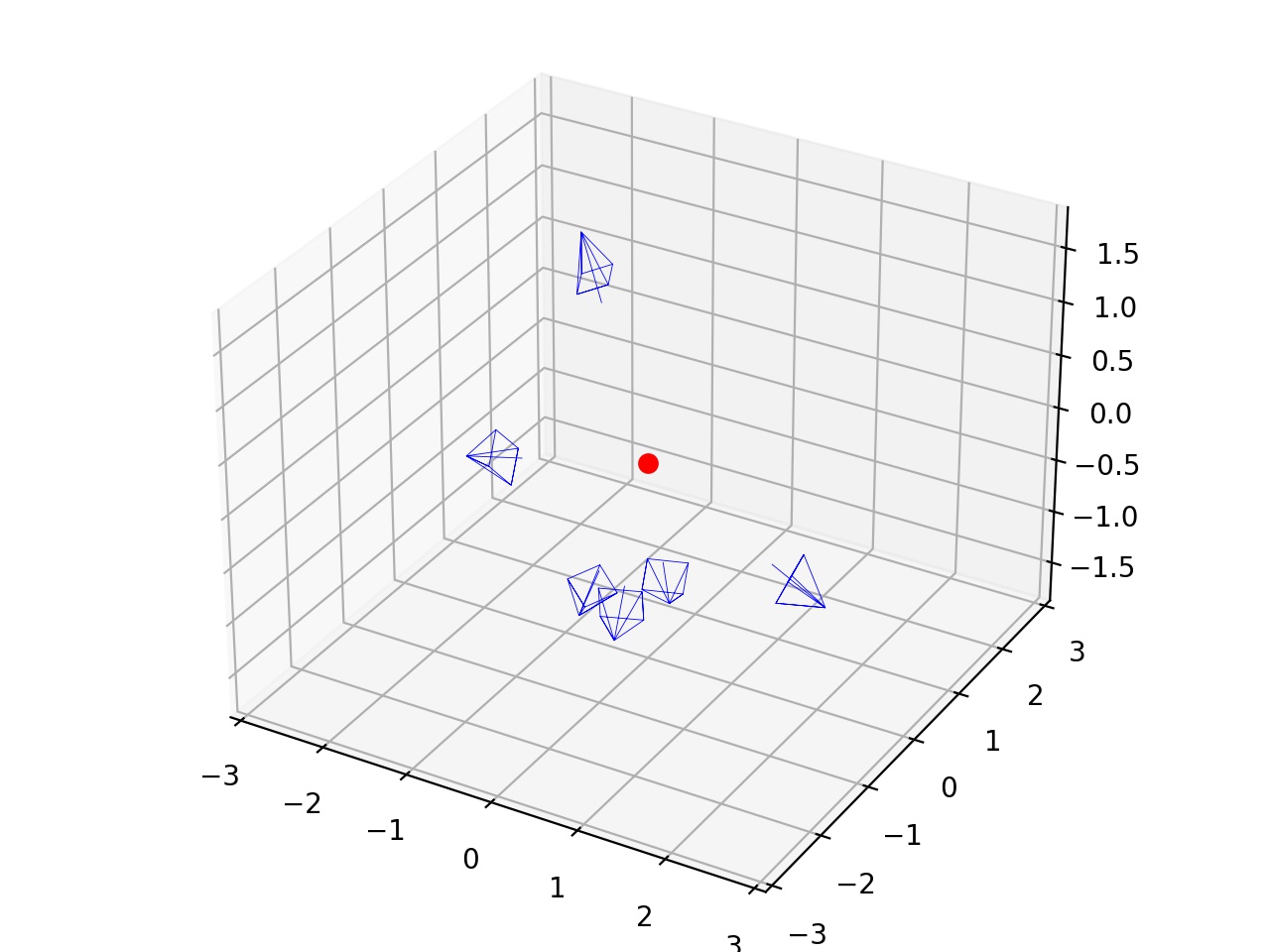} &
\includegraphics[trim= 4cm 2.7cm 4cm 2.2cm , clip, width = 0.135\linewidth]{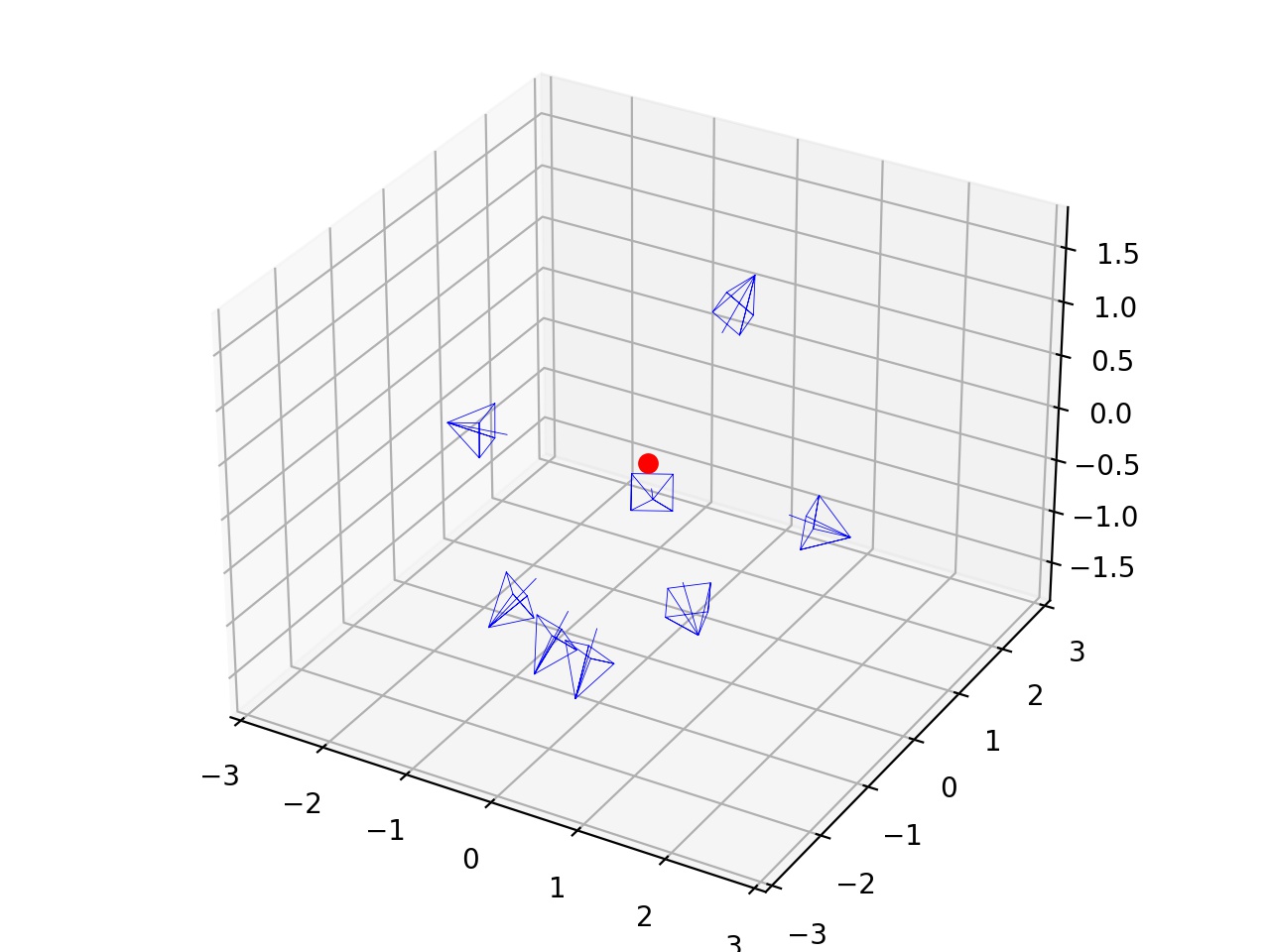} &
\includegraphics[trim= 4cm 2.7cm 4cm 2.2cm , clip, width = 0.135\linewidth]{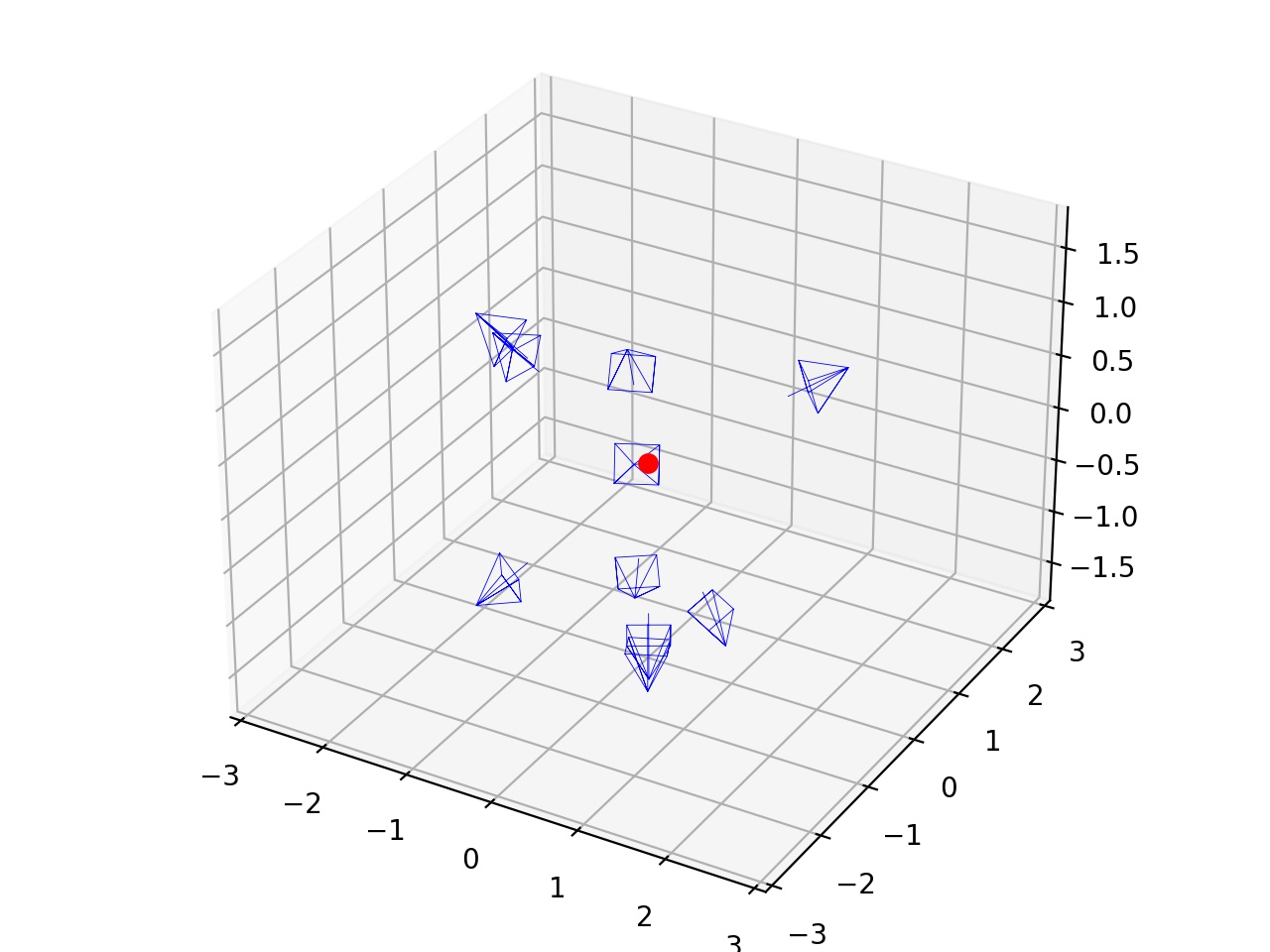} &
\includegraphics[trim= 4cm 2.7cm 4cm 2.2cm , clip, width = 0.135\linewidth]{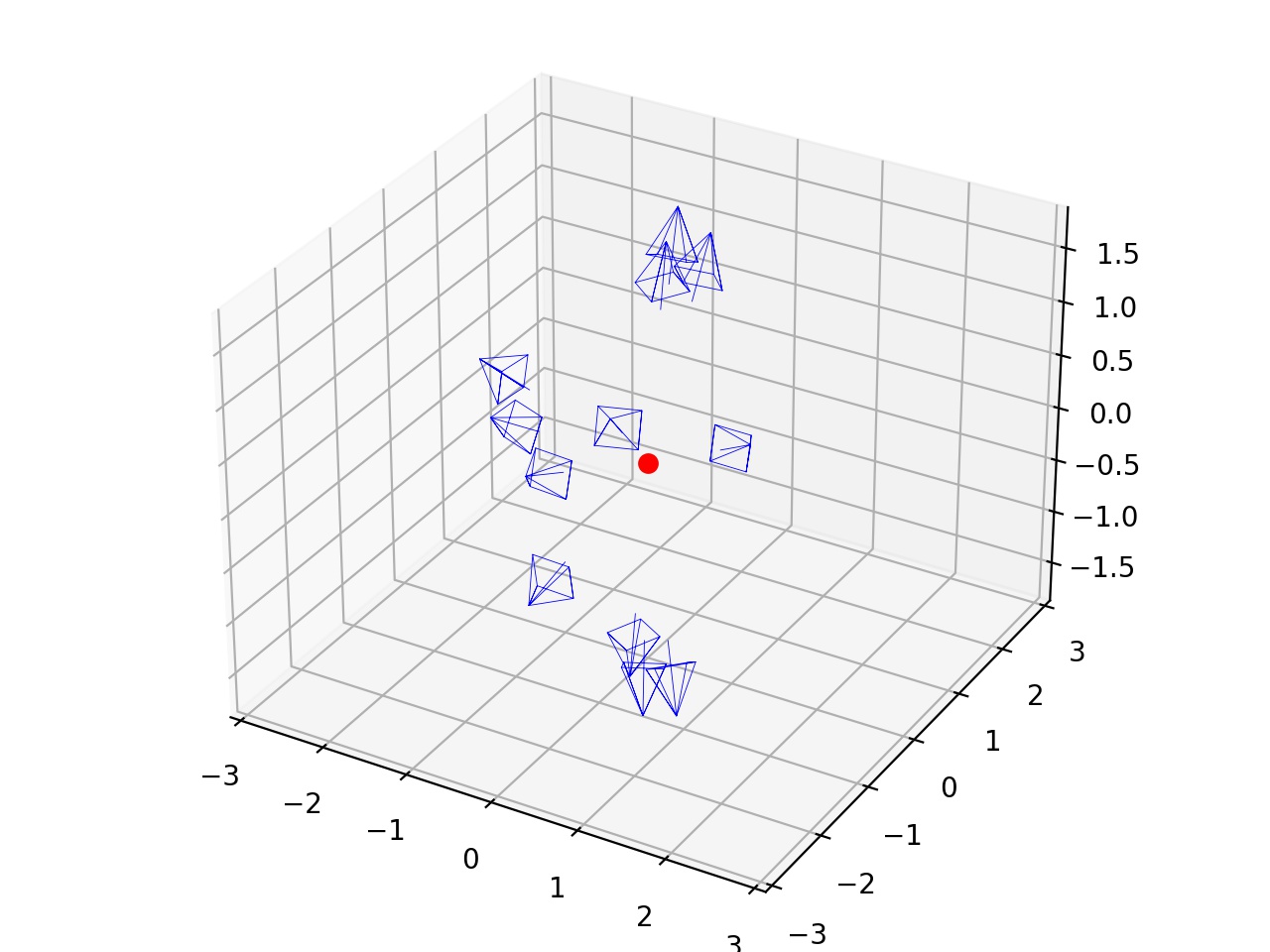} 
\end{tabular}
}
\vspace{2pt}
\caption{  \textbf{Views Configurations.} We show some possible views configurations that can be used with a varying number of views. \textbf{(a):} circular, \textbf{(b):} spherical, \textbf{(c):} random 
}
    \label{fig:views-sup}
\end{figure*}

\mysection{View-Points}
In \figLabel{\ref{fig:views-sup}}, we show the basic views configurations for $M$ views previously used in the literature: circular, spherical, and random. MVTN's learned views are shown later in \specialcell{\ref{sec:ablation-supp}}
Since ViewGCN uses view sampling as a core operation, it requires the number of views to be at least 12, and hence, our MVTN with ViewGCN follows accordingly. 

\mysection{Training MVTN}
We use AdamW \cite{adamw} for our MVTN networks with a learning rate of 0.001. For other training details (\eg training epochs and optimization), we follow the previous works \cite{mvviewgcn,mvcnn} for a fair comparison. The training of MVTN with MVCNN is done in 100 epochs and a batch size of 20, while the MVTN with ViewGCN is performed in two stages as proposed in the official code of the paper \cite{mvviewgcn}. The first stage is 50 epochs of training the backbone CNN on the single view images, while the second stage is 35 epochs on the multi-view network on the $M$ views of the 3D object. We use learning rates of 0.0003 for MVCNN and 0.001 for ViewGCN, and a ResNet-18 \cite{resnet} as the backbone CNN for both baselines and our MVTN-based networks. A weight decay of 0.01 is applied for both the multi-view network and in the MVTN networks. Due to gradient instability from the renderer, we introduce gradient clipping in the MVTN to limit the $\ell_2$ norm of gradient updates to 30 for$\mathbf{G}$. The code for full implementation of MVTN will be made public with the publication of this manuscript

\clearpage \clearpage
\section{Additional Results}

\subsection{Classification and Retrieval Benchmarks}
We provide in Tables \ref{tab:ModelNet40-cls-supp},\ref{tab:Scanobjectnn-supp}, and \ref{tab:retrieval-supp} comprehensive benchmarks of 3D classifications and 3D shape retrieval methods on ModelNet40 \cite{modelnet}, ScanObjectNN \cite{scanobjectnn}, and ShapeNet Core55 \cite{shapenet,shrek17}. These tables include methods that use points as representations as well as other modalities like multi-view and volumetric representations. Our reported results of four runs are presented in each table as ``max (avg $\pm$ std)''.  Note in Table \ref{tab:ModelNet40-cls-supp} how our MVTN improves the previous state-of-the-art in classification (ViewGCN \cite{mvviewgcn}) when tested on the same setup. Our implementations (highlighted using $*$) slightly differ from the reported results in their original paper. This can be attributed to the specific differentiable renderer of Pytorch3D \cite{pytorch3d} that we are using, which might not have the same quality of the non-differentiable OpenGL renderings \cite{opengl} used in their setups.

\subsection{Rotation Robustness}
\begin{figure}[ht]
    \centering
    \includegraphics[width=0.7\linewidth]{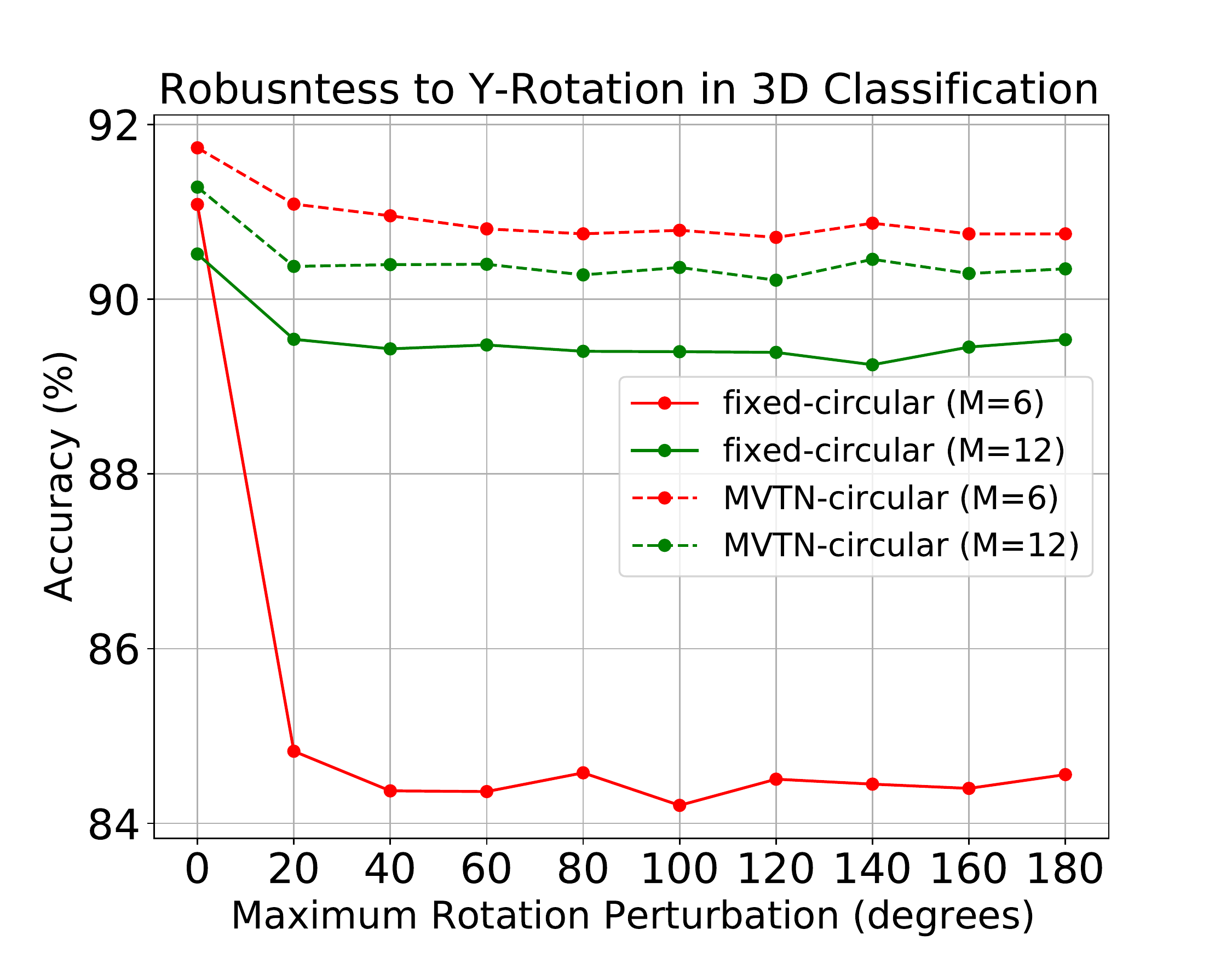}
    \caption{\textbf{Robustness on a Varying Y-Rotation}. We study the effect of varying the maximum rotation perturbation on the classification accuracies on ModelNet40. We compare the performance of circular MVCNN \cite{mvcnn} to our circular-MVTN when it equips MVCNN when the number of views is 6 and 12. Note how MVTN stabilizes the drop in performance for larger Y-rotation perturbations, and the improvement is more significant for the smaller number of views $M$.}
    \label{fig:y-robustness-ablate-sup}
\end{figure}

A common practice in the literature in 3D shape classification is to test the robustness of models trained on the aligned dataset by injecting perturbations during test time \cite{rspointcloud}. We follow the same setup as \cite{rspointcloud} by introducing random rotations during test time around the Y-axis (gravity-axis).
We also investigate the effect of varying rotation perturbations on the accuracy of circular MVCNN when $M=6$ and $M=12$. We note from \figLabel{\ref{fig:y-robustness-ablate-sup}} that using less views leads to higher sensitivity to rotations in general. Furthermore, we note that our MVTN helps in stabilizing the performance on increasing thresholds of rotation perturbations.


\subsection{Occlusion Robustness} \label{sec:occlusion-supp}
\begin{figure}[t]
    \centering
    \includegraphics[trim= {1cm 0 0 0},clip , width=0.7\linewidth]{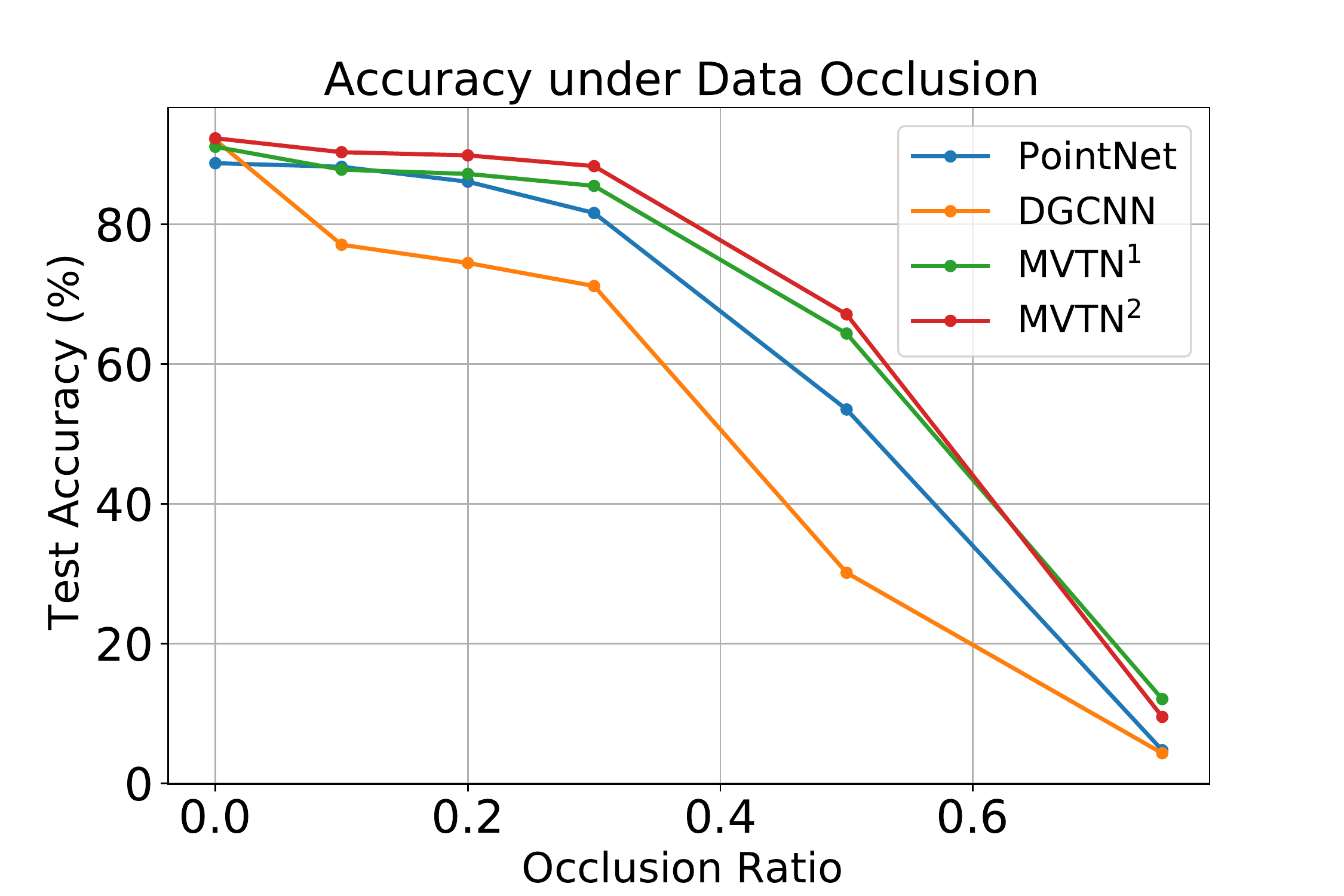}
    \caption{\textbf{Occlusion Robustness of 3D Methods.} We plot test accuracy \vs the Occlusion Ratio of the data to simulate the occlusion robustness of different 3D methods: PointNet \cite{pointnet}, DGCNN \cite{dgcn}, and MVTN. Our MVTN achieves close to 13\% better than PointNet when half of the object is occluded. MVTN$^1$ refers to MVTN with MVCNN as the multi-view network while MVTN$^2$ refers to MVTN with View-GCN as the multi-view network.}
    \label{fig:occlusion}
\end{figure}

\begin{figure*} [] 
\centering
\tabcolsep=0.03cm
\resizebox{0.85\linewidth}{!}{
\begin{tabular}{c|ccccc}

 & \multicolumn{5}{c}{Occlusion Ratio} \\ 
 \textbf{Direction} & 0.1 & 0.2 & 0.3 & 0.5 & 0.75 \\ \midrule

\textbf{+X} &
\includegraphics[trim= 0cm 2cm 0cm 0cm , clip, width = 0.19\linewidth]{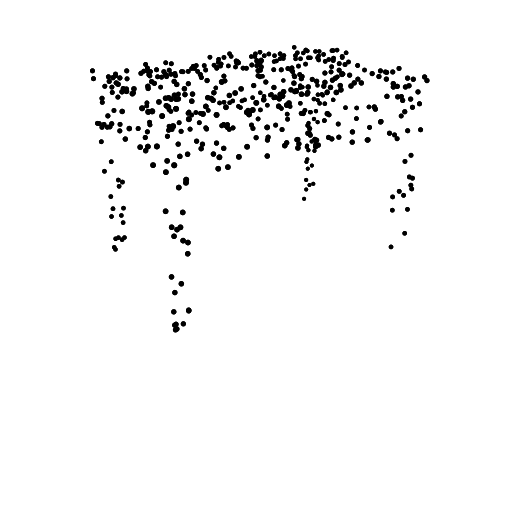} &
\includegraphics[trim= 0cm 2cm 0cm 0cm , clip, width = 0.19\linewidth]{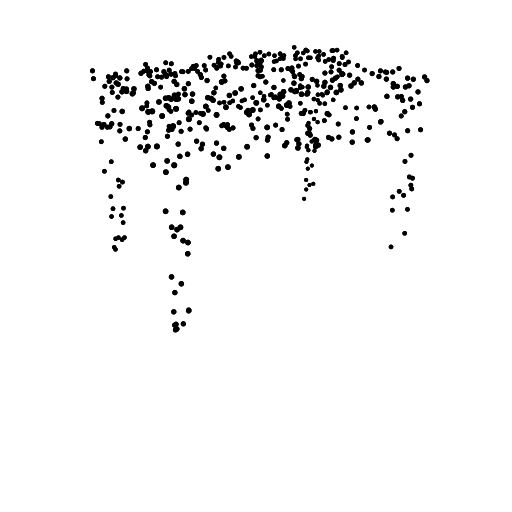} &
\includegraphics[trim= 0cm 2cm 0cm 0cm , clip, width = 0.19\linewidth]{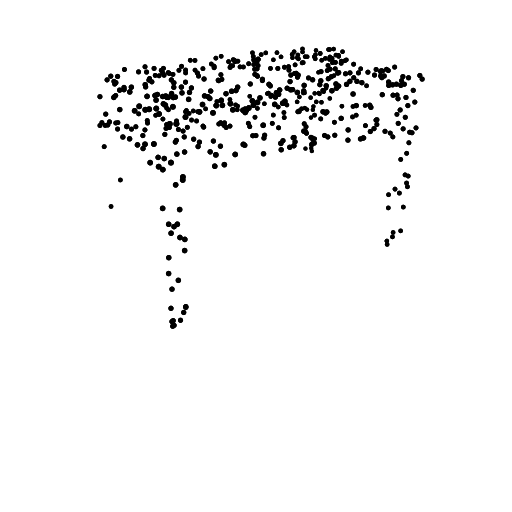} &
\includegraphics[trim= 0cm 2cm 0cm 0cm , clip, width = 0.19\linewidth]{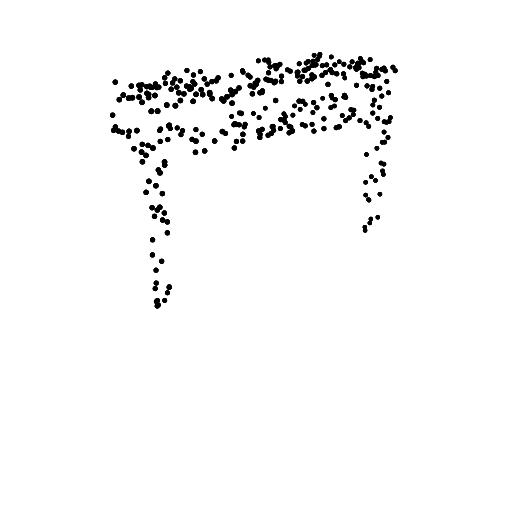} &
\includegraphics[trim= 0cm 2cm 0cm 0cm , clip, width = 0.19\linewidth]{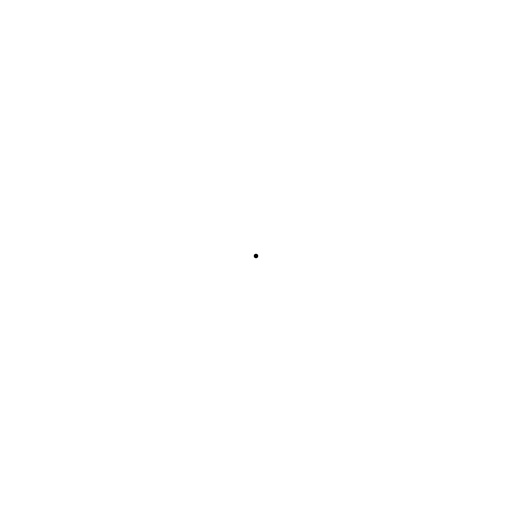} \\ \hline

\textbf{-X} &
\includegraphics[trim= 0cm 2cm 0cm 0cm , clip, width = 0.19\linewidth]{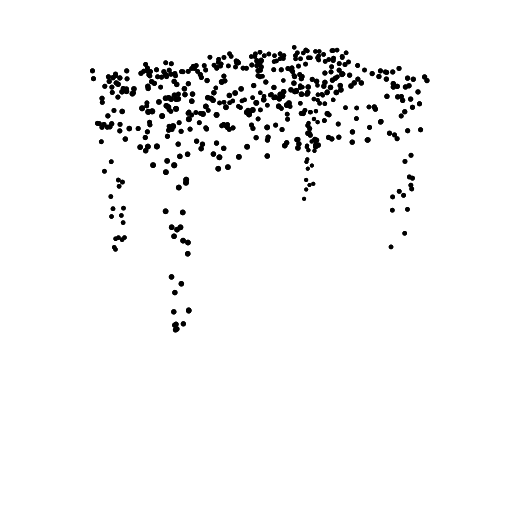} &
\includegraphics[trim= 0cm 2cm 0cm 0cm , clip, width = 0.19\linewidth]{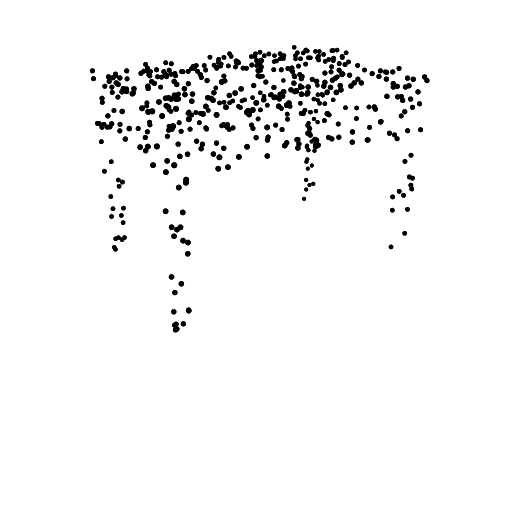} &
\includegraphics[trim= 0cm 2cm 0cm 0cm , clip, width = 0.19\linewidth]{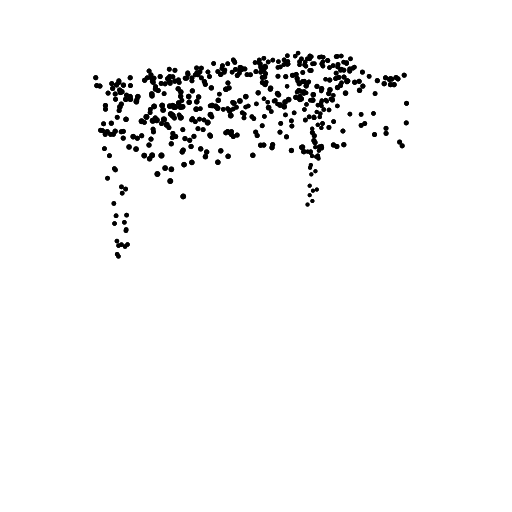} &
\includegraphics[trim= 0cm 2cm 0cm 0cm , clip, width = 0.19\linewidth]{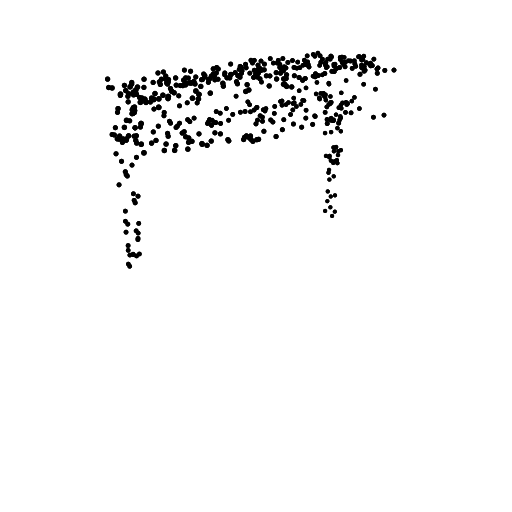} &
\includegraphics[trim= 0cm 2cm 0cm 0cm , clip, width = 0.19\linewidth]{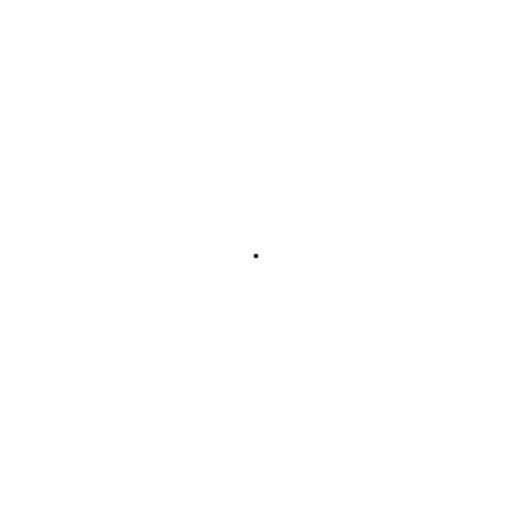} \\ \midrule

\textbf{+Z} &
\includegraphics[trim= 0cm 2cm 0cm 0cm , clip, width = 0.19\linewidth]{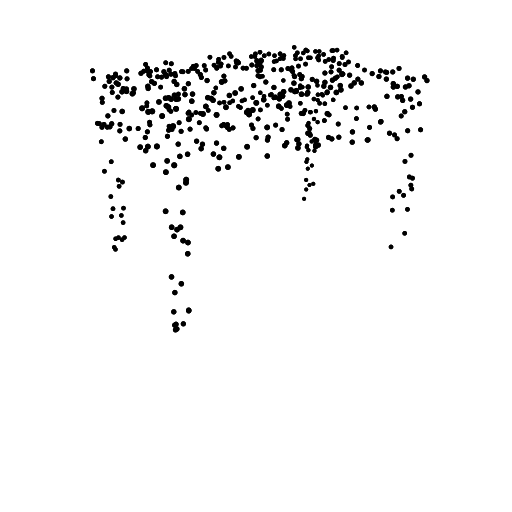} &
\includegraphics[trim= 0cm 2cm 0cm 0cm , clip, width = 0.19\linewidth]{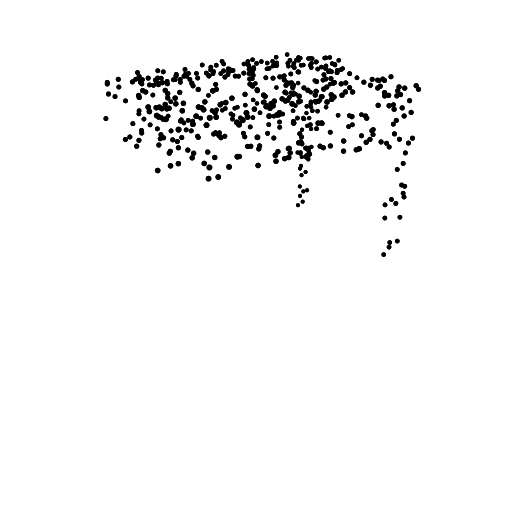} &
\includegraphics[trim= 0cm 2cm 0cm 0cm , clip, width = 0.19\linewidth]{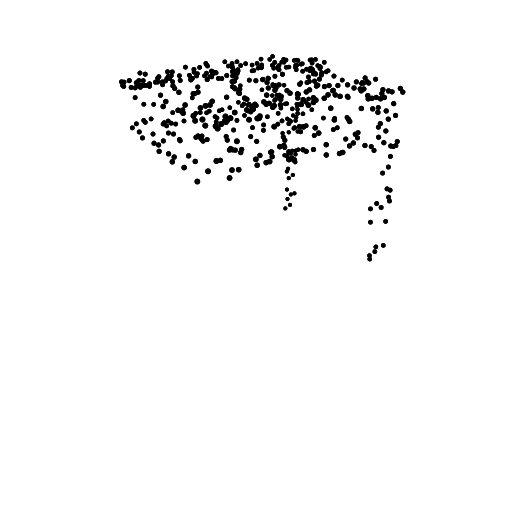} &
\includegraphics[trim= 0cm 2cm 0cm 0cm , clip, width = 0.19\linewidth]{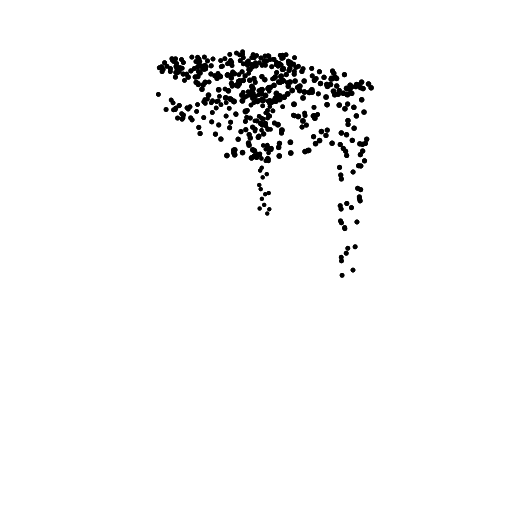} &
\includegraphics[trim= 0cm 2cm 0cm 0cm , clip, width = 0.19\linewidth]{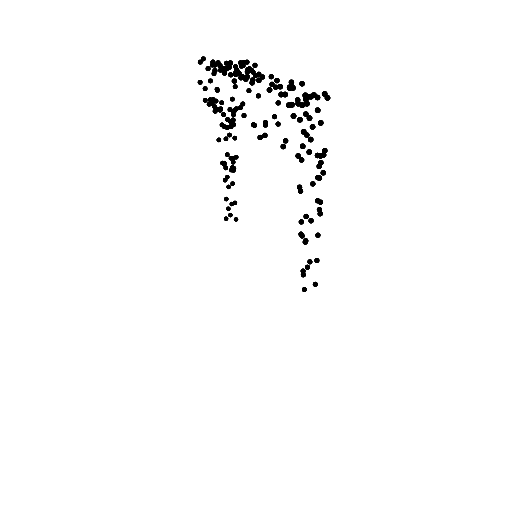} \\ \hline

\textbf{-Z} &
\includegraphics[trim= 0cm 2cm 0cm 0cm , clip, width = 0.19\linewidth]{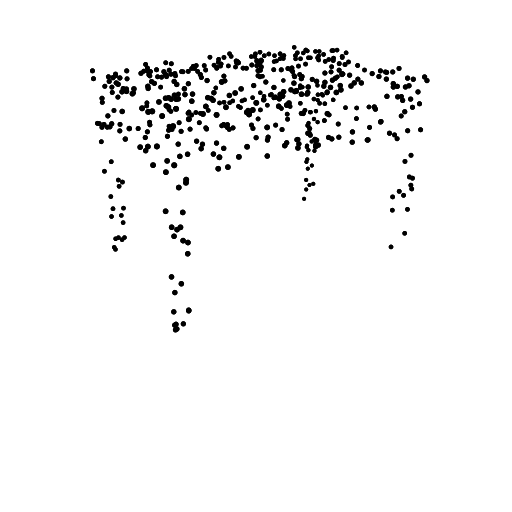} &
\includegraphics[trim= 0cm 2cm 0cm 0cm , clip, width = 0.19\linewidth]{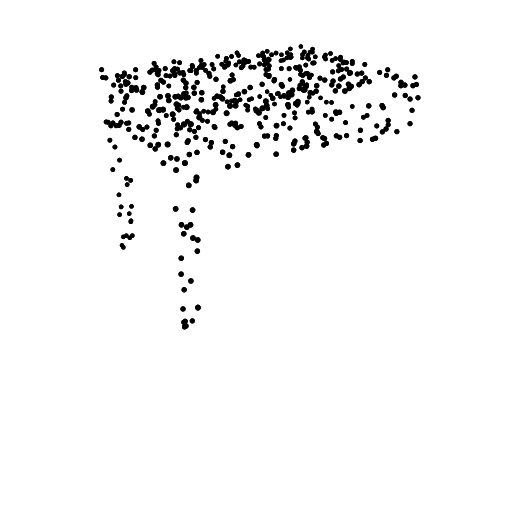} &
\includegraphics[trim= 0cm 2cm 0cm 0cm , clip, width = 0.19\linewidth]{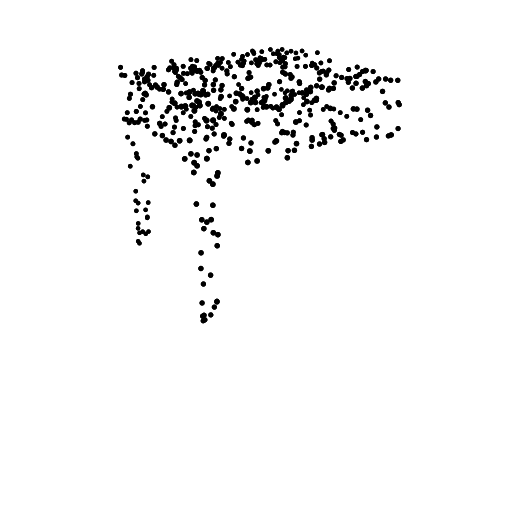} &
\includegraphics[trim= 0cm 2cm 0cm 0cm , clip, width = 0.19\linewidth]{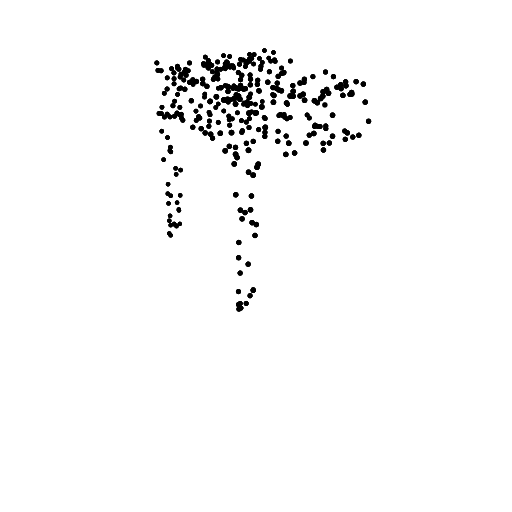} &
\includegraphics[trim= 0cm 2cm 0cm 0cm , clip, width = 0.19\linewidth]{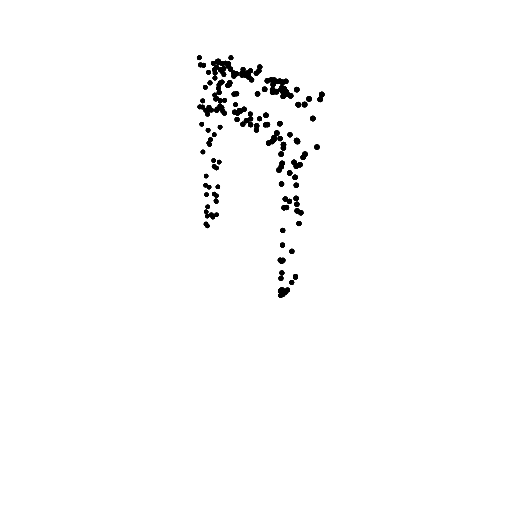} \\ \midrule

\textbf{+Y} &
\includegraphics[trim= 0cm 2cm 0cm 0cm , clip, width = 0.19\linewidth]{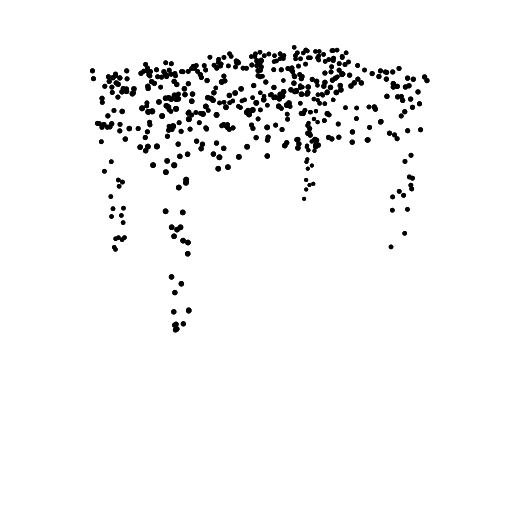} &
\includegraphics[trim= 0cm 2cm 0cm 0cm , clip, width = 0.19\linewidth]{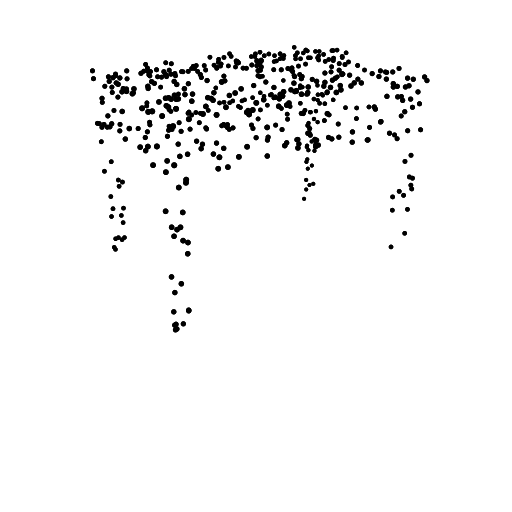} &
\includegraphics[trim= 0cm 2cm 0cm 0cm , clip, width = 0.19\linewidth]{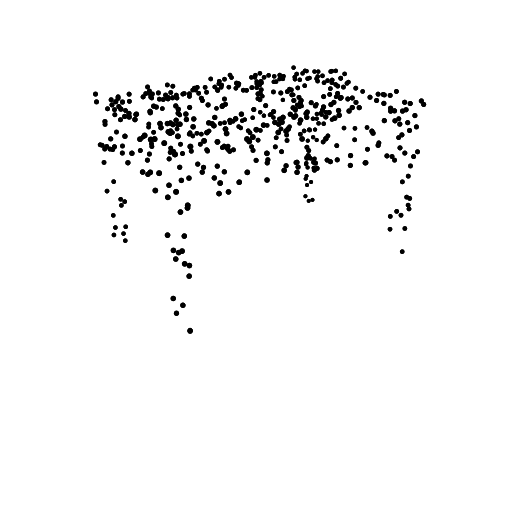} &
\includegraphics[trim= 0cm 2cm 0cm 0cm , clip, width = 0.19\linewidth]{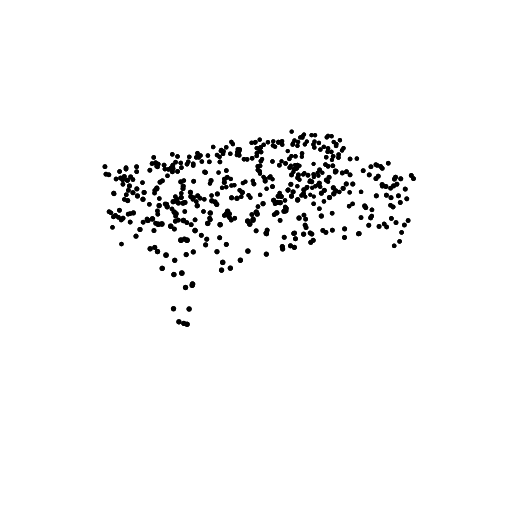} &
\includegraphics[trim= 0cm 2cm 0cm 0cm , clip, width = 0.19\linewidth]{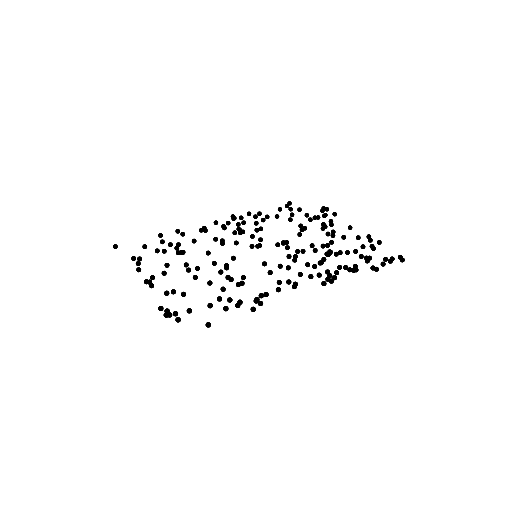} \\ \hline

\textbf{-Y} &
\includegraphics[trim= 0cm 2cm 0cm 0cm , clip, width = 0.19\linewidth]{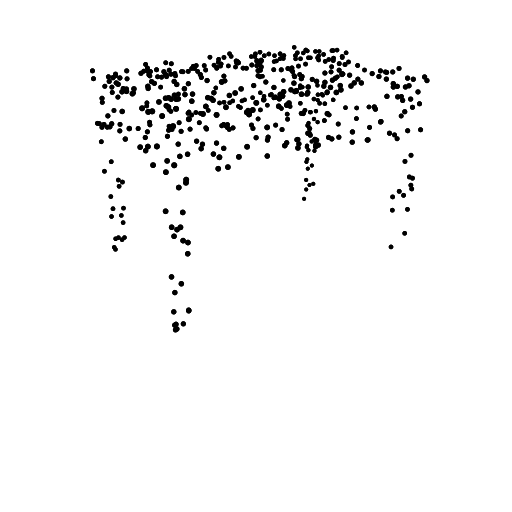} &
\includegraphics[trim= 0cm 2cm 0cm 0cm , clip, width = 0.19\linewidth]{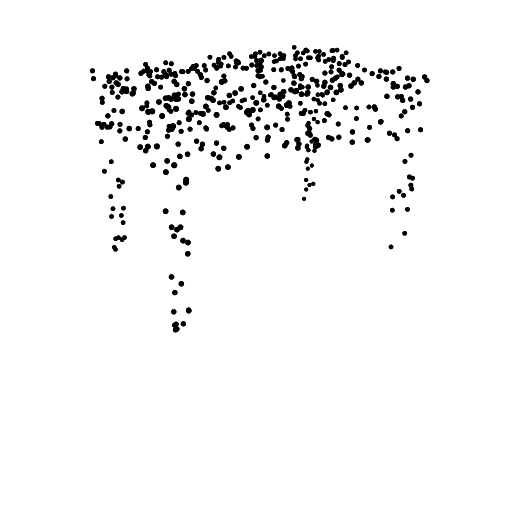} &
\includegraphics[trim= 0cm 2cm 0cm 0cm , clip, width = 0.19\linewidth]{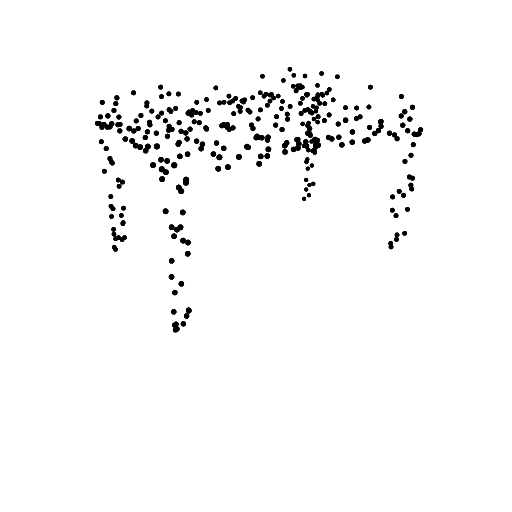} &
\includegraphics[trim= 0cm 2cm 0cm 0cm , clip, width = 0.19\linewidth]{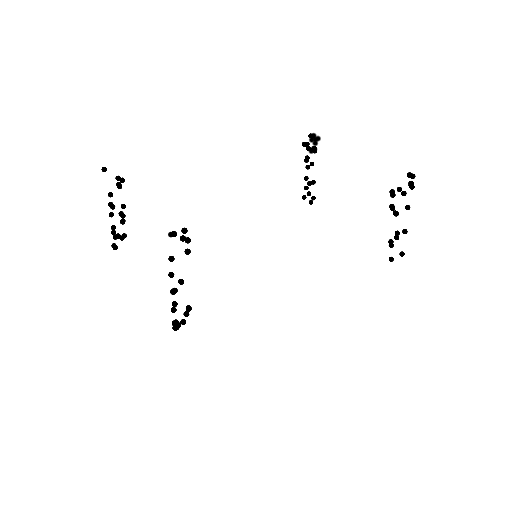} &
\includegraphics[trim= 0cm 2cm 0cm 0cm , clip, width = 0.19\linewidth]{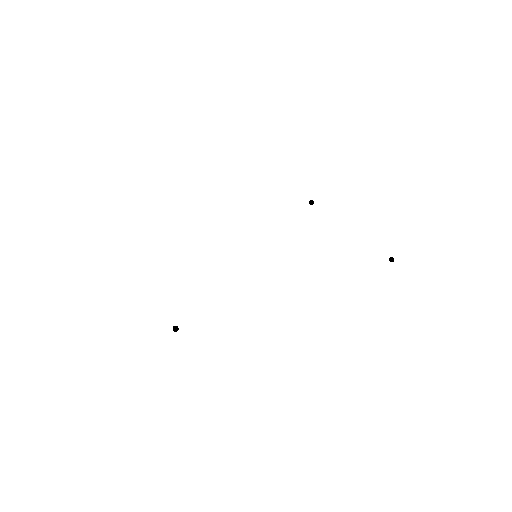} \\ 
 \bottomrule
\end{tabular}
}
\vspace{2pt}
\caption{  \textbf{Occlusion of 3D Objects}: We simulate realistic occlusion scenarios in 3D point clouds by cropping a percentage of the object along canonical directions. Here, we show an object occluded with different ratios and from different directions. 
}
    \label{fig:occlusion-supp}
\end{figure*}

To quantify the occlusion effect due to the viewing angle of the 3D sensor in our setup of 3D classification, we simulate realistic occlusion by cropping the object from canonical directions. We train PointNet \cite{pointnet}, DGCNN \cite{dgcn}, and MVTN on the ModelNet40 point cloud dataset. Then, at test time, we crop a portion of the object (from 0\% occlusion ratio to 75\%) along the $\pm$X, $\pm$Y, and $\pm$Z directions independently. \figLabel{\ref{fig:occlusion-supp}} shows examples of this occlusion effect with different occlusion ratios. We report the average test accuracy (on all the test set) of the six cropping directions for the baselines and MVTN in \figLabel{\ref{fig:occlusion}}. Note how MVTN achieves high test accuracy even when large portions of the object are cropped. Interestingly, MVTN outperforms PointNet \cite{pointnet} by 13\% in test accuracy when half of the object is occluded. This result is significant, given that PointNet is well-known for its robustness \cite{pointnet,advpc}. 

\begin{table*}[t]
\tabcolsep=0.25cm
    \centering
\resizebox{0.75\linewidth}{!}{\begin{tabular}{rccc}
\toprule
 &   & \multicolumn{2}{c}{Classification Accuracy} \\
\multicolumn{1}{c}{Method}       & Data Type & (\textbf{Per-Class})   & (\textbf{Overall}) \\ \midrule
SPH  \cite{sph}                   &  Voxels                 & 68.2 & - \\
LFD     \cite{lfd}                &  Voxels                 & 75.5 & - \\ 
3D ShapeNets \cite{modelnet} &  Voxels                 & 77.3 & - \\
VoxNet \cite{voxnet}     & Voxels                 & 83.0 & 85.9      \\
VRN  \cite{vrn}                      & Voxels                 & - & 91.3      \\
MVCNN-MS \cite{multimvcnn} &        Voxels                 & - & 91.4      \\ 
FusionNet \cite{fusionnet}                & Voxels+MV  & - & 90.8      \\ 
PointNet \cite{pointnet} &  Points                  &       86.2 & 89.2      \\
PointNet++ \cite{pointnet++}   & Points                & - & 91.9      \\
KD-Network \cite{kdnet}&    Points                &     88.5 &     91.8   \\
PointCNN \cite{pc_li2018pointcnn}  & Points            &   88.1   & 91.8      \\
DGCNN \cite{dgcn}             & Points                &  90.2     & 92.2      \\
KPConv\cite{kpconv}  &  Points & -  & 92.9 \\ 
PVNet\cite{pvnet}  &  Points & -  & 93.2 \\ 
PTransformer\cite{pointtransformer}  &  Points & \textbf{90.6} & \textbf{93.7}  \\ \midrule
MVCNN  \cite{mvcnn}         & 12 Views                   & 90.1  & 90.1 \\
GVCNN \cite{mvgvcnn}         & 12 Views                   & 90.7 & 93.1 \\
ViewGCN \cite{mvviewgcn}  & 20 Views   & \textbf{96.5} & \textbf{97.6} \\ 

\midrule
ViewGCN \cite{mvviewgcn}$^*$& 12 views &    90.7 (90.5 $\pm$ 0.2)   &93.0 (92.8 $\pm$ 0.1) \\
ViewGCN \cite{mvviewgcn}$^*$& 20 views &    91.3 (91.0 $\pm$ 0.2)   &93.3 (93.1 $\pm$ 0.2) \\
MVTN (ours)$^*$  & 12 Views       & 92.0 (91.2 $\pm$ 0.6) & \textbf{93.8} (93.4 $\pm$ 0.3) \\
MVTN (ours)$^*$  & 20 Views       & \textbf{92.2} (91.8 $\pm$ 0.3) & 93.5 (93.1 $\pm$ 0.5) \\
\bottomrule
\end{tabular}
}
\vspace{2pt}
    \caption{\textbf{3D Shape Classification on ModelNet40}. We compare MVTN against other methods in 3D classification on ModelNet40 \cite{modelnet}. $^*$ indicates results from our rendering setup (differentiable pipeline), while other multi-view results are reported from pre-rendered views. \textbf{Bold} denotes the best result in its setup. In brackets, we report the average and standard deviation of four runs}
    \label{tab:ModelNet40-cls-supp}
\end{table*}
\begin{table*}[t]
\tabcolsep=0.2cm
    \centering
\resizebox{0.9\linewidth}{!}{\begin{tabular}{rccc}
\toprule
 &  \multicolumn{3}{c}{Classification Overall Accuracy } \\
\multicolumn{1}{c}{Method}& \textbf{Object with Background}  & \textbf{Object Only} & \textbf{PB\_T50\_RS (Hardest)}  \\ \midrule
3DMFV \cite{3Dmfv} &  68.2                  &  73.8  &  63.0  \\
PointNet \cite{pointnet}   & 73.3                &   79.2  & 68.0 \\
SpiderCNN \cite{pc_xu2018spidercnn}&    77.1                &    79.5   &  73.7    \\
PointNet ++ \cite{pointnet++}            & 82.3 & 84.3  &   77.9   \\
PointCNN \cite{pc_li2018pointcnn}  & 86.1 & 85.5  & 78.5 \\
DGCNN \cite{dgcn}  & 82.8 & 86.2  & 78.1 \\ 
SimpleView \cite{simpleview}& - & - & 79.5 \\
BGA-DGCNN \cite{scanobjectnn}   & - & - & 79.7 \\
BGA-PN++ \cite{scanobjectnn}   & - & - & 80.2 \\
\midrule
ViewGCN $^*$  & 91.9 (91.12 $\pm$ 0.5)  & 90.4 (89.7 $\pm$ 0.5) & 80.5 (80.2 $\pm$ 0.4) \\
MVTN (ours)  & \textbf{92.6} (92.5 $\pm$ 0.2)  & \textbf{92.3} (91.7 $\pm$ 0.7) & \textbf{82.8} (81.8 $\pm$ 0.7) \\
\bottomrule
\end{tabular}
}
\vspace{2pt}
    \caption{\textbf{3D Point Cloud Classification on ScanObjectNN}. We compare the performance of MVTN in 3D point cloud classification on three different variants of ScanObjectNN \cite{scanobjectnn}. The variants include object with background, object only, and the hardest variant. $^*$ indicates results from our rendering setup (differentiable pipeline), and we report the average and standard deviation of four runs in brackets.}
    \label{tab:Scanobjectnn-supp}
\end{table*}

\begin{table*}[t]
\tabcolsep=0.25cm
    \centering
\resizebox{0.75\linewidth}{!}{\begin{tabular}{rccc}
\toprule
 &   & \multicolumn{2}{c}{ Shape Retrieval (mAP)} \\
\multicolumn{1}{c}{Method} & Data Type  & \textbf{ModelNet40}  & \textbf{ShapeNet Core} \\ \midrule
ZDFR \cite{zfdrretr} &  Voxels                  &      - & 19.9      \\
DLAN \cite{dlanretr}   & Voxels                &   - & 66.3      \\
SPH  \cite{sph}                   &  Voxels                 & 33.3 & - \\
LFD     \cite{lfd}                &  Voxels                 & 40.9 & - \\ 
3D ShapeNets \cite{modelnet} &  Voxels                 & 49.2 & - \\
PVNet\cite{pvnet}  &  Points & 89.5 & - \\ 
MVCNN  \cite{mvcnn}         & 12 Views                   & 80.2 & 73.5 \\
GIFT \cite{giftretr}&    20 Views                &     - & 64.0      \\
MVFusionNet \cite{mvfusionnet}            & 12 Views                &      - & 62.2      \\
ReVGG \cite{shrek17}  & 20 Views            &     - & 74.9      \\
RotNet \cite{mvrotationnet}  & 20 Views   & - & 77.2 \\ 
ViewGCN \cite{mvviewgcn}  & 20 Views   & - &  78.4 \\ 
MLVCNN \cite{mlvcnn} &   24 Views               & 92.2      & -\\
\midrule
MVTN (ours)  & 12 Views         & \textbf{92.9} (92.4 $\pm$ 0.6) &  \textbf{82.9} (82.4 $\pm$ 0.6) \\
\bottomrule
\end{tabular}
}
\vspace{2pt}
    \caption{\textbf{3D Shape Retrieval}. We benchmark the shape retrieval capability of MVTN on ModelNet40 \cite{modelnet} and ShapeNet Core55 \cite{shapenet,shrek17}. MVTN achieves the best retrieval performance among recent state-of-the-art methods on both datasets with only 12 views. In brackets, we report the average and standard deviation of four runs.}
     \label{tab:retrieval-supp}
\end{table*}

\clearpage \clearpage
\section{Analysis and Insights}

\subsection{Ablation Study} \label{sec:ablation-supp}
This section introduces a comprehensive ablation study on the different components of MVTN, and their effect on test accuracy on the standard ModelNet40 \cite{modelnet}.

\begin{figure}[t]
    \centering
    \includegraphics[,trim=0.5cm 0 1.5cm 0, clip,width=0.63\linewidth]{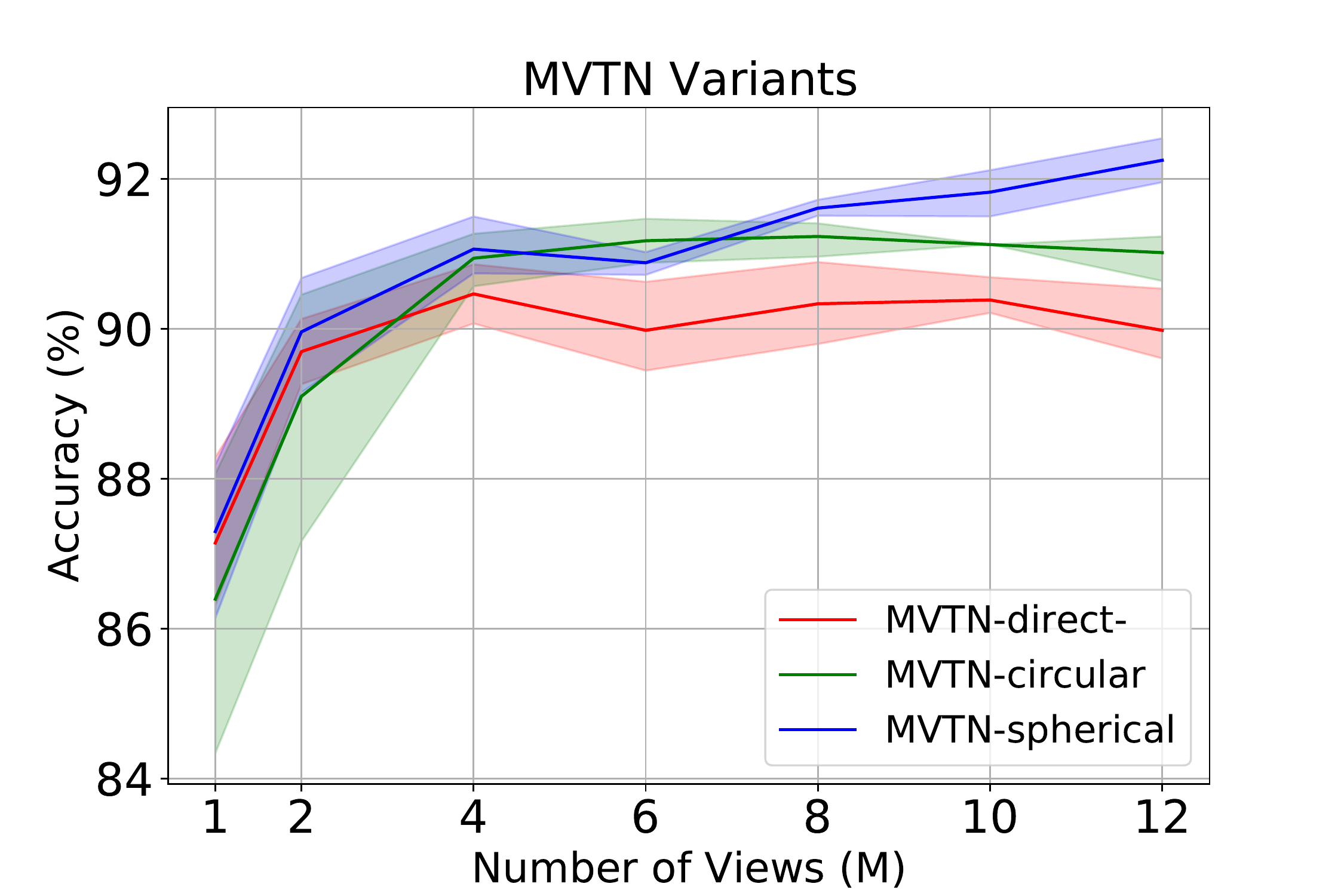}
    \caption{\textbf{Variants of MVTN.} We plot test accuracy vs. the number of views used in training different variants of our MVTN. Note how MVTN-spherical is generally more stable in achieving better performance on ModelNet40. 95\% confidence interval is also plotted on each setup (repeated four times).}
    \label{fig:mvtn-variants}
\end{figure}

\mysection{MVTN Variants} \label{sec:views-supp}
We study the effect of the number of views $M$ on the performance of different MVTN variants (direct, circular, spherical). 
The experiments are repeated four times, and the average test accuracies with confidence intervals are shown in  \figLabel{\ref{fig:mvtn-variants}}. 
 The plots show how learned MVTN-spherical achieves consistently superior performance across a different number of views. Also, note that MVTN-direct suffers from over-fitting when the number of views is larger than four (\ie it gets perfect training accuracy but deteriorates in test accuracy). This can be explained by observing that the predicted view-points tend to be similar to each other for MVTN-direct when the number of views is large. The similarity in views leads the multi-view network to memorize the training but to suffer in testing.

\mysection{Backbone and Late Fusion}
In the main manuscript (Table 6), we study MVTN with ViewGCN as the multi-view network. Here, we study the backbone effect on MVTN with MVCNN as the multi-view network and report all results in Table \ref{tbl:ablation-sup}. The study includes the backbone choice, the point encoder choice, and the type of aggregation method used to combine the multi-view network and the point encoder (late fusion \vs MVTN). Note that including more sophisticated backbones does not improve the accuracy, and using a late fusion mechanism is worse than our MVTN.

\begin{figure}[t]
    \centering
    \includegraphics[width=0.7\linewidth]{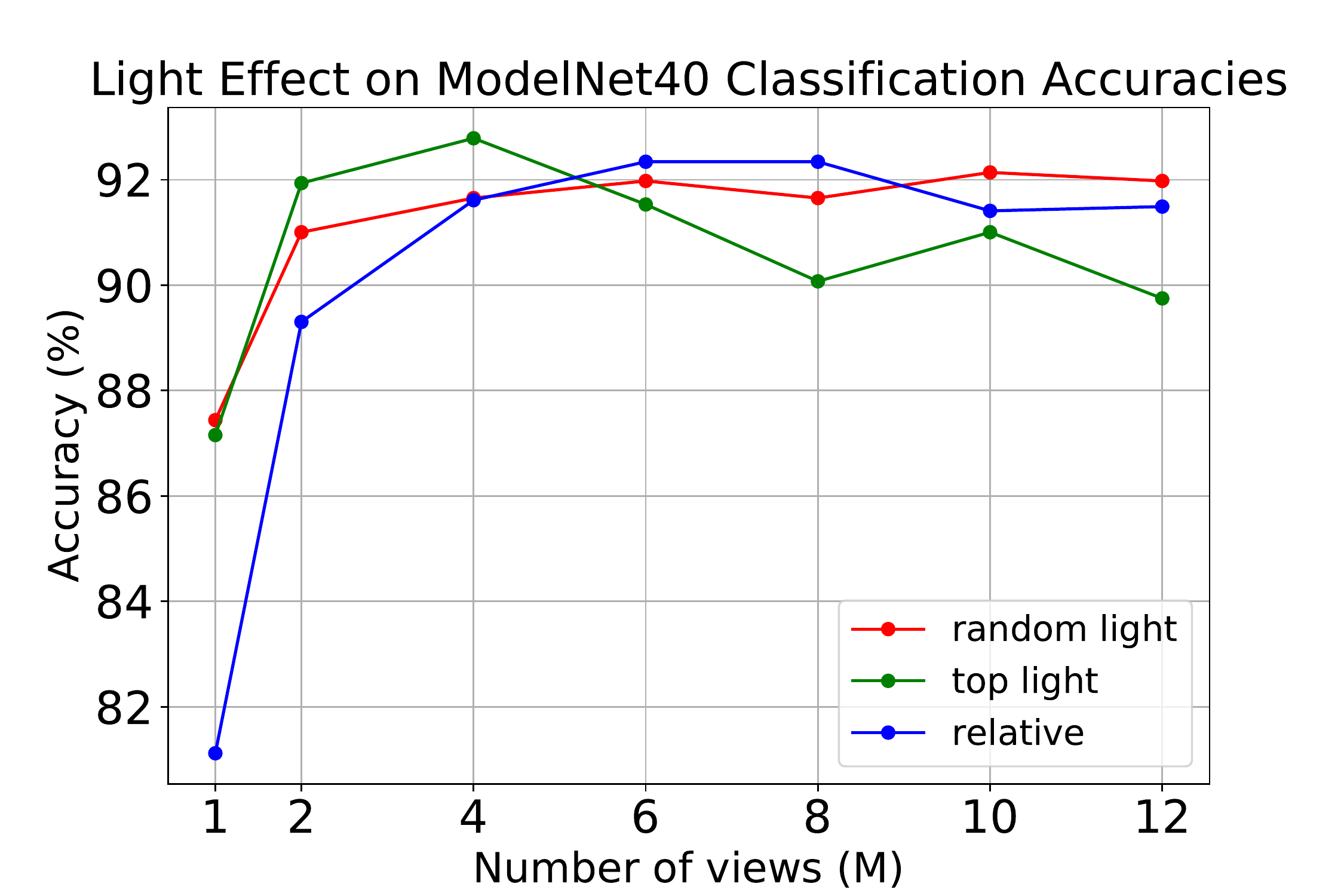}
    \caption{\textbf{Light Direction Effect}. We study the effect of light direction in the performance of the MVTN-circular. We note that randomizing the light direction in training reduce overfitting for larger number of views and leads to better generalization.}
    \label{fig:ablation-light-sup}
\end{figure}

\mysection{Light Direction Effect}
We study the effect of light's direction on the performance of multi-view networks. We note that picking a random light in training helps the network generalize to the test set. Please see \figLabel{\ref{fig:ablation-light-sup}} for the results on circular MVTN with MVCNN when comparing this strategy to fixed light from the top or from camera (\textit{relative}). Note that we use relative light in test time to stabilize the performance.

\begin{figure}[h]
    \centering
  \quad ~~ Effect of Distance \\
    \includegraphics[trim= 0 0 1cm 1.7cm , clip, width=0.68\linewidth]{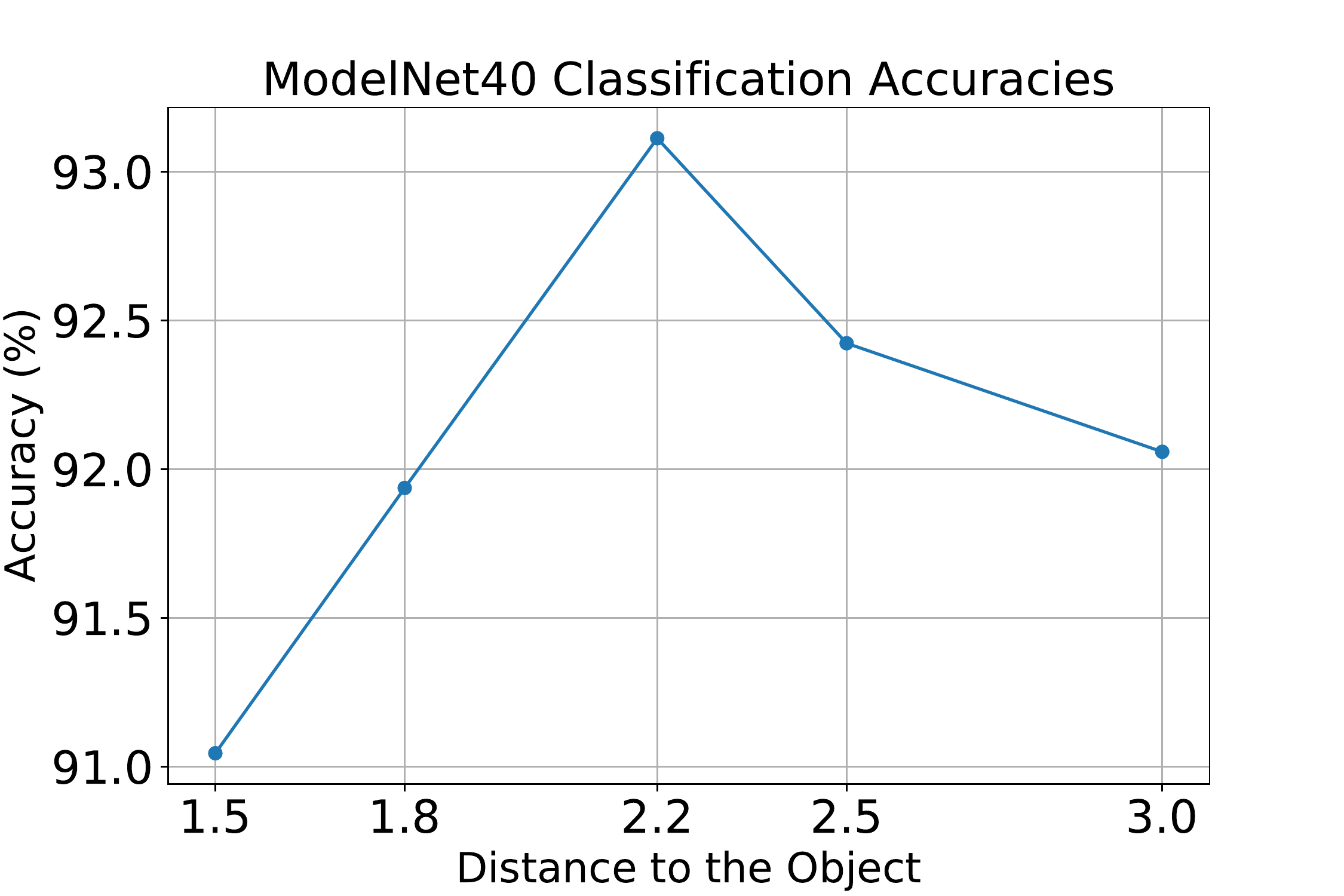}
    \caption{\textbf{Effect of Distance to 3D Object}. We study the effect of changing the distance on training a spherical ViewGCN. We show that the distance of 2.2 units to the center is in between far and close it and gives the best accuracy.}
    \label{fig:ablation-distance-sup}
\end{figure}

\mysection{Learning Distance to the Object}
One possible ablation to the MVTN is to learn the distance to the object. This feature should allow the cameras to get closer to details that might be important to the classifier to understand the object properly. However, we observe that MVTN generally performs worse or does not improve with this setup, and hence, we refrain from learning it. In all of our main experiments, we fixed the distance to $2.2$ units, which is a good middle ground providing best accuracy. Please see \figLabel{\ref{fig:ablation-distance-sup}} for the effect of picking a fixed distance in training spherical ViewGCN.  





\subsection{Transferability of MVTN View-Points}
We hypothesize that the views learned by MVTN are transferable across multi-view classifiers. Looking at results in \figLabel{\ref{fig:views-mvt-sup-1}, \ref{fig:views-mvt-sup-2}}, we believe MVTN picks the best views based on the actual shape and is less influenced by the multi-view network.
This means that MVTN learns views that are more representative of the object, making it easier for \textit{any} multi-view network to recognize it.
As such, we ask the following:
\textit{can we transfer the views MVTN learns under one setting to a different multi-view network?}

To test our hypothesis, we take a 12-view MVTN-spherical module trained with MVCNN as a multi-view network and transfer the predicted views to a ViewGCN multi-view network. In this case, we freeze the MVTN module and only train ViewGCN on these learned but fixed views. ViewGCN with transferred MVTN views reaches $93.1\%$ accuracy in classification. It corresponds to a boost of $0.7\%$ from the $92.4\%$ of the original ViewGCN. Although this result is lower than fully trained MVTN($-0.3\%$), we observe a decent transferability between both multi-view architectures.




\begin{table*}[h]
\footnotesize
\setlength{\tabcolsep}{4pt} 
\renewcommand{\arraystretch}{1.1} 
\centering
\resizebox{0.99\hsize}{!}{
\begin{tabular}{c|cc|cc|cc|cc||c} 
\toprule
\textbf{Views} &  \multicolumn{2}{c|}{\textbf{Backbone}}& \multicolumn{2}{c|}{\textbf{Point Encoder}}& \multicolumn{2}{c|}{\textbf{Setup}}&  \multicolumn{2}{c||}{\textbf{Fusion}} & \multicolumn{1}{c}{\textbf{Results}}\\
 \textbf{number} & \textbf{ResNet18} & \textbf{ResNet50} & \textbf{PointNet\cite{pointnet}} &  \textbf{DGCNN\cite{dgcn}} & \textbf{circular} & \textbf{spherical} &  \textbf{late} & \textbf{MVTN}& \textbf{accuracy}     \\
\midrule
6 & \checkmark & - & \checkmark &   - &  \checkmark  & - &  \checkmark & - & 90.48  \% \\ \hline
6 & \checkmark & - & \checkmark &   - &  \checkmark  & - & - &  \checkmark &  91.13 \% \\ \hline
6 & \checkmark & - & \checkmark &   - &  -  & \checkmark  &  \checkmark & - & 89.51  \% \\ \hline
6 & \checkmark & - & \checkmark &   - &  -  & \checkmark  & - &  \checkmark &  91.94 \% \\ \hline
6 & \checkmark & - & - &  \checkmark &  \checkmark   & - &  \checkmark & - & 87.80  \% \\ \hline
6 & \checkmark & - & - &  \checkmark & \checkmark   & - & - &  \checkmark &  91.49 \% \\ \hline
6 & \checkmark & - & - &  \checkmark &  -  & \checkmark  &  \checkmark & - & 89.82  \% \\ \hline
6 & \checkmark & - & - &  \checkmark &  -  & \checkmark  & - &  \checkmark &  91.29 \% \\  \hline

 6 &   - & \checkmark & \checkmark &   - &  \checkmark  & - &  \checkmark & - & 89.10  \% \\ \hline
 6 &   - & \checkmark & \checkmark &   - &  \checkmark  & - & - &  \checkmark &  90.40 \% \\ \hline
 6 &   - & \checkmark & \checkmark &   - &  -  & \checkmark  &  \checkmark & - & 89.22  \% \\ \hline
 6 &   - & \checkmark & \checkmark &   - &  -  & \checkmark  & - &  \checkmark &  90.76 \% \\ \hline
 6 &   - & \checkmark & - &  \checkmark &  \checkmark   & - &  \checkmark & - & 89.99  \% \\ \hline
 6 &   - & \checkmark & - &  \checkmark & \checkmark   & - & - &  \checkmark &  89.91 \% \\ \hline
 6 &   - & \checkmark & - &  \checkmark &  -  & \checkmark  &  \checkmark & - & 89.95  \% \\ \hline
 6 &   - & \checkmark & - &  \checkmark &  -  & \checkmark  & - &  \checkmark &  90.43 \% \\ \midrule
 
  12 & \checkmark & - & \checkmark &   - &  \checkmark  & - &  \checkmark & - & 87.35\% \\ \hline
12 & \checkmark & - & \checkmark &   - &  \checkmark  & - & - &  \checkmark &  90.68\% \\ \hline
12 & \checkmark & - & \checkmark &   - &  -  & \checkmark  &  \checkmark & - & 88.41\% \\ \hline
12 & \checkmark & - & \checkmark &   - &  -  & \checkmark  & - &  \checkmark &  91.82 \\ \hline
12 & \checkmark & - & - &  \checkmark &  \checkmark   & - &  \checkmark & - & 90.24\% \\ \hline
12 & \checkmark & - & - &  \checkmark & \checkmark   & - & - &  \checkmark &  90.28\% \\ \hline
12 & \checkmark & - & - &  \checkmark &  -  & \checkmark  &  \checkmark & - & 89.83\% \\ \hline
12 & \checkmark & - & - &  \checkmark &  -  & \checkmark  & - &  \checkmark &  91.98\% \\  \hline

 12 &   - & \checkmark & \checkmark &   - &  \checkmark  & - &  \checkmark & - & 86.87\% \\ \hline
 12 &   - & \checkmark & \checkmark &   - &  \checkmark  & - & - &  \checkmark &  88.86\% \\ \hline
 12 &   - & \checkmark & \checkmark &   - &  -  & \checkmark  &  \checkmark & - & 87.16\% \\ \hline
 12 &   - & \checkmark & \checkmark &   - &  -  & \checkmark  & - &  \checkmark &  88.41\% \\ \hline
 12 &   - & \checkmark & - &  \checkmark &  \checkmark   & - &  \checkmark & - & 90.15\% \\ \hline
 12 &   - & \checkmark & - &  \checkmark & \checkmark   & - & - &  \checkmark &  88.37\% \\ \hline
 12 &   - & \checkmark & - &  \checkmark &  -  & \checkmark  &  \checkmark & - & 90.48\% \\ \hline
 12 &   - & \checkmark & - &  \checkmark &  -  & \checkmark  & - &  \checkmark &  89.63\% \\ 
 \bottomrule
\end{tabular}
}
\vspace{2pt}
\caption{  \textbf{Ablation Study}. We study the effect of ablating different components of MVTN on the test accuracy on ModelNet40. Namely, we observe that using more complex backbone CNNs (like ResNet50 \cite{resnet}) or a more complex features extractor (like DGCNN \cite{dgcn}) does not increase the performance significantly compared to ResNet18 and PointNet \cite{pointnet} respectively. Furthermore, combining the shape features extractor with the MVCNN \cite{mvcnn} in \textit{late fusion} does not work as well as MVTN with the same architectures. All the reported results are using MVCNN \cite{mvcnn} as multi-view network.}
\label{tbl:ablation-sup}
\end{table*}

\subsection{MVTN Predicted Views}
\begin{figure}[t]
    \centering
        \includegraphics[width=0.68\linewidth]{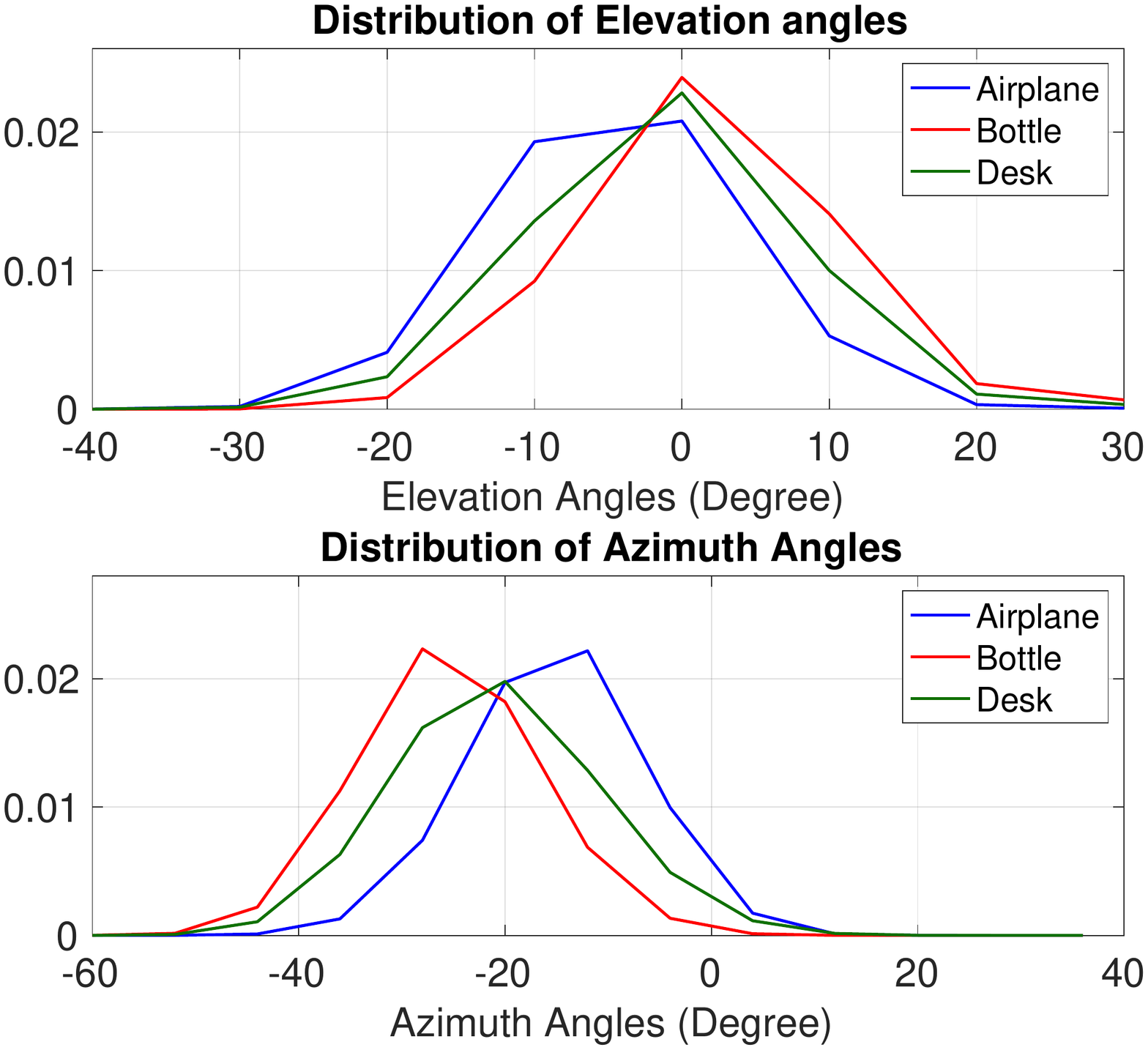}
    \caption{\textbf{Visualizing MVTN learned Views.} We visualize the distribution of azimuth and elevation angles predicted by the MVTN for three different classes. Note that MVTN learns inter-class variations (between different classes) and intra-class variations (on the same class).}
    \label{fig:distribution-sup}
\end{figure}

\begin{figure*} [h] 
\centering
\tabcolsep=0.03cm
\resizebox{0.9\linewidth}{!}{
\begin{tabular}{lc|c}  
\toprule
\textbf{Circular:} \hspace{10pt}  & \includegraphics[trim= 4cm 2.7cm 4cm 2.2cm , clip, width = 0.135\linewidth]{images_qualitative_views_circular_cam.jpg} &
\includegraphics[width = 0.7\linewidth ]{images_qualitative_views_circular_rend.jpg} \\ \hline
\textbf{MVTN-Circular:}  &\includegraphics[trim= 4cm 2.7cm 4cm 2.2cm , clip, width = 0.135\linewidth]{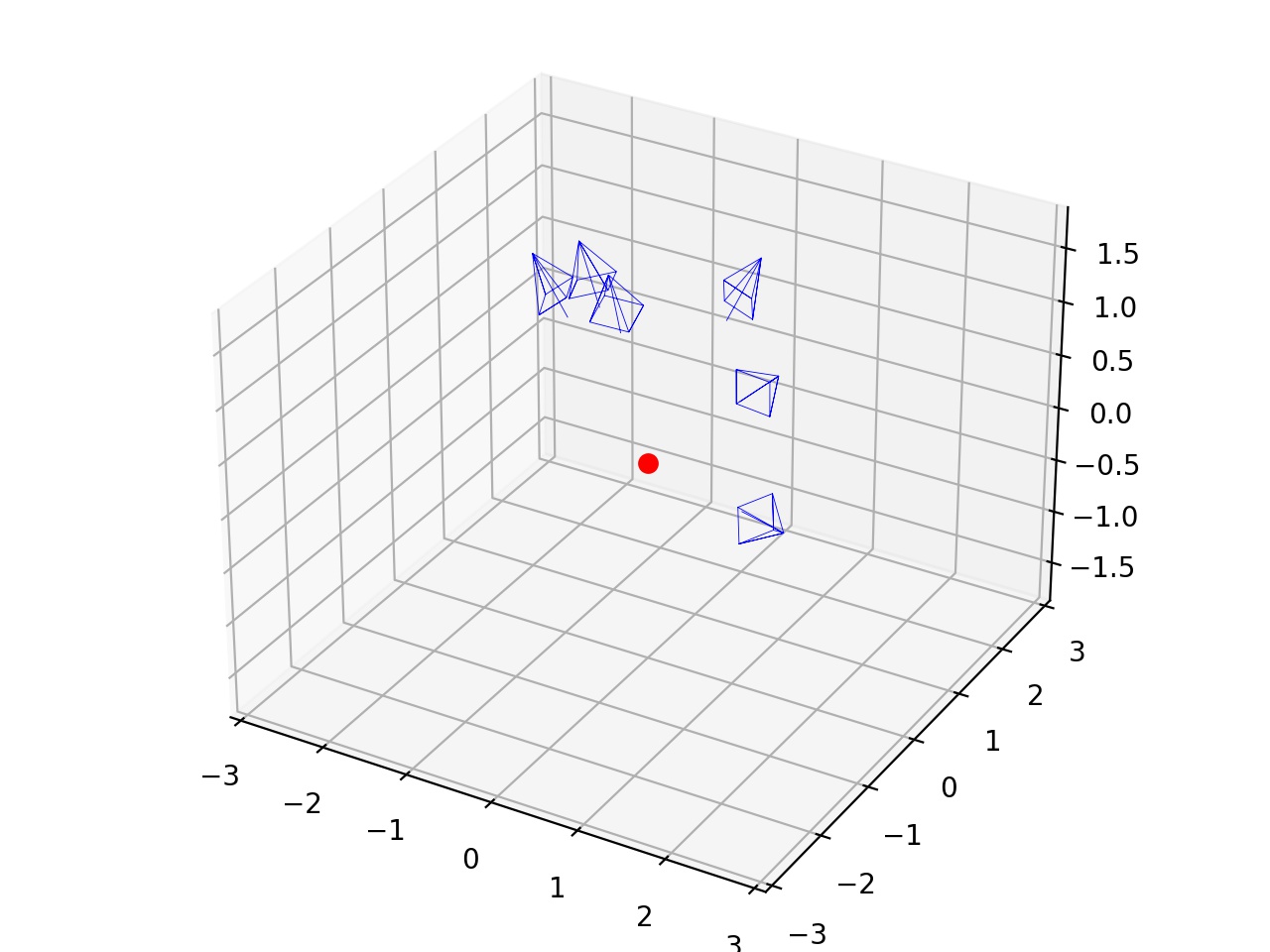} &
\includegraphics[width = 0.7\linewidth]{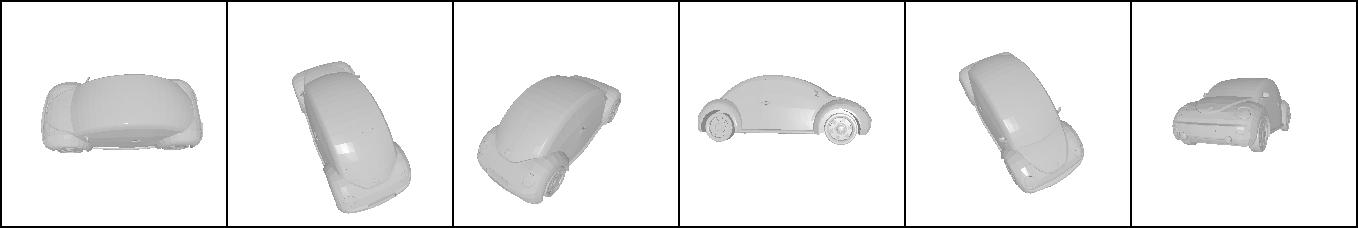} \\ \hline
 \textbf{Spherical:} \hspace{10pt} &\includegraphics[trim= 4cm 2.7cm 4cm 2.2cm , clip, width = 0.135\linewidth]{images_qualitative_views_spherical_cam.jpg} &
\includegraphics[width = 0.7\linewidth]{images_qualitative_views_spherical_rend.jpg} \\ \hline
 \textbf{MVTN-Spherical:}  & \includegraphics[trim= 4cm 2.7cm 4cm 2.2cm , clip, width = 0.135\linewidth]{images_qualitative_views_mvt_spherical_cam.jpg} &
\includegraphics[width = 0.7\linewidth]{images_qualitative_views_mvt_spherical_rend.jpg} \\ \midrule \midrule

\textbf{Circular:} \hspace{10pt}  & \includegraphics[trim= 4cm 2.7cm 4cm 2.2cm , clip, width = 0.135\linewidth]{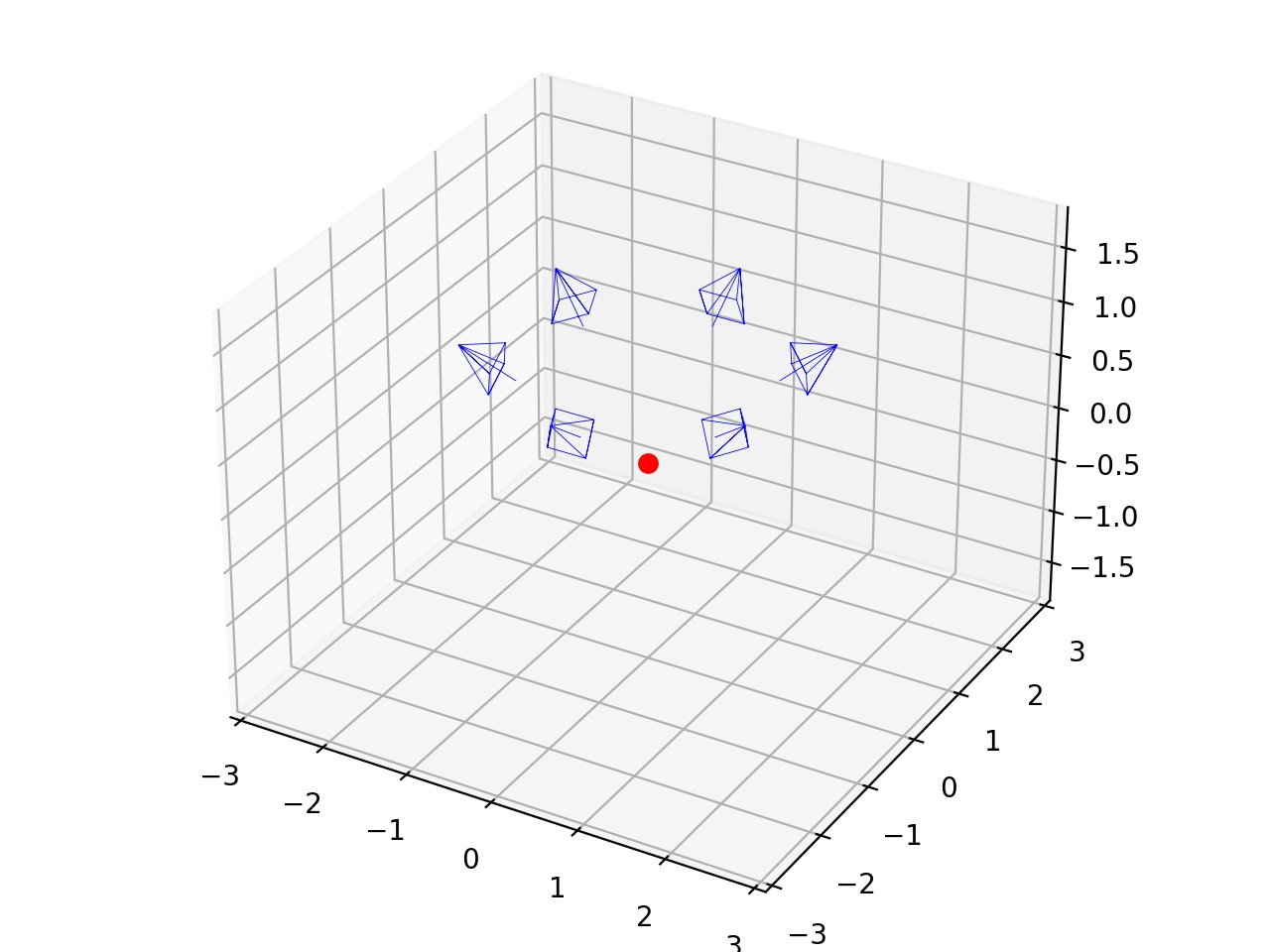} &
\includegraphics[width = 0.7\linewidth ]{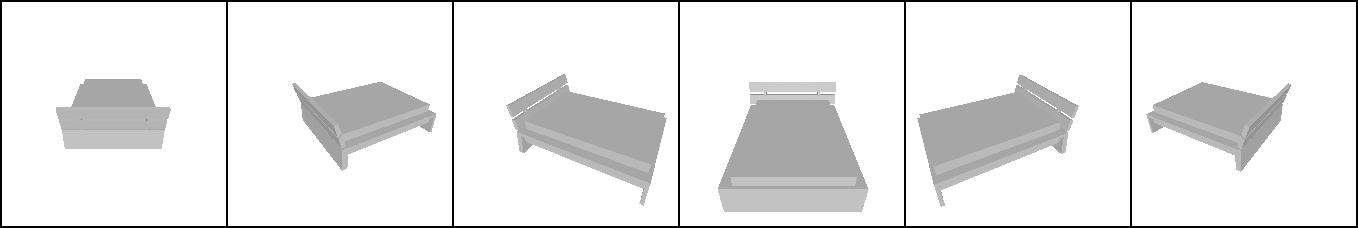} \\ \hline
\textbf{MVTN-Circular:}  &\includegraphics[trim= 4cm 2.7cm 4cm 2.2cm , clip, width = 0.135\linewidth]{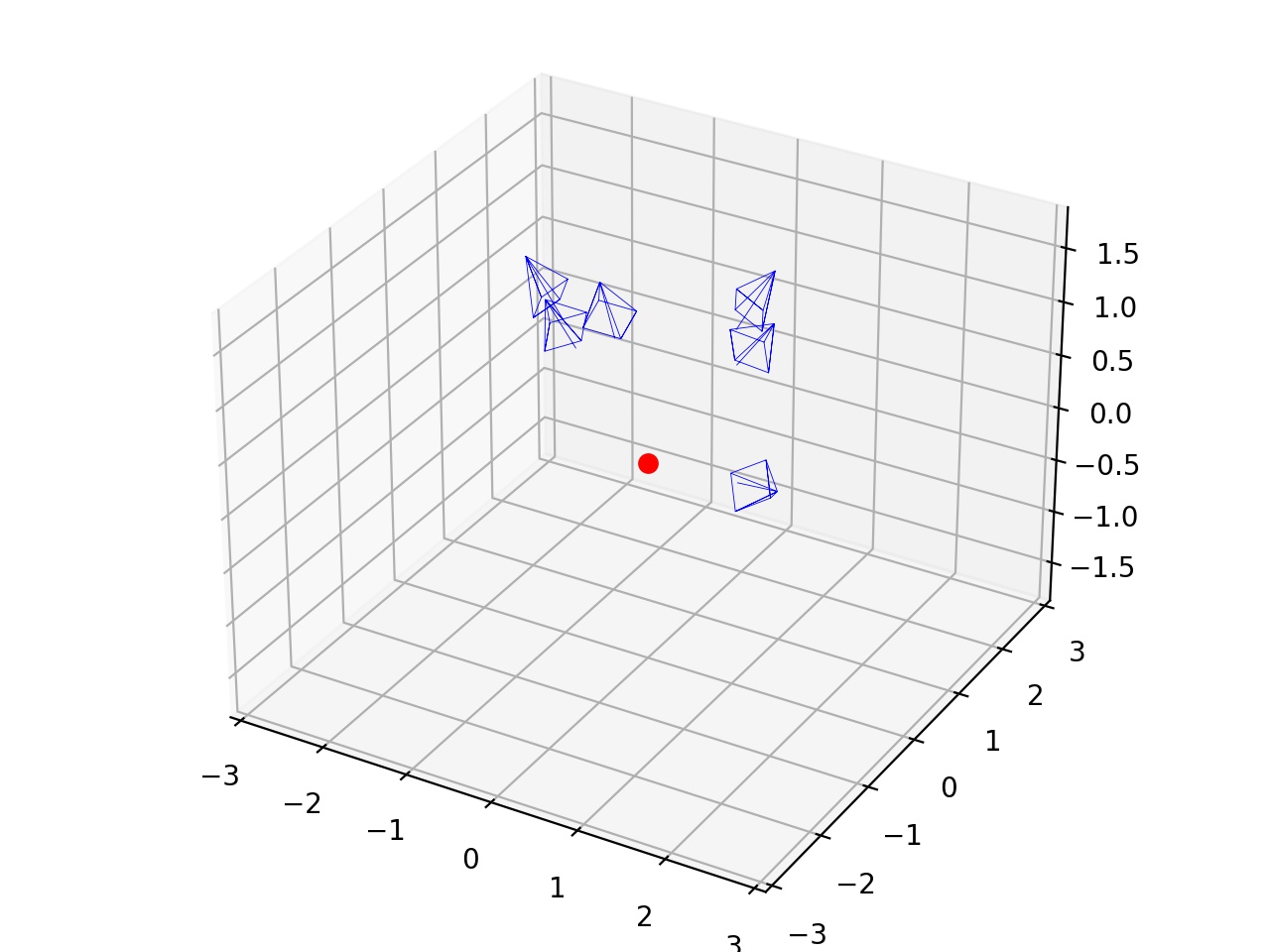} &
\includegraphics[width = 0.7\linewidth]{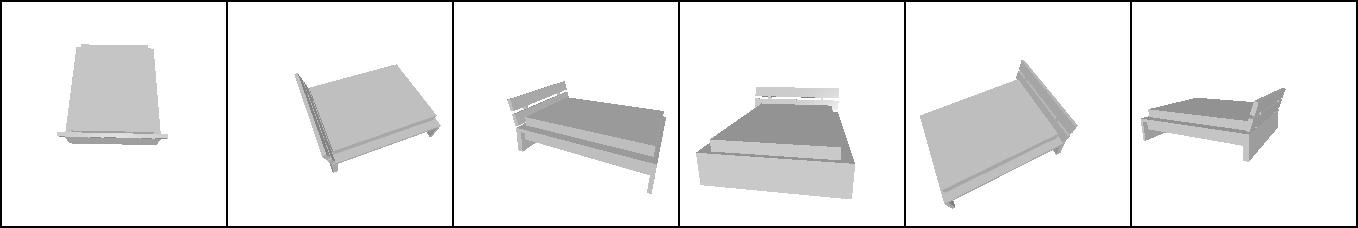} \\ \hline
 \textbf{Spherical:} \hspace{10pt} &\includegraphics[trim= 4cm 2.7cm 4cm 2.2cm , clip, width = 0.135\linewidth]{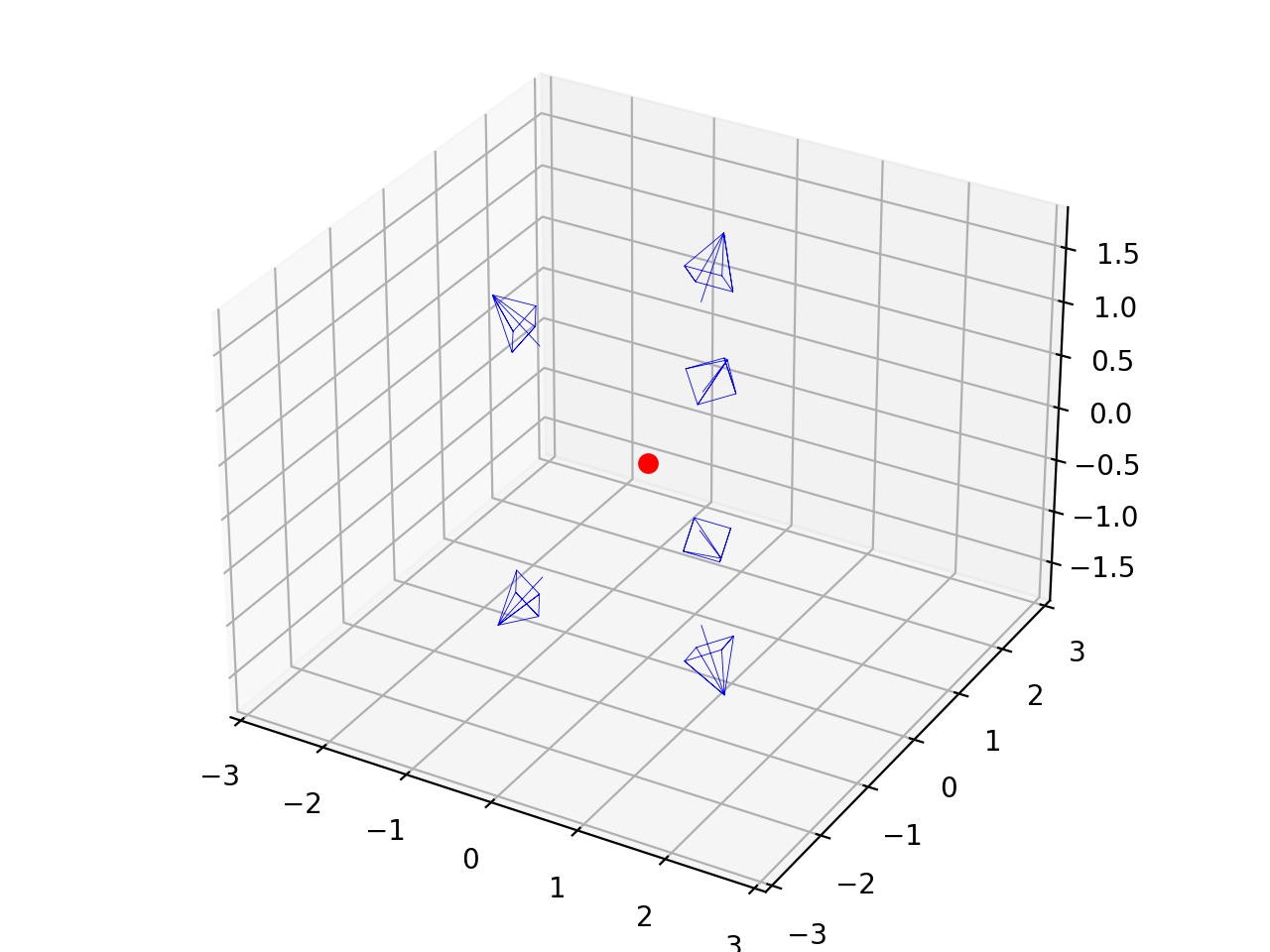} &
\includegraphics[width = 0.7\linewidth]{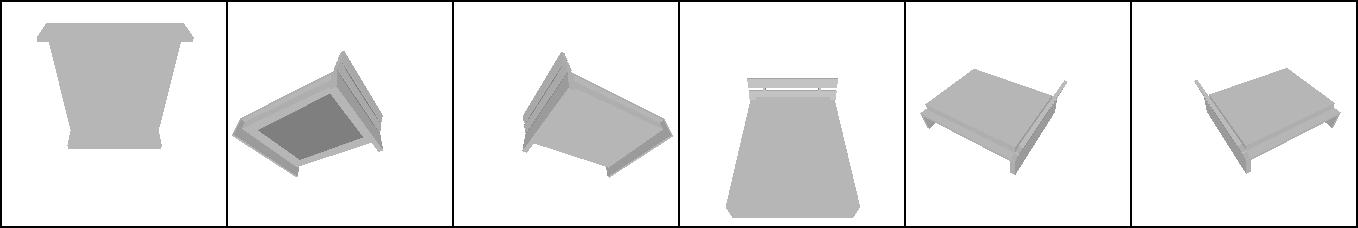} \\ \hline
 \textbf{MVTN-Spherical:}  & \includegraphics[trim= 4cm 2.7cm 4cm 2.2cm , clip, width = 0.135\linewidth]{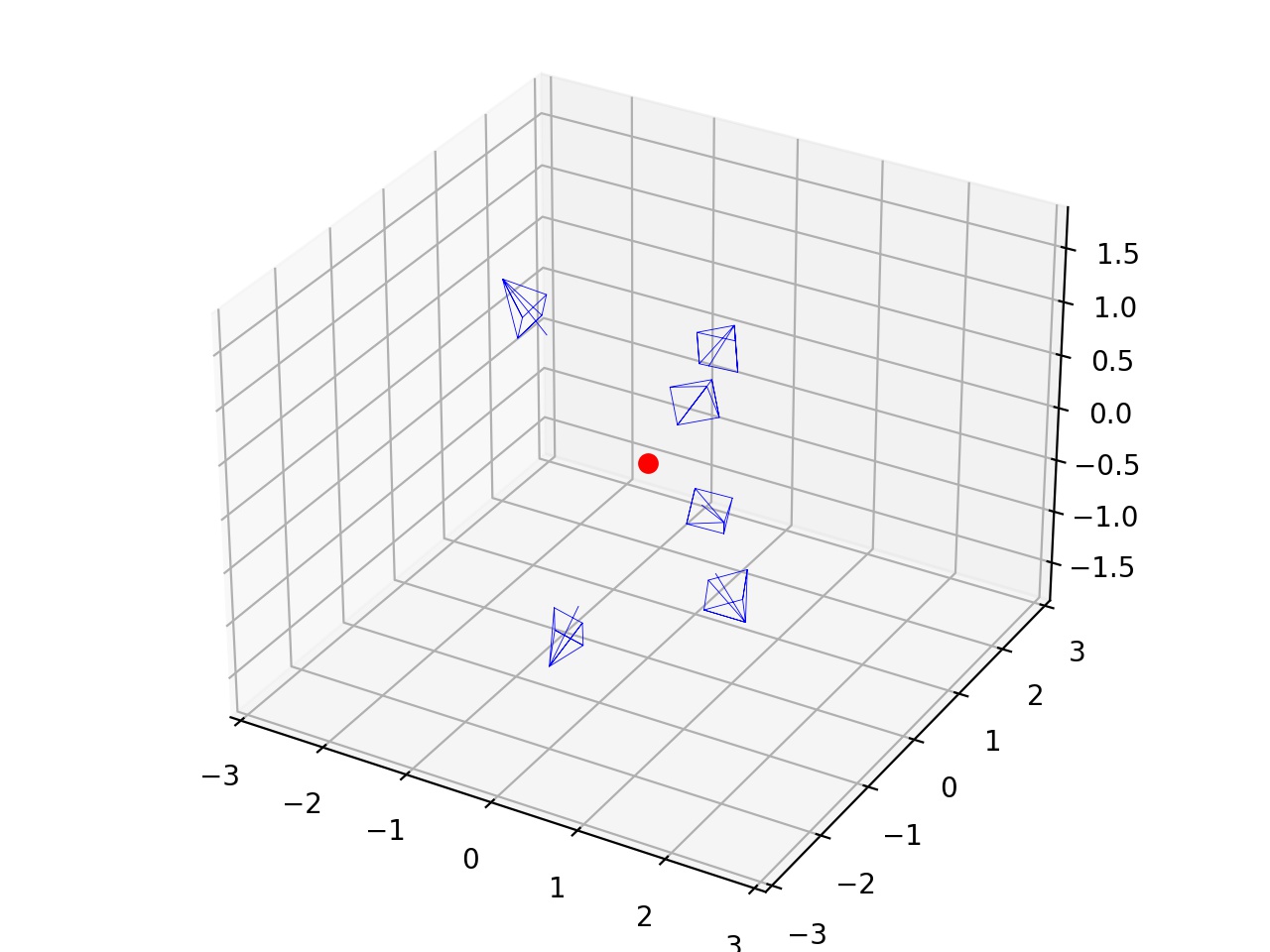} &
\includegraphics[width = 0.7\linewidth]{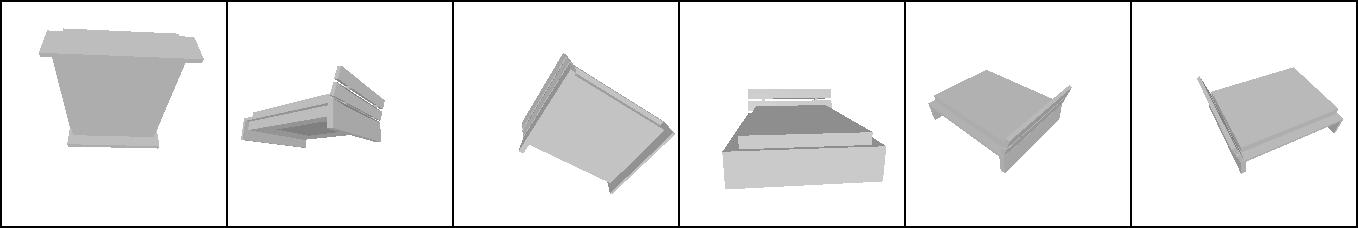} \\ 
\bottomrule
\end{tabular}
}
\vspace{2pt}
\caption{  \textbf{Qualitative Examples for MVTN predicted views (I)}: The view setups commonly followed in the multi-view literature are circular \cite{mvcnn} or spherical \cite{mvviewgcn,mvrotationnet}. The red dot is the center of the object. MVTN-circular/MVTN-spherical are trained to predict the views as offsets to these common configurations. Note that MVTN adjusts the original views to make the 3D object better represented by the multi-view images.}
    \label{fig:views-mvt-sup-1}
\end{figure*}

\begin{figure*} [h] 
\centering
\tabcolsep=0.03cm
\resizebox{0.9\linewidth}{!}{
\begin{tabular}{lc|c}  
\toprule

\textbf{Circular:} \hspace{10pt}  & \includegraphics[trim= 4cm 2.7cm 4cm 2.2cm , clip, width = 0.135\linewidth]{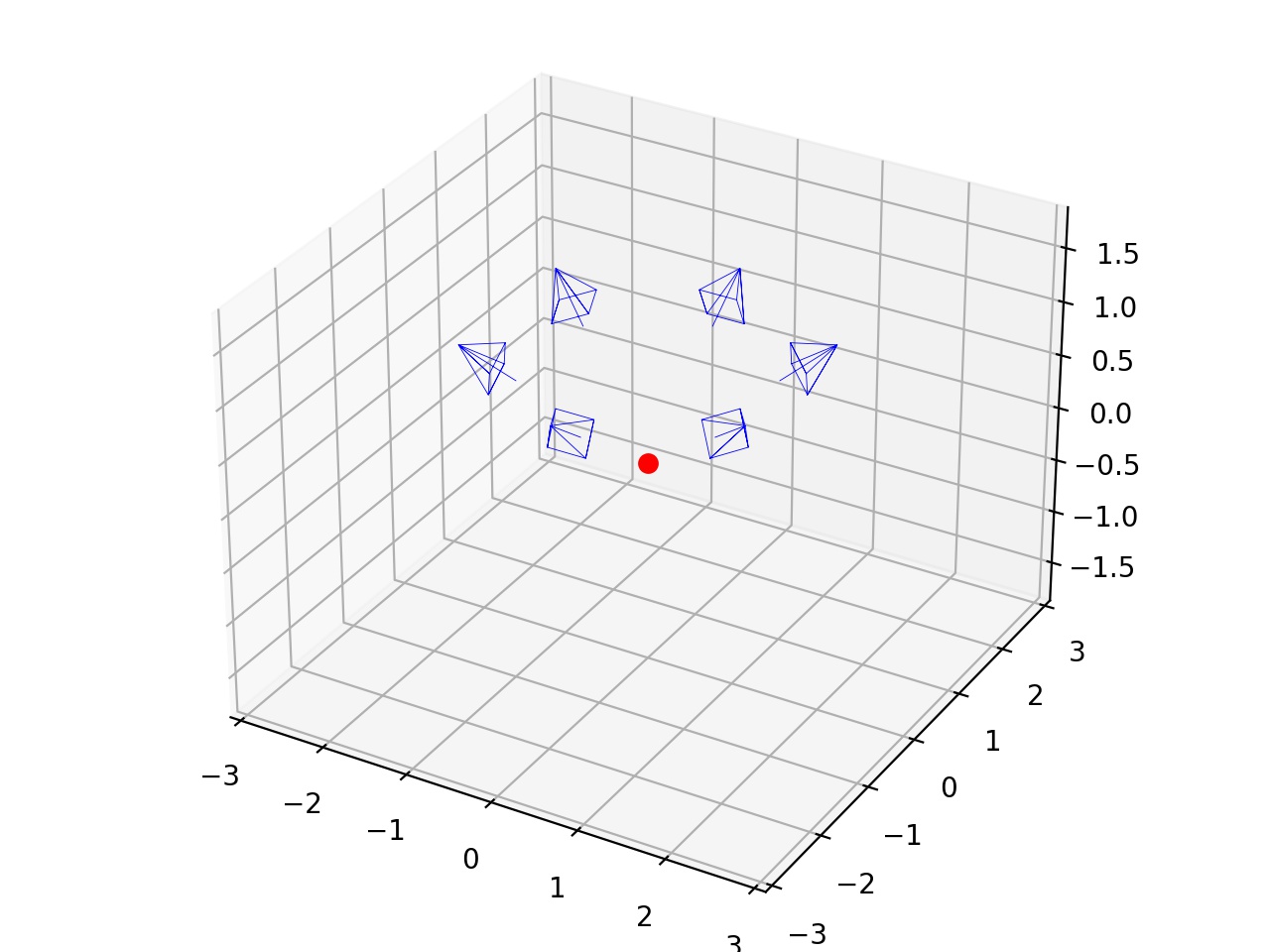} &
\includegraphics[width = 0.7\linewidth ]{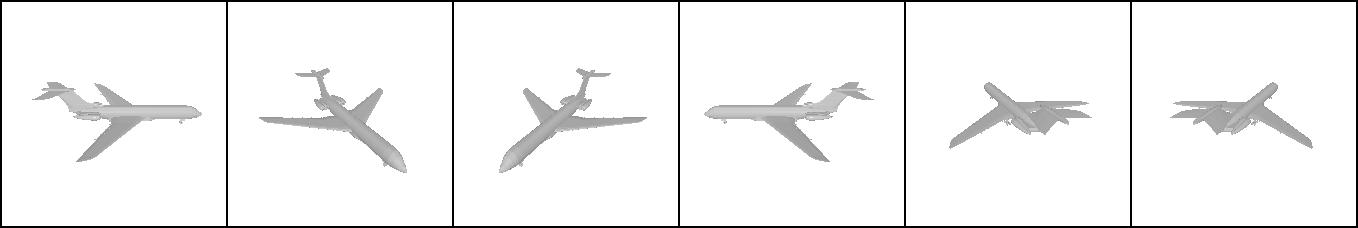} \\ \hline
\textbf{MVTN-Circular:}  &\includegraphics[trim= 4cm 2.7cm 4cm 2.2cm , clip, width = 0.135\linewidth]{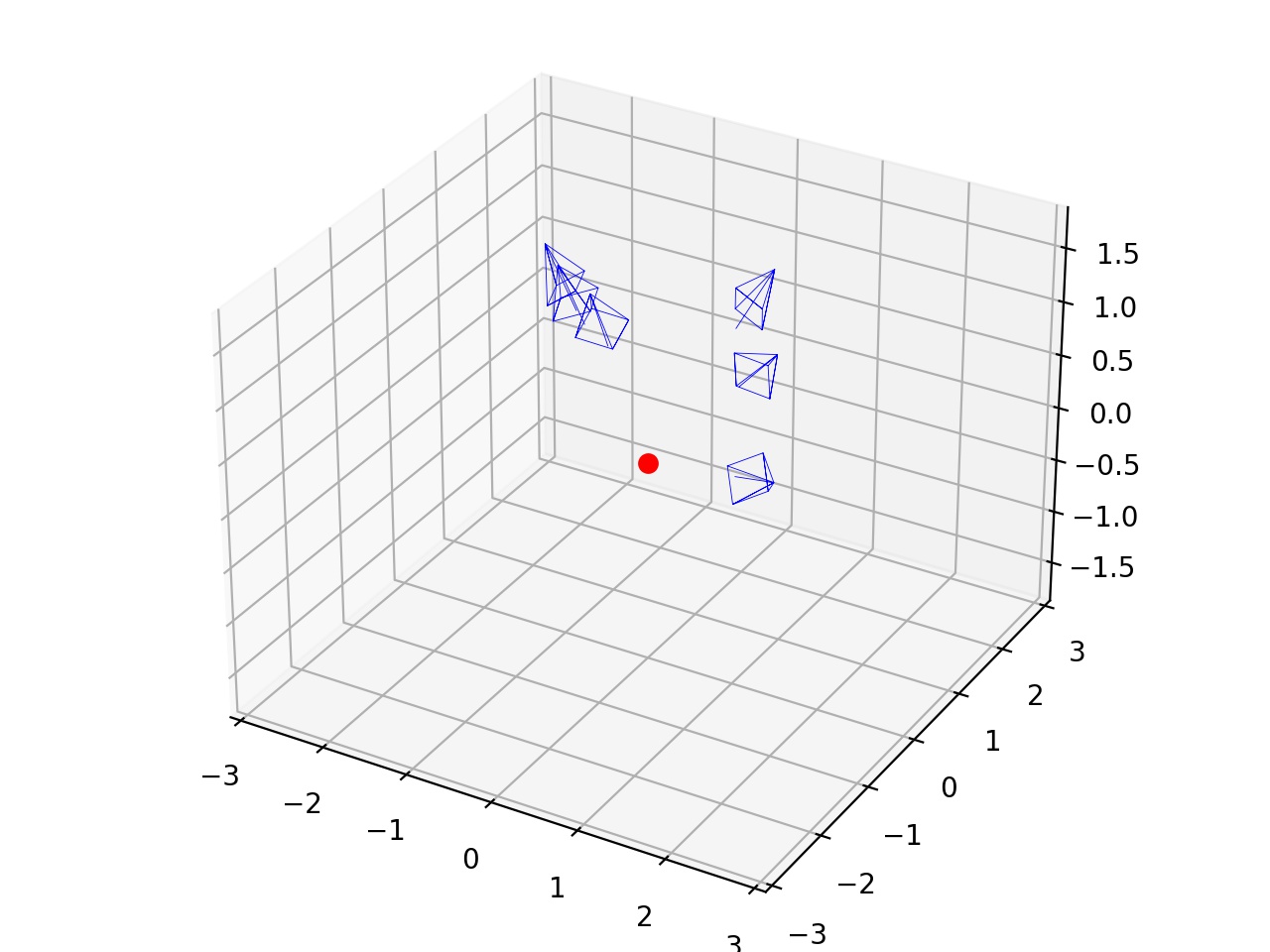} &
\includegraphics[width = 0.7\linewidth]{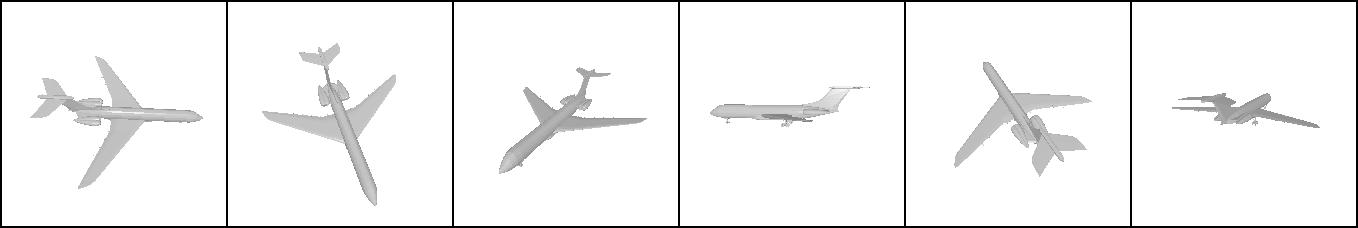} \\ \hline
 \textbf{Spherical:} \hspace{10pt} &\includegraphics[trim= 4cm 2.7cm 4cm 2.2cm , clip, width = 0.135\linewidth]{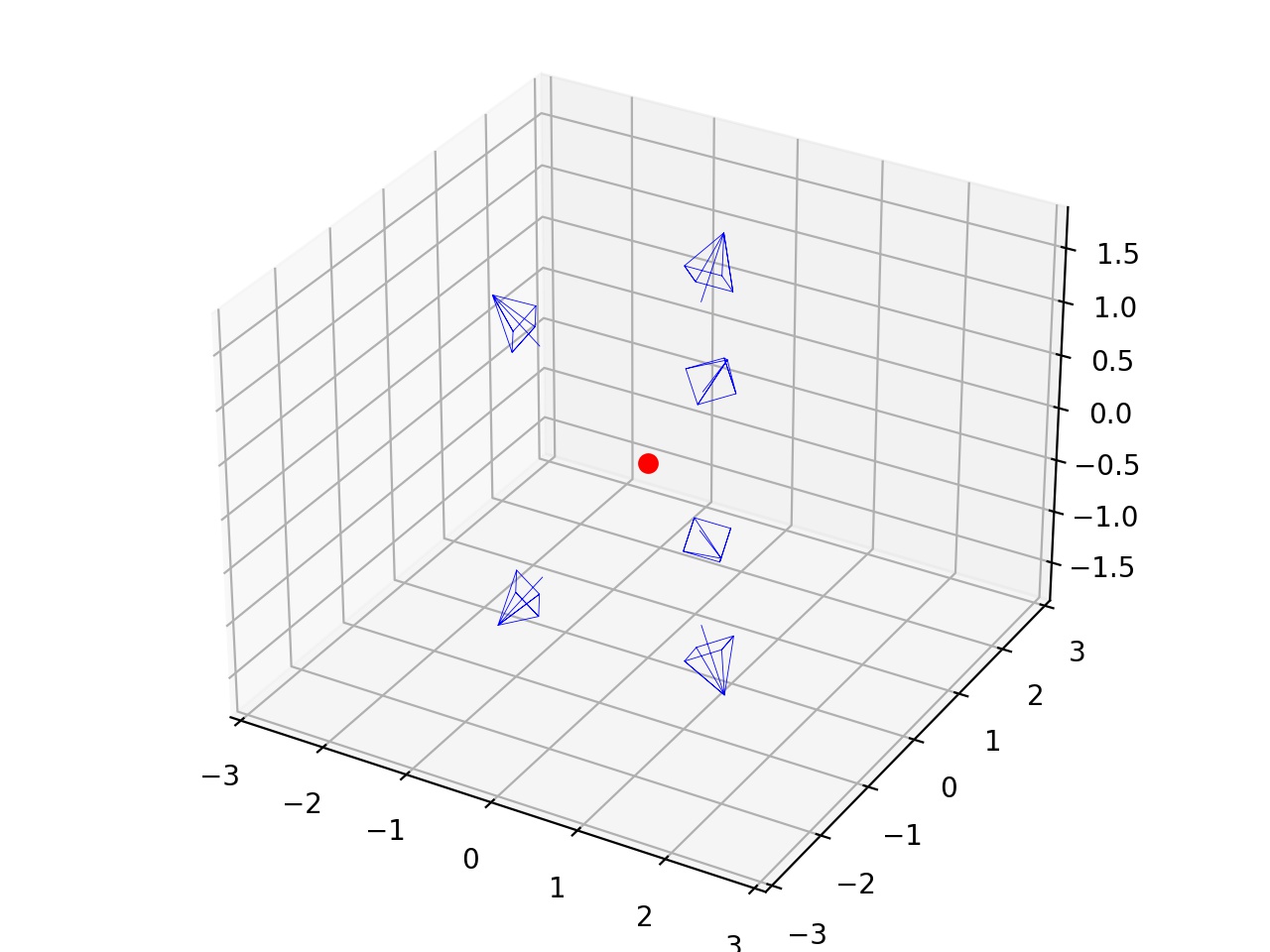} &
\includegraphics[width = 0.7\linewidth]{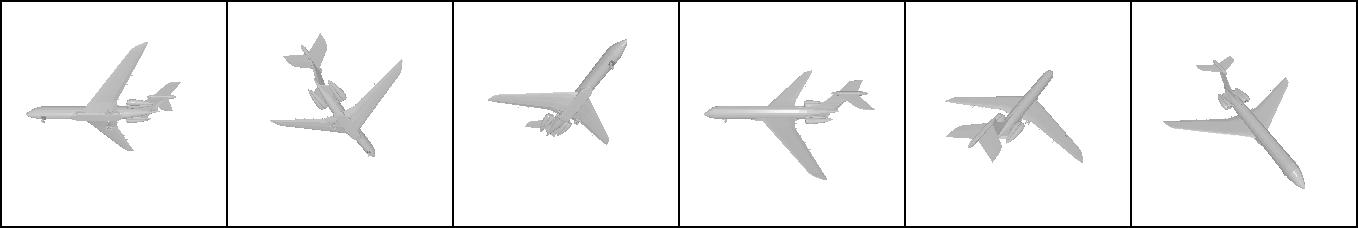} \\ \hline
 \textbf{MVTN-Spherical:}  & \includegraphics[trim= 4cm 2.7cm 4cm 2.2cm , clip, width = 0.135\linewidth]{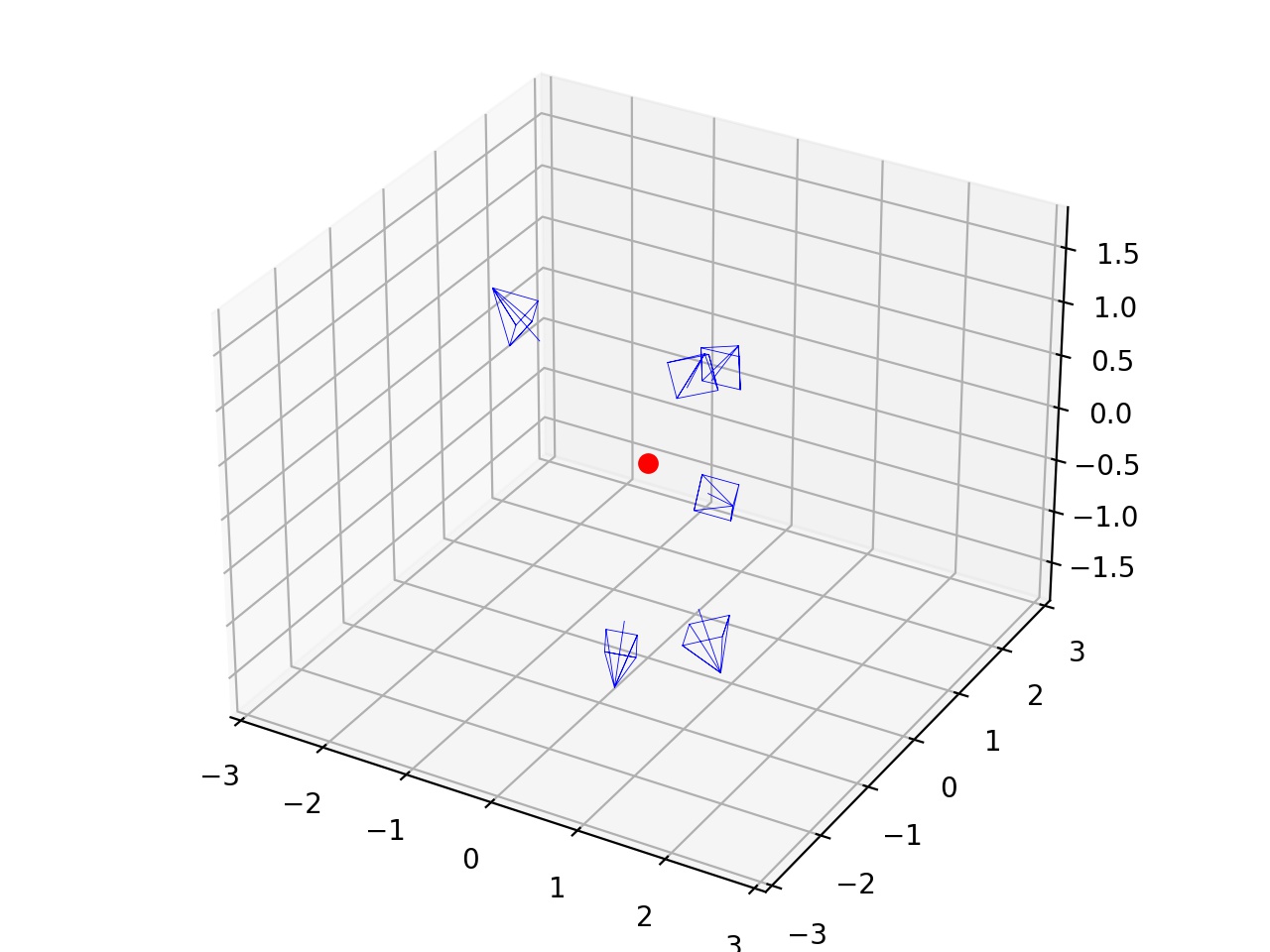} &
\includegraphics[width = 0.7\linewidth]{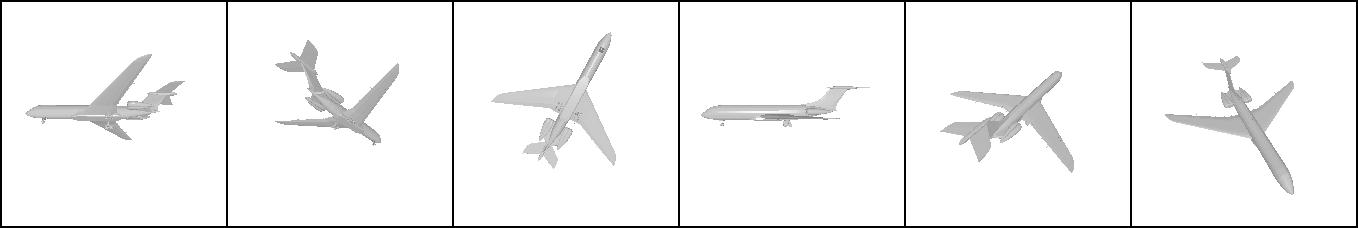} \\ \midrule \midrule

\textbf{Circular:} \hspace{10pt}  & \includegraphics[trim= 4cm 2.7cm 4cm 2.2cm , clip, width = 0.135\linewidth]{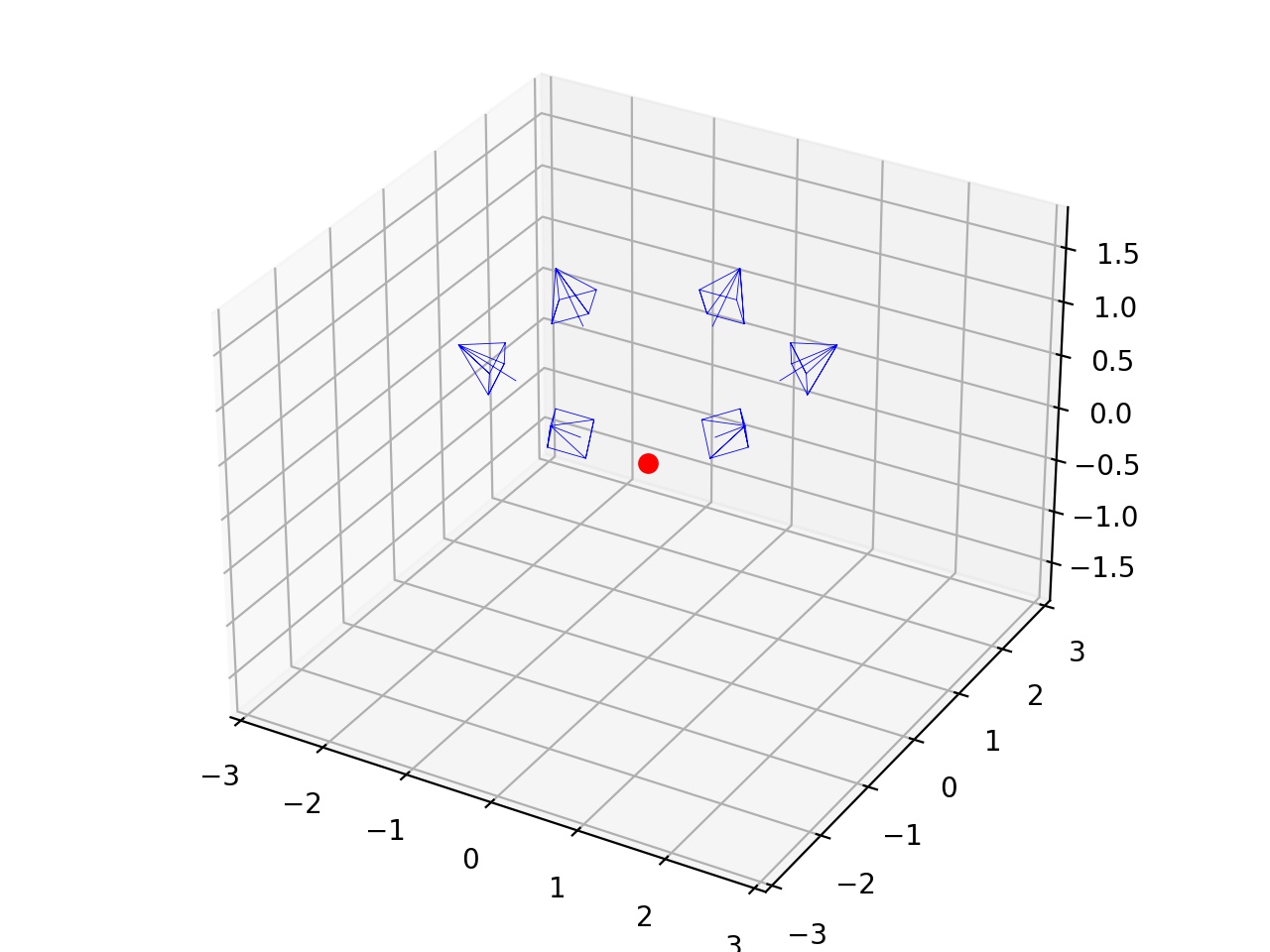} &
\includegraphics[width = 0.7\linewidth ]{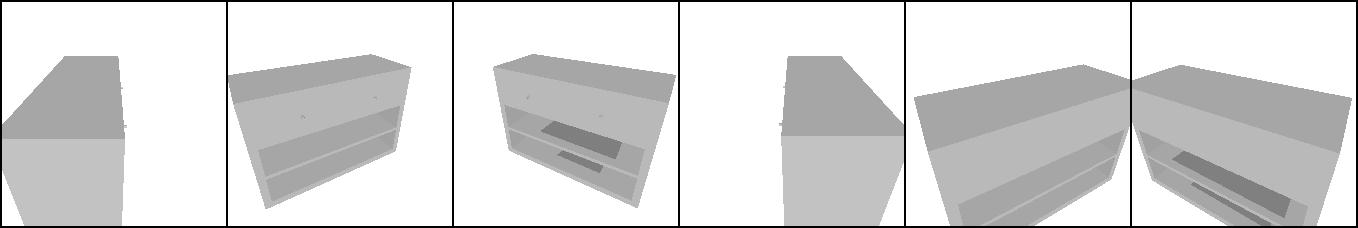} \\ \hline
\textbf{MVTN-Circular:}  &\includegraphics[trim= 4cm 2.7cm 4cm 2.2cm , clip, width = 0.135\linewidth]{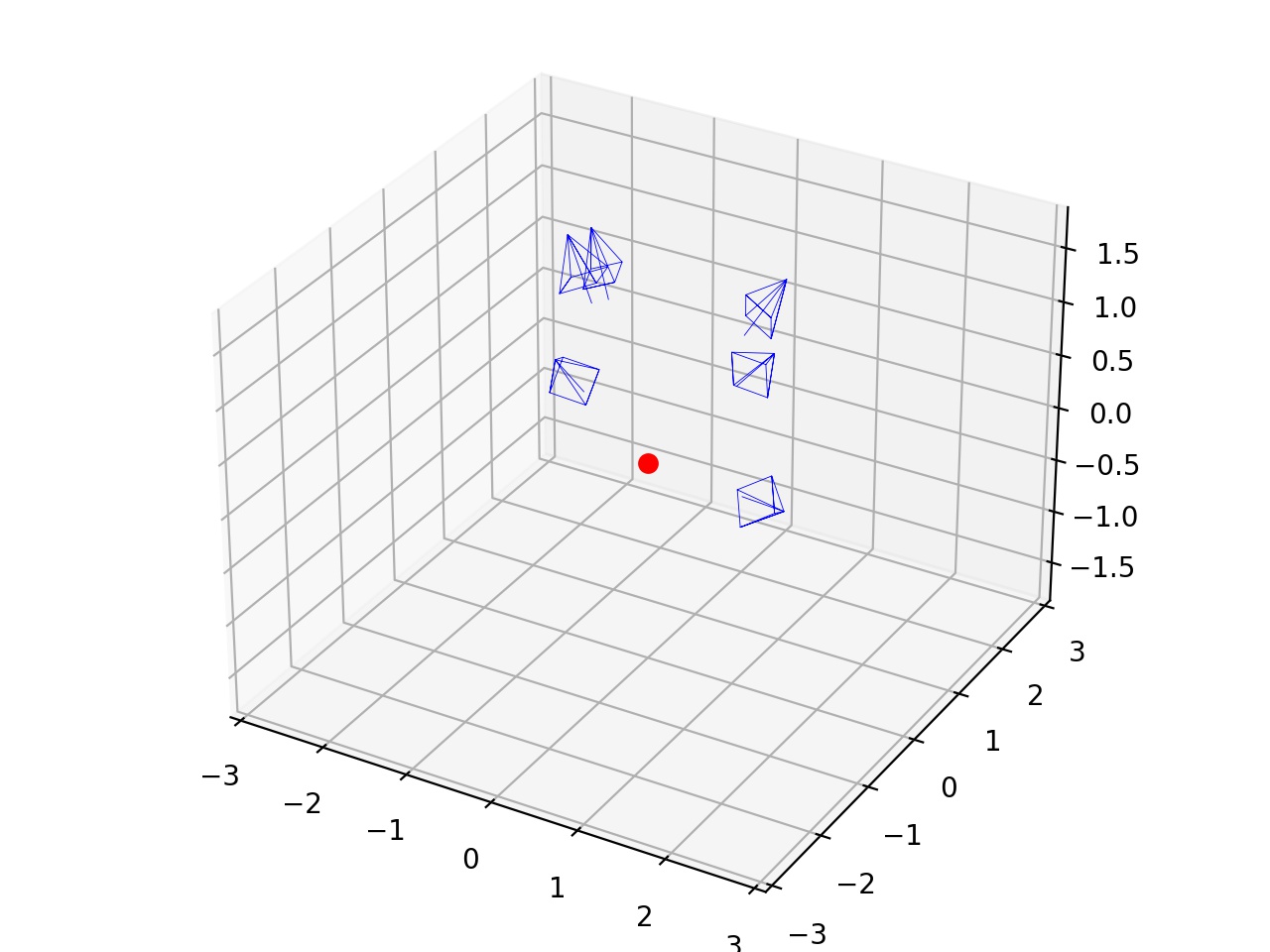} &
\includegraphics[width = 0.7\linewidth]{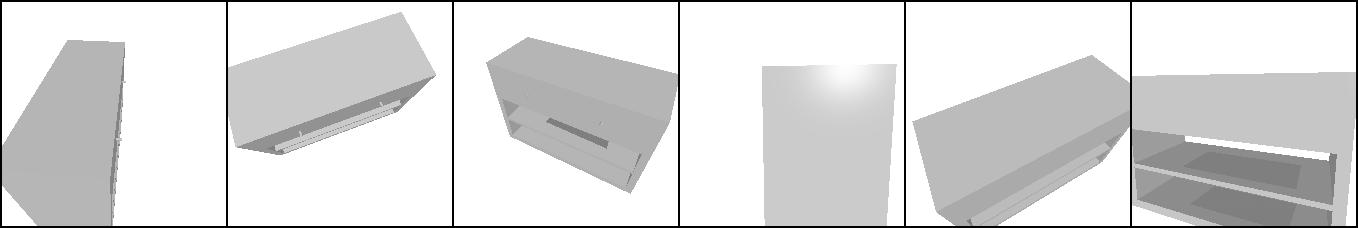} \\ \hline
 \textbf{Spherical:} \hspace{10pt} &\includegraphics[trim= 4cm 2.7cm 4cm 2.2cm , clip, width = 0.135\linewidth]{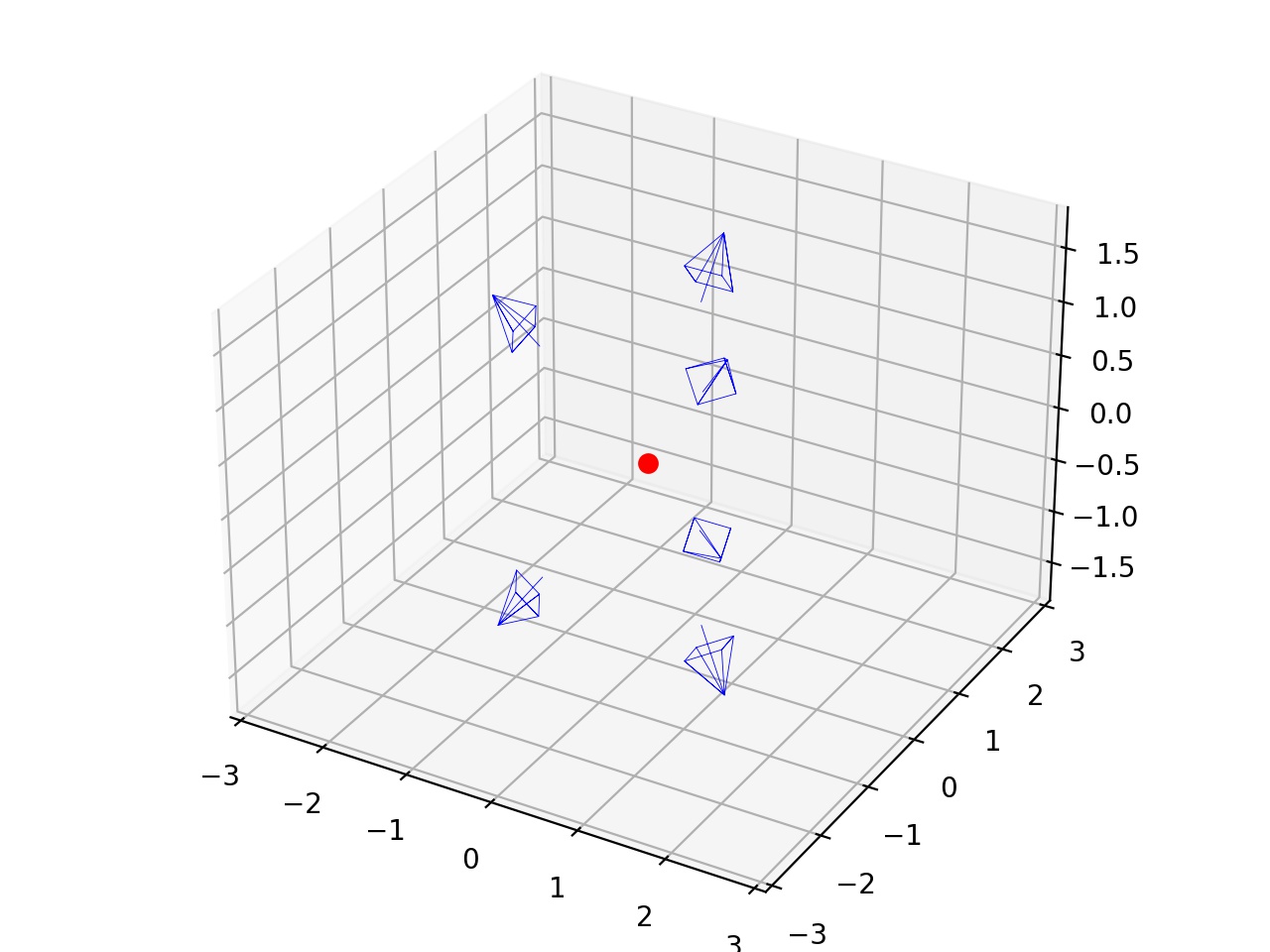} &
\includegraphics[width = 0.7\linewidth]{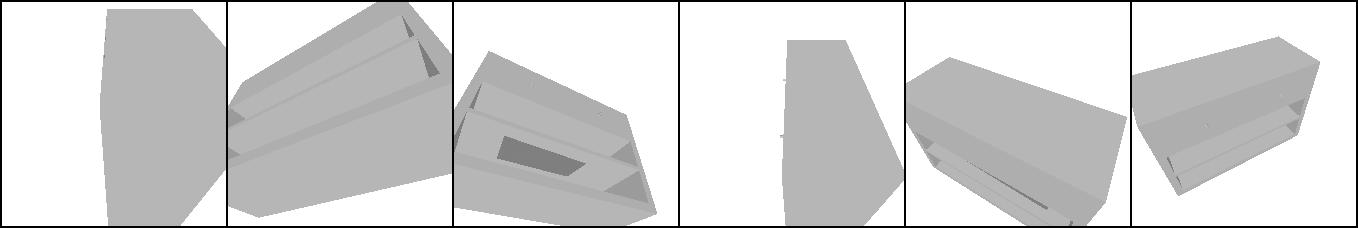} \\ \hline
 \textbf{MVTN-Spherical:}  & \includegraphics[trim= 4cm 2.7cm 4cm 2.2cm , clip, width = 0.135\linewidth]{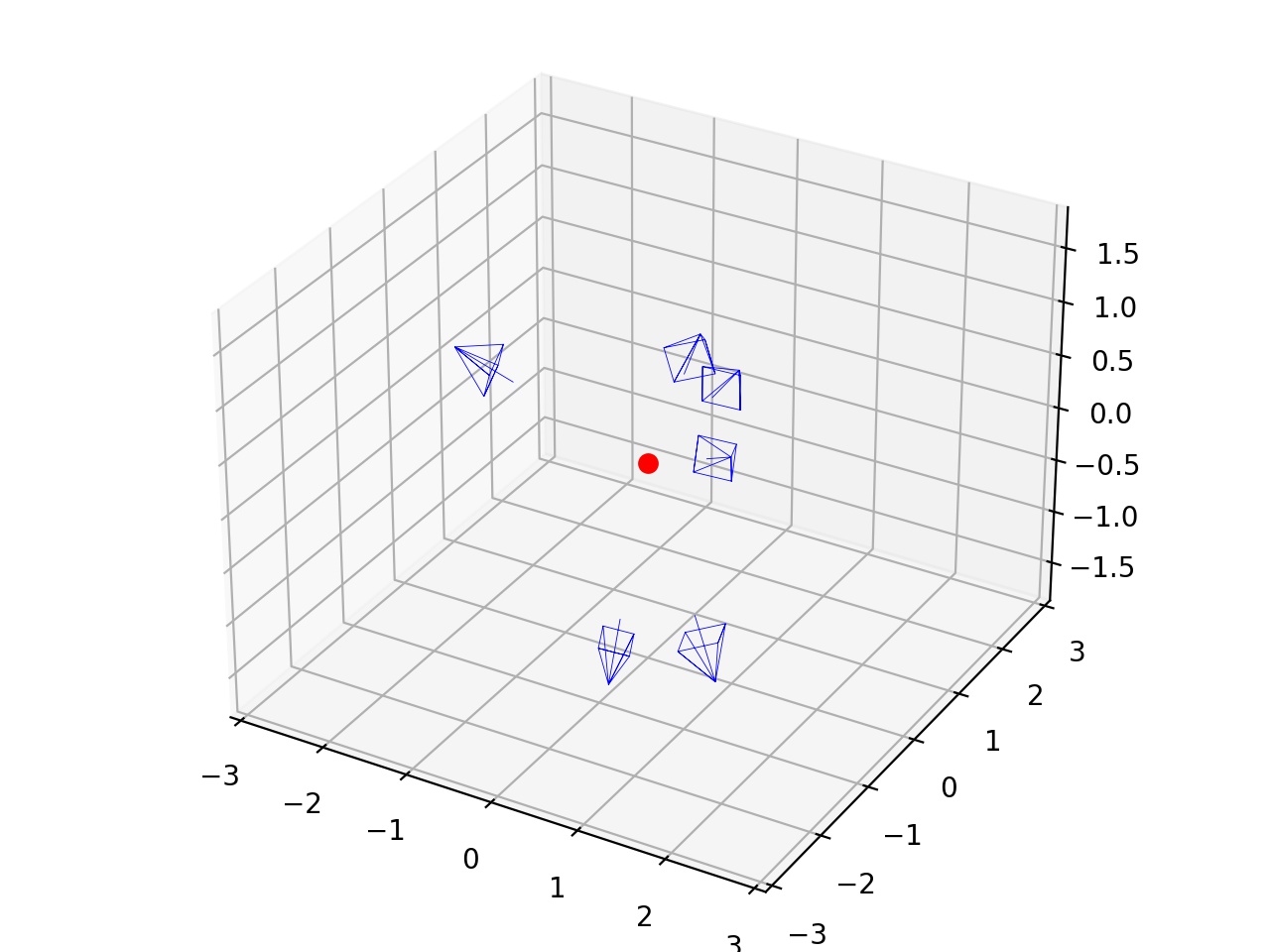} &
\includegraphics[width = 0.7\linewidth]{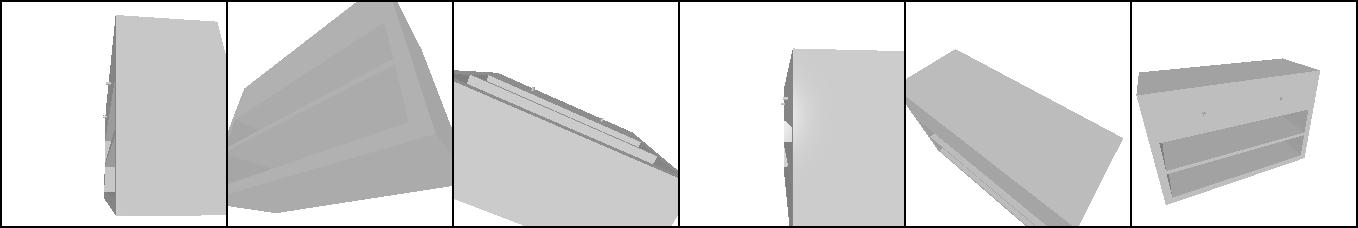} \\
\bottomrule
\end{tabular}
}
\vspace{2pt}
\caption{  \textbf{Qualitative Examples for MVTN predicted views (II)}: The view setups commonly followed in the multi-view literature are circular \cite{mvcnn} or spherical \cite{mvviewgcn,mvrotationnet}. The red dot is the center of the object.  MVTN-circular/MVTN-spherical are trained to predict the views as offsets to these common configurations. Note that MVTN adjusts the original views to make the 3D object better represented by the multi-view images.}
    \label{fig:views-mvt-sup-2}
\end{figure*}

We visualize the distribution of predicted views by MVTN for specific classes in \figLabel{\ref{fig:distribution-sup}}. This is done to ensure that MVTN is learning per-instance views and regressing the same views for the entire class (collapse scenario). We can see that the MVTN distribution of the views varies from one class to another, and the views themselves on the same class have some variance from one instance to another. We also show specific examples for predicted views in \figLabel{\ref{fig:views-mvt-sup-1}, \ref{fig:views-mvt-sup-2}}. Here, we show both the predicted camera view-points and the renderings from these cameras. Note how MVTN shifts every view to better show the discriminative details about the 3D object.
To test that these views are per-instance, we average all the views predicted by our 4-view MVTN for every class and test the trained MVCNN on these fixed per-class views. In this setup, MVTN achieves 90.6\% on ModelNet40, as compared to 91.0\% for the per-instance views and 89\% for the fixed views. 

\subsection{Shape Retrieval Examples}
\begin{figure*} [h] 
\centering
\tabcolsep=0.03cm
\resizebox{0.9\linewidth}{!}{
\begin{tabular}{c|cccccccccc}

\includegraphics[width = 0.09090909090909091\linewidth]{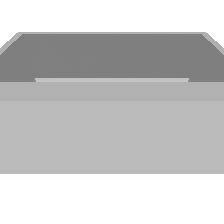} &
\includegraphics[width = 0.09090909090909091\linewidth]{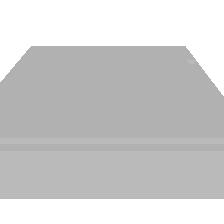}  &
\includegraphics[width = 0.09090909090909091\linewidth]{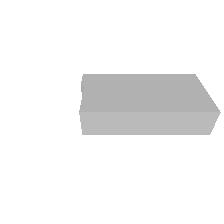}  &
\includegraphics[width = 0.09090909090909091\linewidth]{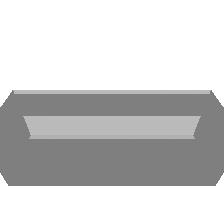}  &
\includegraphics[width = 0.09090909090909091\linewidth]{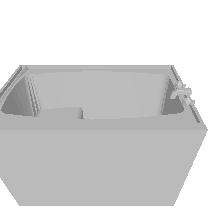}  &
\includegraphics[width = 0.09090909090909091\linewidth]{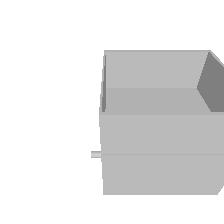}  &
\includegraphics[width = 0.09090909090909091\linewidth]{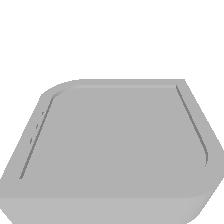}  &
\includegraphics[width = 0.09090909090909091\linewidth]{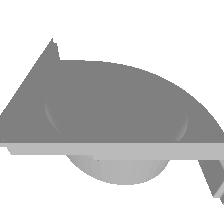}  &
\includegraphics[width = 0.09090909090909091\linewidth]{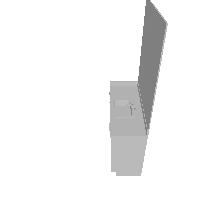}  &
\includegraphics[width = 0.09090909090909091\linewidth]{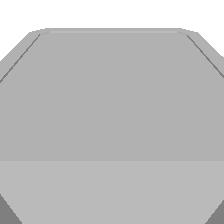}  &
\includegraphics[width = 0.09090909090909091\linewidth]{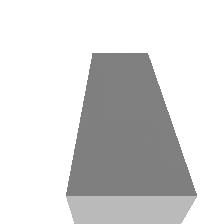}  \\

\includegraphics[width = 0.09090909090909091\linewidth]{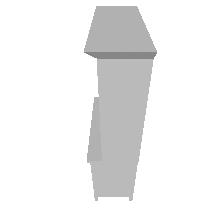} &
\includegraphics[width = 0.09090909090909091\linewidth]{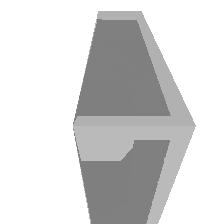}  &
\includegraphics[width = 0.09090909090909091\linewidth]{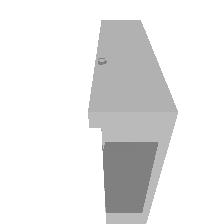}  &
\includegraphics[width = 0.09090909090909091\linewidth]{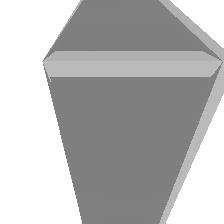}  &
\includegraphics[width = 0.09090909090909091\linewidth]{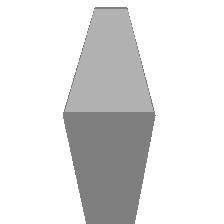}  &
\includegraphics[width = 0.09090909090909091\linewidth]{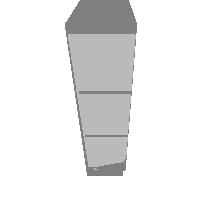}  &
\includegraphics[width = 0.09090909090909091\linewidth]{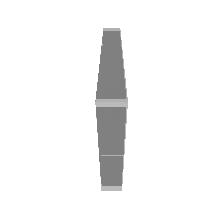}  &
\includegraphics[width = 0.09090909090909091\linewidth]{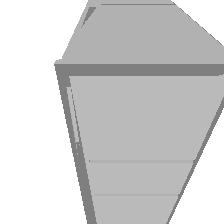}  &
\includegraphics[width = 0.09090909090909091\linewidth]{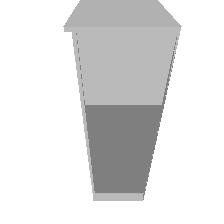}  &
\includegraphics[width = 0.09090909090909091\linewidth]{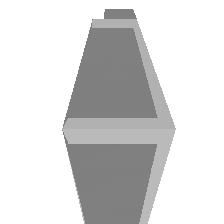}  &
\includegraphics[width = 0.09090909090909091\linewidth]{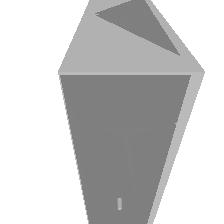}  \\

\includegraphics[width = 0.09090909090909091\linewidth]{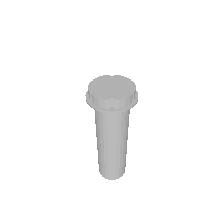} &
\includegraphics[width = 0.09090909090909091\linewidth]{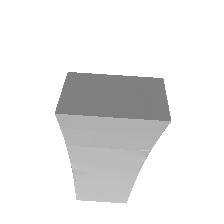}  &
\includegraphics[width = 0.09090909090909091\linewidth]{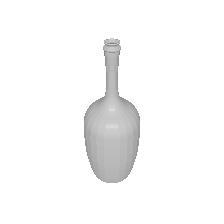}  &
\includegraphics[width = 0.09090909090909091\linewidth]{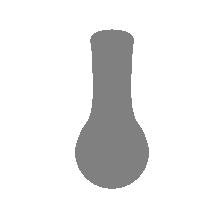}  &
\includegraphics[width = 0.09090909090909091\linewidth]{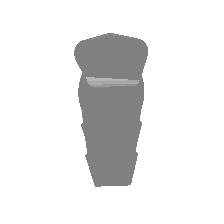}  &
\includegraphics[width = 0.09090909090909091\linewidth]{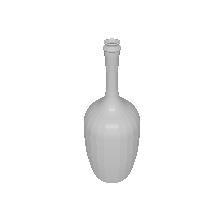}  &
\includegraphics[width = 0.09090909090909091\linewidth]{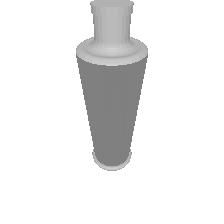}  &
\includegraphics[width = 0.09090909090909091\linewidth]{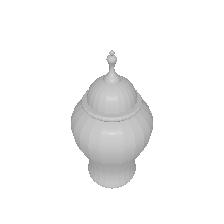}  &
\includegraphics[width = 0.09090909090909091\linewidth]{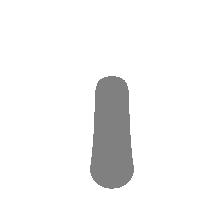}  &
\includegraphics[width = 0.09090909090909091\linewidth]{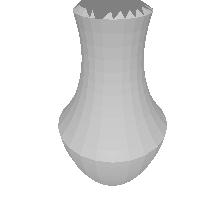}  &
\includegraphics[width = 0.09090909090909091\linewidth]{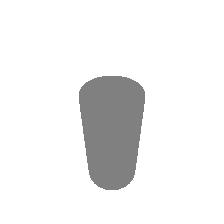}  \\

\includegraphics[width = 0.09090909090909091\linewidth]{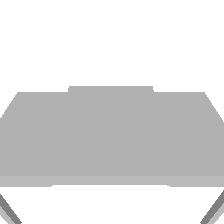} &
\includegraphics[width = 0.09090909090909091\linewidth]{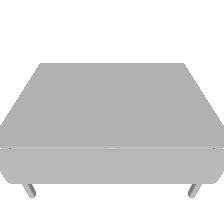}  &
\includegraphics[width = 0.09090909090909091\linewidth]{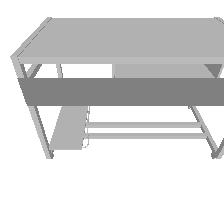}  &
\includegraphics[width = 0.09090909090909091\linewidth]{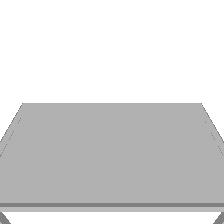}  &
\includegraphics[width = 0.09090909090909091\linewidth]{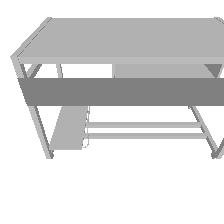}  &
\includegraphics[width = 0.09090909090909091\linewidth]{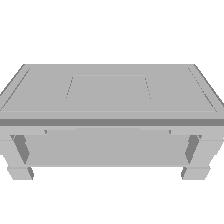}  &
\includegraphics[width = 0.09090909090909091\linewidth]{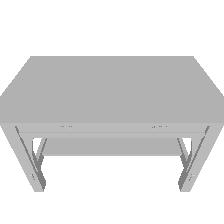}  &
\includegraphics[width = 0.09090909090909091\linewidth]{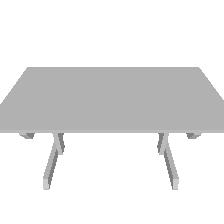}  &
\includegraphics[width = 0.09090909090909091\linewidth]{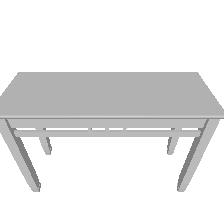}  &
\includegraphics[width = 0.09090909090909091\linewidth]{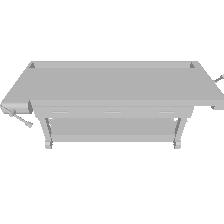}  &
\includegraphics[width = 0.09090909090909091\linewidth]{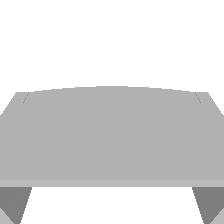}  \\

\includegraphics[width = 0.09090909090909091\linewidth]{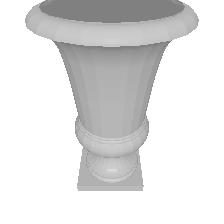} &
\includegraphics[width = 0.09090909090909091\linewidth]{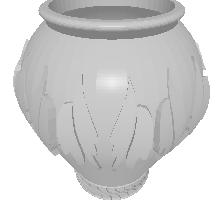}  &
\includegraphics[width = 0.09090909090909091\linewidth]{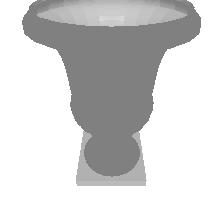}  &
\includegraphics[width = 0.09090909090909091\linewidth]{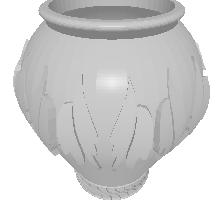}  &
\includegraphics[width = 0.09090909090909091\linewidth]{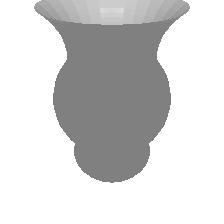}  &
\includegraphics[width = 0.09090909090909091\linewidth]{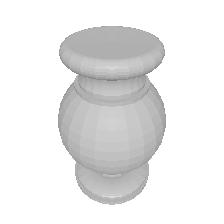}  &
\includegraphics[width = 0.09090909090909091\linewidth]{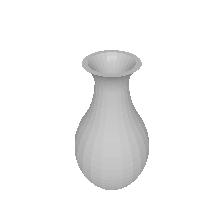}  &
\includegraphics[width = 0.09090909090909091\linewidth]{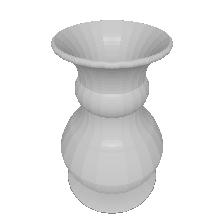}  &
\includegraphics[width = 0.09090909090909091\linewidth]{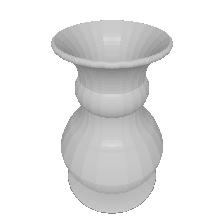}  &
\includegraphics[width = 0.09090909090909091\linewidth]{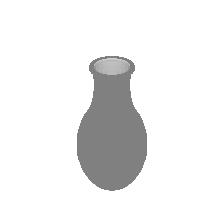}  &
\includegraphics[width = 0.09090909090909091\linewidth]{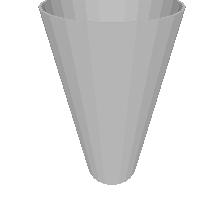}  \\

\includegraphics[width = 0.09090909090909091\linewidth]{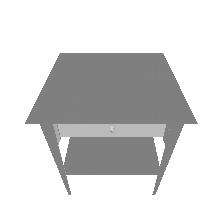} &
\includegraphics[width = 0.09090909090909091\linewidth]{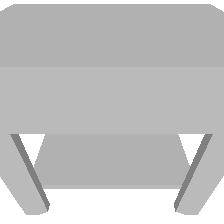}  &
\includegraphics[width = 0.09090909090909091\linewidth]{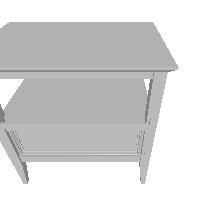}  &
\includegraphics[width = 0.09090909090909091\linewidth]{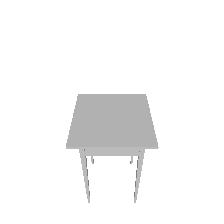}  &
\includegraphics[width = 0.09090909090909091\linewidth]{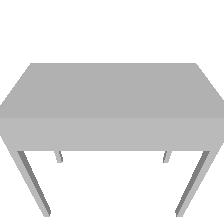}  &
\includegraphics[width = 0.09090909090909091\linewidth]{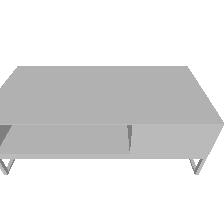}  &
\includegraphics[width = 0.09090909090909091\linewidth]{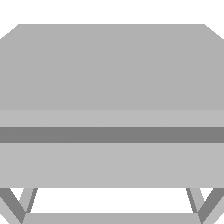}  &
\includegraphics[width = 0.09090909090909091\linewidth]{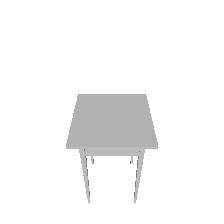}  &
\includegraphics[width = 0.09090909090909091\linewidth]{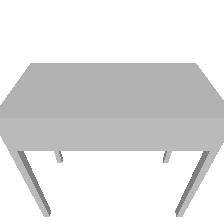}  &
\includegraphics[width = 0.09090909090909091\linewidth]{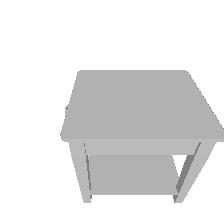}  &
\includegraphics[width = 0.09090909090909091\linewidth]{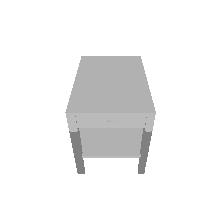}  \\

\includegraphics[width = 0.09090909090909091\linewidth]{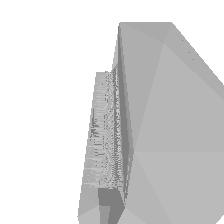} &
\includegraphics[width = 0.09090909090909091\linewidth]{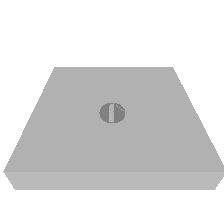}  &
\includegraphics[width = 0.09090909090909091\linewidth]{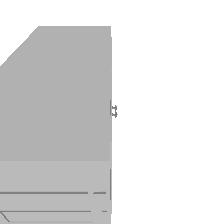}  &
\includegraphics[width = 0.09090909090909091\linewidth]{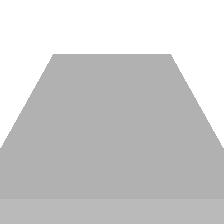}  &
\includegraphics[width = 0.09090909090909091\linewidth]{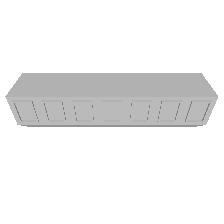}  &
\includegraphics[width = 0.09090909090909091\linewidth]{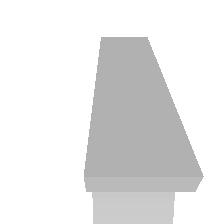}  &
\includegraphics[width = 0.09090909090909091\linewidth]{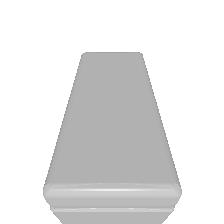}  &
\includegraphics[width = 0.09090909090909091\linewidth]{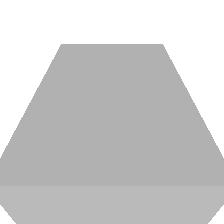}  &
\includegraphics[width = 0.09090909090909091\linewidth]{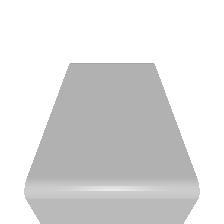}  &
\includegraphics[width = 0.09090909090909091\linewidth]{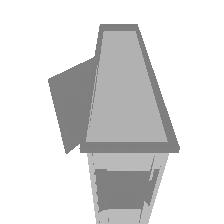}  &
\includegraphics[width = 0.09090909090909091\linewidth]{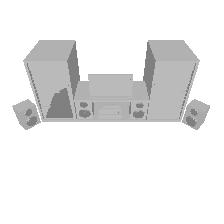}  \\

\includegraphics[width = 0.09090909090909091\linewidth]{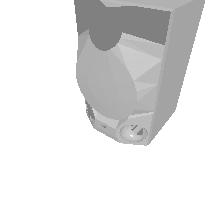} &
\includegraphics[width = 0.09090909090909091\linewidth]{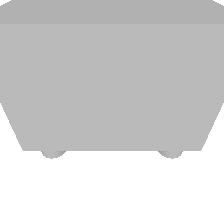}  &
\includegraphics[width = 0.09090909090909091\linewidth]{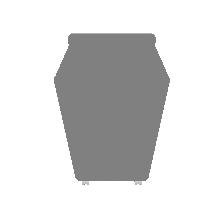}  &
\includegraphics[width = 0.09090909090909091\linewidth]{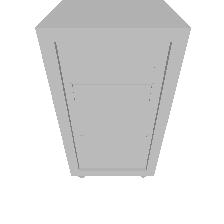}  &
\includegraphics[width = 0.09090909090909091\linewidth]{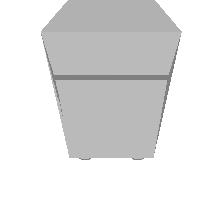}  &
\includegraphics[width = 0.09090909090909091\linewidth]{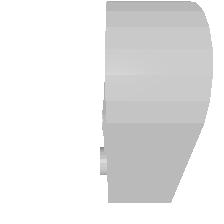}  &
\includegraphics[width = 0.09090909090909091\linewidth]{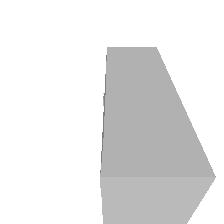}  &
\includegraphics[width = 0.09090909090909091\linewidth]{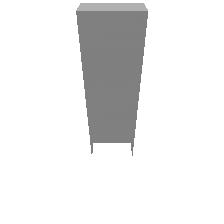}  &
\includegraphics[width = 0.09090909090909091\linewidth]{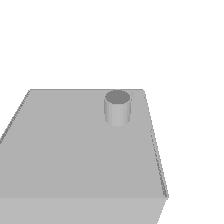}  &
\includegraphics[width = 0.09090909090909091\linewidth]{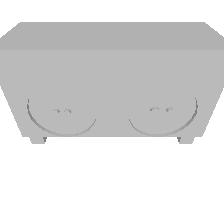}  &
\includegraphics[width = 0.09090909090909091\linewidth]{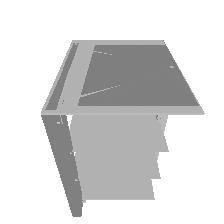}  \\

\includegraphics[width = 0.09090909090909091\linewidth]{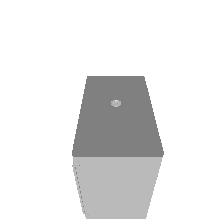} &
\includegraphics[width = 0.09090909090909091\linewidth]{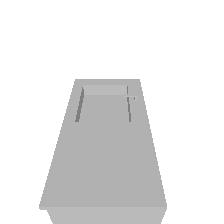}  &
\includegraphics[width = 0.09090909090909091\linewidth]{supimages_retrieval_6996.jpg}  &
\includegraphics[width = 0.09090909090909091\linewidth]{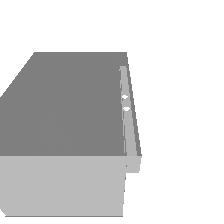}  &
\includegraphics[width = 0.09090909090909091\linewidth]{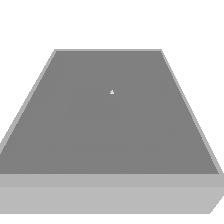}  &
\includegraphics[width = 0.09090909090909091\linewidth]{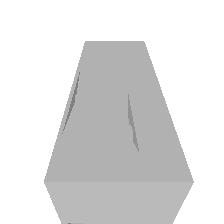}  &
\includegraphics[width = 0.09090909090909091\linewidth]{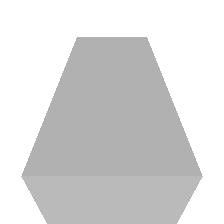}  &
\includegraphics[width = 0.09090909090909091\linewidth]{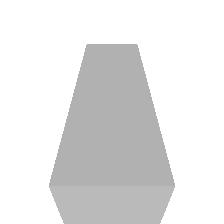}  &
\includegraphics[width = 0.09090909090909091\linewidth]{supimages_retrieval_4585.jpg}  &
\includegraphics[width = 0.09090909090909091\linewidth]{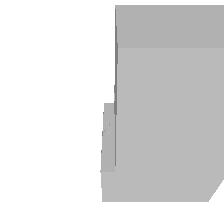}  &
\includegraphics[width = 0.09090909090909091\linewidth]{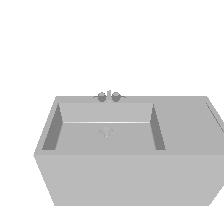}  \\

\includegraphics[width = 0.09090909090909091\linewidth]{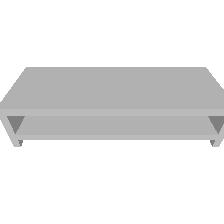} &
\includegraphics[width = 0.09090909090909091\linewidth]{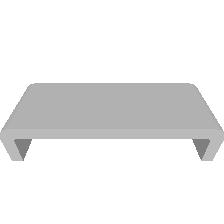}  &
\includegraphics[width = 0.09090909090909091\linewidth]{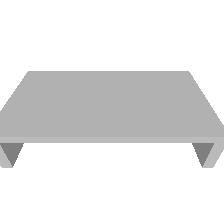}  &
\includegraphics[width = 0.09090909090909091\linewidth]{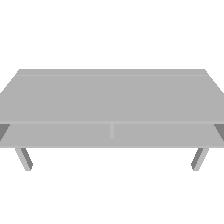}  &
\includegraphics[width = 0.09090909090909091\linewidth]{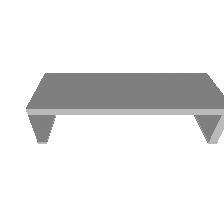}  &
\includegraphics[width = 0.09090909090909091\linewidth]{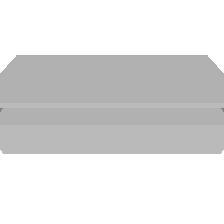}  &
\includegraphics[width = 0.09090909090909091\linewidth]{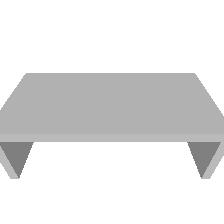}  &
\includegraphics[width = 0.09090909090909091\linewidth]{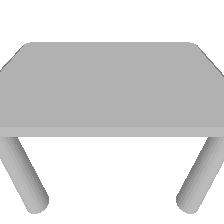}  &
\includegraphics[width = 0.09090909090909091\linewidth]{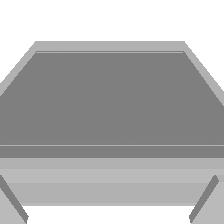}  &
\includegraphics[width = 0.09090909090909091\linewidth]{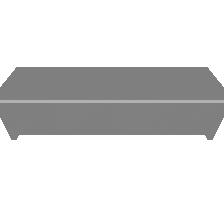}  &
\includegraphics[width = 0.09090909090909091\linewidth]{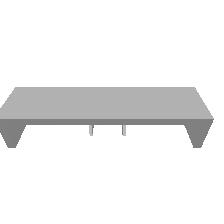}  \\

\includegraphics[width = 0.09090909090909091\linewidth]{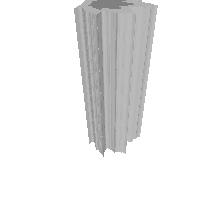} &
\includegraphics[width = 0.09090909090909091\linewidth]{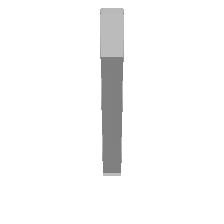}  &
\includegraphics[width = 0.09090909090909091\linewidth]{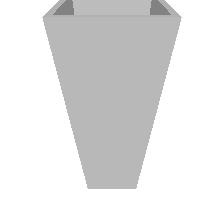}  &
\includegraphics[width = 0.09090909090909091\linewidth]{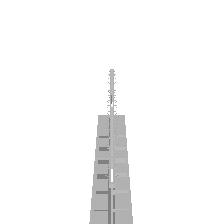}  &
\includegraphics[width = 0.09090909090909091\linewidth]{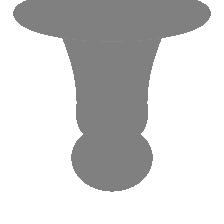}  &
\includegraphics[width = 0.09090909090909091\linewidth]{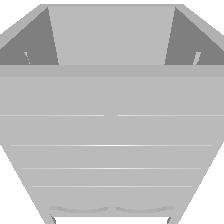}  &
\includegraphics[width = 0.09090909090909091\linewidth]{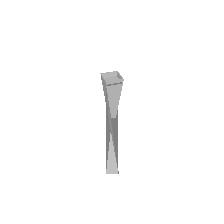}  &
\includegraphics[width = 0.09090909090909091\linewidth]{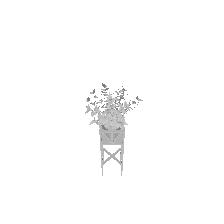}  &
\includegraphics[width = 0.09090909090909091\linewidth]{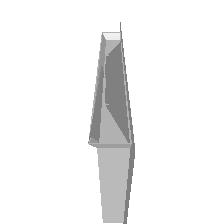}  &
\includegraphics[width = 0.09090909090909091\linewidth]{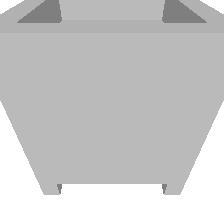}  &
\includegraphics[width = 0.09090909090909091\linewidth]{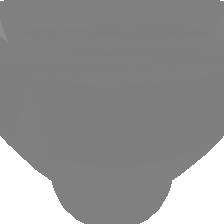}  \\ \hline
\end{tabular}
}
\vspace{2pt}
\caption{  \textbf{Qualitative Examples for Object Retrieval}: \textit{(left):} we show some query objects from the test set. \textit{(right)}: we show top ten retrieved objects by our MVTN from the training set.
}
    \label{fig:imgs-retr-sup}
\end{figure*}

We show qualitative examples of our retrieval results using the MVTN-spherical with ViewGCN in \figLabel{\ref{fig:imgs-retr-sup}}. Note that the top ten retrieved objects for all these queries are positive (from the same classes of the queries). 

\clearpage \clearpage

\end{appendices}

\clearpage \clearpage
\bibliography{sn-bibliography}

\end{document}